\documentclass[10pt,journal,compsoc]{IEEEtran}
\usepackage{graphicx}
\usepackage{amsmath}
\usepackage{amssymb}
\usepackage{booktabs}
\usepackage{multirow}
\usepackage{xcolor}

\newcommand{\eg}{e.g., }
\newcommand{\ie}{i.e., }
\newcommand{\etal}{et al. }

\usepackage[pagebackref,breaklinks,colorlinks]{hyperref}

\usepackage[capitalize]{cleveref}
\crefname{section}{Sec.}{Secs.}
\Crefname{section}{Section}{Sections}
\Crefname{table}{Table}{Tables}
\crefname{table}{Tab.}{Tabs.}

\ifCLASSOPTIONcompsoc
  \usepackage[nocompress]{cite}
\else
  \usepackage{cite}
\fi
\ifCLASSINFOpdf
\else
\fi

\usepackage{caption}
\begin{document}
%
\title{
Booster: a Benchmark for Depth from Images of Specular and Transparent Surfaces
}
%
%
%
%

\author{
        Pierluigi~Zama~Ramirez, 
        Alex~Costanzino,
        Fabio~Tosi, 
        Matteo~Poggi, \\
        Samuele~Salti, 
        Stefano~Mattoccia, 
        and~Luigi~Di~Stefano
       }

\IEEEtitleabstractindextext{%
\begin{abstract}
    Estimating depth from images nowadays yields outstanding results, both in terms of in-domain accuracy and generalization. However, we identify two main challenges that remain open in this field: dealing with non-Lambertian materials and effectively processing high-resolution images. Purposely, we propose a novel dataset that includes accurate and dense ground-truth labels at high resolution, featuring scenes containing several specular and transparent surfaces.
    Our acquisition pipeline leverages a novel deep space-time stereo framework, enabling easy and accurate labeling with sub-pixel precision.
    The dataset is composed of 606 samples collected in 85 different scenes,
    each sample includes both a high-resolution pair (12 Mpx) as well as an unbalanced stereo pair (Left: 12 Mpx, Right: 1.1 Mpx), typical of modern mobile devices that mount sensors with different resolutions. Additionally, we provide manually annotated material segmentation masks and 15K unlabeled samples. The dataset is composed of a train set and two test sets, the latter devoted to the evaluation of stereo and monocular depth estimation networks. Our experiments highlight the open challenges and future research directions in this field.
\end{abstract}

}


\twocolumn[{
\renewcommand\twocolumn[1][]{#1}
\maketitle
\begin{center}
    \vspace{-1cm}
    \setlength{\tabcolsep}{2pt}
    \begin{tabular}{c c c c c}
        \includegraphics[width=0.17\textwidth]{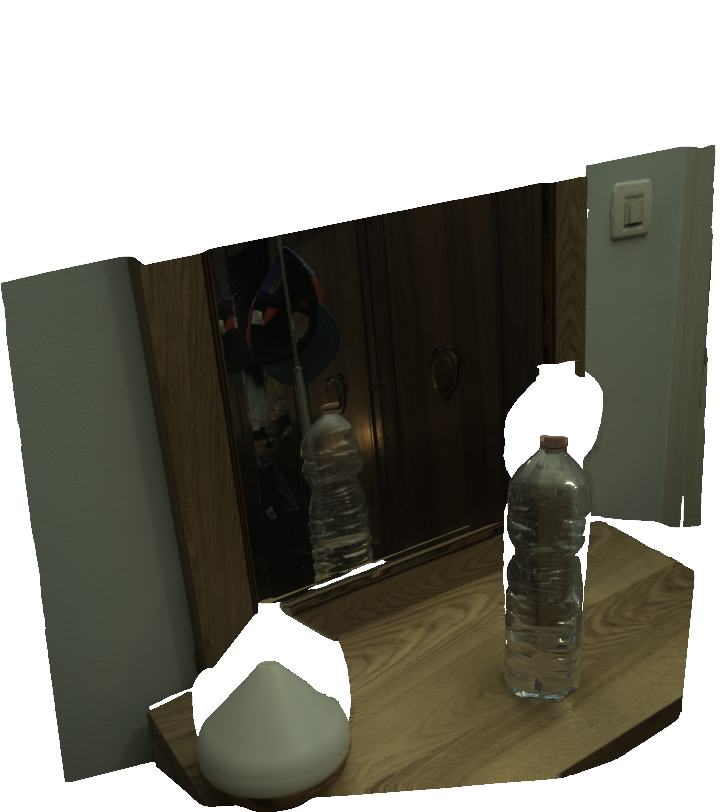} &
        \includegraphics[width=0.17\textwidth]{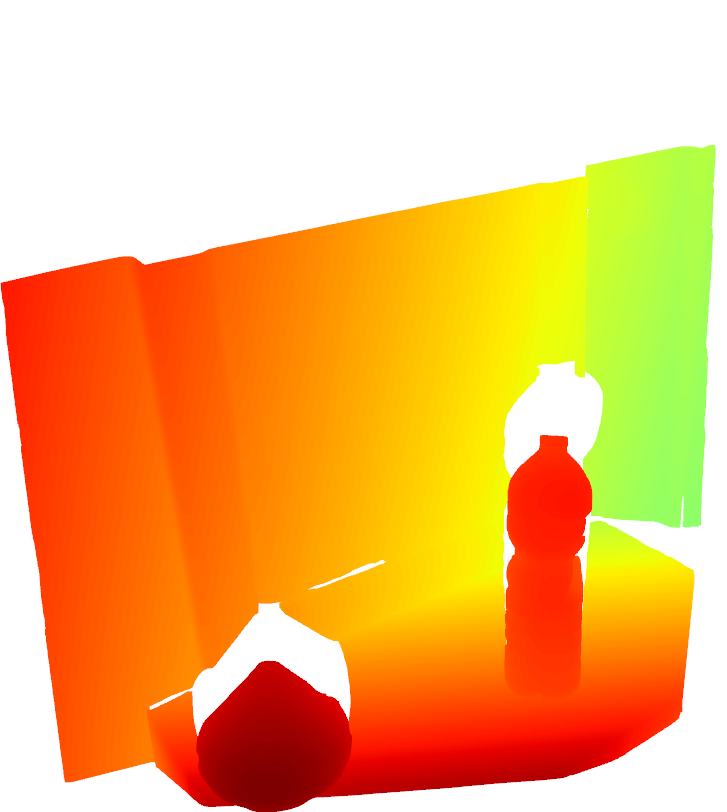} &
        \includegraphics[width=0.17\textwidth]{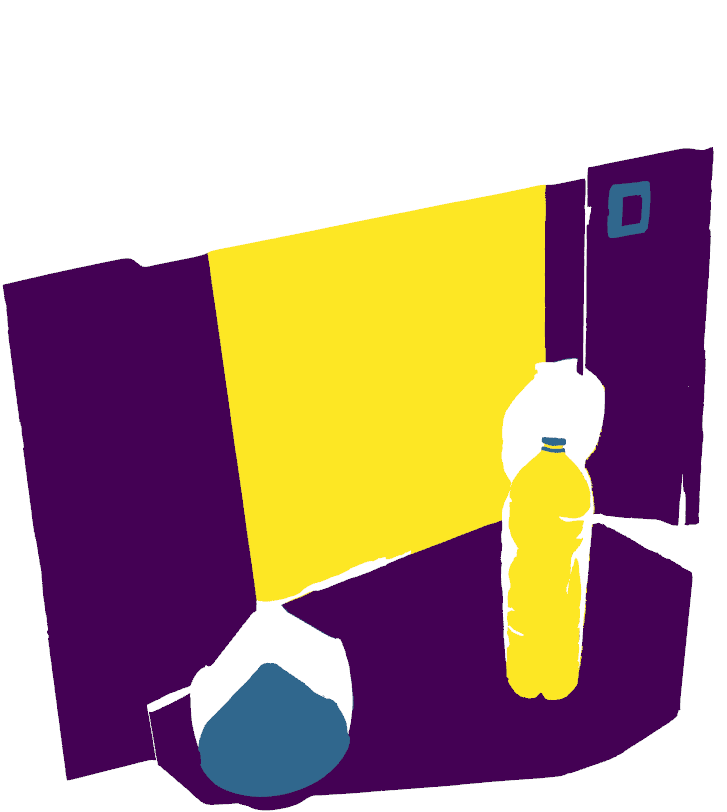} &
        \includegraphics[width=0.17\textwidth]{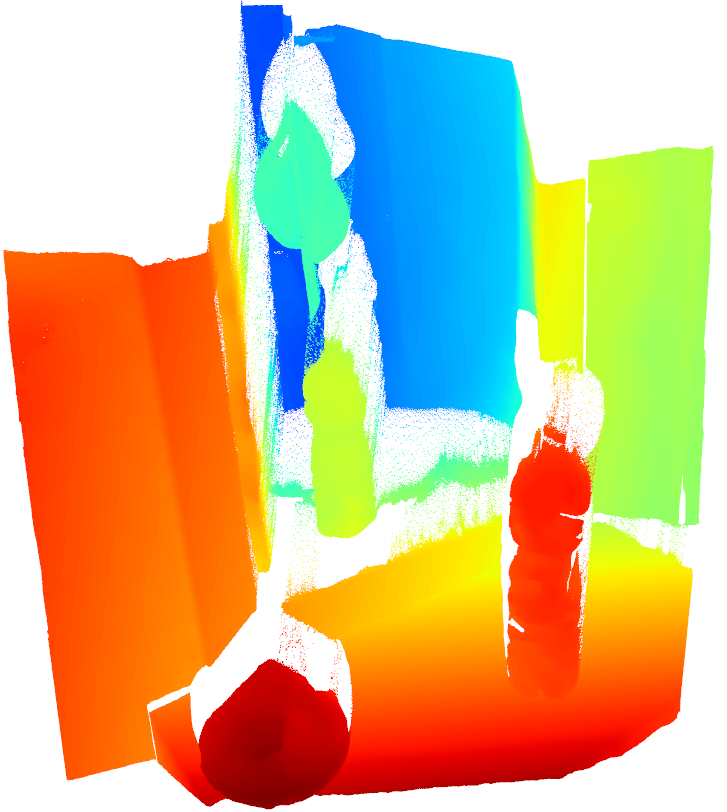} &\includegraphics[width=0.17\textwidth]{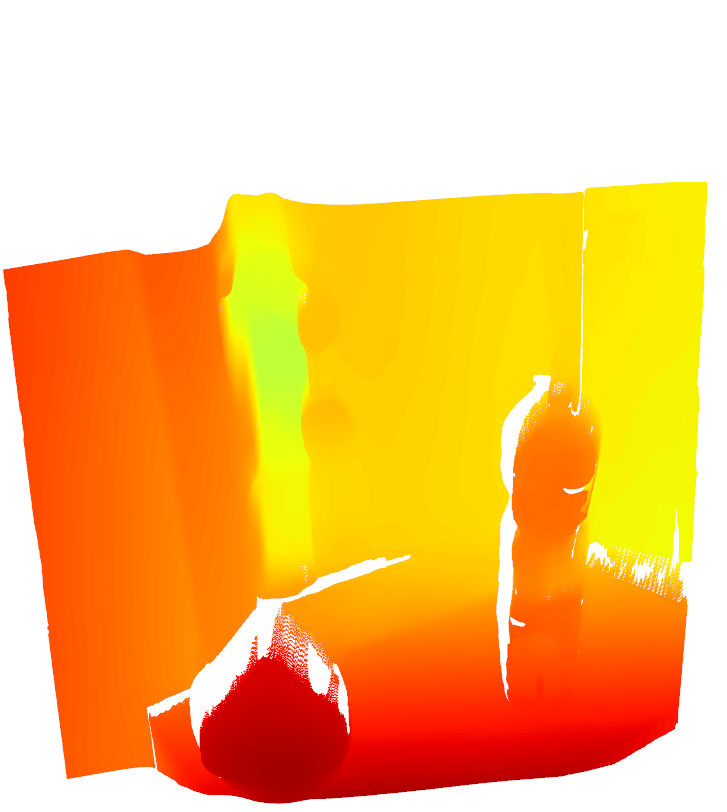}
        \\
        \small \textbf{RGB} & \small \textbf{GT} & \small \textbf{Materials} & \small \textbf{Stereo} &  \textbf{Mono} \\
    \end{tabular}
    \label{fig:teaser}
\end{center}
\small \hypertarget{fig:teaser}{Fig. 1.} \textbf{Exemplar scene from Booster.} Our dataset contains images acquired in indoor environments with specular and transparent objects such as the mirror (RGB column). For each scene, we provide dense ground-truth (GT column) and dense  segmentation masks that identify the most challenging materials (Materials column). State-of-the-art stereo  \cite{li2022practical} (Stereo column) and monocular \cite{Ranftl_2021_ICCV} (Mono column) networks struggle in these scenes, highlighting the open challenges in monocular and stereo depth estimation.
\vspace{1cm}
}]

\IEEEdisplaynontitleabstractindextext

%
\IEEEpeerreviewmaketitle

\IEEEraisesectionheading{\section{Introduction}\label{sec:intro}}



\IEEEPARstart{R}{}ecovering depth information from images is a long-standing problem in computer vision, thoroughly investigated because of its many advantages compared to alternative methodologies -- mostly based on active sensors -- such as the lower cost and intrusiveness in the sensed environments.
Among image-based approaches, two of the cheapest and most studied strategies are stereo matching \cite{scharstein2002taxonomy,poggi2021synergies} and, recently, monocular depth estimation \cite{zhao2020monocular, bhoi2019monocular}, or \textit{depth-from-mono}.
Both approaches already yield astonishing results thanks to the availability of challenging benchmarks \cite{Geiger2012CVPR,Menze2015CVPR,scharstein2014high,schoeps2017cvpr} on top of which the research community competes for the highest ranks.
All the top-performing approaches for mono \cite{miangoleh2021boosting} and stereo \cite{lipson2021raft,li2022practical} are based on deep learning and produce incredibly accurate results. For instance, the best stereo networks achieve error rates close to 1\% on popular benchmarks such as KITTI 2012 and 2015 \cite{Geiger2012CVPR,Menze2015CVPR} or ETH3D \cite{schoeps2017cvpr}. Monocular networks are slightly behind in performance due to the more difficult and ill-posed setup, but still provide very good results \cite{midas, miangoleh2021boosting}.
Should this evidence suggest that depth estimation, thanks to deep learning,  can be considered a solved problem? As shown in Fig. \hyperref[fig:teaser]{1}, we believe that this is definitely not the case and, rather, it is time for the community to focus on the \textbf{open challenges} left unsolved in the field. In particular, we identify two of such challenges, namely i) non-Lambertian surfaces and ii)  high-resolution images. 

Regarding the former, there are many materials and surfaces representing a hurdle to most depth estimation methodologies. 
On the one hand, finding corresponding pixels belonging to transparent or specular surfaces is extremely challenging when dealing with stereo, and might even be an inherently ill-posed problem in many cases, e.g., can we really find the depth of a mirror by matching pixels?
On the other hand, monocular networks estimate depth based on high-level cues such as scene context or relative object size. Thus, in theory, they could estimate meaningful depth values also for non-Lambertian surfaces. Indeed, if we look at the depth predicted for the mirror in Fig.  \hyperref[fig:teaser]{1}, the monocular network is more accurate than the stereo counterpart.
Nonetheless, there are still large errors caused by the lack of accurate ground-truths for such objects in most real datasets used for training. Even though we find labels for such surfaces in synthetic datasets obtained with graphic simulation, they are too unrealistic, and monocular networks trained on them do not generalize well to the real world. Thus, we cannot train a network that is robust to non-Lambertian objects to date, but we believe this can be achieved by having the availability of a large amount of properly labeled training samples.

Concerning the latter, when considering higher-resolution images, only a few networks can properly handle high-resolution images.
Indeed, processing large images is computationally expensive, as well as it requires a much larger receptive field compared to the one of most state-of-the-art networks.
Indeed, when looking at the performance of top-ranking stereo networks on higher resolution datasets, such as the Middlebury 2014 \cite{scharstein2014high} benchmark, we can notice much higher errors than when facing lower resolution datasets. Additionally, when the image dimensions increase, the disparity range and the number of occluded and untextured pixels make finding the exact corresponding pixels much more difficult.
Even in the case of monocular networks, processing higher-resolution images is much more complex and requires networks that feature larger receptive fields and reason at multiple context levels. However, this problem is investigated only by a few approaches \cite{miangoleh2021boosting}, and there is still a large room for improvement. 

Finally, we highlight a further challenge introduced by the camera setup mounted on modern devices, \eg smartphones, which are built up of sensors with different specifications such as resolution and focal length. 
In such a setting, one may wish to recover a depth map at the highest of the resolutions of two cameras mounted on the device, \ie solve an \textit{unbalanced} stereo problem.  However, such a research direction has been barely explored so far \cite{liu2020visually,aleotti2021neural}.

The open challenges just discussed drove us towards collecting Booster \cite{booster}, a high-resolution dataset framing scenes containing many objects and surfaces that are either specular or transparent, as well as very large untextured regions.
Instead of limiting our attention to the stereo-matching problem, in this paper, we extend our dataset with brand-new labeled samples, and we use them to benchmark state-of-the-art monocular networks.
The labels provided by Booster are obtained by leveraging a 
\textit{deep} space-time stereo pipeline \cite{davis2003spacetime} which combines disparity estimates computed from multiple static images -- up to 100 -- acquired under a variety of texture patterns projected onto the scene from different directions and after having carefully painted all non-Lambertian surfaces.
We employ a state-of-the-art, pre-trained deep network, RAFT-Stereo \cite{lipson2021raft}, to compute and accumulate disparity maps through time within the space-time framework.
Moreover, we clean all the final maps to remove outliers and artifacts ensuring high-quality disparity labels.
We point out that we provide depth labels for the closest surfaces only -- \ie we do not provide annotations for surfaces being behind transparent objects or reflected over mirrors. 
As such, our dataset mainly addresses scenarios dealing with autonomous driving, obstacle avoidance, and robotic manipulation, in which the distance to the closest object / person / obstacle is critical, and thus is the one we are interested in.
The main contributions of our work are:
\begin{itemize}
\item We propose a novel dataset consisting of both high-resolution and unbalanced stereo pairs featuring a large collection of labeled non-Lambertian objects. We have acquired 85 scenes under different illuminations, yielding 606 \textit{balanced} stereo pairs at 12 Mpx and 606 \textit{unbalanced} pairs, each consisting of 12 Mpx and 1.1 Mpx images. The latter setup provides the first-ever dataset for unbalanced stereo matching, as prior work is limited to simulation experiments \cite{liu2020visually,aleotti2021neural}.
Samples are annotated with dense ground-truth disparities and grouped into 228 training images, 191 test images for the stereo benchmark, and  187 for the monocular depth benchmark. 
The dataset along with an online evaluation service are available at \url{https://cvlab-unibo.github.io/booster-web/}. To ensure a fair evaluation of future methods we withheld ground-truths of the test splits, and we provide only the left images in the monocular case.

\item Images are labeled in a semi-automatic manner employing a deep space-time stereo framework, which joins the use of modern stereo networks \cite{lipson2021raft} with the classical space-time stereo framework \cite{davis2003spacetime}. 

\item Together with ground-truth disparities, we provide manually labeled segmentation maps to identify and rank hard-to-match materials based on specularity and transparency. This feature is conducive to studying the open challenges addressed in this paper. 
Additionally, we provide 15K raw pairs -- in balanced and unbalanced settings -- to foster further advances in self-supervised approaches.

\item We evaluate the latest and most accurate, state-of-the-art  stereo  \cite{chang2018psmnet,zhang2019ga,yang2019hierarchical,cheng2020hierarchical} and monocular \cite{midas,Ranftl_2021_ICCV,miangoleh2021boosting} networks, respectively, on the stereo and mono test splits of our dataset, as trained by their authors. Our evaluation highlights the open challenges that wait to be faced by the community. 

\end{itemize}

\section{Related Work}

We survey existing works most relevant to ours.


\textbf{Traditional and Deep Stereo.} Stereo matching has a long history. At first, algorithms were classified into \textit{local} and \textit{global} ones, according to the specific pipeline they implemented \cite{scharstein2002taxonomy}. 
Several methods such as \cite{di2004fast,yoon2006adaptive,yang2012non,de2011linear,hosni2012fast} have been developed, with Semi-Global Matching (SGM) \cite{hirschmuller2007stereo} becoming by far the most popular.
When deep learning conquered the scene, the first attempts to infuse it into stereo algorithms were aimed at replacing the single steps of the conventional pipeline \cite{scharstein2002taxonomy} with small neural networks, \eg to perform matching cost computation \cite{zbontar2016stereo,Chen_2015_ICCV,luo2016efficient}, optimization \cite{seki2016patch,seki2017sgm-net} or refinement \cite{batsos2018recresnet,gidaris2017detect,aleotti2021neural}.
However, end-to-end deep soon became the main solution for stereo \cite{mayer2016large,Kendall_2017_ICCV,Pang_2017_ICCV_Workshops}, ranking on top of the KITTI 2012 \cite{Geiger2012CVPR} and 2015 \cite{Menze2015CVPR} benchmarks.

\begin{figure*}[t]
    \centering
    \includegraphics[width=\textwidth]{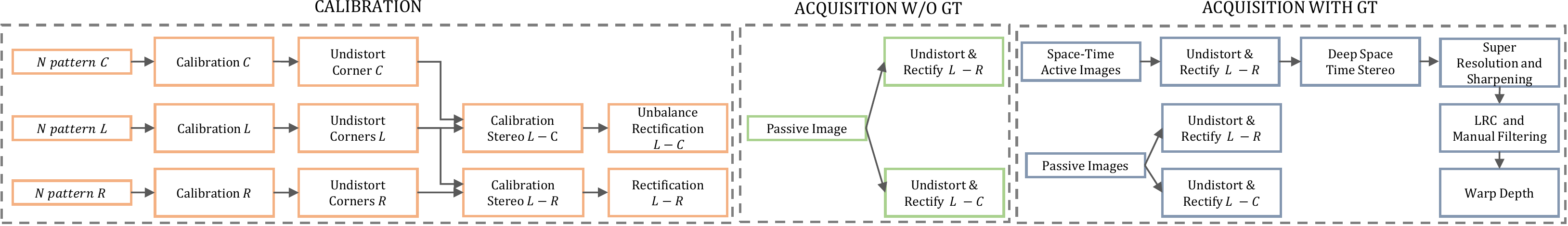}
    \vspace{-0.25cm}\caption{\textbf{Dataset acquisition overview.} Our collection pipeline is made of three, main phases. Left (orange): offline calibration of our trinocular rig and the two stereo systems $L-C$ and $L-R$. Middle (green): image acquisition without ground-truth. Right (grey): acquisition of textured images, used to obtain ground-truth labels.}
    \label{fig:pipeline}
\end{figure*}

Through the years, a large number of custom architectures have been proposed \cite{Liang_2018_CVPR,chang2018psmnet,zhang2019ga,yang2019hierarchical,cheng2020hierarchical,song2018edgestereo,yang2018segstereo,dovesi2020real,Tonioni_2019_CVPR,poggi2021synergies}.
Moreover, further research trends focusing on self-supervised learning strategies \cite{Tonioni_2017_ICCV,tonioni2020unsupervised,Zhou_2017_ICCV,wang2019unos,lai19cvpr,aleotti2020reversing,Poggi2021continual} and zero-shot generalization across datasets \cite{cai2020matchingspace,zhang2019domaininvariant,aleotti2021neural,liu2022graftnet,chuah2022itsa} emerged, as well as the unbalanced stereo task \cite{liu2020visually,aleotti2021neural} aimed at estimating disparity from a pair of images captured from heterogeneous cameras.

\textbf{Stereo benchmarks.} 
The existence of several benchmarks over which to evaluate stereo algorithms made possible the rapid evolution of stereo networks.
While early datasets counted only a few dozen samples, acquired in indoor and controlled environments \cite{scharstein2002taxonomy,scharstein2003high,hirschmuller2007evaluation,scharstein2007learning}, larger and more complex stereo datasets appeared in the last 10 years, starting with KITTI 2012 \cite{geiger2010efficient} and 2015 \cite{Menze2015CVPR}, featuring driving environments and annotated by means of a Velodyne LiDAR sensor, followed by Middlebury 2014 \cite{scharstein2014high}, consisting of indoor scenes, captured at up to 6 Mpx and annotated through pattern projection, and ETH3D \cite{schops2017multi} collecting both indoor and outdoor scenes.
Following the success of KITTI, more driving-themed datasets have been released through the years, such as DrivingStereo \cite{yang2019drivingstereo}, Argoverse \cite{Argoverse}, Apolloscape \cite{huang2019apolloscape} and DSEC \cite{Gehrig21ral}.
However, the accuracy of state-of-the-art stereo networks on these datasets is very high, highlighting that most of the challenges they features are indeed almost solved.
On the contrary, we will highlight that these models struggle with Booster.

\textbf{Monocular Depth Estimation.} Pioneer works aiming at estimating depth from a single image rely on hand-crafted features, encoding perceptual cues such as object size, texture density, or linear perspective meaningful of depth~\cite{Saxena2008}. 
Deep learning made possible huge advances on this task, by enabling direct learning of depth-related priors from annotated data \cite{chen2016single, eigen2014depth,laina2016deeper, Ramamonjisoa_2020_CVPR, wang2020cliffnet}. The key to the rapid evolution of this research trend is in the availability of large-scale datasets \cite{Chen_2020_CVPR, li2018megadepth, Xian2018DepthWeb,Zamir_2018_CVPR,midas,Ranftl_2021_ICCV} annotated with depth labels, as well as the development of self-supervised strategies \cite{godard2017unsupervised,gonzales2020forgetlidar,godard2019monodepth2,Zhou_2017_ICCV,watson2019depthhints,guo2018learning,jiang2018self,johnston2020self,Tosi_2019_CVPR,Tosi_2020_CVPR,Poggi_2020_CVPR} to compensate for the lack of annotations.
Similarly to what was observed for stereo, a further challenge concerns high-resolution images. On this track, Miangoleh \etal \cite{miangoleh2021boosting} proposed a framework for restoring the high-frequency details in the estimated depth maps by changing the input of a pre-trained monocular network and then merging the multiple estimations obtained accordingly. 

\textbf{Monocular Depth Estimation Benchmarks.}
Compared to stereo, monocular depth estimation is a much more recent research trend that bloomed when several datasets annotated with depth were already available. Among them, the NYUv2 dataset \cite{SilbermanECCV12}
and the KITTI raw sequences \cite{Geiger2012CVPR} are the most used as representatives of indoor and outdoor environments, respectively. In particular, the raw KITTI dataset has been used according to the split proposed by Eigen \etal \cite{eigen2014depth}, then being followed by a dedicated benchmark \cite{uhrig2017sparsity}. 
Following MiDaS \cite{midas}, more datasets have been included as part of a robust benchmark aimed at measuring the robustness of monocular depth networks, i.e. ETH3D \cite{schops2017multi}, TUM \cite{sturm12iros} and Sintel \cite{sintel}.
Again, none of these datasets features non-Lambertian objects. When they are present, ground-truth labels are not available because of the failure of the active sensors used to produce annotations. On the contrary, Booster features some very challenging yet accurately annotated non-Lambertian objects, as well as images at much higher resolution.

\label{sec:pipeline}

\section{Processing pipeline}

In this section, we report in detail the single steps implemented in our collection pipeline, including camera calibration, image acquisition, and ground-truth annotation. In Fig. \ref{fig:pipeline} we sketch an overview of the whole pipeline.

\subsection{Camera setup.} 
\begin{figure}
    \centering
    \begin{tabular}{cc}
        \includegraphics[clip,trim=0 6cm 0 0,width=0.24\textwidth]{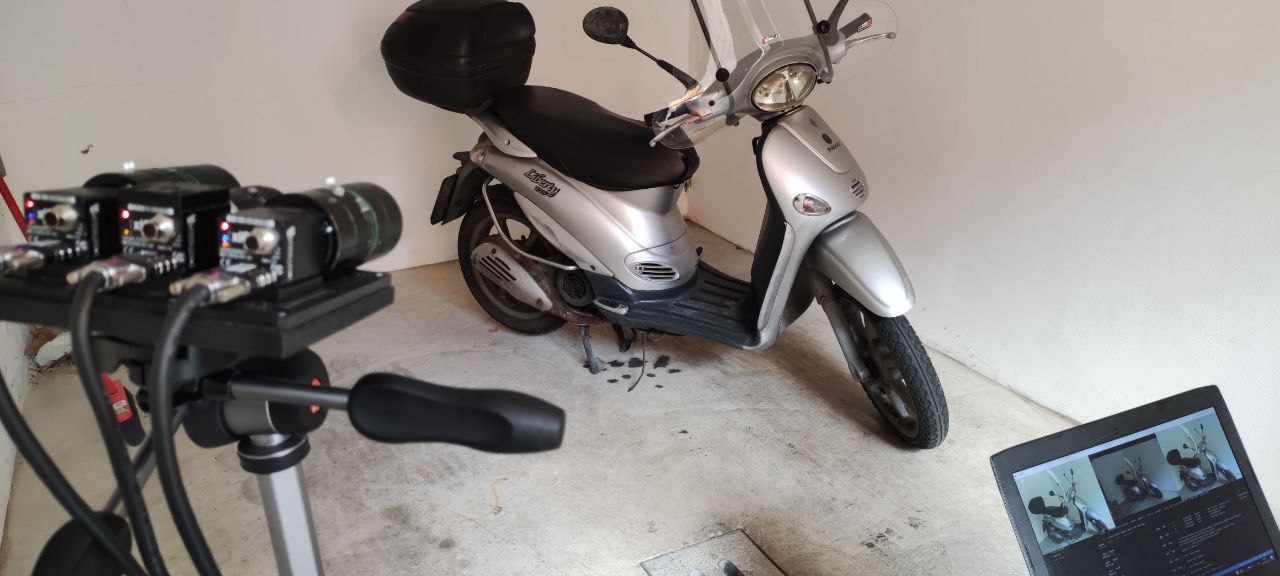} & 
        \multirow{3}{*}{
        \begin{tabular}{c}
            \includegraphics[width=0.15\textwidth]{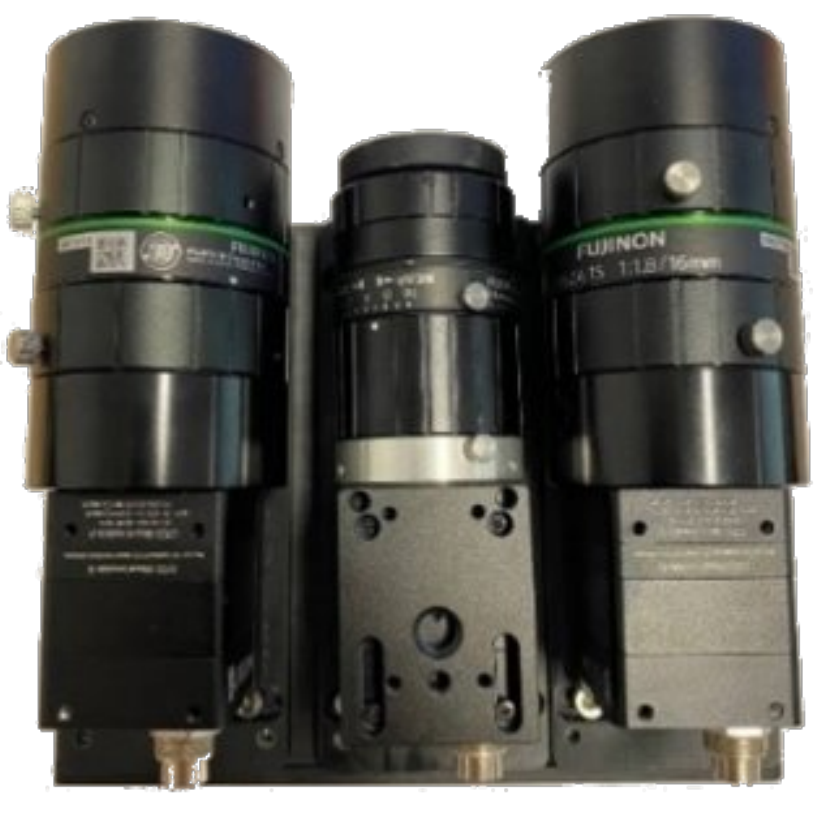} \\
            \small $L$ \quad\quad $C$ \quad\quad $R$ \\
        \end{tabular}} \\
        \includegraphics[clip,trim=0 12cm 0 4cm,width=0.24\textwidth]{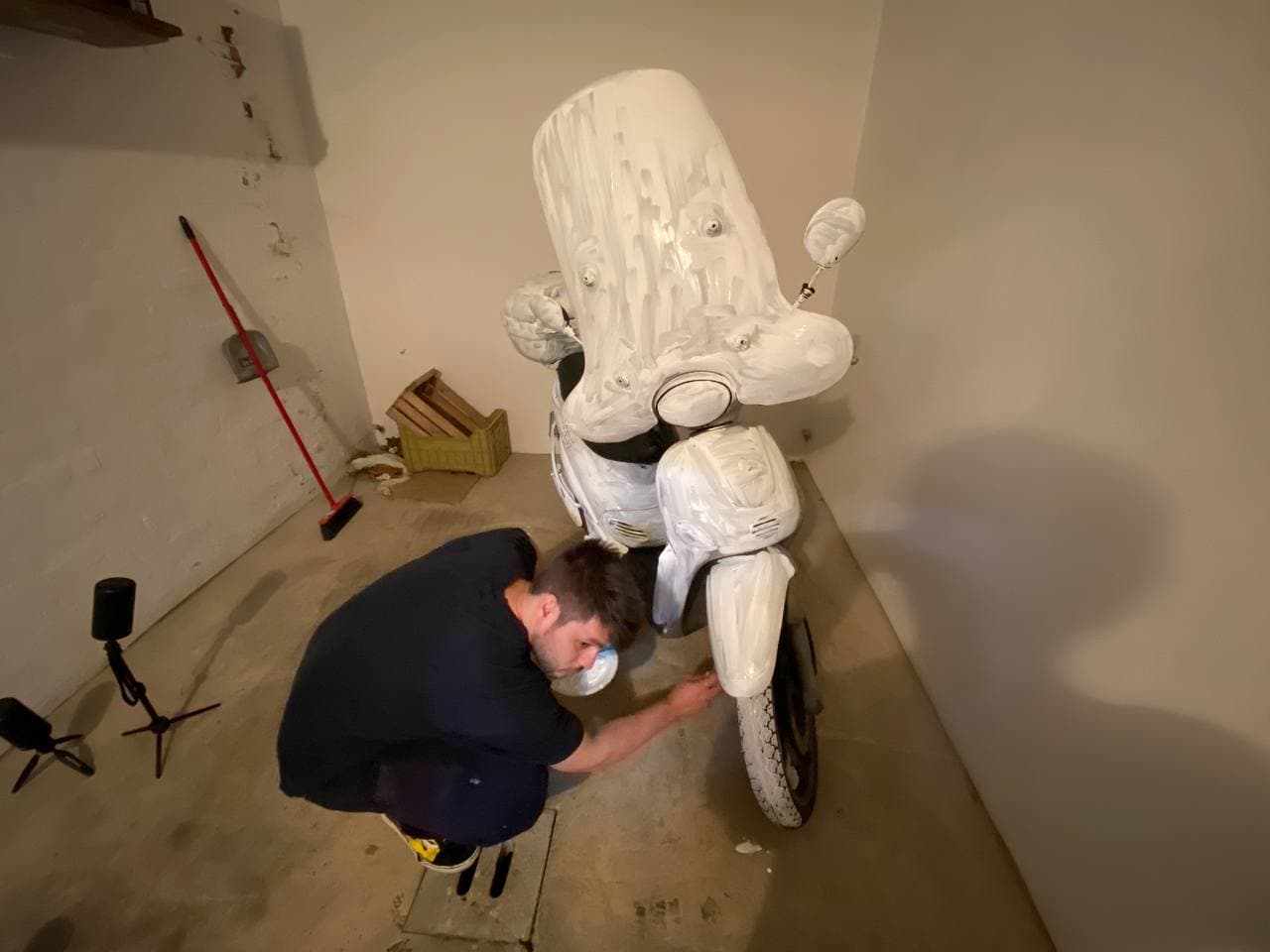} \\
        \includegraphics[clip,trim=0 6cm 0 0,width=0.24\textwidth]{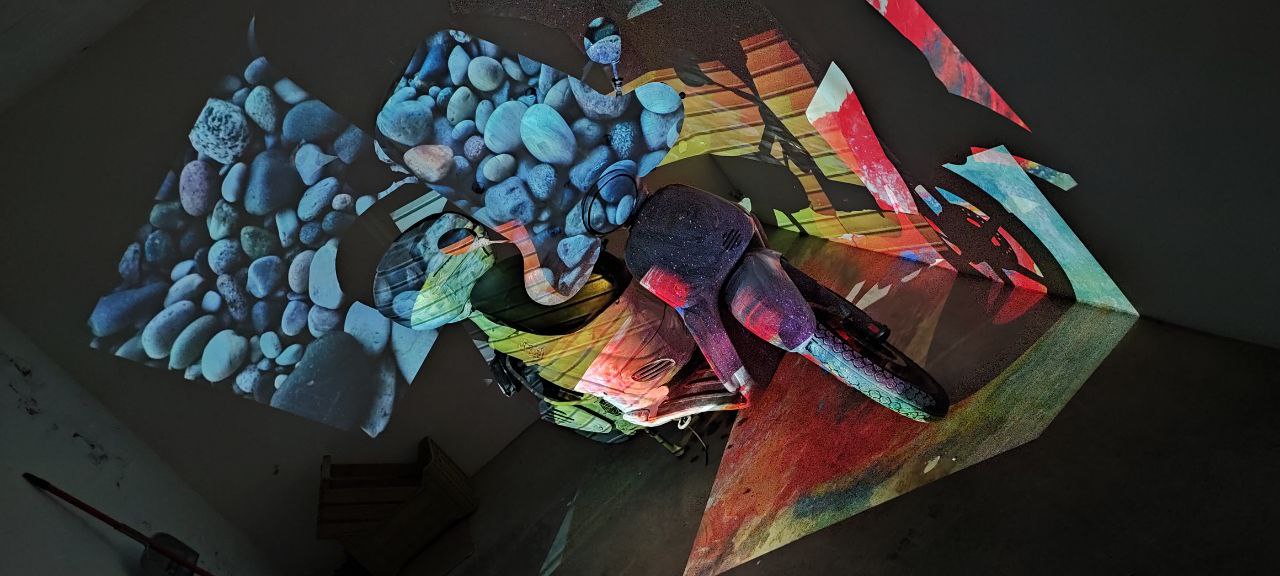}
        \\ 
    \end{tabular}
    \vspace{-0.25cm}\caption{\textbf{Cameras setup and acquisition.} On the left: i) passive stereo pairs collection, ii) painting of reflective/transparent materials, iii) textured stereo pairs acquisition. On the right, top-view of our camera rig, with $L$ and $R$ being two 12 Mpx sensors, and $C$ a wide-angle 2.3 Mpx sensor.}
    \label{fig:rig}
\end{figure}

Our dataset has been collected with a custom stereo rig, mounting 2 high-resolution cameras with Sony IMX253LQR-C 12.4 Mpx sensors and a third camera with a Sony IMX174LQJ-C 2.3 Mpx sensor positioned in between the two high-res cameras. The three cameras are depicted in Fig. \ref{fig:rig}, on the right, and we denote them as $L$, $C$, and $R$ from left to right, respectively. As such, our rig allows us to collect balanced stereo pairs from ($L$,$R$) cameras and unbalanced ones from ($L$,$C$), with camera $L$ collecting the reference image in both pairs.
The balanced and unbalanced setups feature a baseline of $\sim$ 8 and 4 centimeters.

\subsection{Camera calibration.}
As for standard stereo cameras, an offline calibration procedure is necessary before collecting images with our rig. Given our peculiar setup, we first calibrate the single cameras, then perform stereo calibration to obtain both rectified balanced and unbalanced pairs.

\begin{figure*} [t]
    \centering
    \renewcommand{\tabcolsep}{2pt}   
     \scalebox{0.90}{
    \begin{tabular}{ccc|cc|cc}
    \multicolumn{3}{c}{\textit{Raw Images}} & \multicolumn{2}{c}{\textit{Balanced Stereo}} & \multicolumn{2}{c}{\textit{Unbalanced Stereo}} \\
    \includegraphics[width=0.16\textwidth]{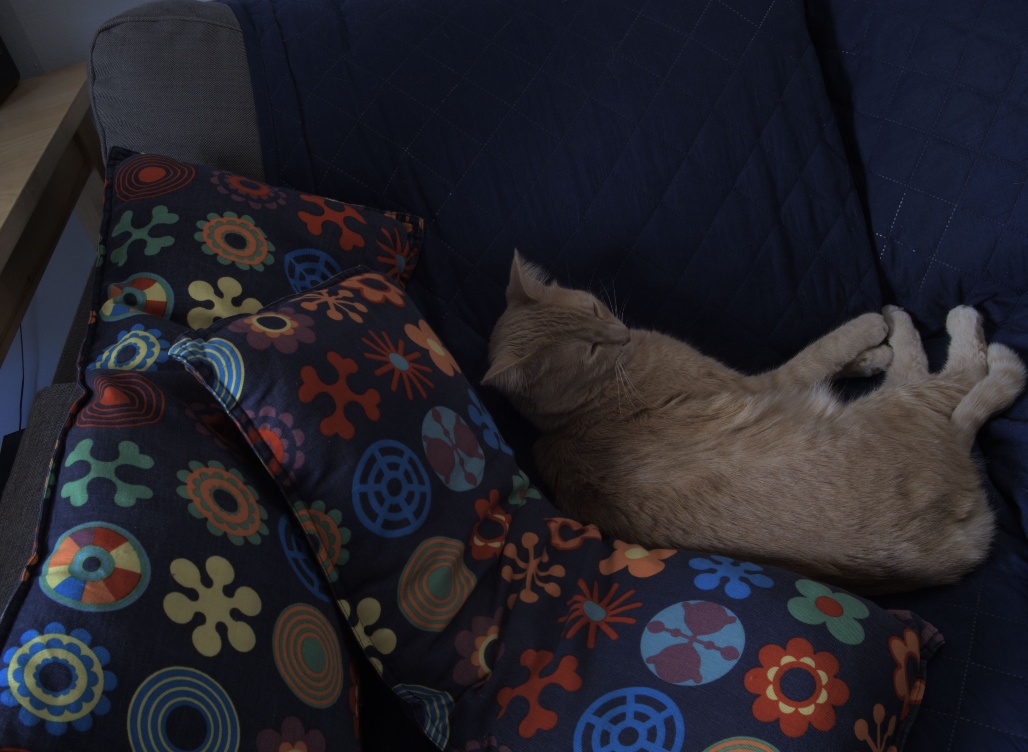}  &
    \includegraphics[width=0.12\textwidth]{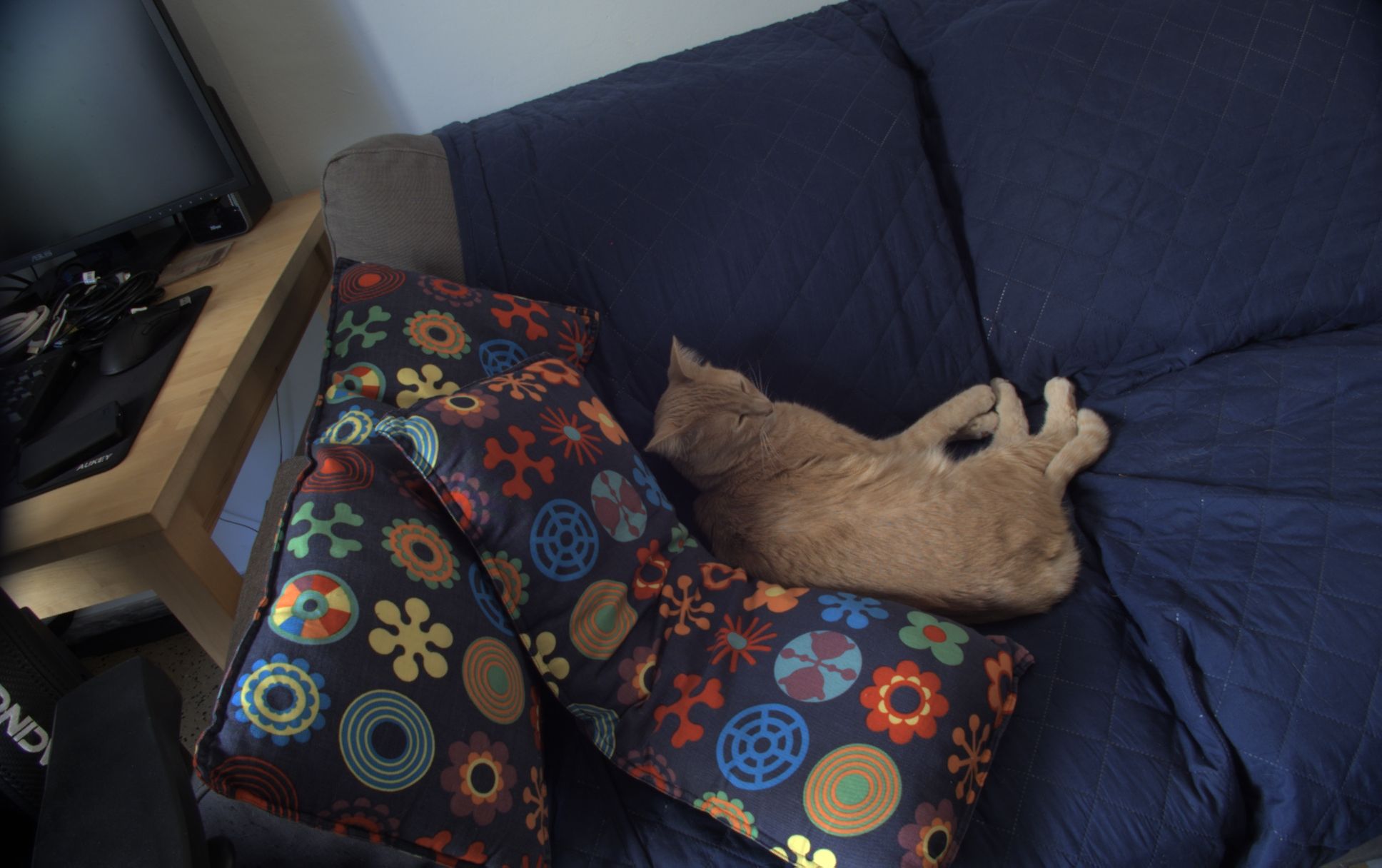} &
    \includegraphics[width=0.16\textwidth]{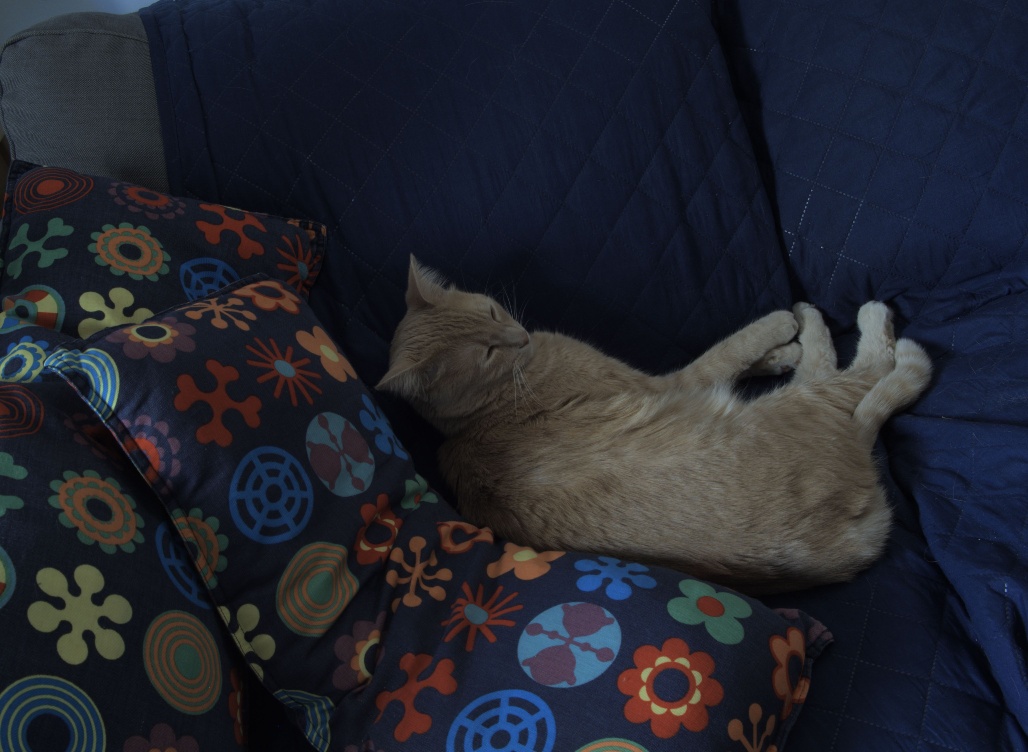} &
    \includegraphics[width=0.16\textwidth]{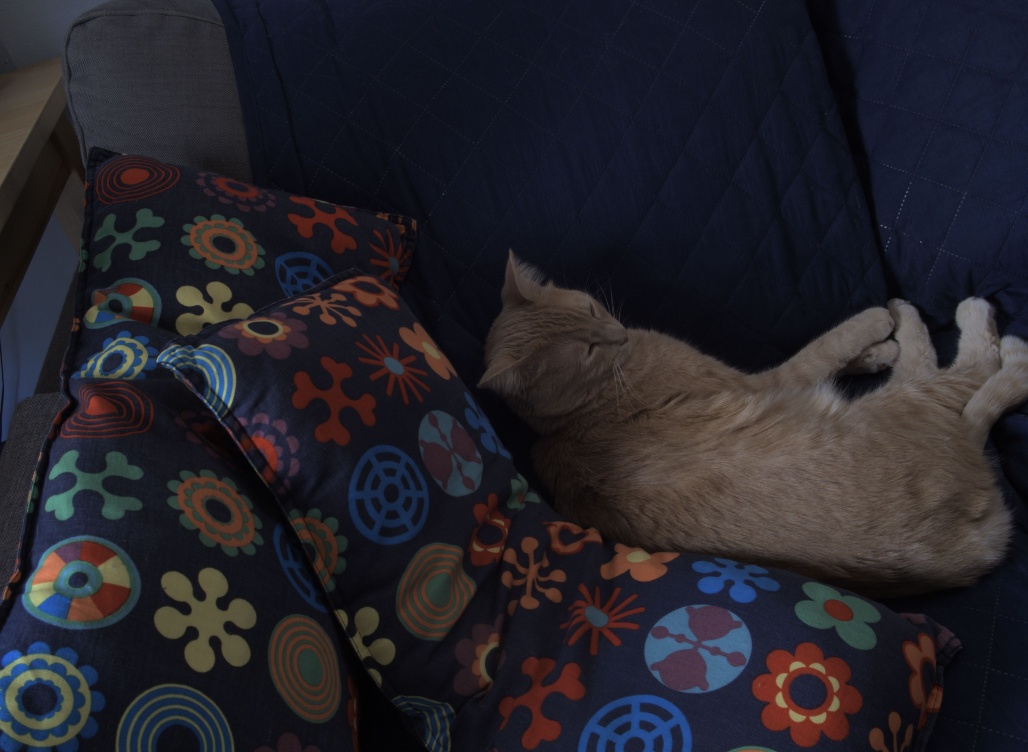}  &
    \includegraphics[width=0.16\textwidth]{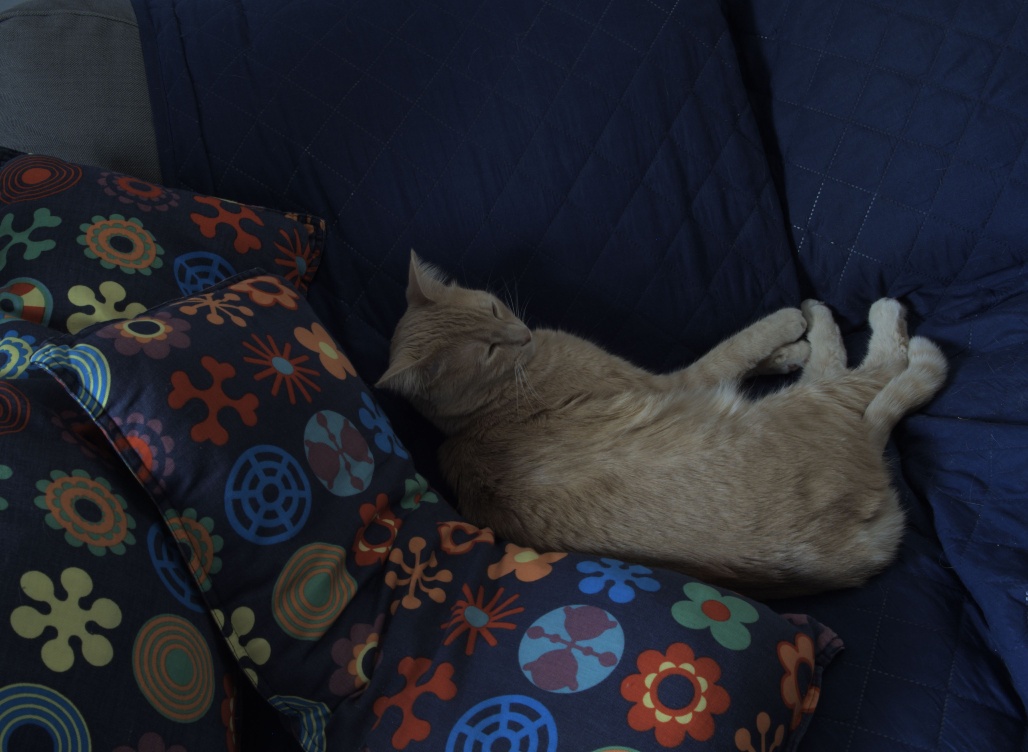}  &
    \includegraphics[width=0.16\textwidth]{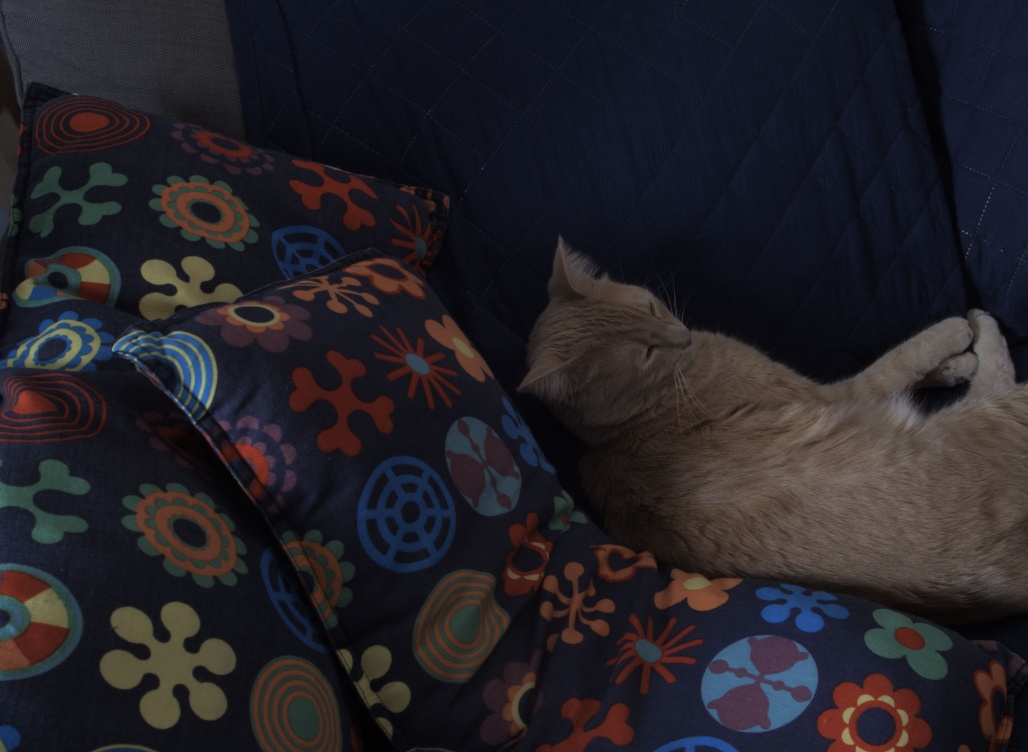}  &
    \includegraphics[width=0.08\textwidth]{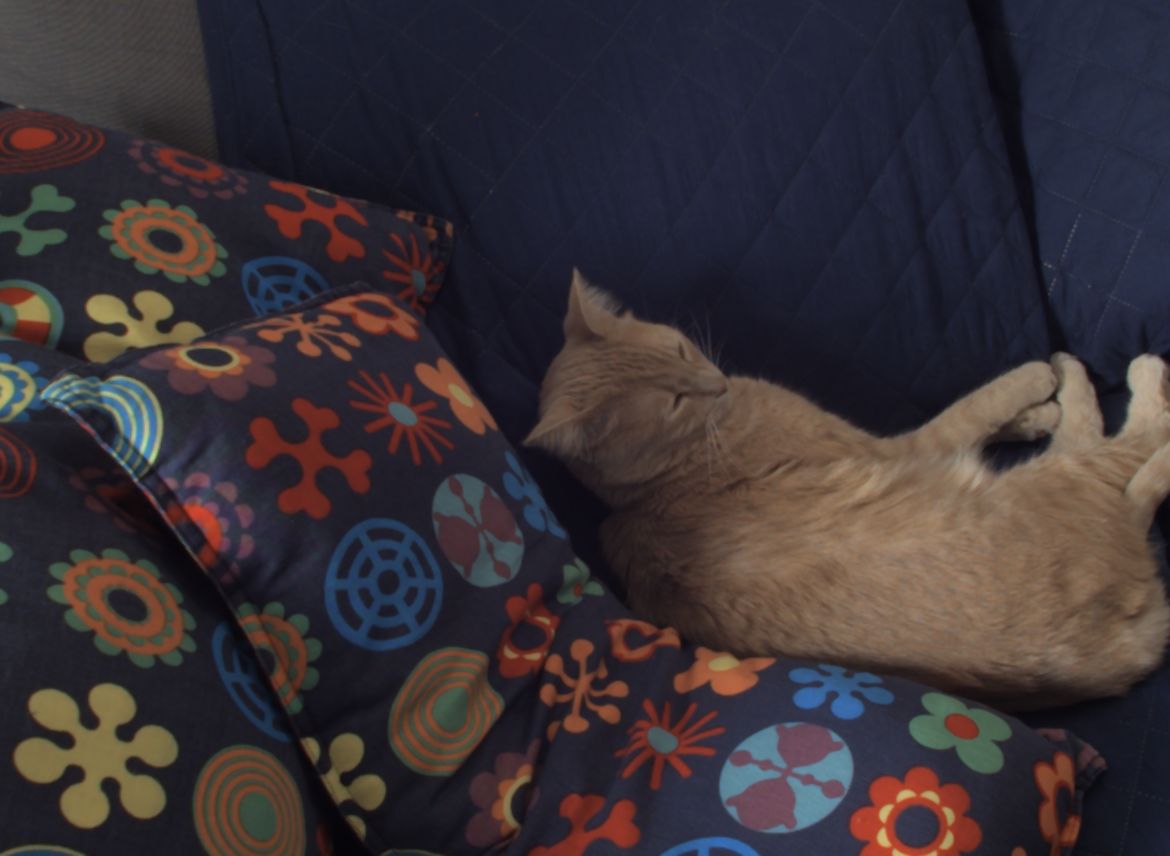}  \\
     \small{$L$: 4112$\times$3008} & 
     \small{$C$: 1936$\times$1216} & 
     \small{$R$: 4112$\times$3008} & 
     \small{$L_{LR}$: 4112$\times$3008} & 
     \small{$R_{LR}$: 4112$\times$3008} &
     \small{$L_{LC}$: 4112$\times$3008} & 
     \small{$C_{LC}$: 1170$\times$856} \\
    \end{tabular}
    }
    \caption{\textbf{Balanced/unbalanced rectification examples.} From left to right: $L$, $R$, and $C$ raw images from our rig; $L_{LR}$,$R_{LR}$ rectified balanced stereo pair from the $L-R$ setup; $L_{LC}$,$C_{LC}$ unbalanced rectified stereo pair from the $L-C$ setup.}
    \label{fig:raw_vs_rectified}
\end{figure*}

\textbf{Calibration of individual cameras.}
We start by calibrating each camera separately, according to the pinhole model. To this aim, we acquire N images (\ie 15) framing a known pattern (\ie a chessboard) using our rig.
The distortion-free projective transformation of a pinhole camera is defined as: 
\begin{equation}
    p = A[\mathcal{R}\mathcal{T}]P_w
\end{equation}
with $P_w$ being a 3D point coordinates in the world reference frame (WRF), $p$ a 2D pixel in the image plane, $A$ the intrinsic parameters matrix, and $\mathcal{R}$, $\mathcal{T}$ the rotation and translation applied from the world reference frame (WRF) to the camera reference frame (CRF), respectively. We model lens distortion by means of a vector $Dist=(k_1,k_2, k_3,p_1,p_2)$ with $k_1, k_2, k_3$ denoting the radial distortion parameters and $p1, p2$ the tangential distortion parameters, following OpenCV conventions.
We calibrate independently
each camera to estimate their intrinsic matrices $A_L$, $A_C$, $A_R$ and lens distortion parameters $Dist_L$, $Dist_C$, $Dist_R$ of the $L$, $C$, and $R$ cameras, respectively. Then, we undistort images to then perform a stereo calibration of the two stereo rigs, \ie, the $L-C$ and $L-R$ pairs, by estimating the rotations $R_{LC}$, $R_{LR}$ and translations $T_{LC}$, $T_{LR}$, from the $L$ to $C$, and $L$ to $R$ camera reference systems, respectively.

\textbf{Balanced Stereo Calibration.}
In order to perform stereo matching in a canonical manner, \ie by looking for correspondences on horizontal epipolar lines, we need to estimate the rectification transformations (homographies) to be applied to both images to \textit{rectify} them. For the $L-R$ \textit{balanced} stereo system, we can exploit standard rectification practices since the resolution is the same for both images. Specifically, we use the OpenCV implementation to estimate the new intrinsic matrix $A^{LR}_L$ and $A^{LR}_R$, and the rotations $R^{LR}_L$ and $R^{LR}_R$ of $L$ and $R$ to map the initial image plane into the rectified image plane.
Accordingly, we use this information to obtain rectified stereo pair $L_{LR}, R_{LR}$.

\textbf{Unbalanced Stereo Rectification.}
In order to rectify images collected by $L-C$ cameras pair, we implement the \textit{unbalanced rectification} scheme outlined in \cite{aleotti2021neural}. This strategy allows for warping the images so that they are rectified when brought to the same resolution, either by means of up-sampling or down-sampling operations solely.
For this purpose, we identify the camera with the smallest horizontal field-of-view $HFOV$ as $j$, and the other one as $i$.
\begin{equation}
\begin{cases}
i=L, j=C & \text{if } HFOV_C < HFOV_L  \\
i=C, j=L & \text{if } HFOV_L < HFOV_C \\
\end{cases}
\end{equation}

By acting on the intrinsic parameters of $i$, we simulate a crop and scale of the images collected by it to make them match the $HFOV$, Aspect Ratio ($AR$), and size of images collected by $j$, and then estimate the rectification transformation based on these parameters. For this purpose, we obtain new width and height $\hat{W}_i$ and $\hat{H}_i$ as

\begin{equation}
\hat{W}_i = 2 \tan \frac{HFOV_j}{2}f_i
\hspace{1cm}
\hat{H}_i = \frac{H_j}{W_j} \hat{W}_i
\end{equation}
and crop images collected by $i$ accordingly. Then, we modify the intrinsic parameters of $i$ to simulate the crop and resize needed to match the resolution of $j$ as follows: 

\[
\hat{A}_i =
  \begin{bmatrix}
    f_x^i \cdot \frac{W_j}{\hat{W_i}} & 0 & (u_0^i - \frac{W_i - \hat{W}_i}{2}) \cdot \frac{W_j}{\hat{W_i}}  \\
    0 & f_y^i \cdot \frac{H_j}{\hat{H_i}} & (v_0^i - \frac{H_i - \hat{H}_i}{2}) \cdot \frac{H_j}{\hat{H_i}} \\
    0 & 0 & 1 \\
  \end{bmatrix}
\]

This approach is equivalent to estimating rectification homographies for two cameras collecting images both with height $H_j$ and width $W_j$, finding the new intrinsic  $A^{LC}_L$ and $A^{LC}_R$, and the rotations
$R^{LC}_L$, $R^{LC}_R$, of $L$ and $C$ to map the initial image plane into the rectified one. 
Then, we rescale $A^{LC}_i$ with vertical and horizontal scale factors equal to $\frac{\hat{H}_i}{H_j}$ and $\frac{\hat{W}_i}{W_j}$, respectively, in order to adjust the focal length and piercing point of the camera. Finally, we can rectify the unbalanced pair to obtain $L_{LC}$ and $C_{LC}$.
Fig. \ref{fig:raw_vs_rectified} shows an example of images before and after the rectification procedures.

\subsection{Image acquisition.} After the cameras have been calibrated, we start collecting our dataset. 
Purposely, we place our rig in front of a specific scene for which we want to collect stereo images and, subsequently, obtain ground-truth labels. 
For each acquisition, before starting, we properly set up the stage to allow the capture of one or multiple objects/surfaces embodying some of the open challenges we aim to address with our dataset. Then, the acquisition procedure is made of three main steps, visually presented in Fig. \ref{fig:rig} on right: 1) passive images acquisition -- we capture a set of balanced and unbalanced stereo images under varying lighting conditions; 2) scene painting -- we cover any non-Lambertian surface in the scene with paint or spray, to allow for projecting textures over them; 3) textured image acquisition -- we flood the environment with random patterns projected from multiple directions by six portable projectors, and acquire a hundred images by varying the projected texture. We use colored textures to augment the matching capability of a state-of-the-art deep stereo network since we observed empirically that colorful patterns look more distinctive for deep networks than white-black banded patterns often used for this purpose \cite{davis2003spacetime}. 
From each scene, we obtain a set of passive stereo pairs -- balanced and unbalanced -- under different illumination conditions, and a larger set of textured images. We release the former group as the actual dataset, while the latter is used only to produce ground-truth disparities. 

\subsection{Deep space-time stereo estimation.} The textured images we collected are then processed by means of a novel deep space-time stereo pipeline to infer accurate disparity maps to be used as ground-truth for the corresponding passive pairs.
We implement such a framework by leveraging a state-of-the-art, pre-trained stereo network excelling at generalizing across domains. Since any of the challenging surfaces have been painted and textured during the acquisition phase, such a network will provide some very high-quality initial disparities. Moreover, since multiple textured pairs have been collected, we exploit all of them to improve the predicted labels.
Specifically, since any stereo network encodes matching likelihood inside a \textit{cost-volume}, we aggregate the cost volumes obtained from each independent, textured pair. Consequently, we can integrate the distinctiveness introduced by the single patterns and compensate for the potential lack of texture occurring in some portions of the scenes in single acquisitions.
We select RAFT-Stereo \cite{lipson2021raft} as the engine in our deep space-time stereo framework that uses the dot product between left and right features $\mathbf{f}$ and $\mathbf{g}$ to measure the similarity between them.
Built on this principle, RAFT-Stereo produces a 3D correlation volume to store the matching probability between any pixel in the reference image and all those laying on the same horizontal scanline in the target image:

\begin{equation}
    \mathbf{C}_{ijk} = \sum_h \mathbf{f}_{ijk} \cdot \mathbf{g}_{ikh}, \quad\quad \mathcal{C} \in \mathbb{R}^{H\times W\times W}
\end{equation}
Then, the model recursively predicts a disparity map $\mathbf{d}_i$ by looking up the correlation volume. Such prediction is implemented as a recurrent neural network $\Theta$, fed in input with reference image features $\mathbf{f}$, additional context features $\mathbf{c}$, disparity $\mathbf{d}_{i-1}$ estimated at the previous iteration and correlation scores $\mathbf{C}$ sampled according to $\mathbf{d}_{i-1}$

\begin{equation}
    \mathbf{d}_i = \Theta(\mathbf{f},\mathbf{c},\mathbf{d}_{i-1},\mathbf{C}) 
\end{equation}
After a fixed number of iterations, we obtain the final disparity map $\mathbf{d}$.
Given $T$ textured stereo pairs, we build an accumulated correlation volume $\mathbf{C}^*$ by averaging the single correlation volumes computed from $\mathbf{f}^t$ and $\mathbf{g}^t$ extracted from stereo pair $t$

\begin{equation}
    \mathbf{C}^*_{ijk} = \frac{1}{T} \sum_t \sum_h \mathbf{f}^t_{ijk} \cdot \mathbf{g}^t_{ikh}, \quad\quad \mathcal{C} \in \mathbb{R}^{H\times W\times W}
\end{equation} 
Then, from this aggregated volume, we estimate a set of disparity maps from any given stereo pair by forwarding it to $\Theta$ together with reference features $\mathbf{f}^t$
and context features $\mathbf{c}^t$ from the single, reference image from pair $t$

\begin{equation}
    \mathbf{d}^t_i = \Theta(\mathbf{f}^t,\mathbf{c}^t,\mathbf{d}^t_{i-1},\mathbf{C}^*) 
\end{equation}
From the estimated disparity maps $\mathbf{d}^t$, we compute their average and obtain an initial ground-truth disparity map $\mathbf{d}^*$ together with an uncertainty guess $\mathbf{u}^*$ in the form of their variance.
This pipeline generates accurate ground-truth maps up to half the resolution of our textured images, \ie about 3 Mpx, since RAFT-Stereo cannot deal with larger resolutions and disparity ranges.

As a result, our deep space-time stereo pipeline produces a set of accurate disparity maps, one for each collected scene. However, these labels still require additional processing.

\begin{figure*}
    \centering
    \renewcommand{\tabcolsep}{3pt}
    \begin{tabular}{ccccc}
        \small \textit{RGB \& Mask} & \small \textit{Raft Passive} & \small \textit{Raft Spacetime} & \small \textit{SR \& Sharpening} & \small \textit{Manual Filtering}\\ 
        \includegraphics[width=0.18\textwidth]{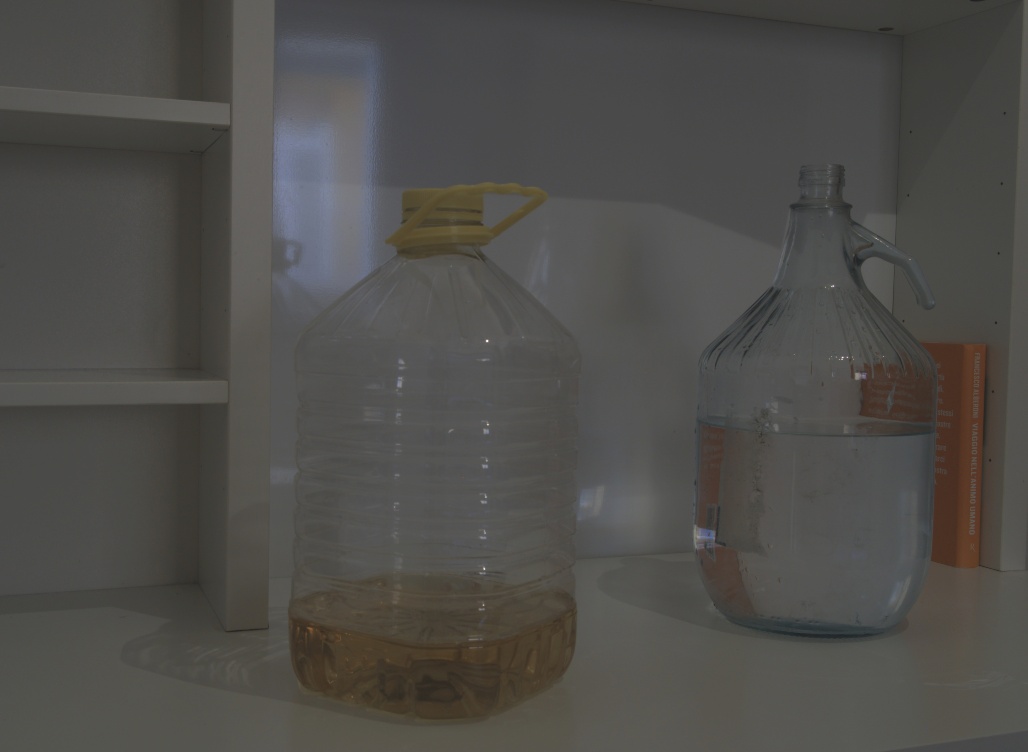} &
        \includegraphics[width=0.18\textwidth]{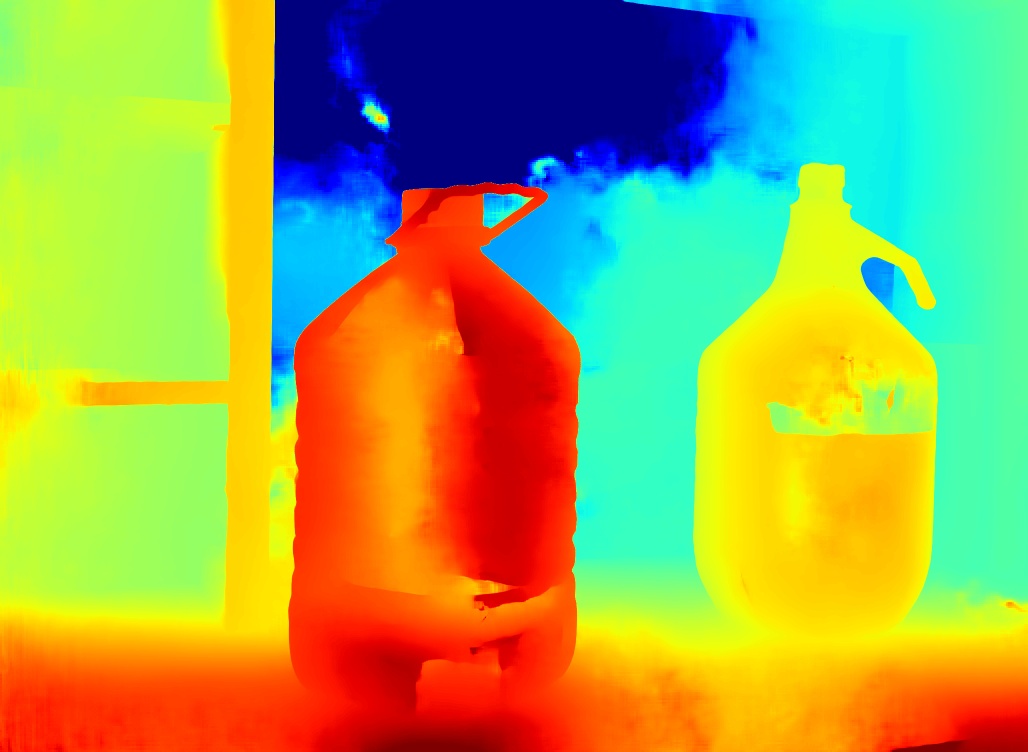} &
        \includegraphics[width=0.18\textwidth]{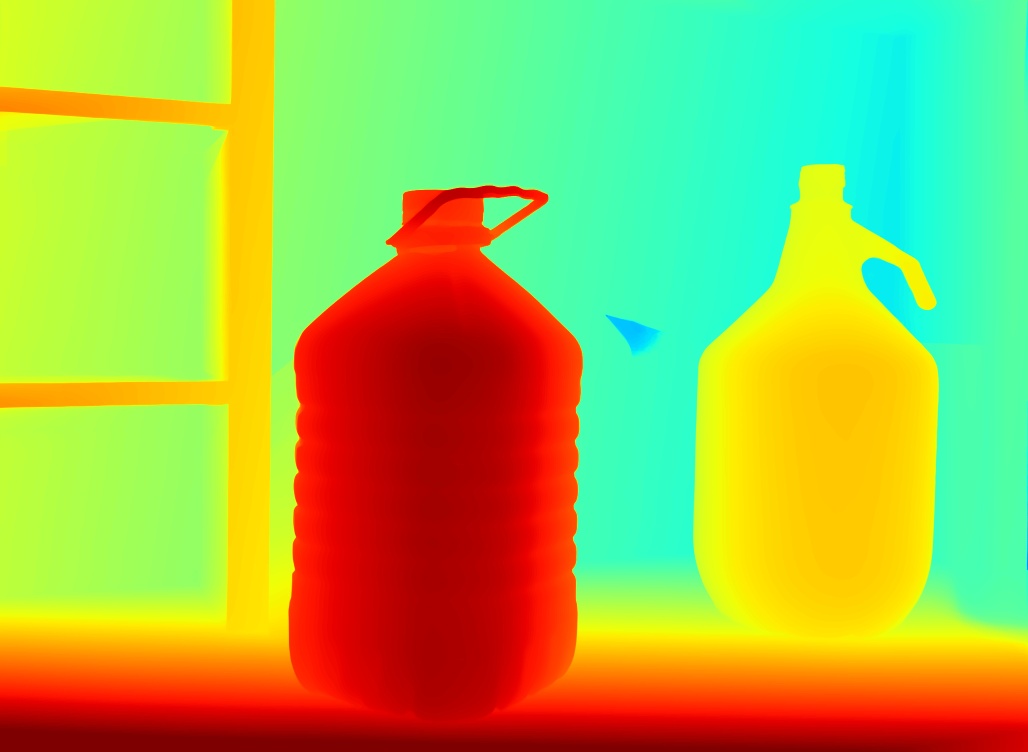} &
        \includegraphics[width=0.18\textwidth]{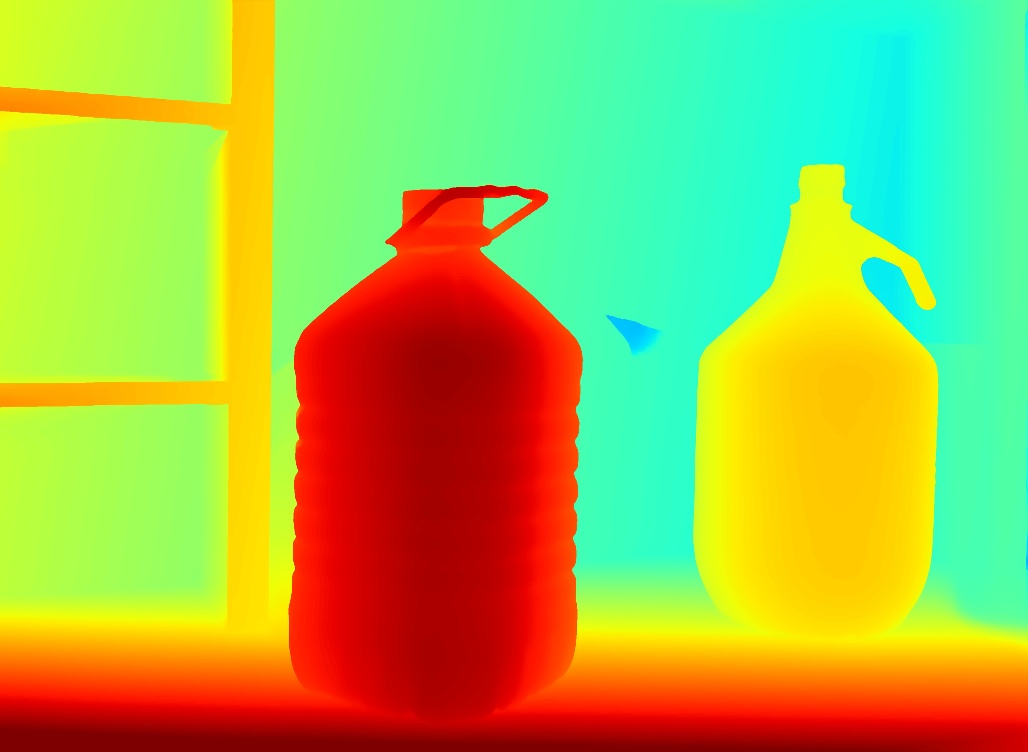} &
        \includegraphics[width=0.18\textwidth]{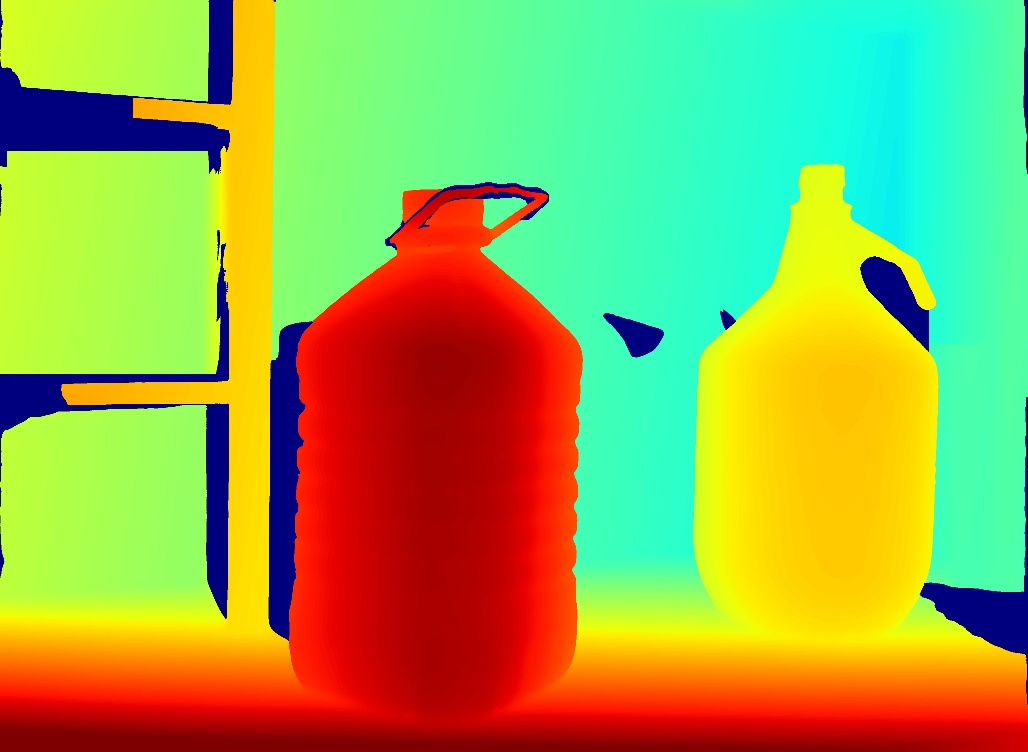} \\
        \includegraphics[width=0.18\textwidth]{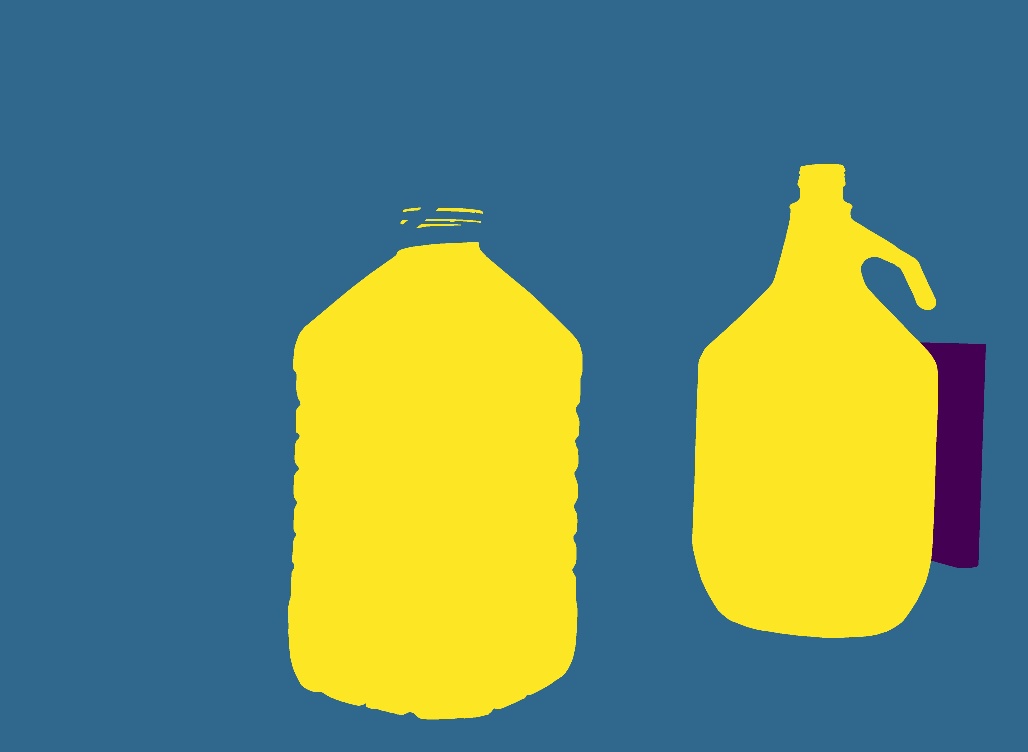} &
        \includegraphics[width=0.18\textwidth]{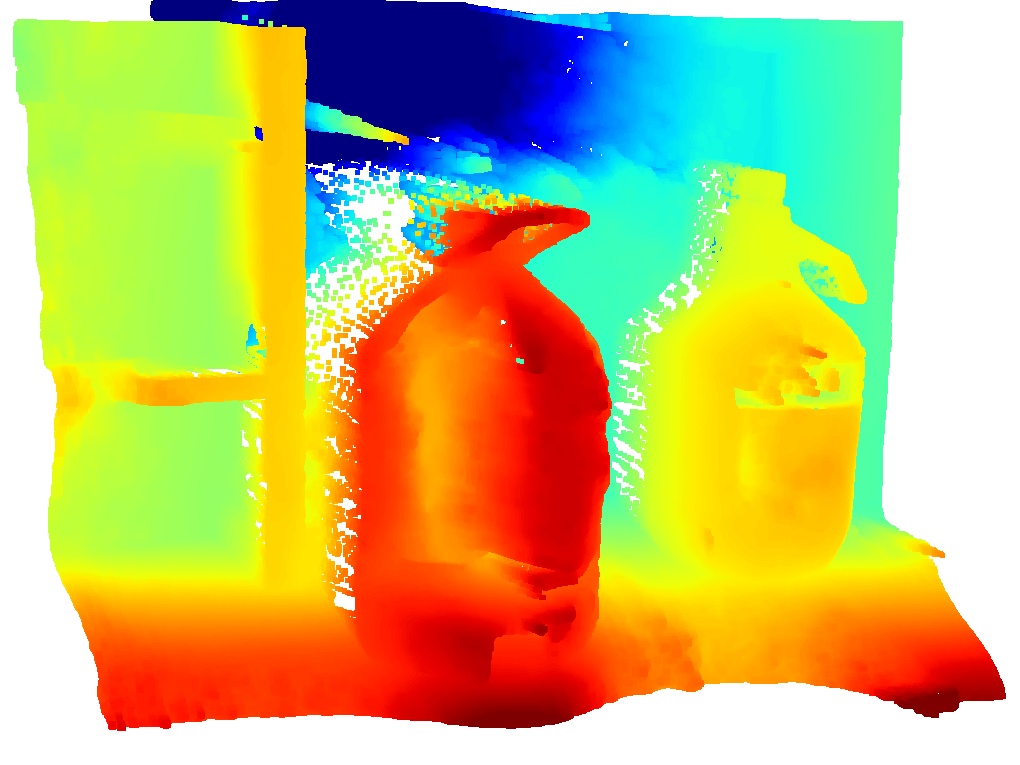} &
        \includegraphics[width=0.18\textwidth]{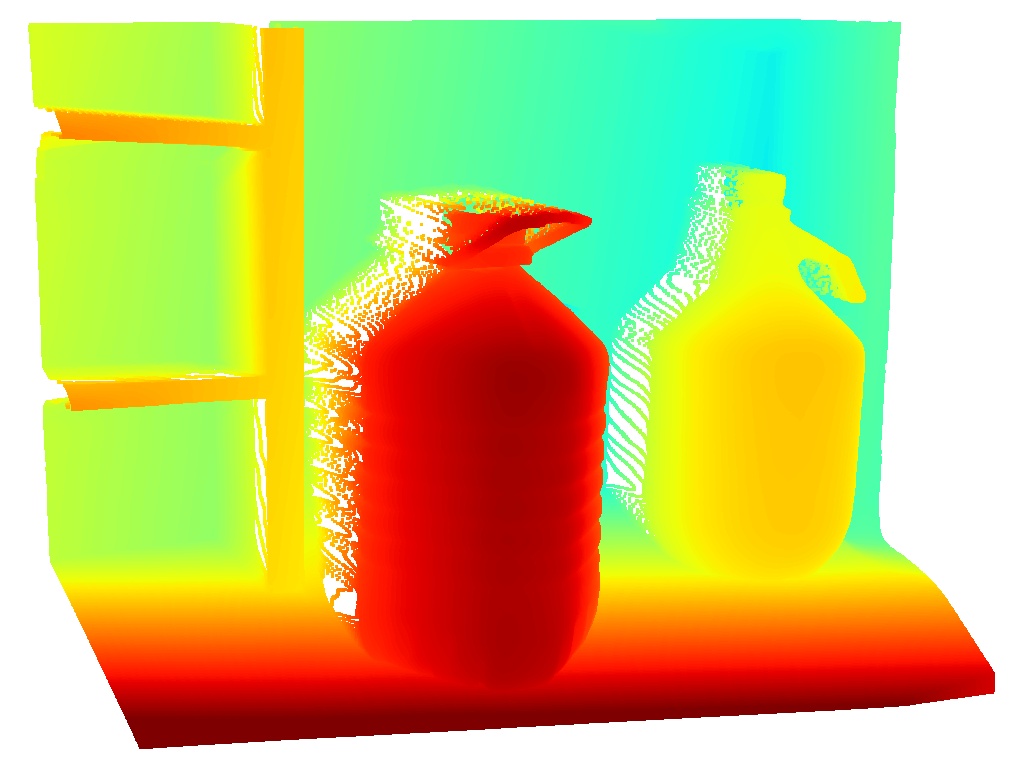} &
        \includegraphics[width=0.18\textwidth]{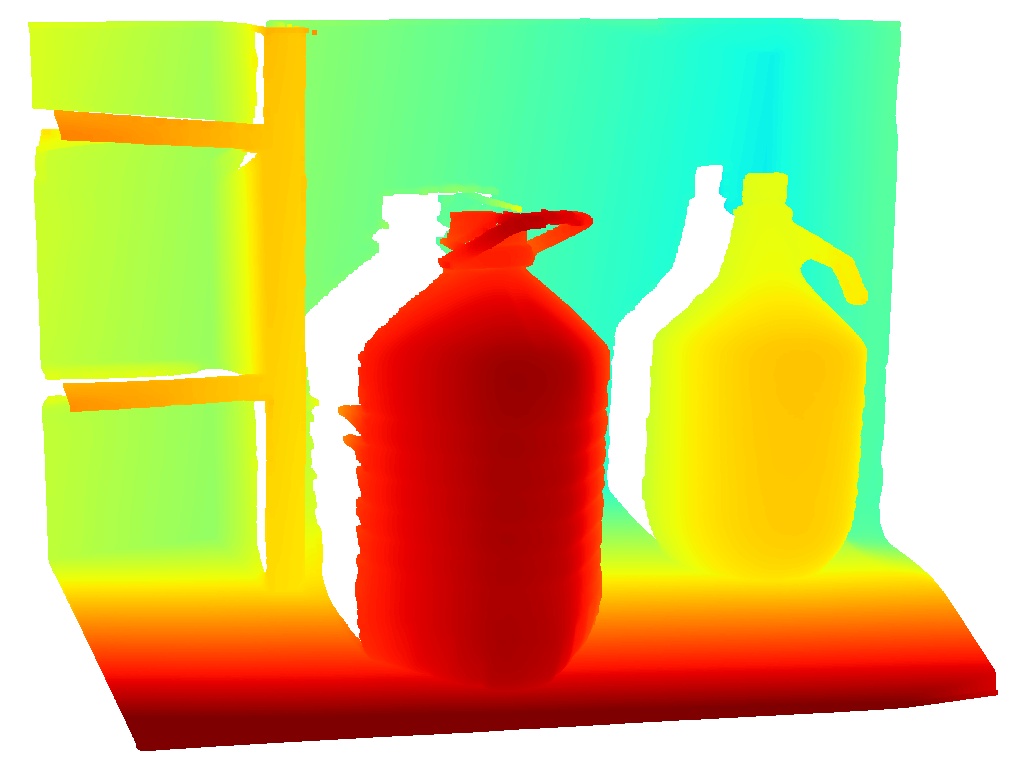} &
        \includegraphics[width=0.18\textwidth]{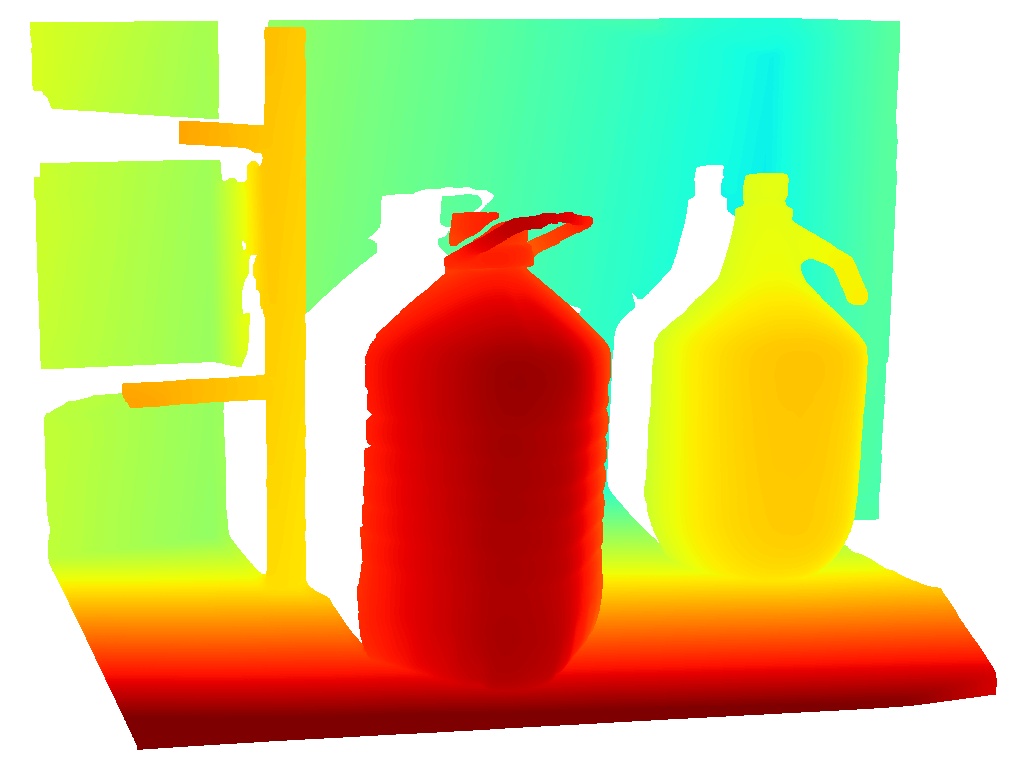} \\
    \end{tabular}
    \vspace{-0.25cm}\caption{\textbf{Data annotation pipeline.} From left to right: RGB reference image (top) and material segmentation (bottom), disparity maps (top) and point clouds (bottom) obtained by RAFT-Stereo on the passive pairs, followed by those produced by our deep space-time stereo framework, the super-resolution \& sharpening procedure, and the final, manual cleaning.}
    \label{fig:annotation_pipeline}
\end{figure*}
\begin{figure*}[t]
    \centering
    \includegraphics[width=0.75\textwidth]{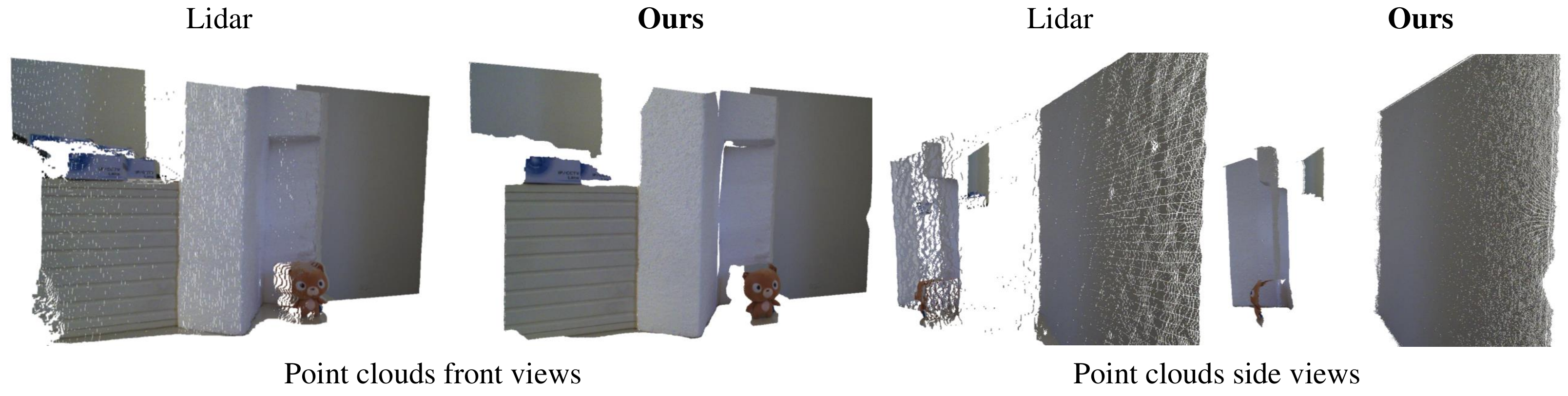}
    \caption{\textbf{Cross-verification with Intel L515 LiDAR.} Front (columns 1-2) and side (columns 3-4) 3D visualization of a sample scene. Columns 1,3: point cloud acquired by a LiDAR sensor. Columns 2,4: point cloud reconstructed by our framework.}
    \label{fig:lidar_vs_ours}
\end{figure*}

\subsection{Super-resolution and sharpening.} 
Despite being very accurate, disparity labels produced so far are yet imperfect under two main viewpoints, 1) the resolution, which is half of the real image resolution, and 2) the quality of depth discontinuities, which are a common concern in disparity maps predicted by deep networks because of over-smoothing \cite{chen2019over,Tosi2021CVPR}. To deal with both at once, we enhance the quality of our initial labels by using the neural disparity refinement methodology proposed in \cite{aleotti2021neural}.
Starting from pre-trained weights \cite{aleotti2021neural}, we overfit a single instance of the neural disparity refinement network on each scene for about 300 steps, assuming the disparity map itself as both input and ground-truth. This strategy allows for maintaining accurate disparity values at high resolution, as well as sharpening depth boundaries.

To improve sub-pixel precision, we replaced the original prediction mechanism \cite{aleotti2021neural} with SMD output head Tosi \etal \cite{Tosi2021CVPR}. Indeed, the former introduced undesired artifacts in our early experiments, while the latter better preserved the correct disparity values. Accordingly, each instance of the neural disparity refinement network is optimized to infer a bimodal Laplacian distribution with weight $\pi$ and modes $(\mu_1, b_1)$ and $(\mu_2, b_2)$.

\begin{equation}
    \mathbf{p}(d) = \frac{\pi}{2\mathbf{b}_1}e^{-\frac{d-\mathbf{\mu}_1}{\mathbf{b}_1}} + \frac{1-\pi}{2\mathbf{b}_2}e^{-\frac{d-\mathbf{\mu}_2}{\mathbf{b}_2}}
\end{equation}
The trained network is then used to produce a sharpened disparity map $\mathbf{d}^*$ at full resolution, by exploiting the continuous representation of the network itself, by selecting the mode with the highest density value.

\subsection{Manual cleaning and filtering.} 
Finally, we manually clean the full-resolution disparity map obtained so far to remove any remaining artifacts. This is carried out by projecting the disparity map into a 3D point cloud to visualize structural errors in the geometry of the scene. Then, we use the variance map $\mathbf{u}^*$ as guidance, easing the detection of most of the artifacts. When removing points from the point cloud, the corresponding pixels are then filtered out from the disparity map. 
After manual cleaning, we apply a $35\times35$ bilateral filter -- with $\sigma_\text{color}=5$ and $\sigma_\text{dist}=50$ -- to smooth objects surfaces, leading to the final disparity map $\mathbf{d}^*$. Then, we can obtain depth by triangulation, knowing the focal length of the reference camera $L$ and the baseline between $L$ and $R$.

Fig. \ref{fig:annotation_pipeline} summarizes the steps of the pipeline described so far, highlighting how the quality of 3D reconstructions yielded by our annotations improves after each step.

\subsection{Accuracy assessment.} 

In this subsection, we inquire about the quality of our disparity labels, according to two main criteria.

\textbf{Planar regions check.}
Following Scharstein \etal{} \cite{scharstein2014high}, we measure the accuracy of our ground-truth annotations on planar regions. Accordingly, we manually select portions in the images containing planar surfaces and fit a plane to the disparities over each of them. Then, we measure the residuals between the fitted plane equation and the ground-truth disparities. The lower the error, the more accurate the disparities.
We perform this test over 153 planar regions in our dataset, achieving an average residual error of 0.053 pixels, which turns out comparable to the error reported for the Middlebury 2014 dataset (0.032) -- the latter obtained by applying an explicit sub-pixel refinement based on plane fitting, not performed in our case.

\textbf{Cross-verification with a LiDAR sensor.}
We further investigate the quality of our ground-truth labels by acquiring some sample scenes, both with our custom rig and an Intel Realsense L515 LiDAR. Before collection, we calibrate the two to obtain the relative pose between the cameras and LiDAR.
Then, we get disparity from our deep space-time pipeline, we project it over the Realsense depth maps according to the estimated pose and compute their pixel-wise difference.
The two are in close agreement, as about 82 \% of the measurements differ by less than 1 cm and the RMSE between these inliers is about 3.3 mm, suggesting that our labels are metrically consistent -- \ie no scale or shift is introduced by the deep network used in our pipeline.

We dig further and compute residuals on fitted planes for both: we obtain 0.12 and 0.05 for the LiDAR and our method, respectively, thus showing that our method produces disparity maps that are less noisy than depth measurements by the Realsense on planar surfaces. In Fig. \ref{fig:lidar_vs_ours}, we can appreciate the quality of the 3D reconstruction of a sample scene enabled by our technique and the LiDAR.



\subsection{Further processing}

On the labels obtained so far -- \ie disparities between the high-resolution pairs -- we can further elaborate additional annotations, for the balanced and unbalanced setup. In the former case, we can extract occlusion masks, in the latter we can obtain the corresponding disparity labels referred to the $L-C$ pairs by exploiting calibration.

\textbf{Left-right consistency (balanced setup).} 
On top of our ground-truth labels, we can identify pixels that are occluded in the reference image by means of a left-right consistency check. To this aim, we run twice the processing pipeline described so far, including the manual cleaning, producing two disparity maps for each scene, $\mathbf{d}^*_L$ and $\mathbf{d}^*_R$, respectively, for the left and right images. Then, any pixel at coordinates $(x,y)$ in $\mathbf{d}^*_L$ is filtered out if the absolute difference with its match $x-\mathbf{d}_L(x,y),y$ in $\mathbf{d}^*_R$ is larger than a threshold, set to 2 pixels:

\begin{equation}
    |\mathbf{d}_L(x,y)-\mathbf{d}_R(x-\mathbf{d}_L(x,y),y)| > 2
\end{equation}
The same procedure is performed in the opposite direction, on top of $\mathbf{d}^*_R$, by removing any pixel at coordinate $(x,y)$ after comparison with pixel $(x+\mathbf{d}^*_R(x,y),y)$ on the left disparity map. 

So far, the output of our annotation pipeline consists of two high-resolution ground-truth disparity maps, respectively, for the left and right images of the balanced setup.

\begin{figure}
    \centering
    \includegraphics[width=0.5\linewidth]{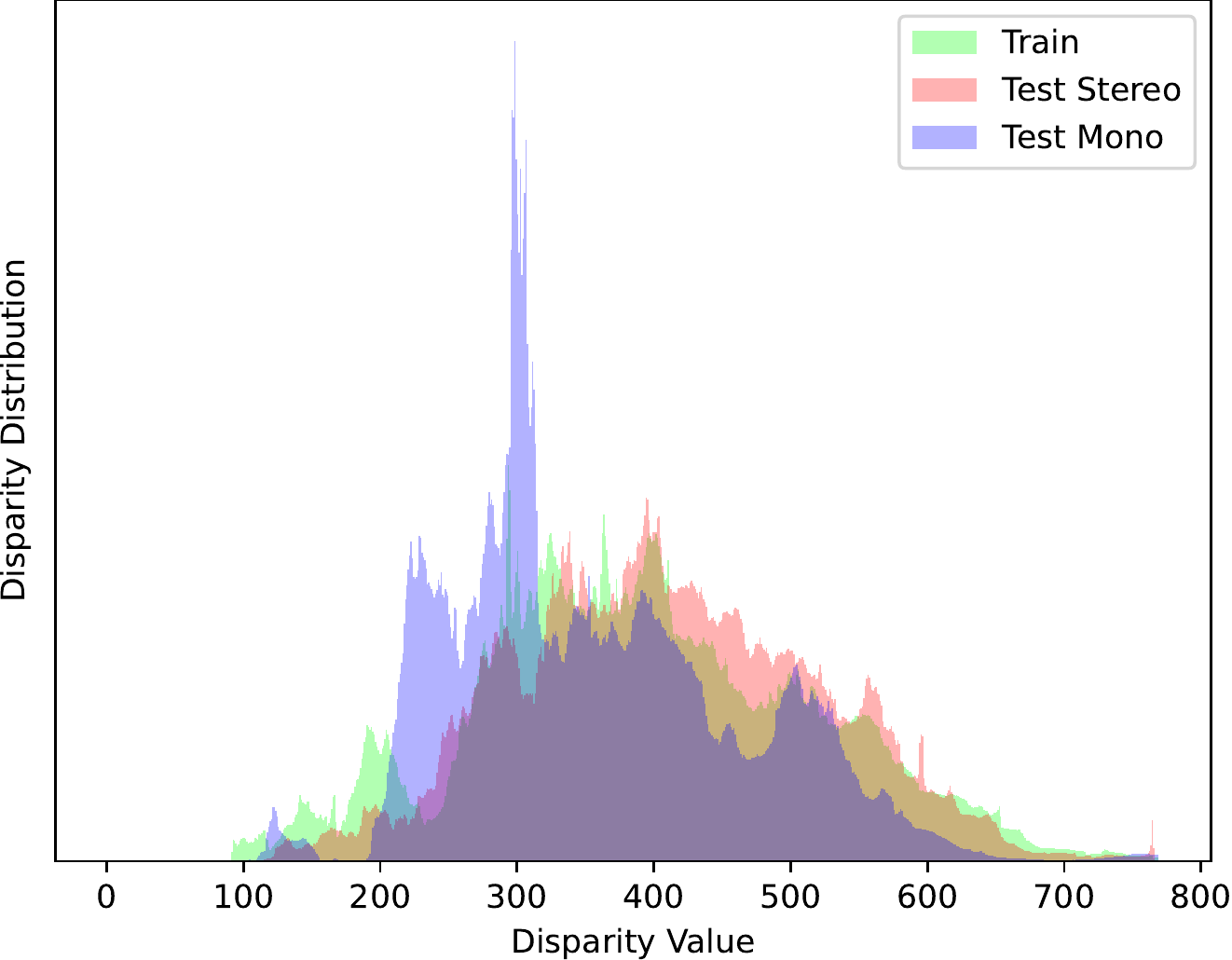}
    \caption{\textbf{Disparity Distribution.} Train, Test Mono, and Test Stereo disparity value distributions. Values extracted from the high-resolution ground-truth maps before warping.}
    \label{fig:disp_hist}
\end{figure}

\textbf{Warping (unbalanced setup).}
Ground-truth disparity maps obtained so far are aligned with balanced stereo pairs for cameras $L-R$. 
We exploit the camera calibration to obtain annotation for the unbalanced pairs for the $L-C$ stereo system. 
Since the rectification transformation applied to the raw images to obtain rectified pairs is a homography (\ie it consists of modifying the intrinsic parameters and applying a rotation), we can easily perform a backward warping of the ground-truth labels
from the left images of the $L-R$ rectified pairs to align them to the left images of $L-C$ ones. Specifically, given pixels coordinates $(u,v)$ in the $L_{LR}$ image, we compute coordinates $(u',v')$ in $L_{LC}$ as:


\begin{equation}\label{eq:mapping}
\begin{pmatrix}
ku' \\
kv' \\
k 
\end{pmatrix} = A^{LC}_L R^{LC}_L {R^{LR}_L}^{-1} {A^{LR}_L}^{-1}
\begin{pmatrix}
u \\ 
v \\ 
1
\end{pmatrix}
\end{equation}

By knowing this mapping, we can perform a backward warping to obtain $Disp_L^{LC}$ from $Disp_L^{LR}$. 
However, we also need to change the disparity values according to the 3D rotation between the two cameras and the different baseline characterizing the unbalanced setup before warping. 
Thus, given the disparity map $Disp_L^{LR}$, we first transform it to the corresponding depth map $D_L^{LR}$ as follows:

\begin{equation}
    D_L^{LR}= \frac{f_{LR}b_{LR}}{Disp_L^{LR}}
\end{equation}

with $f_{LR}$ being the focal length of $L_{LR}$ and $b_{LR}$ the baseline of the stereo system $L-R$.
Then, we back-project each pixel of $L_{LR}$ to 3D using $D_L^{LR}$ and we rotate it accordingly to relative rotation between $L-R$ and $L-C$, obtaining the pixel in the $L_{LC}$ reference frame:

\begin{equation}
\begin{pmatrix}
x' \\
y' \\
z'
\end{pmatrix}
= R^{LC}_L {R^{LR}_L}^{-1} D_L^{LR} {A^{LR}_L}^{-1}
\begin{pmatrix}
u \\ 
v \\ 
1
\end{pmatrix}
\end{equation}

Accordingly, we can create a depth map $D_L^{LR \rightarrow LC}$ for which any pixel $(u,v)$ contains the depth value of the corresponding pixel aligned in the $L_{LC}$ reference frame, $z'$.
Then, we backward warp the depth values:

\begin{equation}
    D_L^{LC} = \phi(D_L^{LR \rightarrow LC})
\end{equation}

with $\phi$ being the backward warping operation applying the mapping function defined at Eq. \ref{eq:mapping} and $D_L^{LC}$ the depth map aligned with $L_{LC}$.
Finally, we convert depth back into disparity for the $L_{LC}$ image as:

\begin{equation}
    Disp_L^{LC}= \frac{f_{LC}b_{LC}}{D_L^{LC}}
\end{equation}

with $f_{LC}$ and $b_{LC}$ being the focal length of $L_{LC}$ and the baseline of the $L-C$ stereo system.

\subsection{Segmentation masks.} 
Finally, we manually annotate images to identify challenging surfaces, \ie transparent or specular, by producing segmentation masks. 
We categorize materials considering their visual appearance in relation to their interaction with light. In particular, when creating the classes, we focused on the visual aspect that an object would have when reflected or transmitted by a surface. We emphasize this because we argue that one of the main problems that stereo and monocular algorithms face when dealing with these materials is that they are deceived by reflected or transmitted objects that resemble real ones, leading to incorrect correspondences in the case of stereo, and a misinterpretation of the scene in general. Based on the above considerations, we defined four classes of materials, where higher classes correspond to higher levels of difficulty for depth algorithms, as follows:

\textit{Class 3:} Surfaces almost allowing regular reflection and/or transmission, as seen with mirrors and transparent glass. Reflected and transmitted objects on these surfaces closely resemble real ones, making depth estimation challenging for both monocular and stereo algorithms.

\textit{Class 2:} surfaces where light interacts with the material, causing partial distortion of transmitted and reflected objects in shape and color. However, the reflected or transmitted image still retains a resemblance to the real object. Examples of such surfaces include colored/textured glass or irregular, rough specular surfaces.

\textit{Class 1:} Surfaces causing significant distortions when light interacts with them. The reflected/transmitted light still depends on the camera viewpoint, but transmitted and reflected objects cannot be distinguished. Examples include rough metals, glossy ceramic tiles, and frosted glass.

\textit{Class 0:} Completely opaque or diffuse surfaces, fully absorbing or diffusely reflecting light, with no transmission through the material.

We annotate pixels in the reference image of the $L-R$ pair, and warp it on the corresponding reference image of the unbalanced pair $L-C$, as described in the previous subsection. 
An example of segmentation masks annotating the images in our dataset is reported in Fig. \ref{fig:annotation_pipeline}.

\begin{figure}
    \centering
    \includegraphics[width=0.5\linewidth]{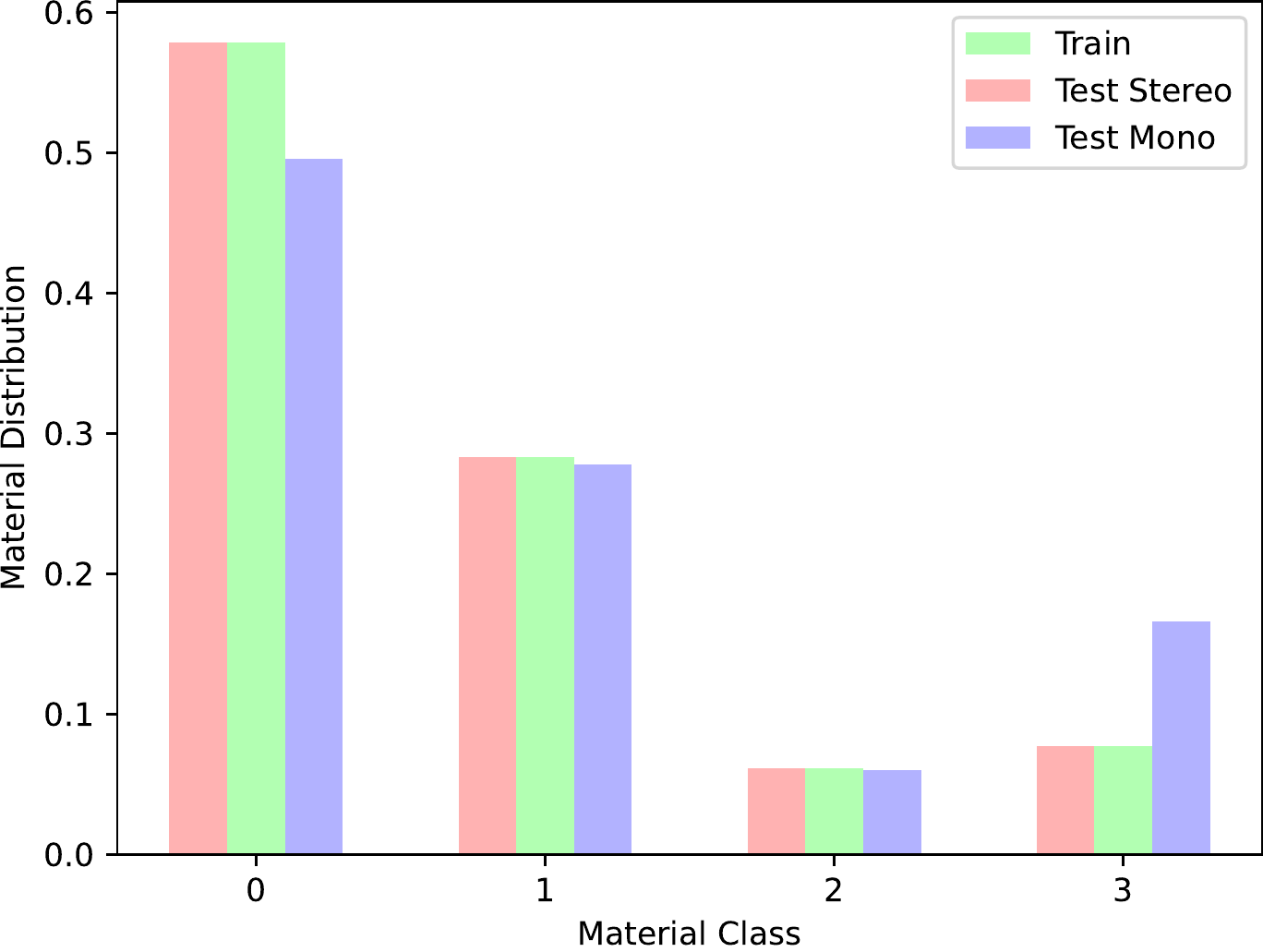}
    \caption{\textbf{Material Distribution.} Train, Test Mono, and Test Stereo material class distributions.}
    \label{fig:mat_hist}
\end{figure}
\begin{figure}[t]
    \centering
    \renewcommand{\tabcolsep}{1pt}
    \begin{tabular}{cccccc}
    \multicolumn{2}{c}{\scriptsize  \textit{Balanced Setup}} & & \scriptsize  \textit{Unbalanced Setup} & & \scriptsize  \textit{Illuminations} \\
    \includegraphics[width=0.10\textwidth]{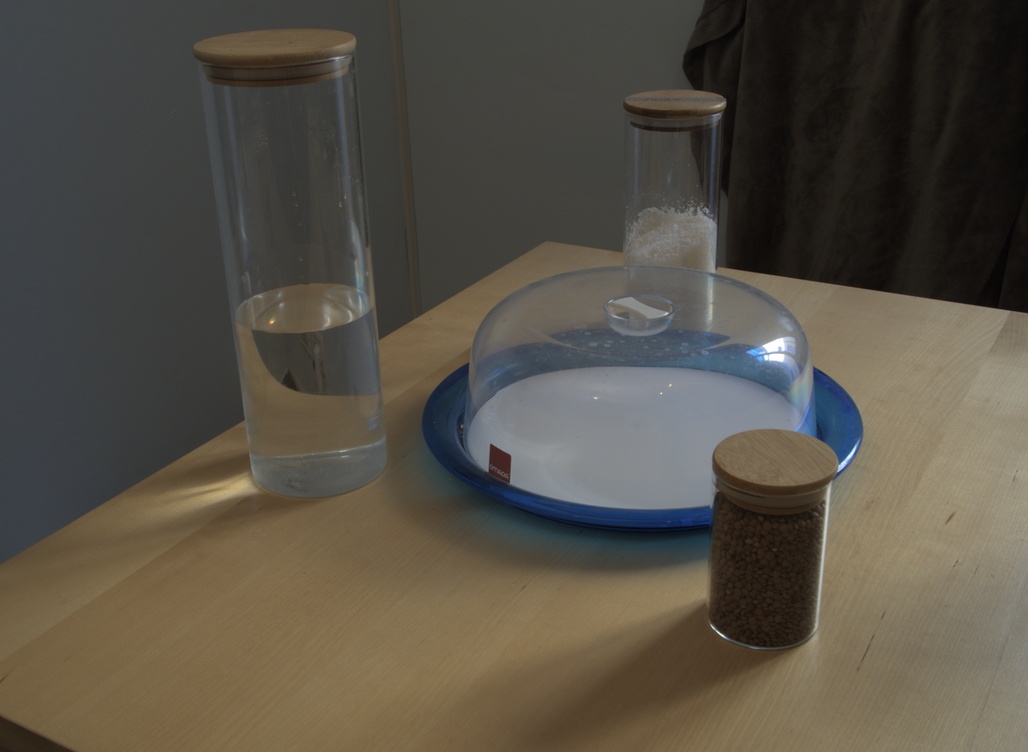} &
    \includegraphics[width=0.10\textwidth]{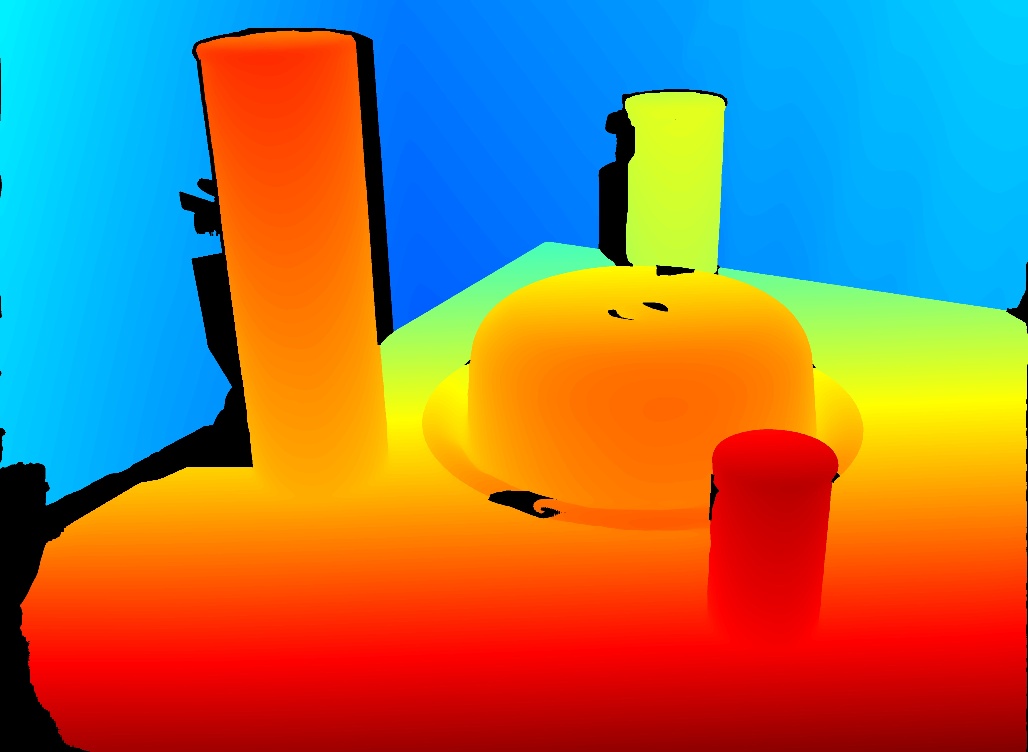} & &
    \includegraphics[width=0.10\textwidth]{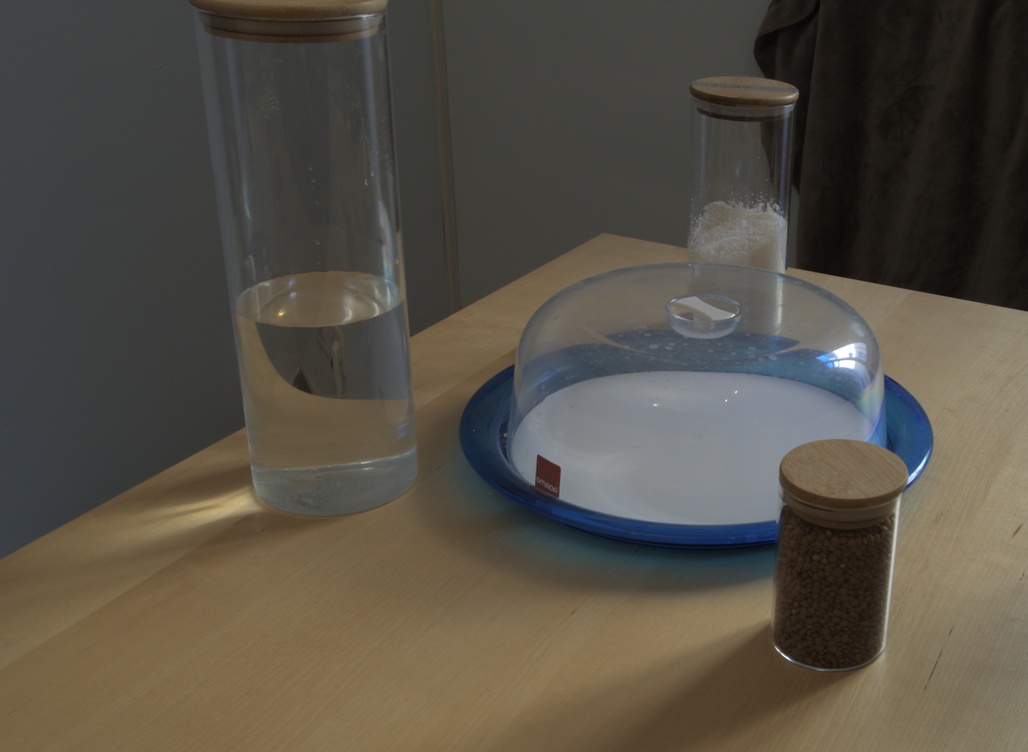} & &
    \includegraphics[width=0.10\textwidth]{images/dataset_sample/camera_00_02/im0.jpg} \\
    
    \includegraphics[width=0.10\textwidth]{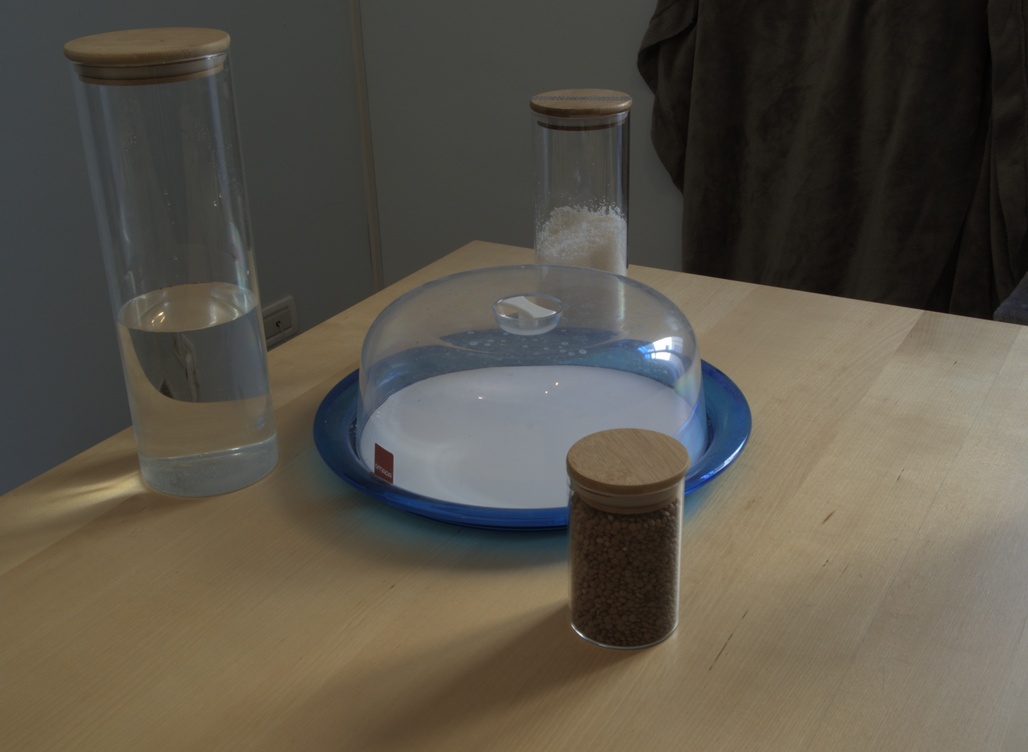} & 
    \includegraphics[width=0.10\textwidth]{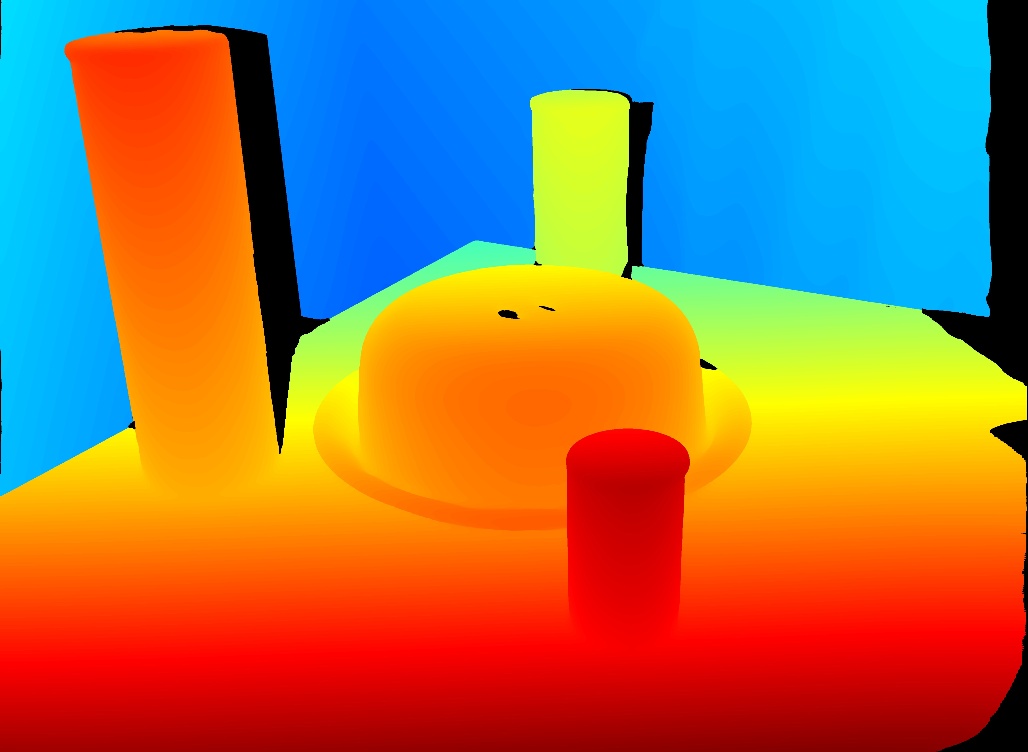} & &
    \includegraphics[width=0.04\textwidth]{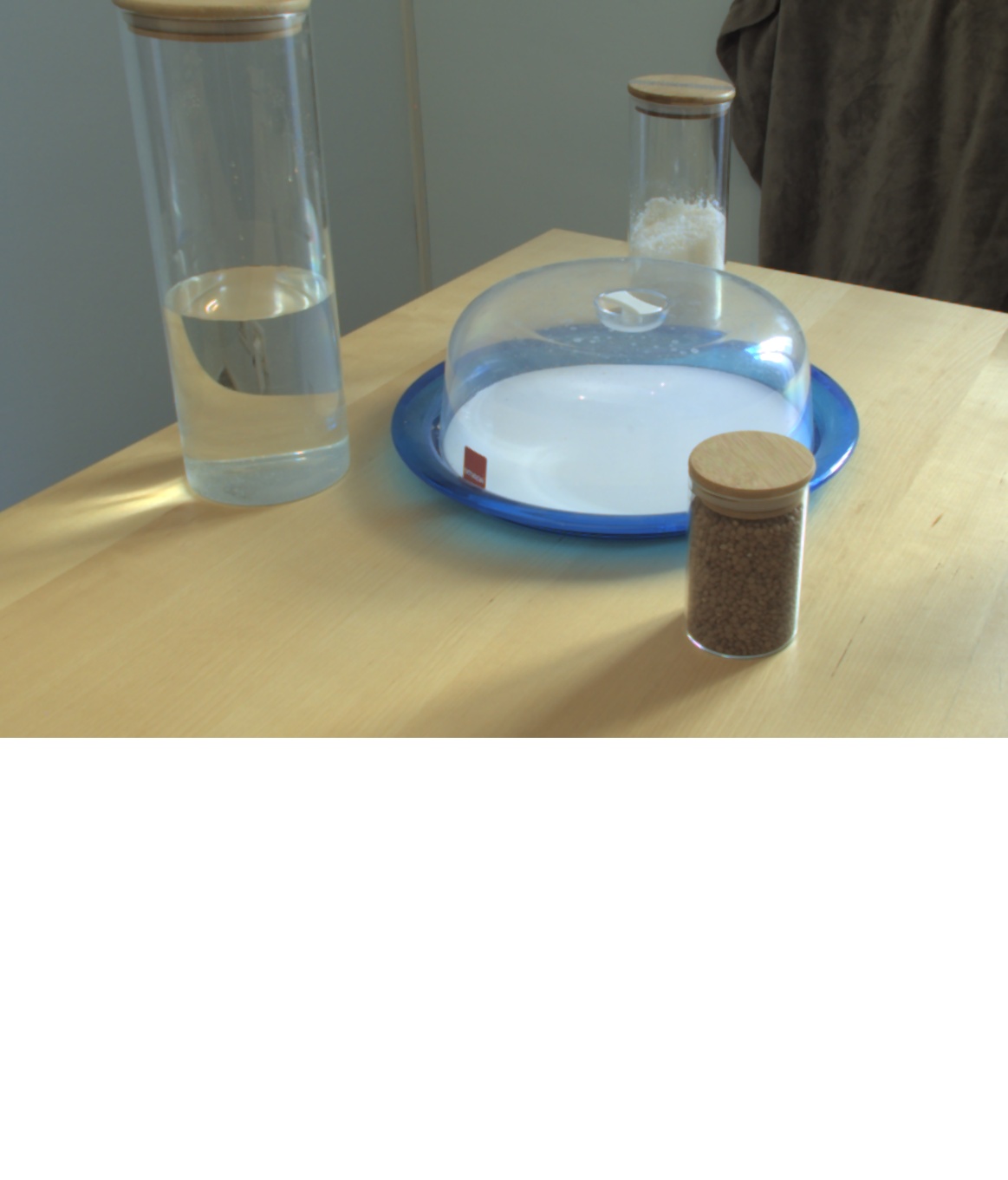} & &
    \includegraphics[width=0.10\textwidth]{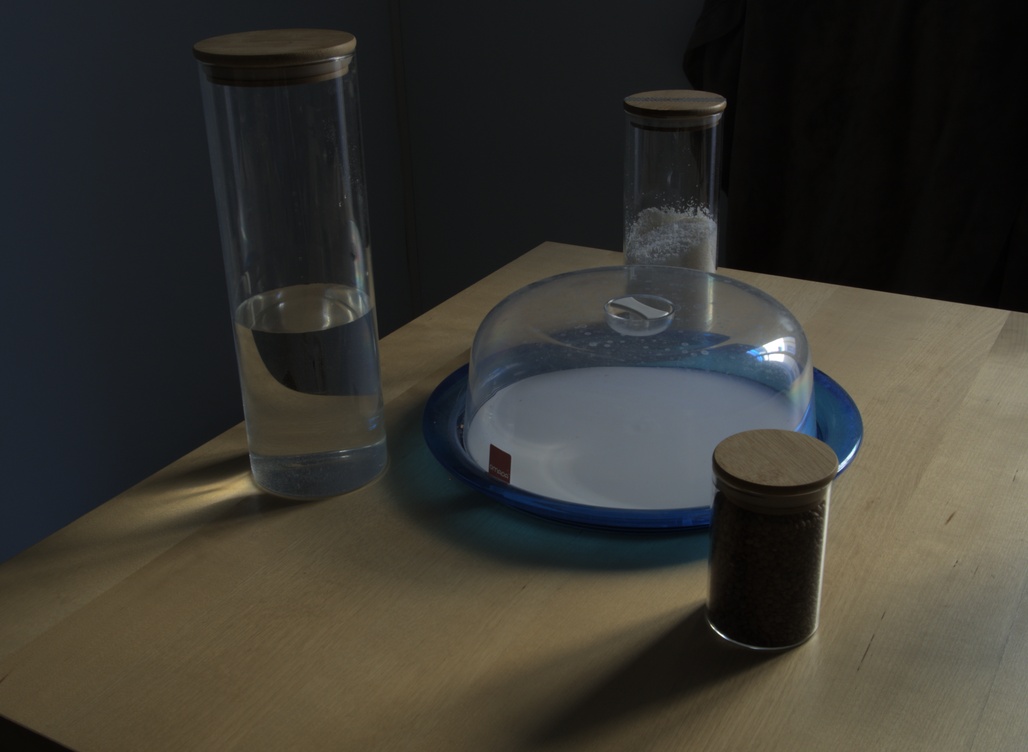} \\
    
    \includegraphics[width=0.10\textwidth]{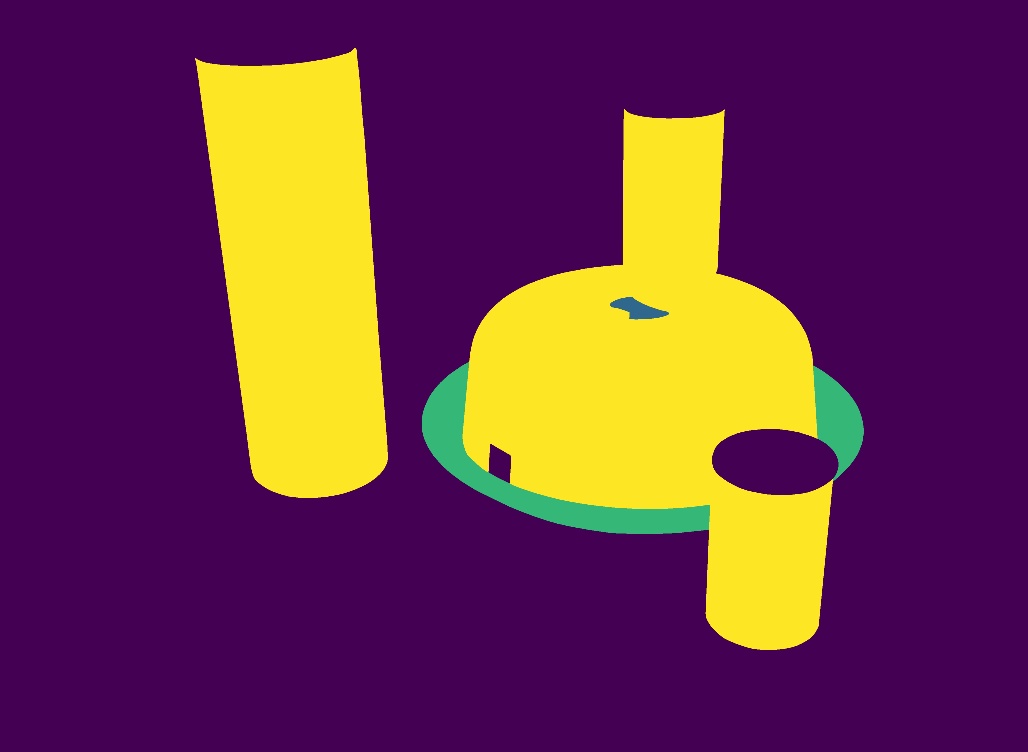} &
    \includegraphics[width=0.10\textwidth]{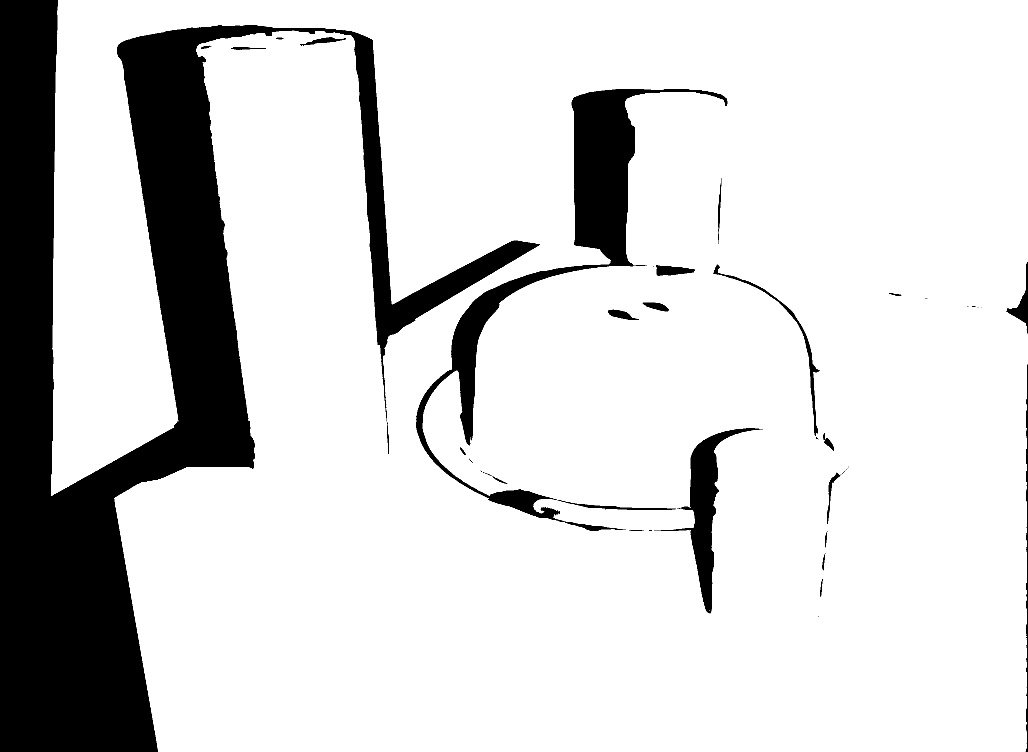} & &
    \includegraphics[width=0.10\textwidth]{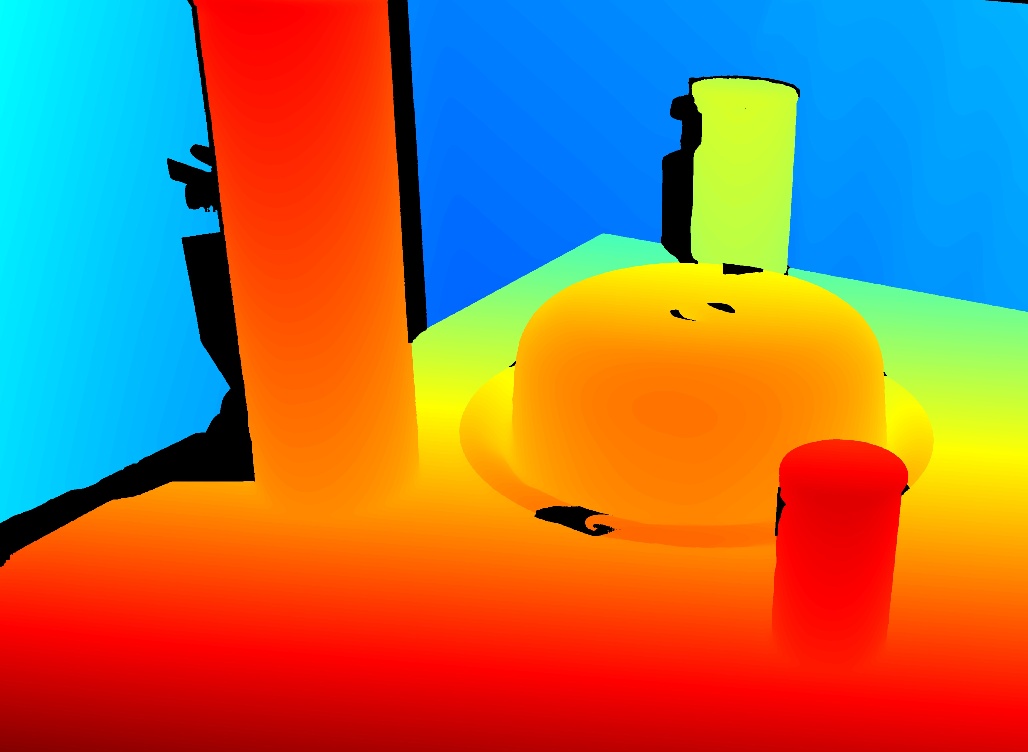} & &
    \includegraphics[width=0.10\textwidth]{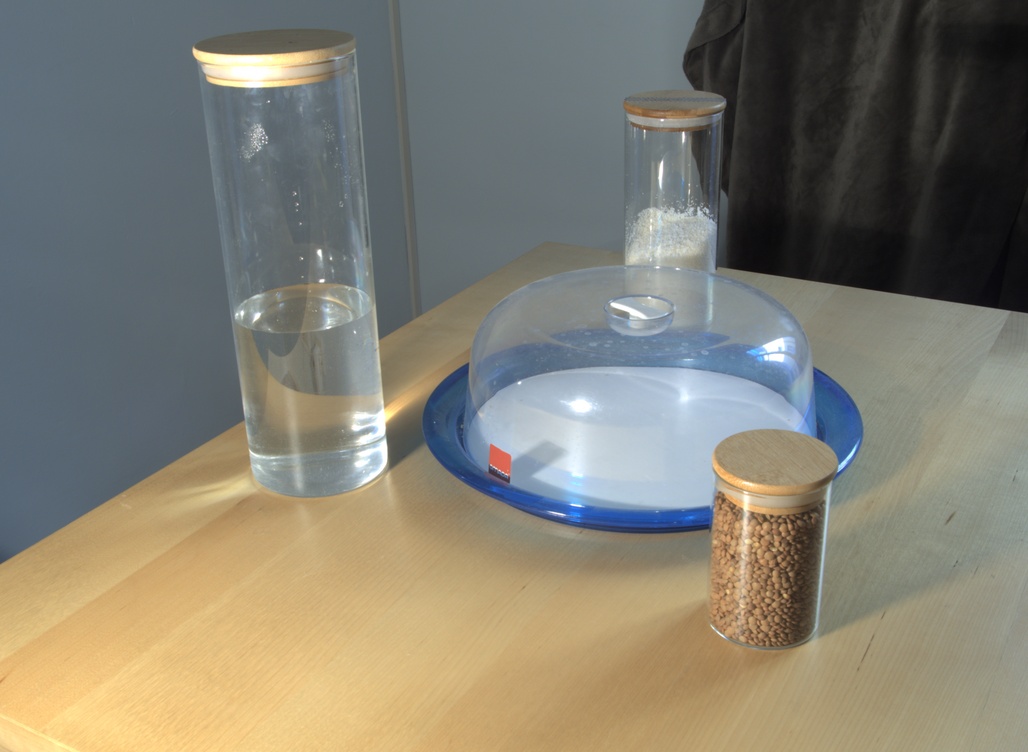} \\\\

    \includegraphics[width=0.10\textwidth]{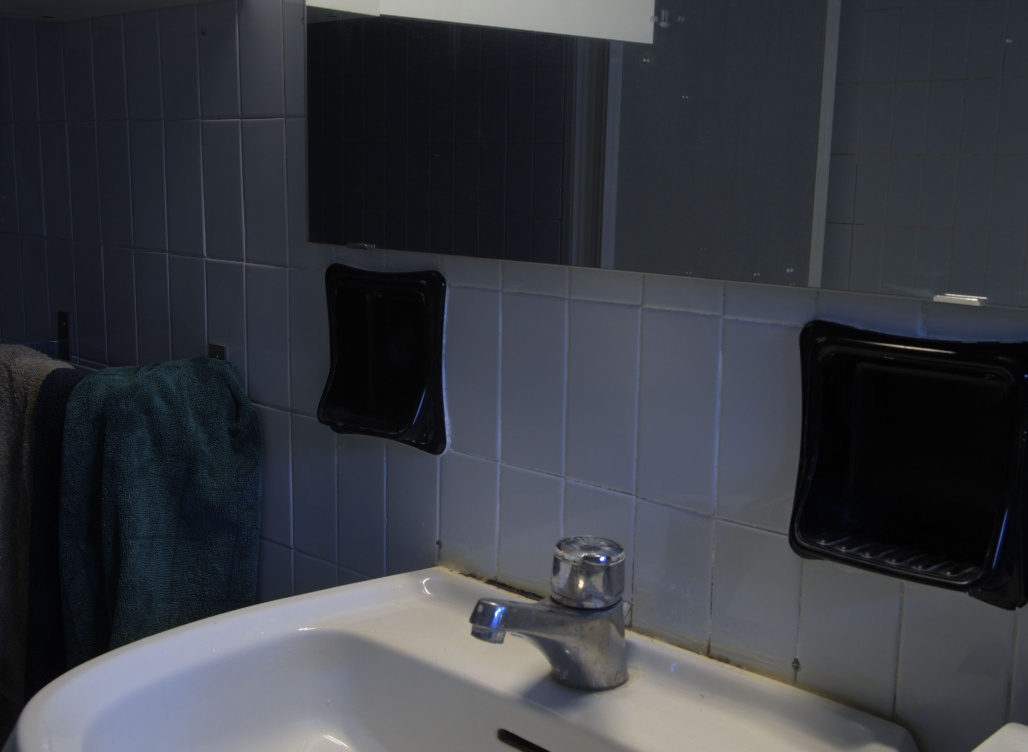} &
    \includegraphics[width=0.10\textwidth]{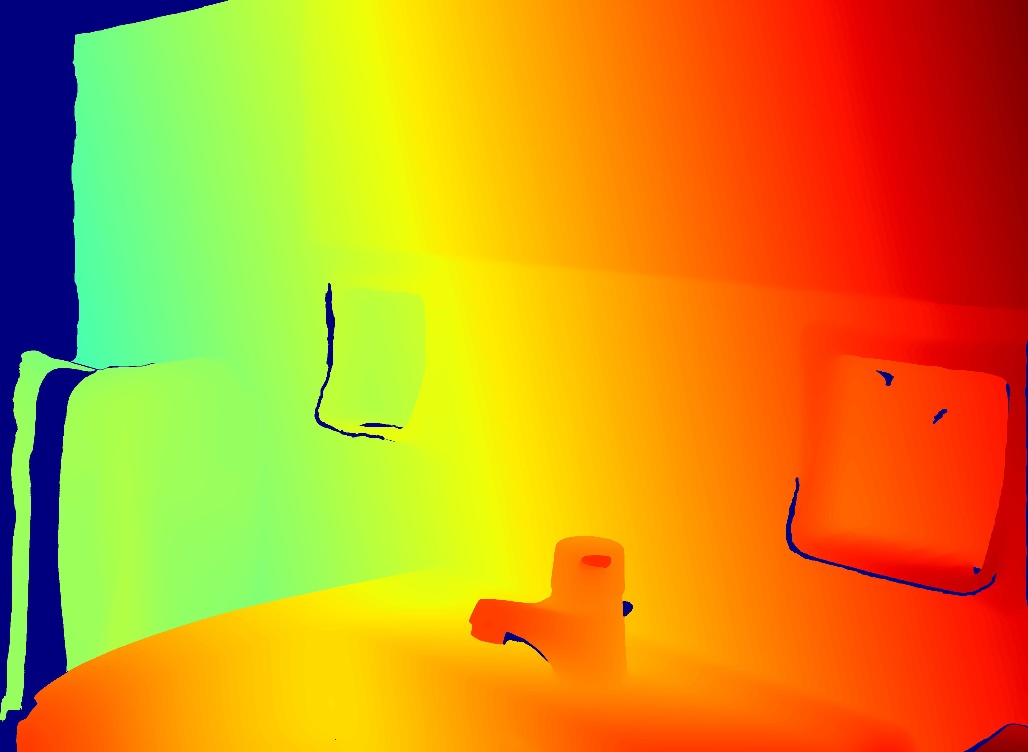} & &
    \includegraphics[width=0.10\textwidth]{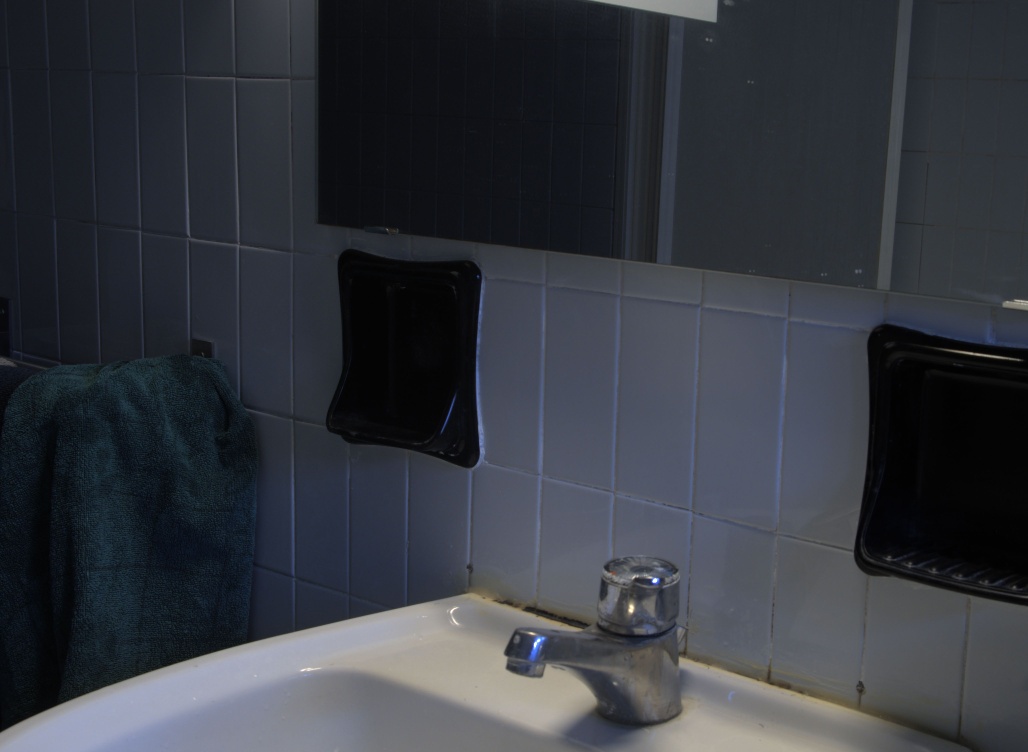} & &
    \includegraphics[width=0.10\textwidth]{images_supp/samples/balanced_Mirror1_camera_00_im0.jpg} \\
    \includegraphics[width=0.10\textwidth]{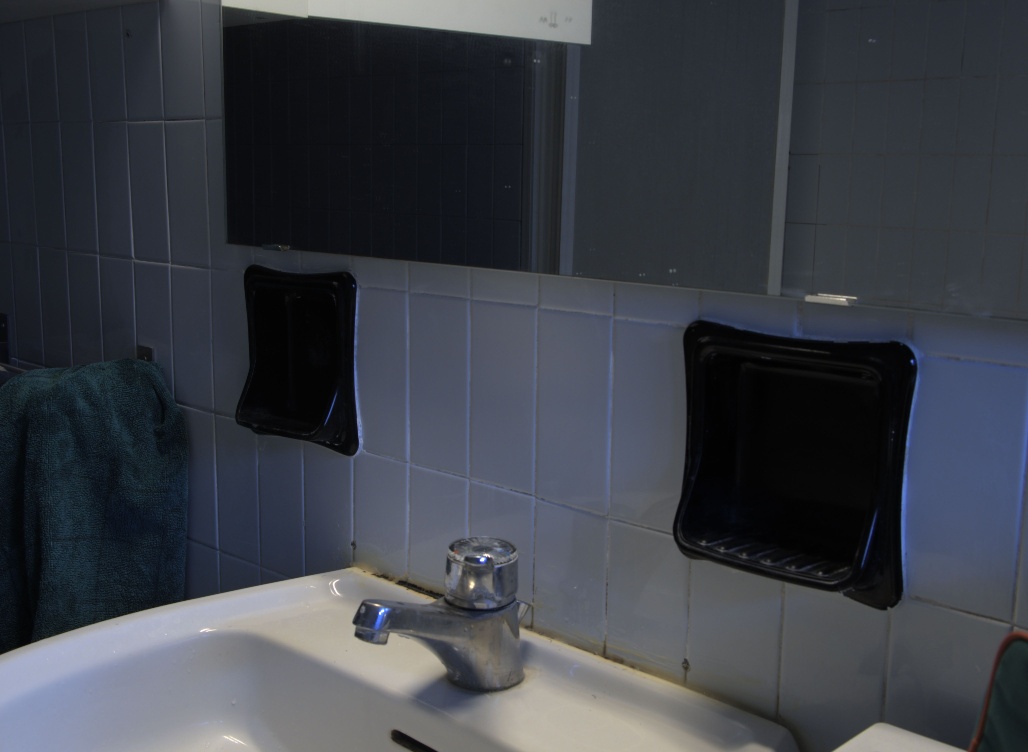} &
    \includegraphics[width=0.10\textwidth]{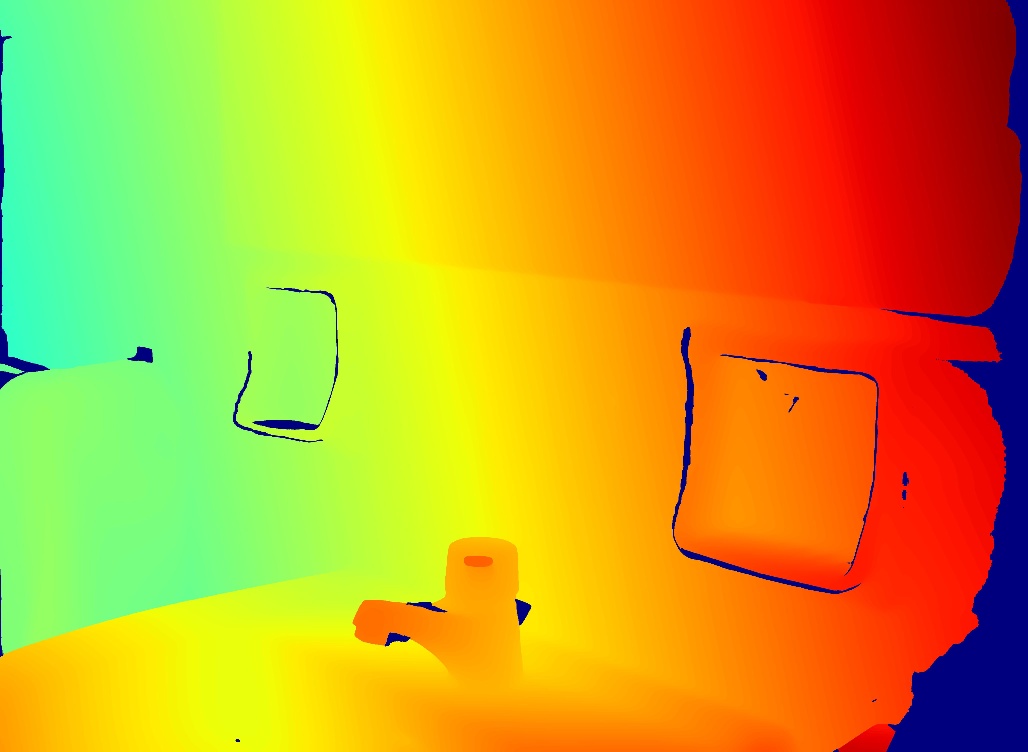} & &
    \includegraphics[width=0.10\textwidth]{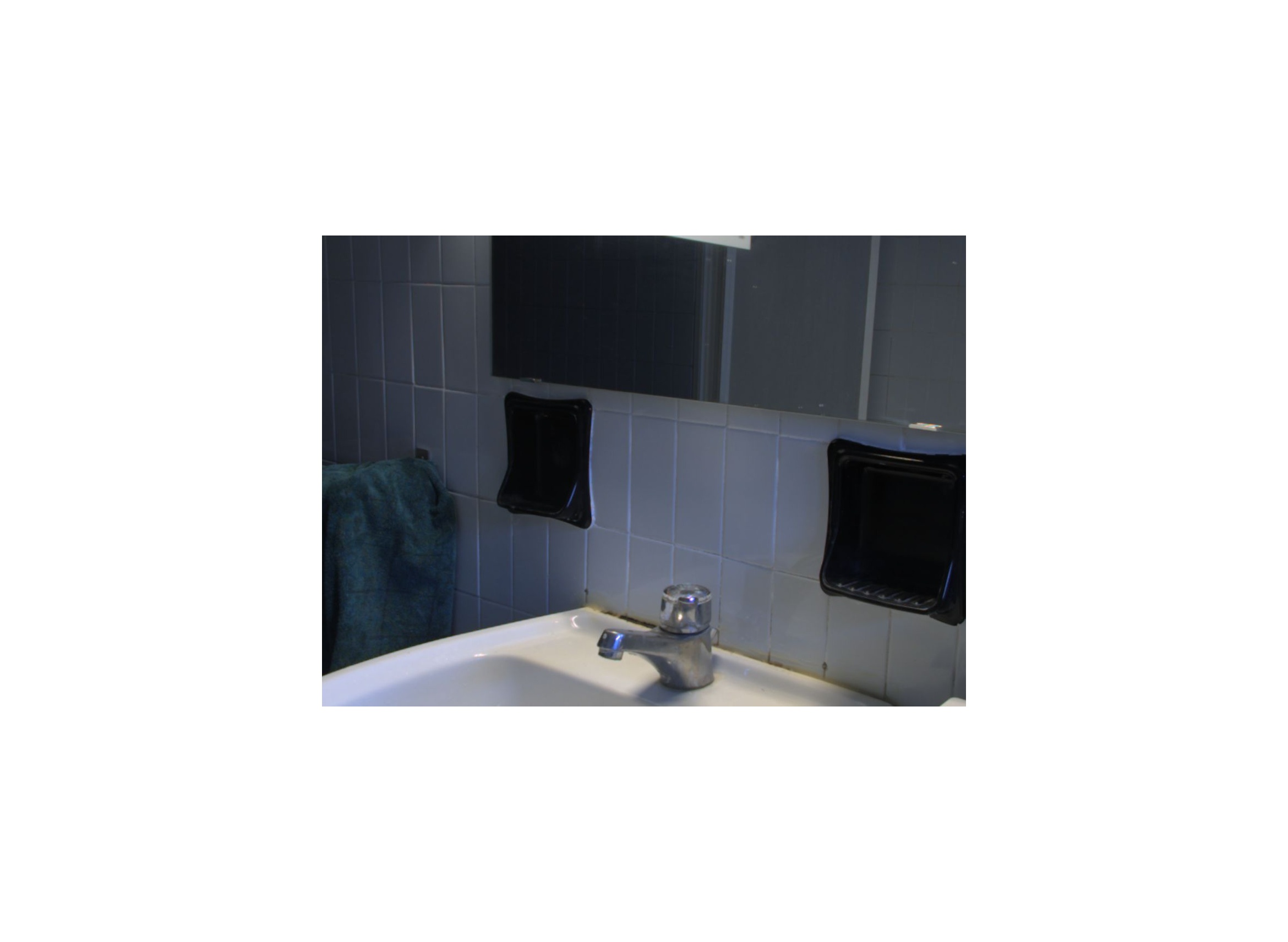} & &
    \includegraphics[width=0.10\textwidth]{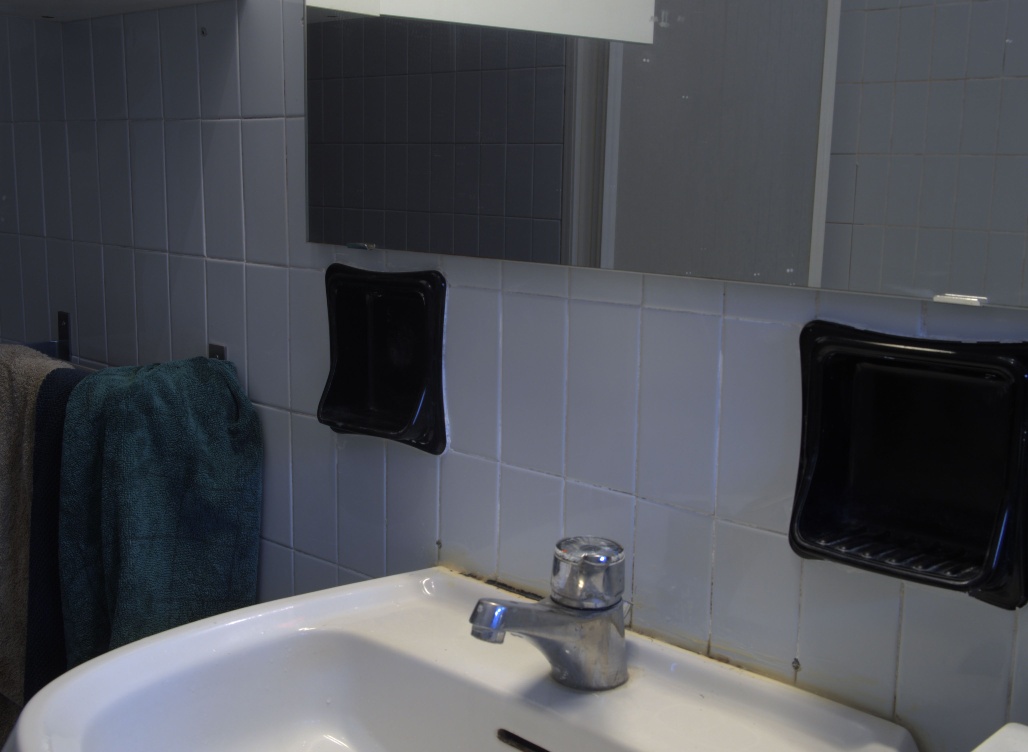} \\
    \includegraphics[width=0.10\textwidth]{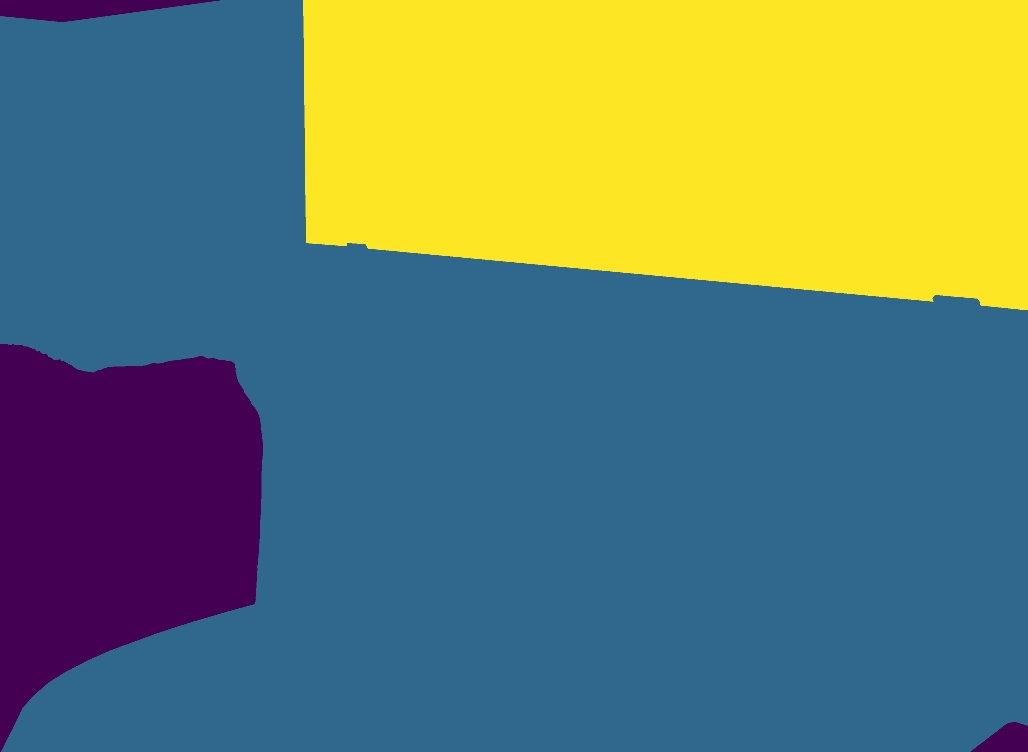} &
    \includegraphics[width=0.10\textwidth]{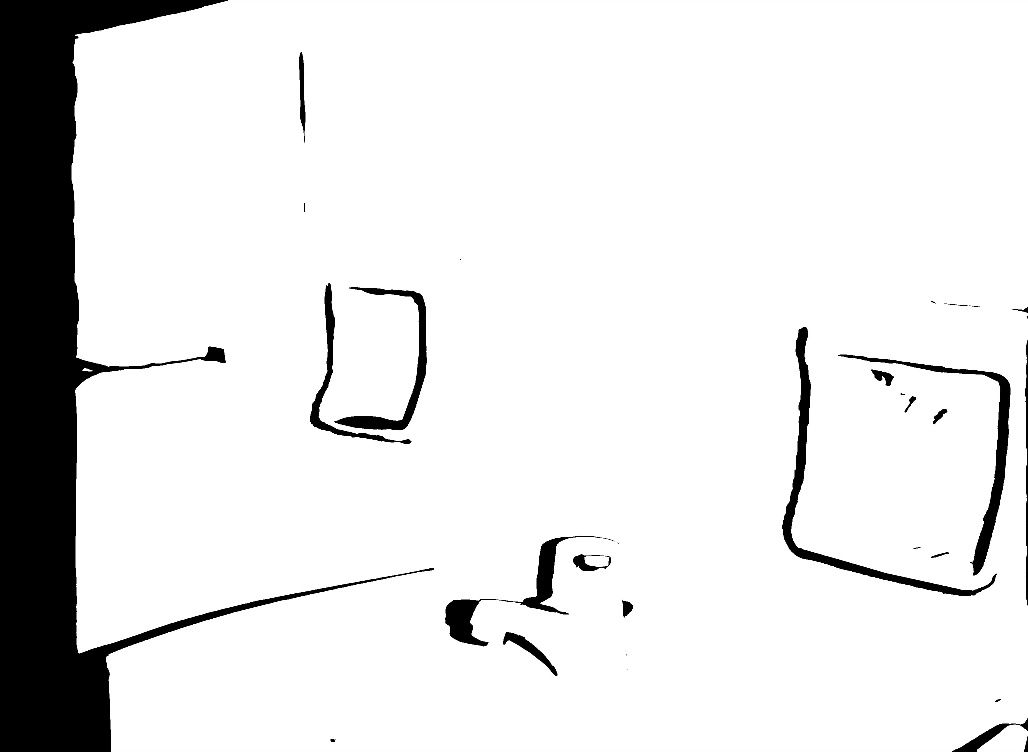} & &
    \includegraphics[width=0.10\textwidth]{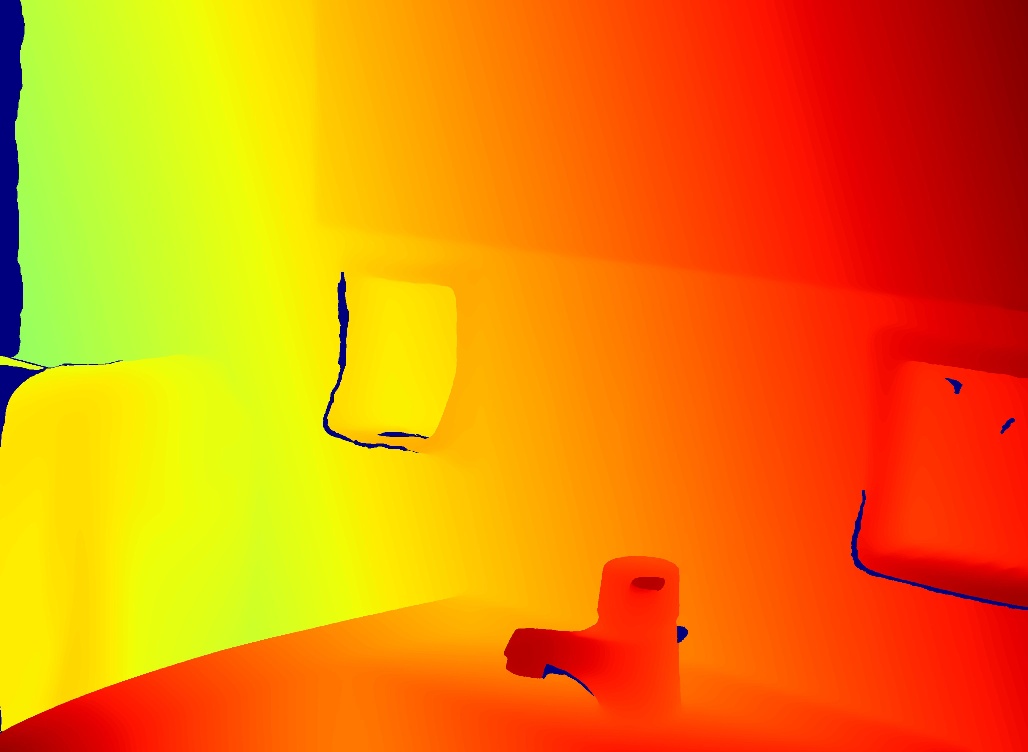} & &
    \includegraphics[width=0.10\textwidth]{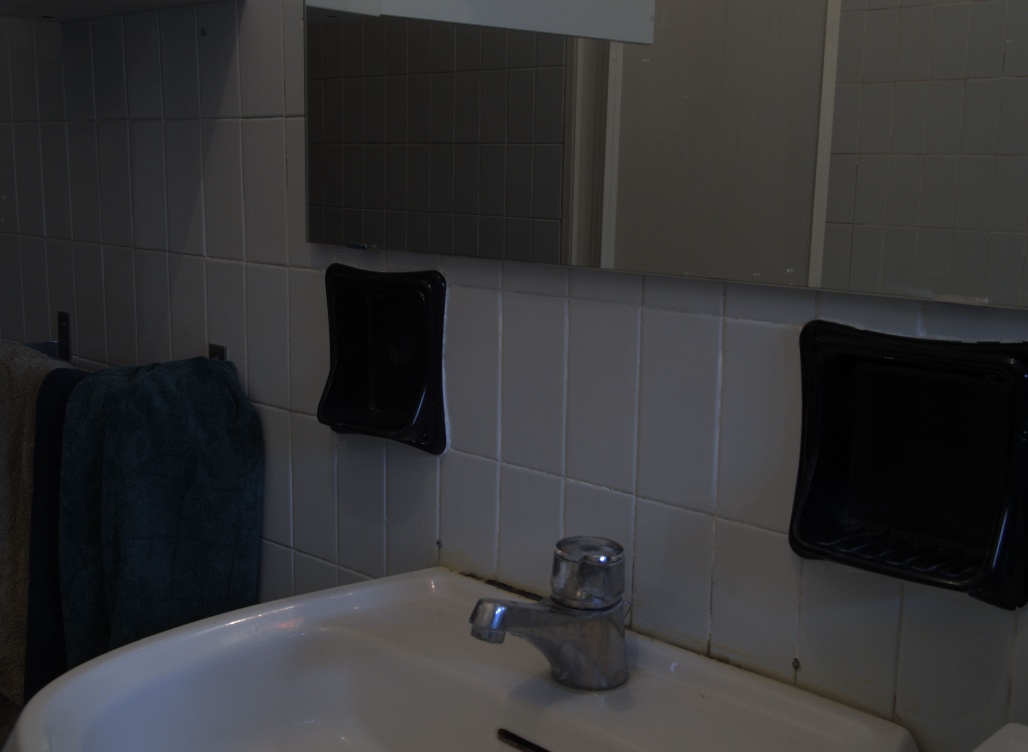} \\
    
    \end{tabular}
    \vspace{-0.25cm}\caption{\textbf{Data samples overview (stereo).} Columns 1 and 2: data available in the balanced setup: 12 Mpx stereo pair, material segmentation mask, left and right disparity maps, left-right consistency mask. Column 3: data available in the unbalanced setup (12 Mpx -  1.1 Mpx  image pair, high-res  disparity map associated with the  12 Mpx  image ). Column 4: 12 Mpx images acquired under different illuminations.}
    \label{fig:booster_sample}
\end{figure}

\begin{figure}[t]
    \centering
    \renewcommand{\tabcolsep}{1pt}
    \begin{tabular}{cccc}
    {\scriptsize  \textit{Reference Image}} & \scriptsize  \textit{Depth} & \scriptsize  \textit{Materials} & \scriptsize  \textit{Illuminations} \\
    \includegraphics[width=0.10\textwidth]{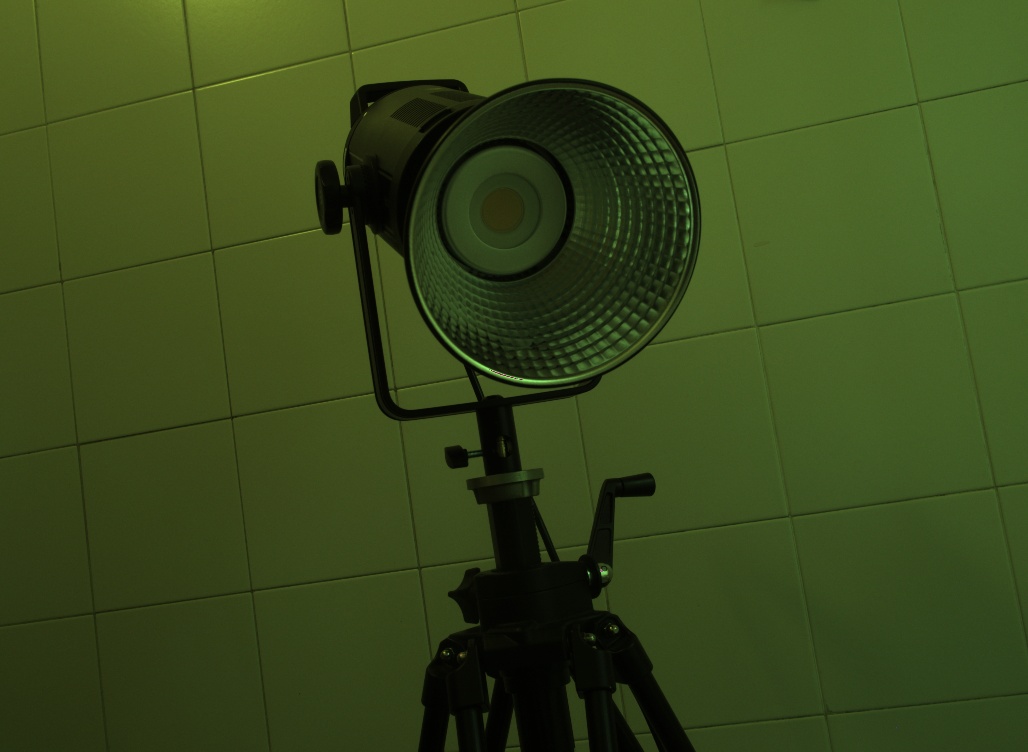} &
    \includegraphics[width=0.10\textwidth]{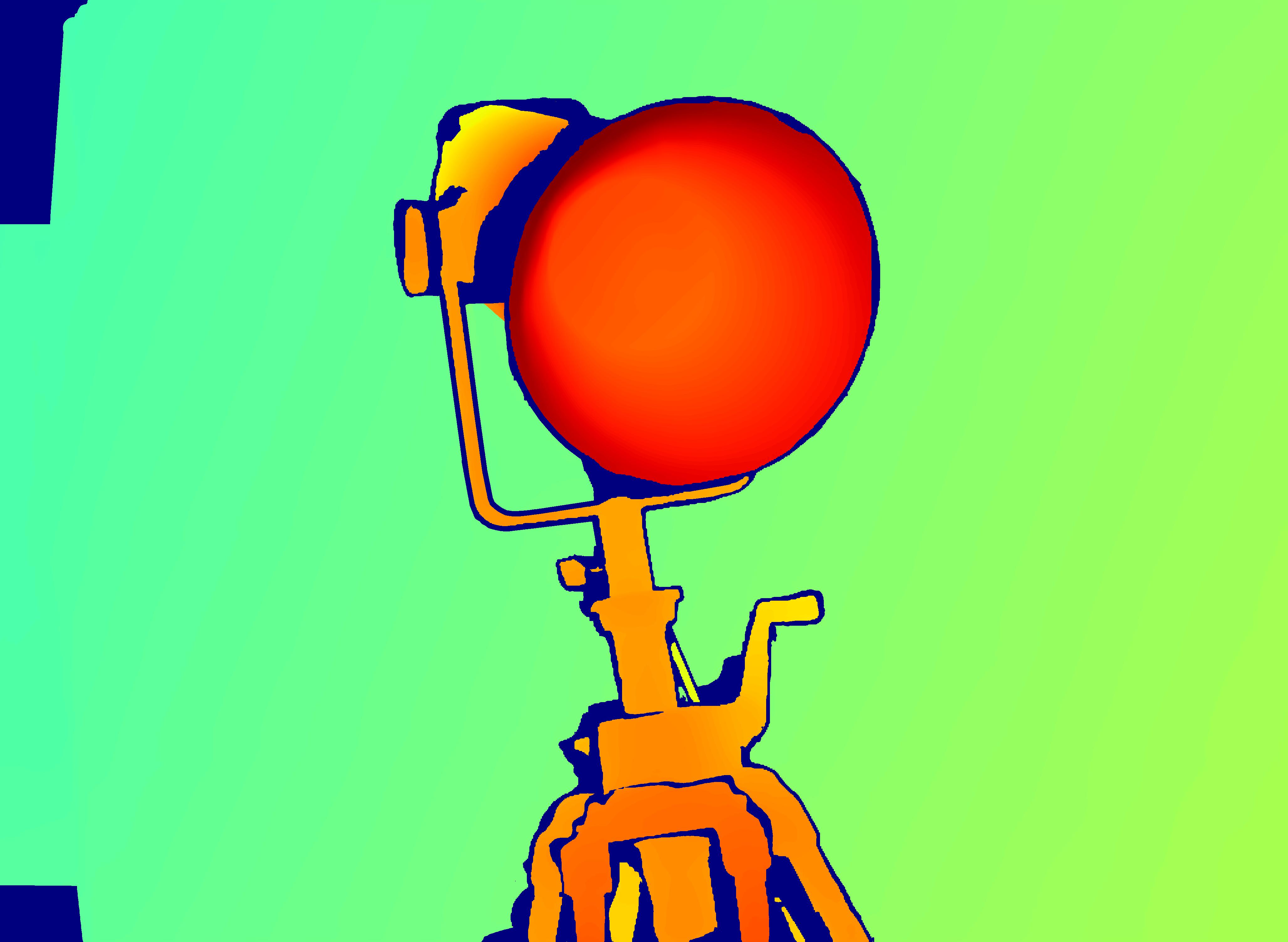} &
    \includegraphics[width=0.10\textwidth]{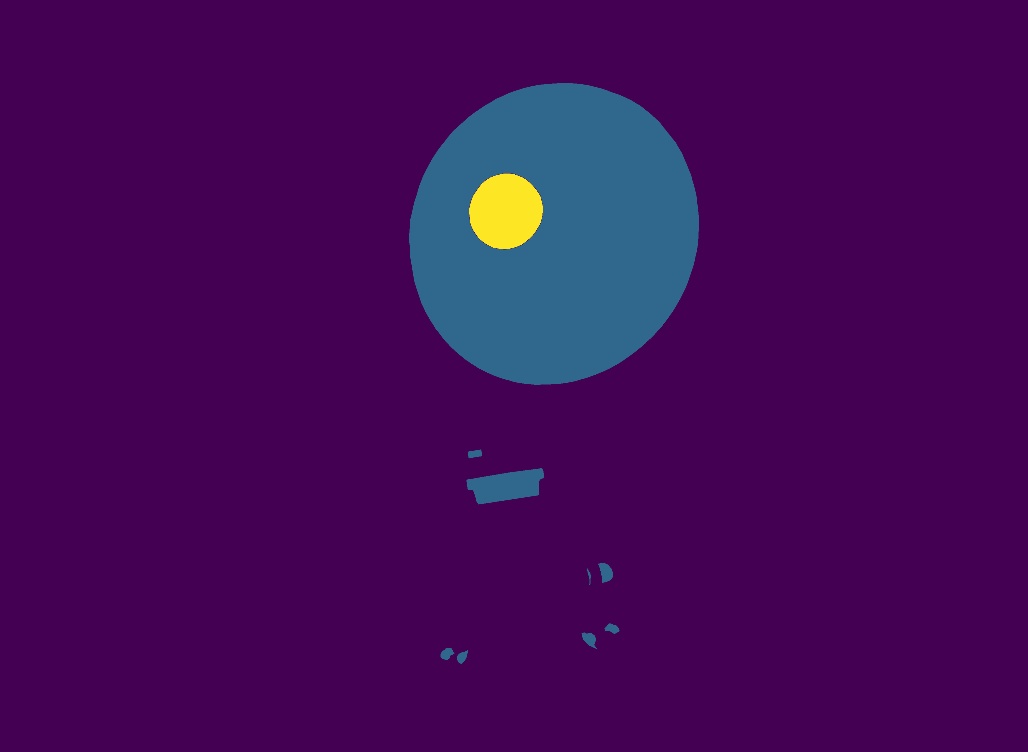} &
    \includegraphics[width=0.10\textwidth]{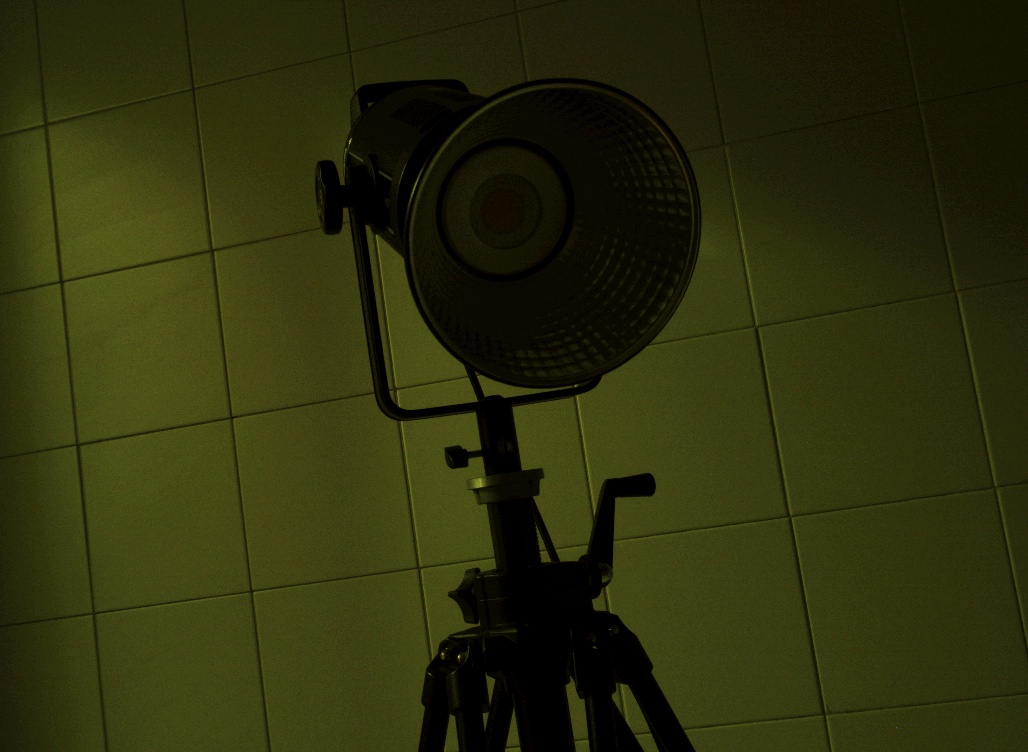} \\
    \includegraphics[width=0.10\textwidth]{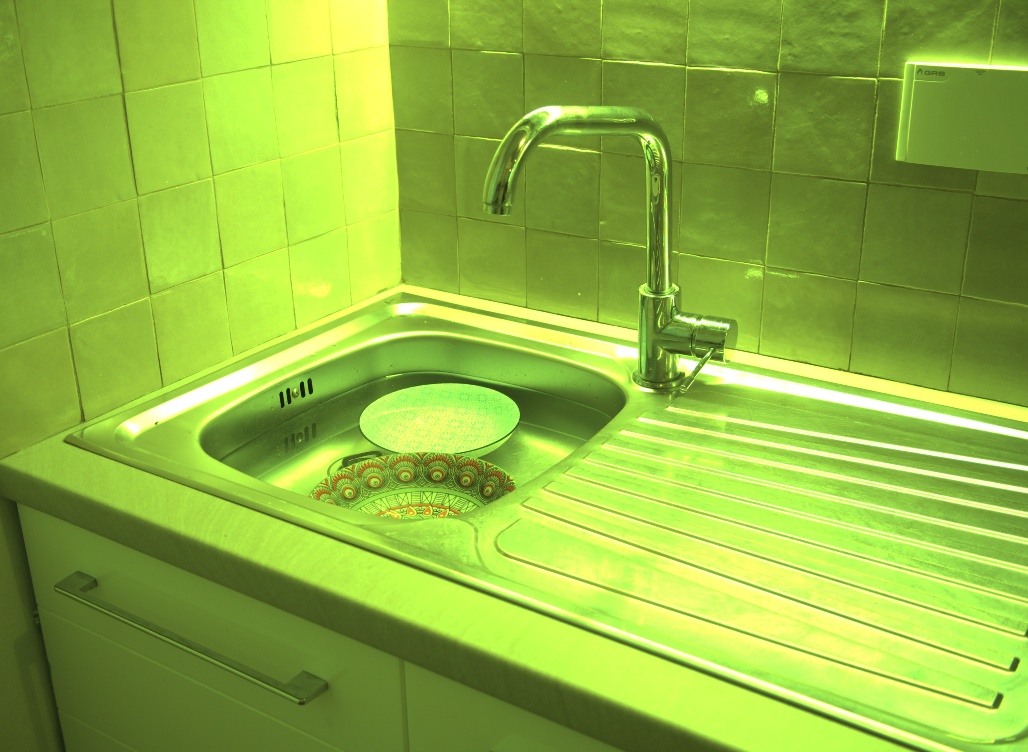} &
    \includegraphics[width=0.10\textwidth]{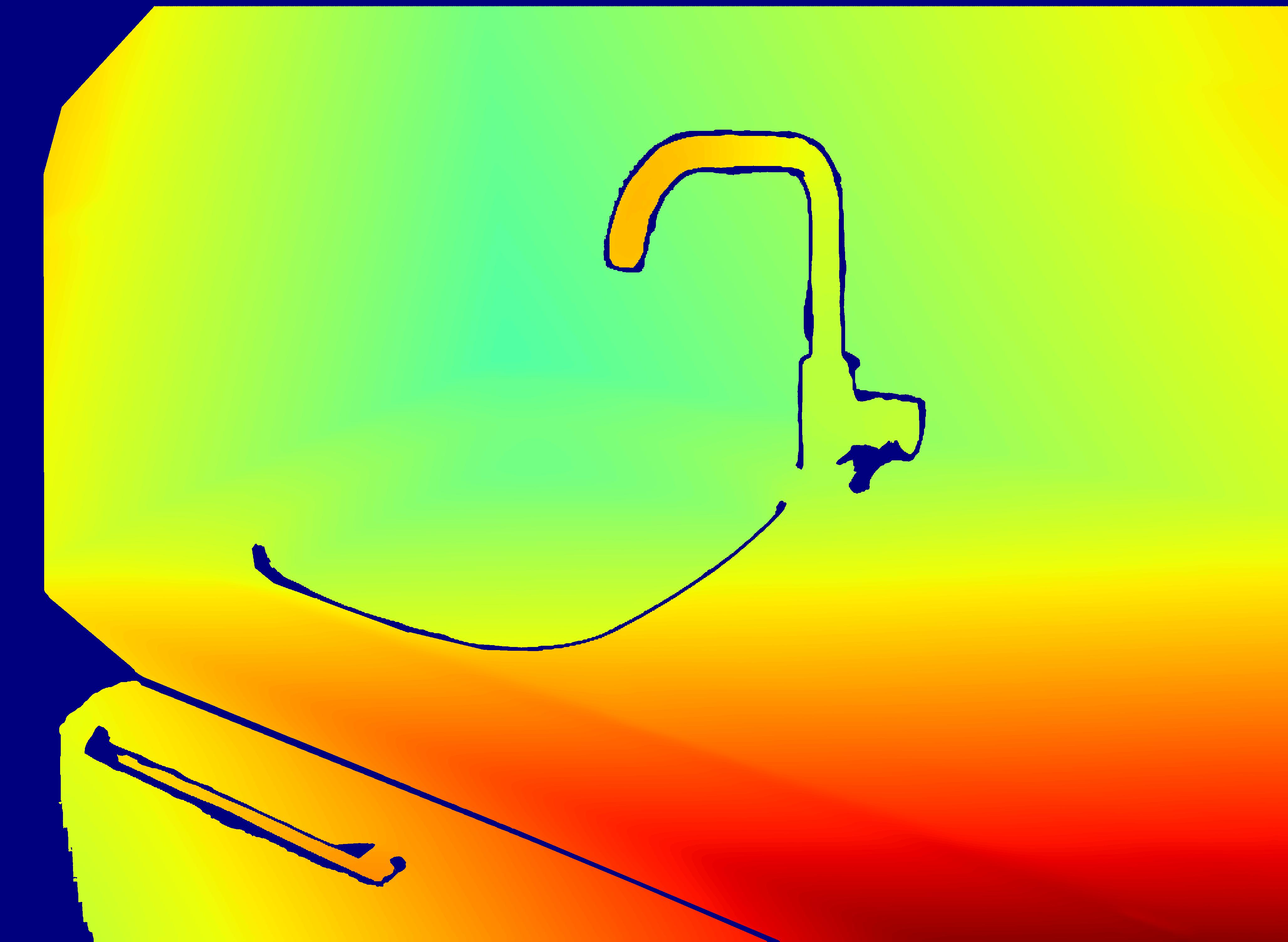} &
    \includegraphics[width=0.10\textwidth]{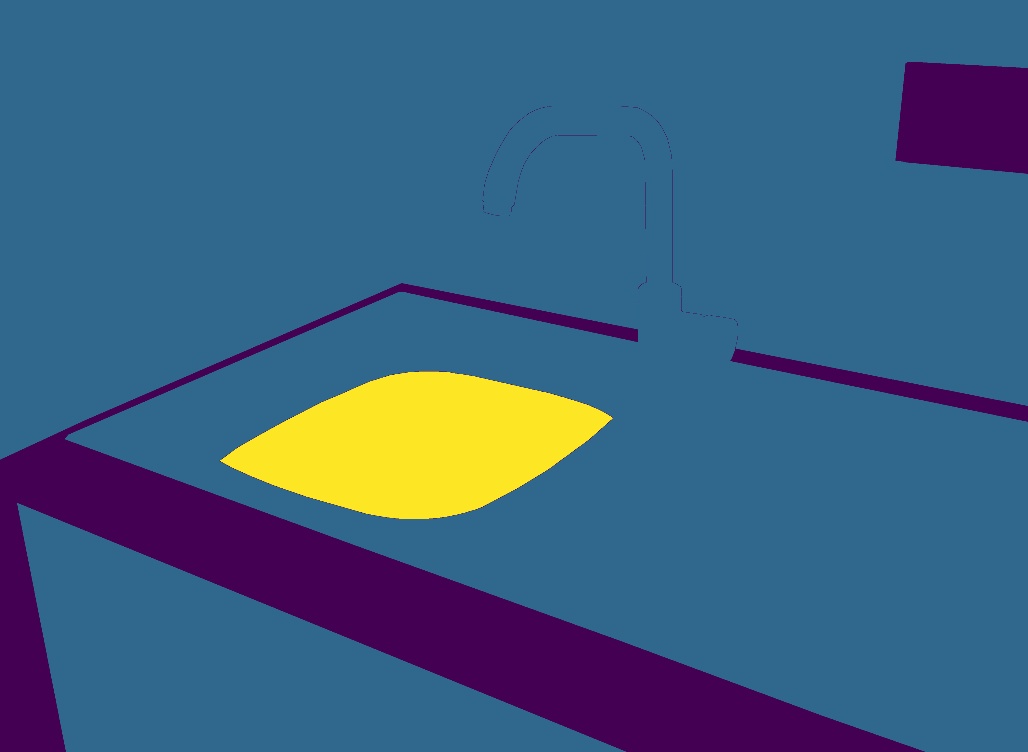} &
    \includegraphics[width=0.10\textwidth]{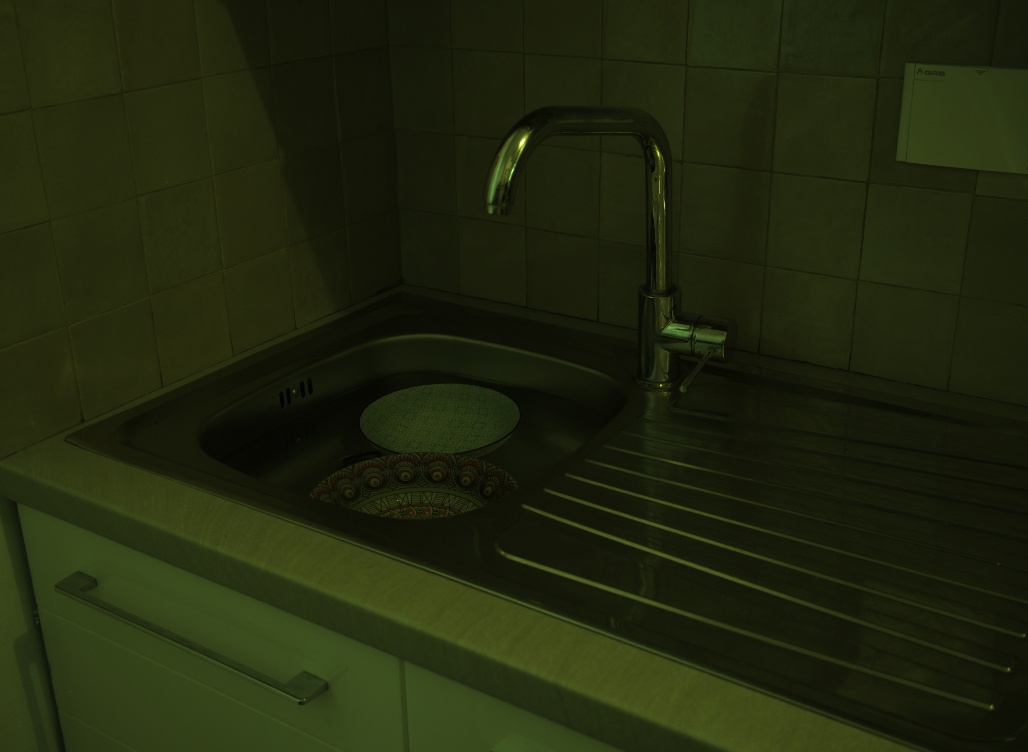} \\
    \includegraphics[width=0.10\textwidth]{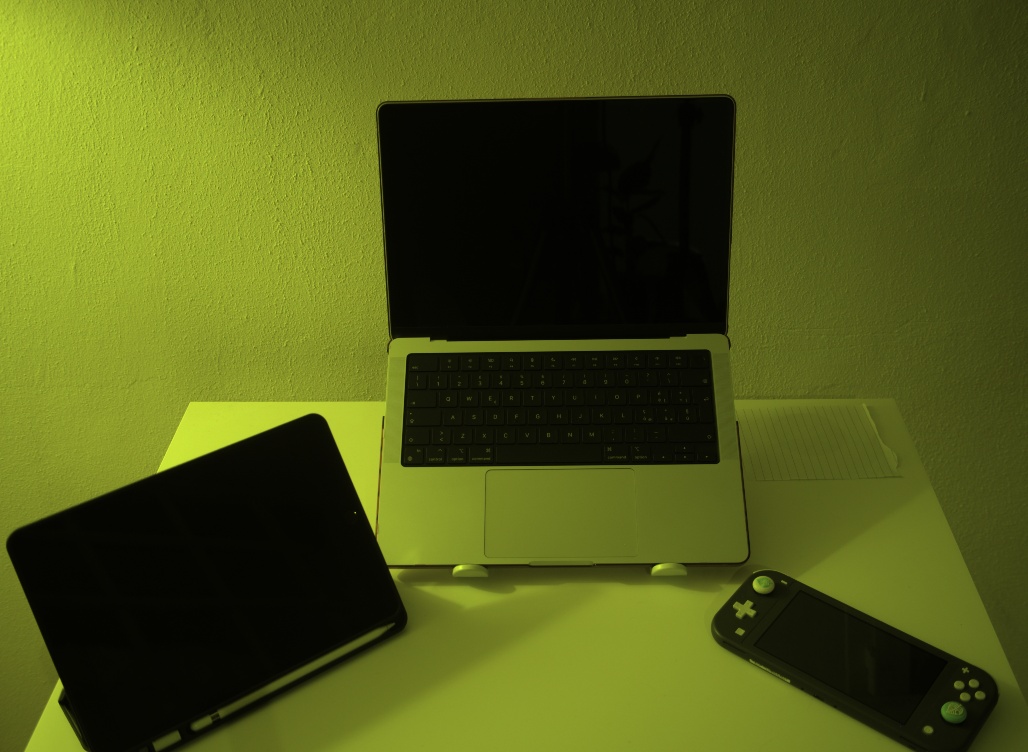} &
    \includegraphics[width=0.10\textwidth]{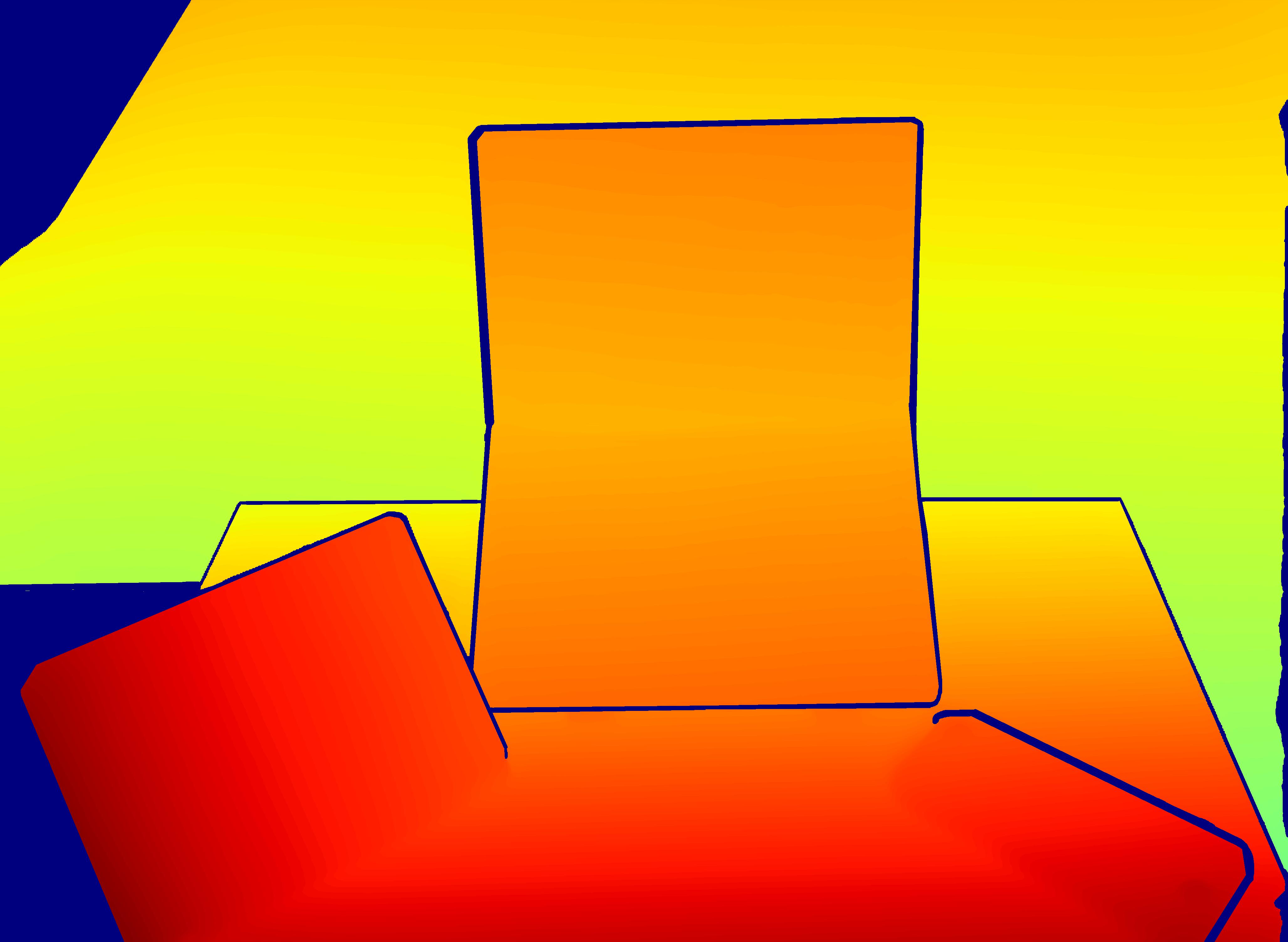} &
    \includegraphics[width=0.10\textwidth]{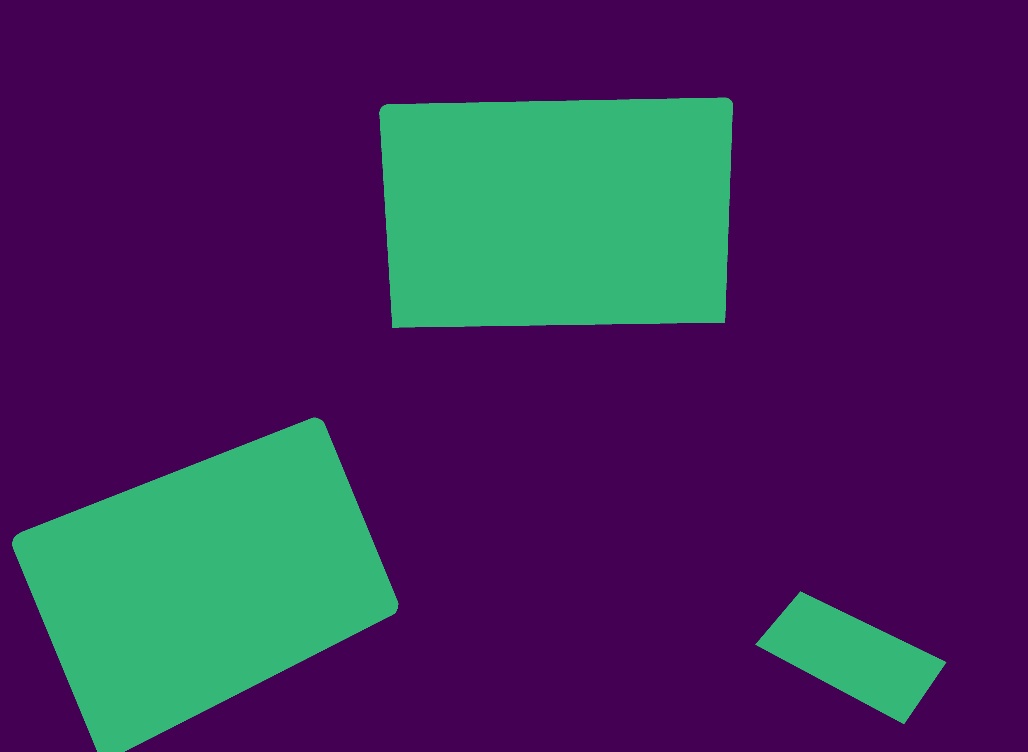} &
    \includegraphics[width=0.10\textwidth]{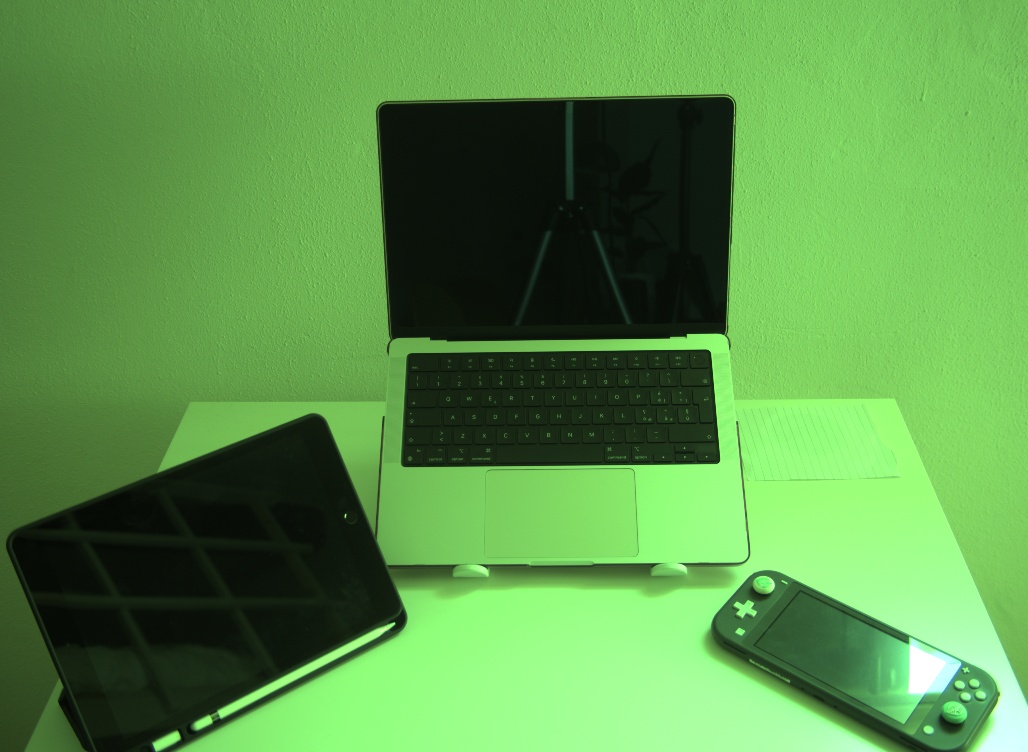} \\  
    \end{tabular}
    \vspace{-0.25cm}\caption{\textbf{Data samples overview (mono).} From left to right: reference image, ground-truth depth, material segmentation mask, additional images under different illumitations.
    }
    \label{fig:booster_sample_mono}
\end{figure}

\section{The Booster Dataset}

We describe  in detail here the Booster dataset organization, as well as the benchmarks and metrics used for evaluation.

\begin{figure}[t]
    \centering
    \renewcommand{\tabcolsep}{2pt}
    \begin{tabular}{cccc}
    \multicolumn{2}{c}{\scriptsize \textit{Balanced Stereo Pair}} &  \multicolumn{2}{c}{\scriptsize \textit{Unbalanced Stereo Pair}} \\ \includegraphics[width=0.24\linewidth]{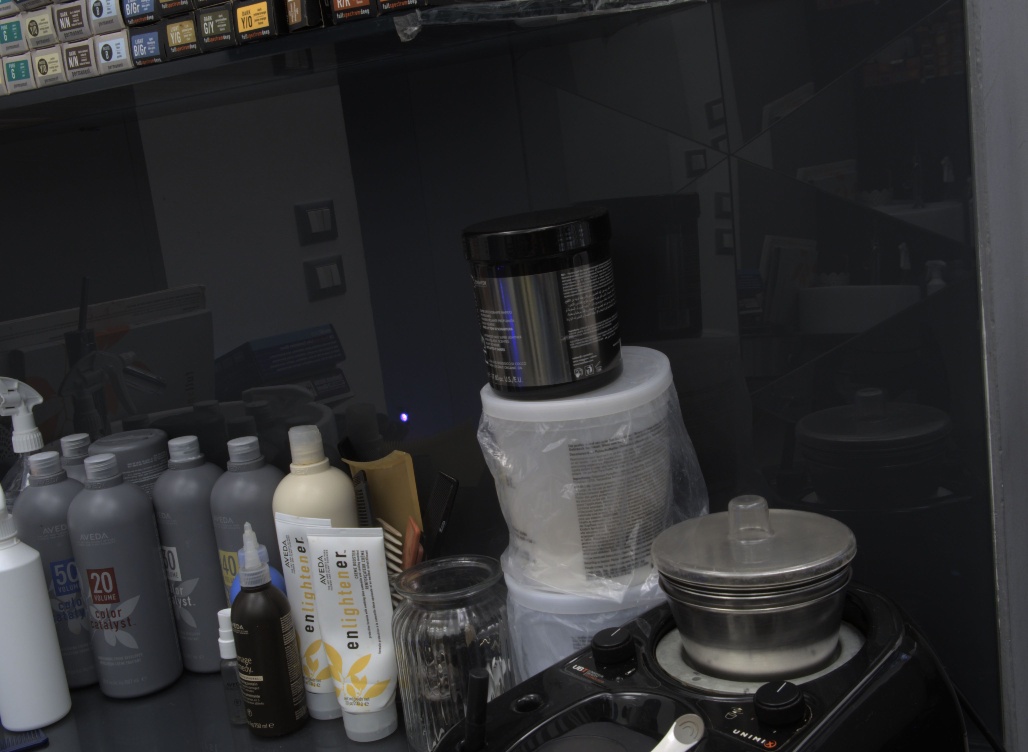} &   \includegraphics[width=0.24\linewidth]{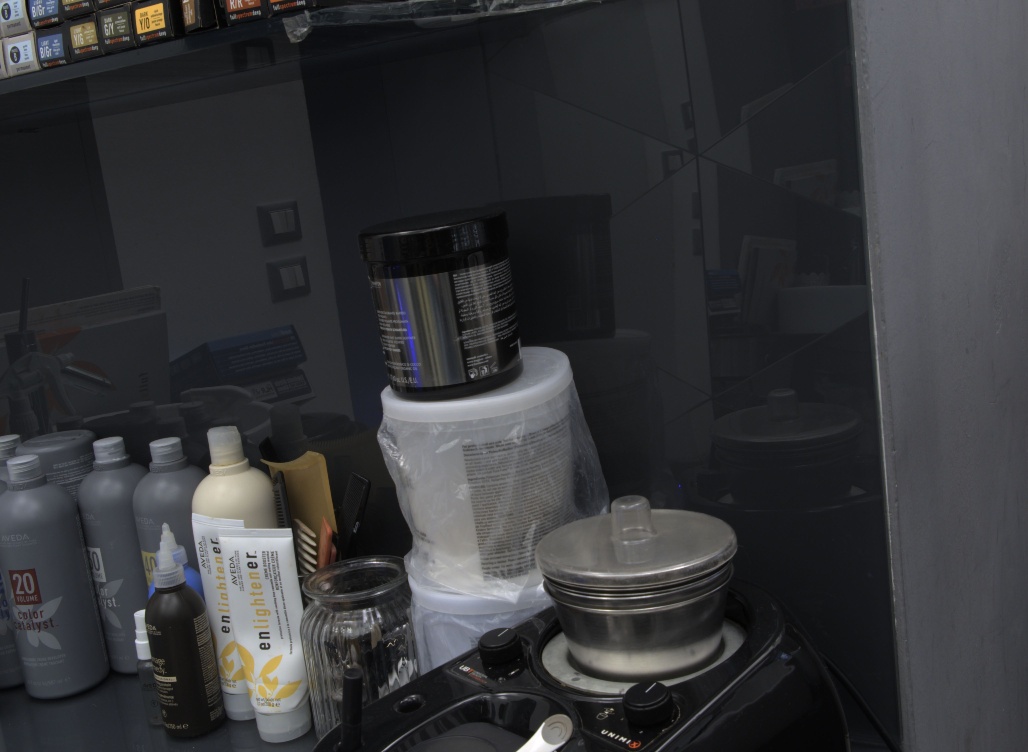} & \includegraphics[width=0.24\linewidth]{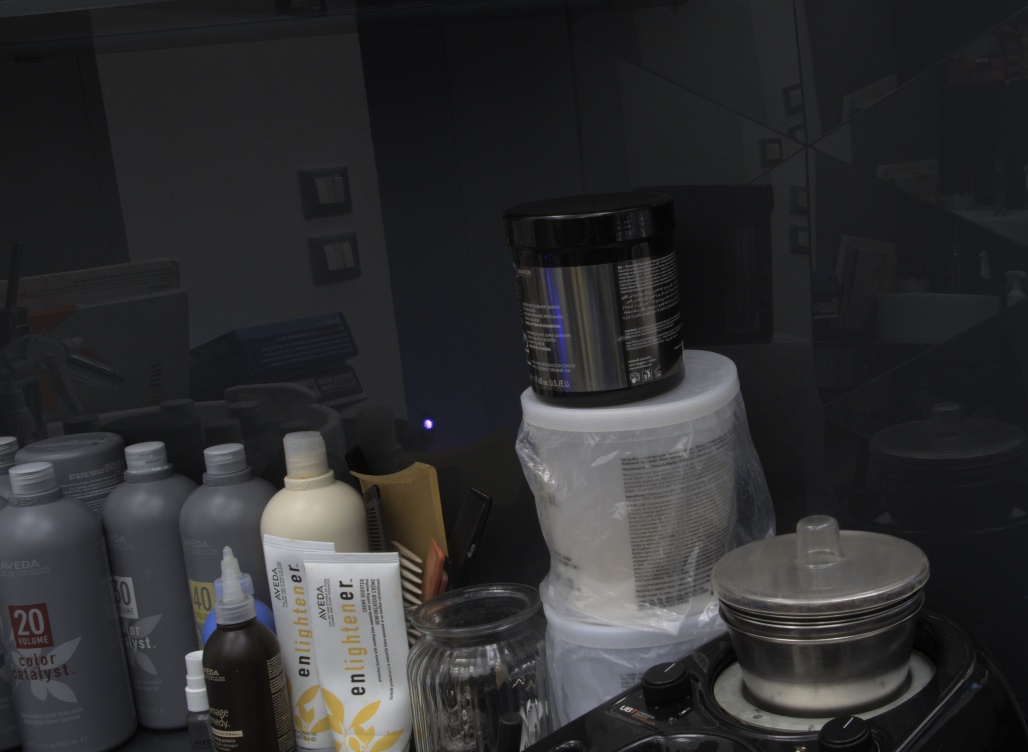} & \includegraphics[width=0.12\linewidth]{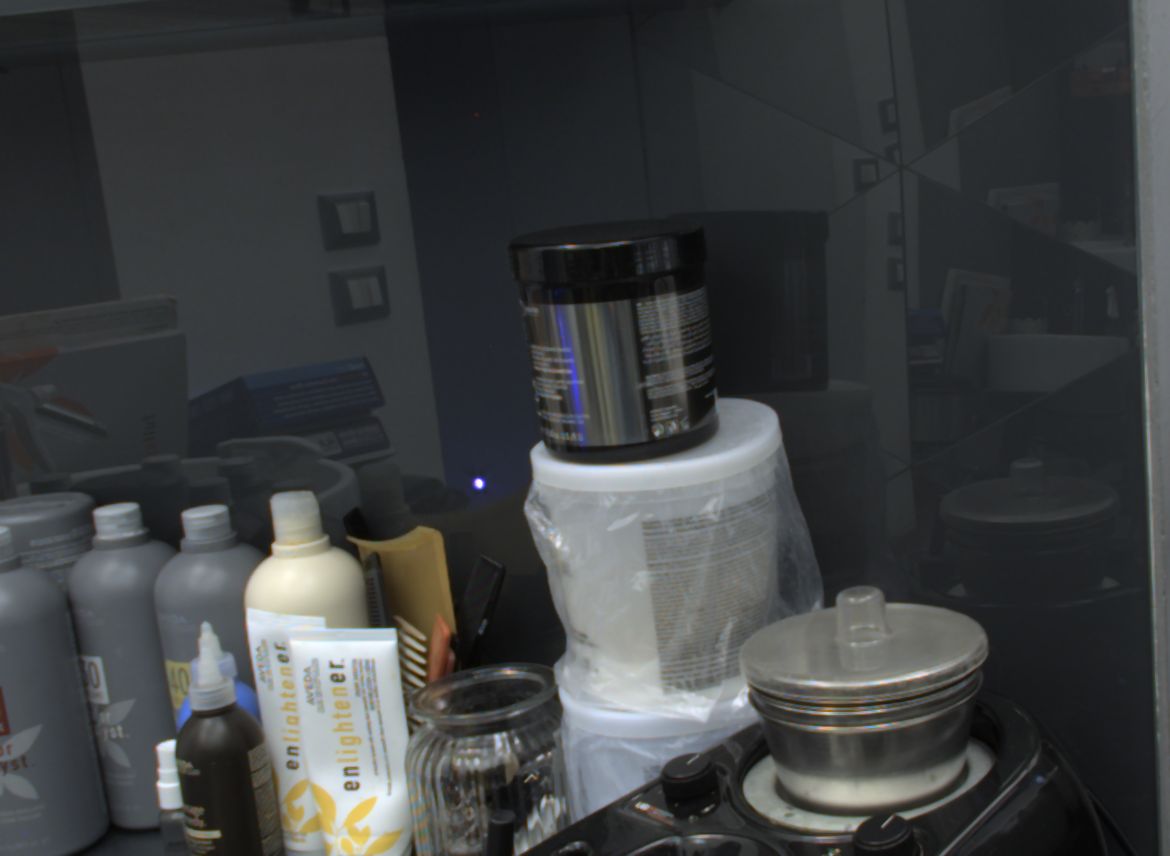} \\ \includegraphics[width=0.24\linewidth]{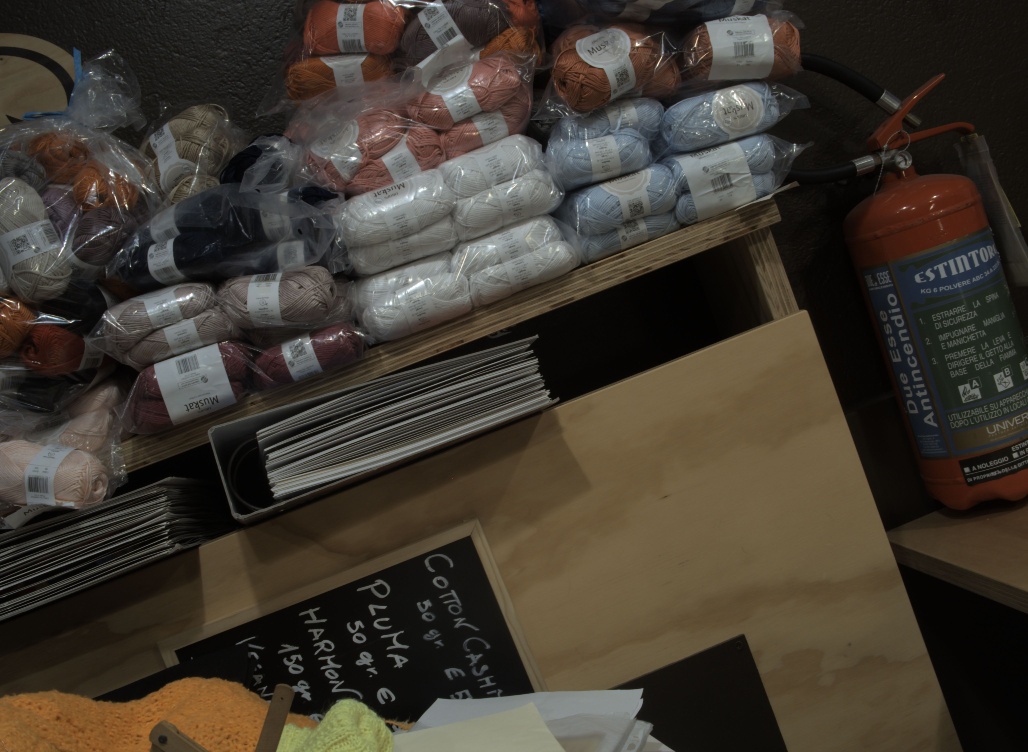} & \includegraphics[width=0.24\linewidth]{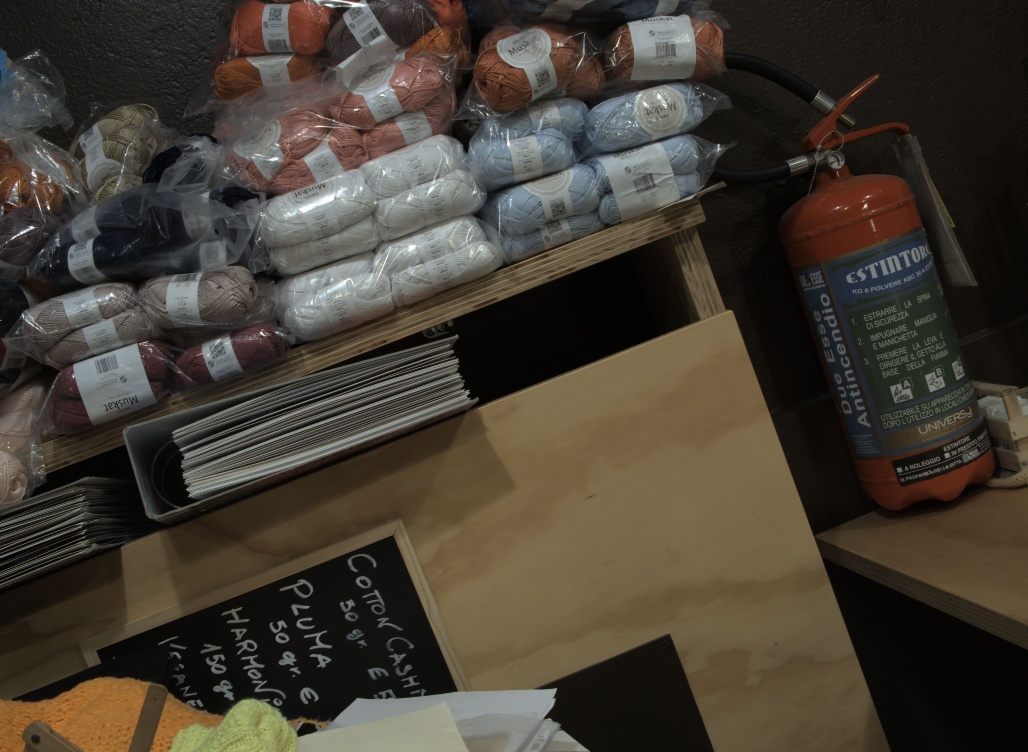} & \includegraphics[width=0.24\linewidth]{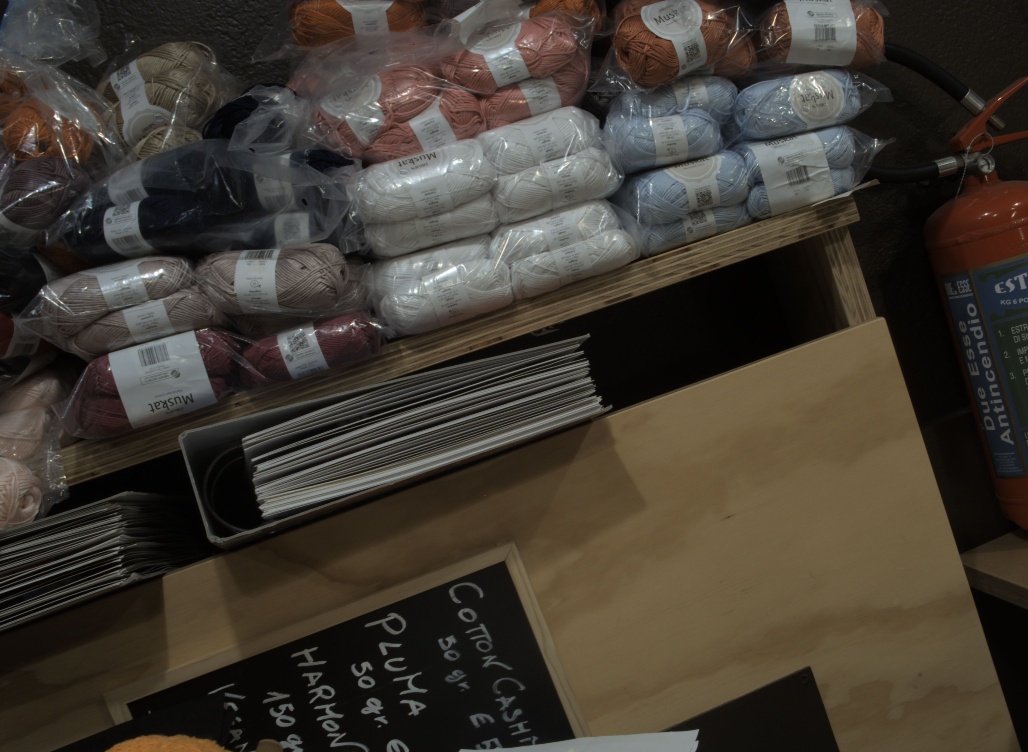} & \includegraphics[width=0.12\linewidth]{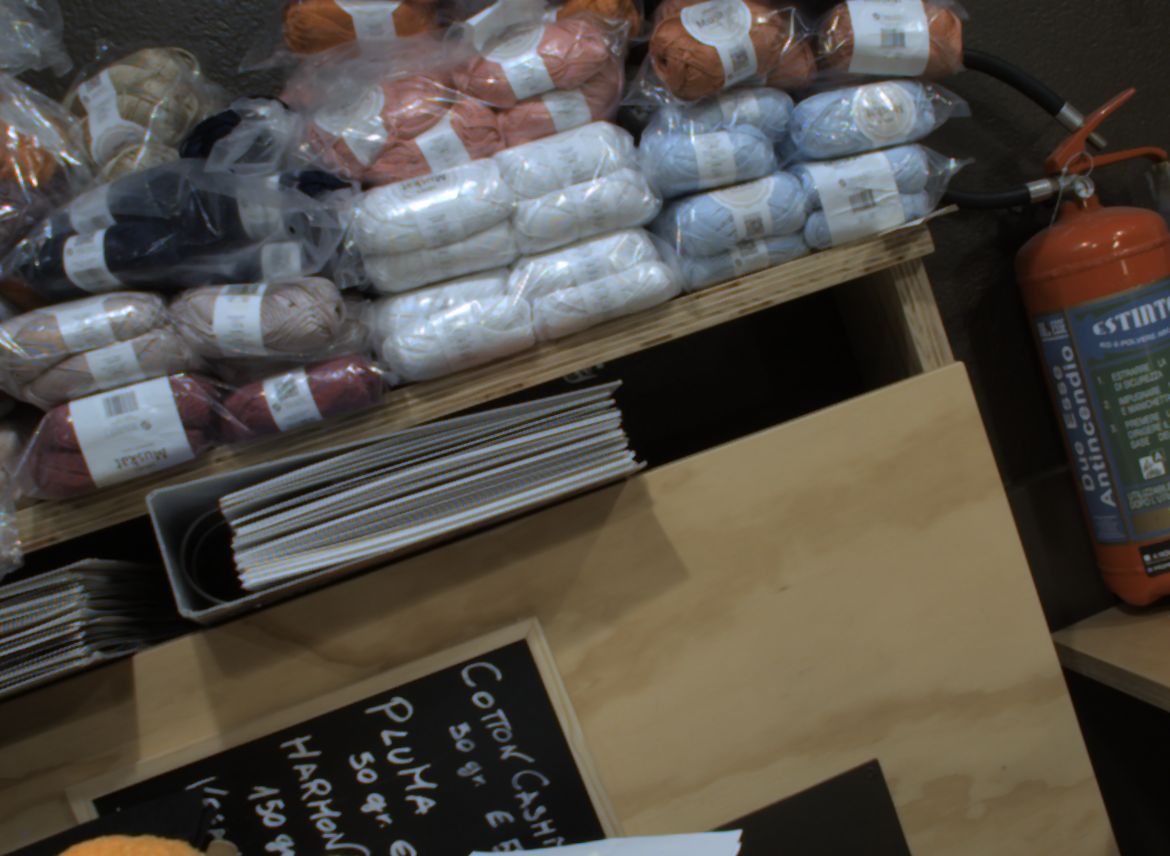} \\ \includegraphics[width=0.24\linewidth]{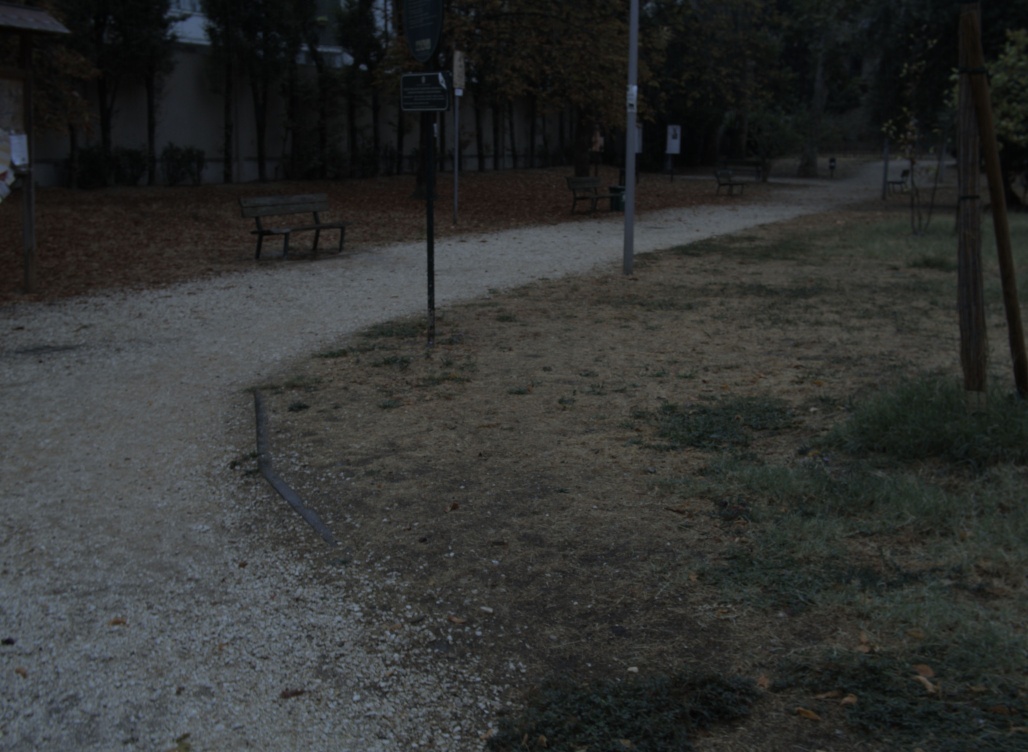} & \includegraphics[width=0.24\linewidth]{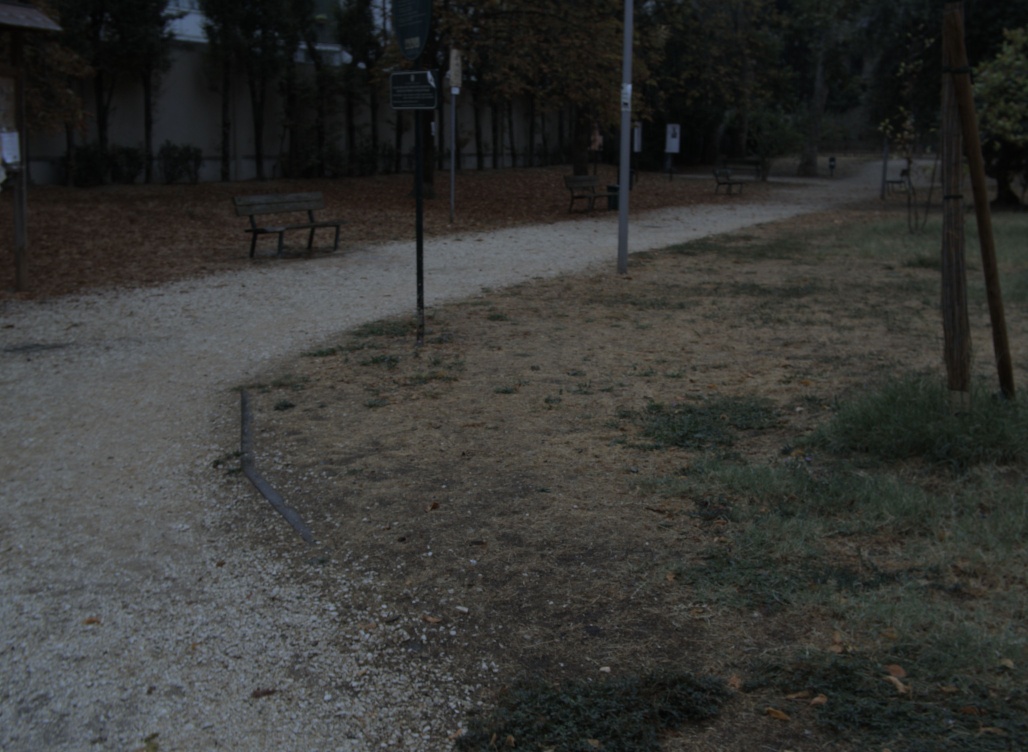} & \includegraphics[width=0.24\linewidth]{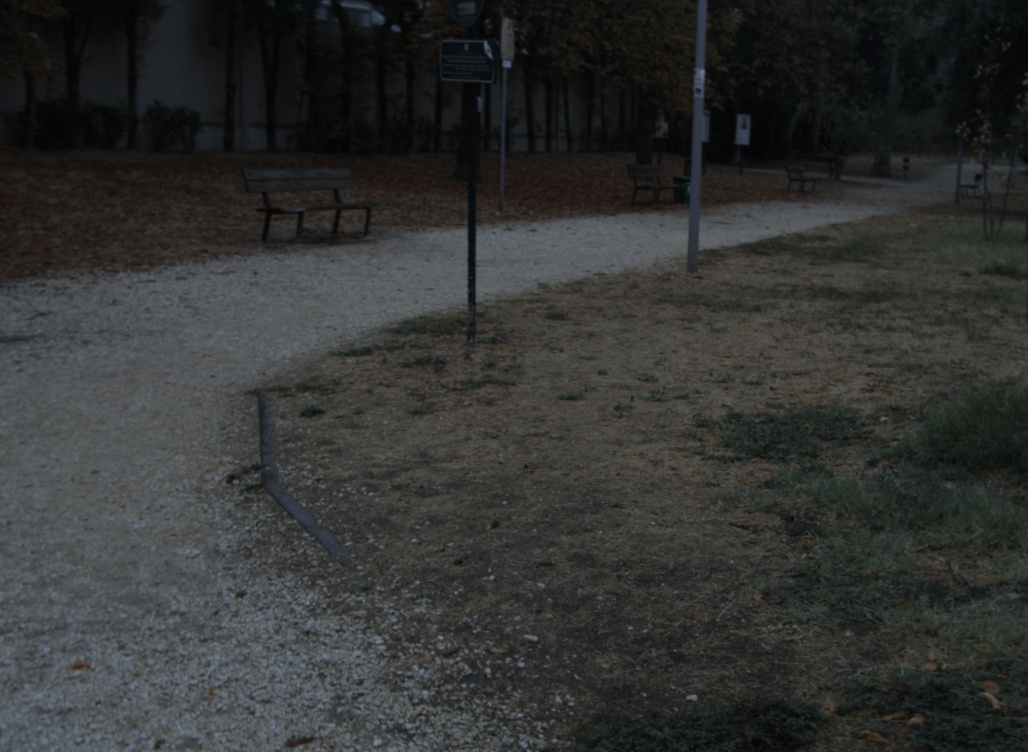} & \includegraphics[width=0.12\linewidth]{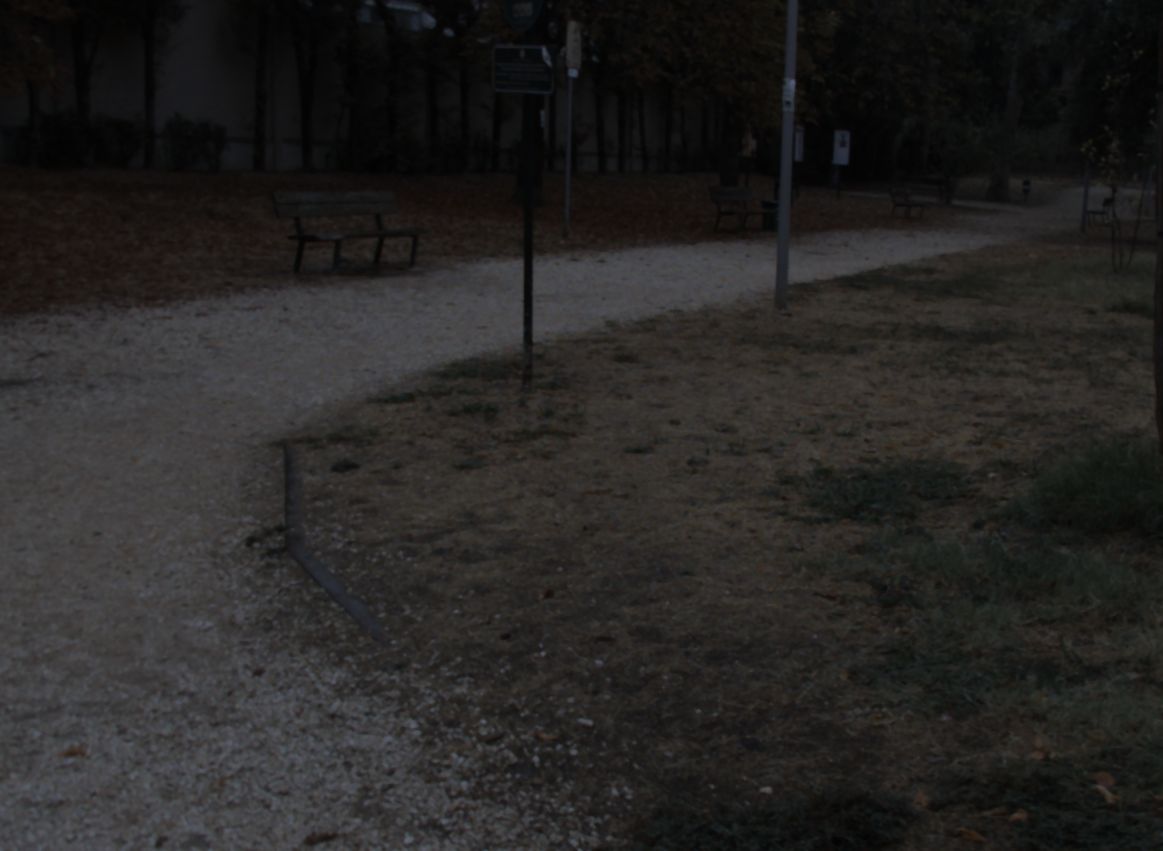} \\
    \end{tabular}
    \caption{\textbf{Booster passive images.} From left to right: left balanced, right balanced, left unbalanced, right unbalanced.}
    \label{fig:passives}
\end{figure}

\textbf{Dataset composition and splits.} We set up our camera rig and projectors in 85, different indoor scenes. In each, we collected multiple passive stereo images under different illumination conditions, thus obtaining a total of 606 samples, then annotated with dense annotations by means of the pipeline detailed in Sec. \ref{sec:pipeline}.
Respectively, we split the 85 scenes we collected into 38 for training, 26 and 21 for testing stereo and monocular networks -- dubbed \textit{Test Stereo} and \textit{Test Mono}. Accordingly, Booster counts 228 training images, 191 testing stereo images, and 187 testing monocular images. 
For images belonging to the training split, we release all the available data, \eg balanced and unbalanced stereo images, material segmentation, disparity ground-truth, etc. 
For those belonging to the testing splits, we release only RGB images and the calibration file. Concerning Test Stereo, we release both balanced and unbalanced pairs, while we release only the high-resolution left image from the $L-R$ pair in Test Mono, thus avoiding the possibility of exploiting the stereo pairs to gain an unfair advantage in the monocular benchmark. 
The splits are defined to obtain a variety of environments within the same split yet to have similar context across them (\eg all splits contain a scene framing a mirror) to balance the distribution of challenging materials and objects. \cref{fig:mat_hist} shows that the material class distributions of pixels belonging to the train, Test Mono, and Test Stereo split are similar.
Moreover, we show in \cref{fig:disp_hist} that disparity values are distributed consistently across the three splits.

\textbf{Benchmarks.}
Three main benchmarks are defined in Booster: \textit{Balanced} stereo, including pairs at 12 Mpx, \textit{Unbalanced} stereo, featuring equally many 12 Mpx - 1.1 Mpx pairs, and the \textit{Monocular} benchmark, containing only single 12Mpx images.
The three benchmarks share the same training split. Test Stereo is used for the unbalanced and balanced stereo setups, while Test Mono is used for the monocular benchmark only.
The \textit{Unbalanced} stereo represents the first-ever real benchmark for unbalanced stereo matching, a task studied so far only by simulating the unbalanced setup by resizing one of the two images of a balanced pair of same-resolution stereo images \cite{liu2020visually,aleotti2021neural}.
Fig. \ref{fig:booster_sample} shows two examples of scenes featured in the stereo benchmark of Booster, highlighting the annotations we provide for each.
Fig. \ref{fig:booster_sample_mono} reports some examples taken from the monocular benchmark, showing collected images and annotations.

\begin{table*}[t]
\centering
\renewcommand{\tabcolsep}{8pt}
\scalebox{0.8}{
\begin{tabular}{ccc}

 \begin{tabular}{c}
    \multirow{3}{*}{\rotatebox{90}{Full res.}} \\ 
 \end{tabular}
 \begin{tabular}{ll}
 \\
 \toprule
 & Input \\
 Model & Res.\\
 \midrule
 SGM \cite{hirschmuller2007stereo} & Q \\
 MC-CNN \cite{zbontar2016stereo} & Q \\
 LEAStereo \cite{cheng2020hierarchical} & Q \\
 CFNet \cite{shen2021cfnet} & Q \\
 HSMNet \cite{yang2019hierarchical} & Q \\
 RAFT-Stereo$^*$ \cite{lipson2021raft} & Q \\
 RAFT-Stereo \cite{lipson2021raft} & Q \\
 CREStereo \cite{li2022practical} & Q \\
 \midrule
 SGM \cite{hirschmuller2007stereo} & H \\
 HSMNet \cite{yang2019hierarchical} & H \\
 SGM+Neural Ref. \cite{aleotti2021neural} & H \\
 RAFT-Stereo$^*$ \cite{lipson2021raft} & H \\
 RAFT-Stereo \cite{lipson2021raft} & H \\
 CREStereo \cite{li2022practical} & H \\
 \midrule
 HSMNet \cite{yang2019hierarchical} & F \\
 \bottomrule
 \end{tabular}
 &
 \begin{tabular}{rrrr | rr }
 \multicolumn{6}{c}{All pixels} \\
 \toprule
 bad-2 & bad-4 & bad-6 & bad-8 & MAE & RMSE \\
 $\downarrow$ (\%) & $\downarrow$ (\%) & $\downarrow$ (\%) & $\downarrow$ (\%) & $\downarrow$ (px.) & $\downarrow$ (px.) \\
 \midrule
 80.35 & 66.89 & 58.09 & 52.21 & 57.01 & 119.21 \\ 
 88.09 & 66.30 & 47.77 & 40.53 & 31.23 & 62.98 \\ 
 70.86 & 55.41 & 47.56 & 42.25 & 27.61 & 51.72 \\ 
 61.34 & 48.33 & 42.22 & 38.34 & 27.60 & 51.62 \\ 
 66.95 & 48.05 & 37.46 & 31.14 & 20.97 & 42.72 \\ 
 40.27 &  27.54 &  22.83 &  20.13 & 17.08 &  36.30 \\ 
 35.64 & 23.62 & 19.61 & 17.43 & 16.28 & 34.64 \\ 
 \bfseries 33.07 & \bfseries 21.36 & \bfseries 17.35 & \bfseries 15.09 & \bfseries  12.56 & \bfseries 29.45\\ 
 \midrule
 76.61 & 64.72 & 58.34 & 54.37 & 71.68 & 133.35 \\ 
 53.75 & 36.47 & 28.71 & 24.50 & 19.17 & 42.00 \\ 
 78.54 & 63.20 & 53.77 & 46.87 & 31.82 & 67.02 \\ 
 46.31 & 35.49 & 30.98 & 28.15 & 23.95 & 49.94 \\ 
 37.35 & 27.38 & 23.98 & 21.94 & 20.95 & 42.00 \\ 
 40.86 & 31.24 & 27.57 & 25.31 & 24.05 & 50.93 \\ 
 \midrule
 50.85 & 36.53 & 30.77 & 27.56 & 30.82 & 68.97 \\ 
 \bottomrule
 \end{tabular}
 &
 \begin{tabular}{rrrr | rr }
 \multicolumn{6}{c}{Cons pixels} \\
 \toprule
 bad-2 & bad-4 & bad-6 & bad-8 & MAE & RMSE \\
 $\downarrow$ (\%) & $\downarrow$ (\%) & $\downarrow$ (\%) & $\downarrow$ (\%) & $\downarrow$ (px.) & $\downarrow$ (px.) \\
 \midrule
 78.40 & 63.70 & 54.13 & 47.79 & 41.28 & 91.86 \\ 
 87.64 & 64.20 & 44.24 & 36.70 & 27.56 & 57.34 \\ 
 69.15 & 53.17 & 45.42 & 40.24 & 26.36 & 49.52 \\ 
 59.13 & 46.02 & 40.08 & 36.36 & 25.72 & 48.55 \\ 
 65.23 & 45.86 & 35.36 & 29.31 & 20.93 & 42.42 \\ 
 38.65 &  26.49 &  22.25 &  19.84 &  17.13 & 35.76\\ 
 34.35 & 23.04 & 19.46 & 17.47 & 16.43 & 34.27 \\ 
 \bfseries  30.50 & \bfseries 19.48 & \bfseries 16.08 & \bfseries 14.20 & \bfseries 12.47 &  \bfseries 29.02 \\ 
 \midrule
 74.18 & 61.17 & 54.25 & 49.99 & 55.25 & 106.55 \\ 
 51.25 & 34.06 & 26.78 & 23.01 & 18.92 & 41.28 \\ 
 78.35 & 60.59 & 49.59 & 42.50 & 30.92 & 68.37 \\ 
 44.02 & 33.59 & 29.49 & 26.95 & 23.25 & 48.11 \\ 
 35.88 & 26.59 & 23.59 & 21.75 & 20.81 & 41.08 \\ 
 37.51 & 28.45 & 25.38 & 23.55 & 23.59 & 49.93 \\ 
 \midrule
 48.11 & 33.88 & 28.50 & 25.61 & 30.02 & 66.79 \\ 
 \bottomrule
 \end{tabular}
 
 \\
 \\
 
 \begin{tabular}{c}
    \multirow{3}{*}{\rotatebox{90}{Quarter res.}} \\ 
 \end{tabular}
 \begin{tabular}{ll}
 \\
 \toprule
 & Input \\
 Model & Res.\\
 \midrule
 SGM \cite{hirschmuller2007stereo} & Q \\
 MC-CNN \cite{zbontar2016stereo} & Q \\
 LEAStereo \cite{cheng2020hierarchical} & Q \\
 CFNet \cite{shen2021cfnet} & Q \\
 HSMNet \cite{yang2019hierarchical} & Q \\
 RAFT-Stereo$^*$ \cite{lipson2021raft} & Q \\
 RAFT-Stereo \cite{lipson2021raft} & Q \\
 CREStereo \cite{li2022practical} & Q \\
 \bottomrule
 \end{tabular}
 &
 \begin{tabular}{rrrr | rr }
 \multicolumn{6}{c}{All pixels} \\
 \toprule
 bad-2 & bad-4 & bad-6 & bad-8 & MAE & RMSE \\
 $\downarrow$ (\%) & $\downarrow$ (\%) & $\downarrow$ (\%) & $\downarrow$ (\%) & $\downarrow$ (px.) & $\downarrow$ (px.) \\
 \midrule
 52.76 & 39.43 & 33.11 & 29.26 & 14.64 & 30.68 \\ 
 40.33 & 30.36 & 25.64 & 22.25 & 7.82 & 15.85 \\ 
 42.21 & 30.23 & 24.37 & 20.43 & 6.89 & 12.92 \\ 
 38.31 & 29.53 & 24.70 & 21.34 & 6.89 & 12.89 \\ 
 31.11 & 20.25 & 15.92 & 13.23 & 5.24 & 10.67 \\ 
 20.13 &  15.13 &  12.85 &  11.05 &  4.27 &  9.05 \\ 
 17.43 & 13.49 & 11.59 & 10.10 & 4.07 & 8.64 \\ 
 \bfseries 15.13 & \bfseries 10.70 & \bfseries 8.91 & \bfseries 7.57 & \bfseries 3.15 & \bfseries 7.40\\ 
 \bottomrule
 \end{tabular}
 &
 \begin{tabular}{rrrr | rr }
 \multicolumn{6}{c}{Cons pixels} \\
 \toprule
 bad-2 & bad-4 & bad-6 & bad-8 & MAE & RMSE \\
 $\downarrow$ (\%) & $\downarrow$ (\%) & $\downarrow$ (\%) & $\downarrow$ (\%) & $\downarrow$ (px.) & $\downarrow$  (px.) \\
 \midrule
 48.42 & 34.18 & 27.58 & 23.63 & 10.75 & 24.05 \\ 
 36.50 & 26.50 & 21.84 & 18.79 & 6.90 & 14.43 \\ 
 40.19 & 28.68 & 23.21 & 19.50 & 6.58 & 12.36 \\ 
 36.32 & 27.85 & 23.24 & 20.05 & 6.42 & 12.11 \\ 
 29.25 & 19.47 & 15.70 & 13.23 & 5.22 & 10.59 \\ 
 19.82 &  15.19 &  12.98 &  11.17 &  4.28 &  8.91 \\ 
 17.46 & 13.70 & 11.82 & 10.31 & 4.10 & 8.54 \\ 
 \bfseries 14.21 & \bfseries 10.49 & \bfseries 8.96 & \bfseries 7.65 & \bfseries 3.12 & \bfseries 7.28  \\ 
 \bottomrule
 \end{tabular}
 
 \end{tabular}
 }
 \caption{\textbf{Results on the Booster Balanced benchmark.} Stereo networks load weights provided by their authors. We evaluate on full-resolution ground-truth maps, or by downsampling them to quarter resolution. Best scores in \textbf{bold}.}
 \label{tab:stereo_tournament}
\end{table*}

\textbf{Unlabeled samples.} We collect and release 15K additional samples -- in both balanced and unbalanced settings -- by walking in several indoor and outdoor environments. We hope these images can foster the development of learning approaches not requiring ground-truth labels. Some examples of these unlabeled images are shown in \cref{fig:passives}.

\begin{table}[t]
\centering
\scalebox{0.8}{

\begin{tabular}{cc}
 \begin{tabular}{c}
    \multirow{3}{*}{\rotatebox{90}{Full res.}} \\ 
 \end{tabular}
 \begin{tabular}{l}
 \\
 \toprule
 \\
 Model \\
 \midrule
 LEAStereo \\
 LEAStereo (ft) \\
 \midrule
 CFNet \\
 CFNet (ft) \\
 \midrule
 RAFT-Stereo \\
 RAFT-Stereo (ft) \\
  \midrule
 CRESTereo \\ 
 CREStereo (ft)  \\ 
 \bottomrule
 \end{tabular}
 
 &
 
 \begin{tabular}{rrrr | rr }
 \multicolumn{6}{c}{All pixels} \\
 \toprule
 bad-2 & bad-4 & bad-6 & bad-8 & MAE & RMSE \\
 $\downarrow$ (\%) & $\downarrow$ (\%) & $\downarrow$ (\%) & $\downarrow$ (\%) & $\downarrow$ (px.) & $\downarrow$  (px.) \\
 \midrule
 70.86 & 55.41 & 47.56 & 42.25 & 27.61 & 51.72 \\ 
 \bfseries 62.27 & \bfseries 41.96 & \bfseries 32.10 & \bfseries 26.28 & \bfseries 20.66 & \bfseries 47.29 \\
 \midrule
 \bfseries 61.34 & 48.33 & 42.22 & 38.34 & 27.60 & 51.62 \\ 
 66.94 & \bfseries 46.07 & \bfseries 35.50 & \bfseries 29.74 & \bfseries 19.65 & \bfseries 43.00 \\ 
 \midrule
 35.64 & 23.62 & 19.61 & 17.43 & 16.28 & 34.64 \\ 
 \bfseries 34.67 & \bfseries 19.22 & \bfseries 13.68 & \bfseries 10.82 & \bfseries 5.26 & \bfseries 13.04 \\ 
 \midrule
 33.07 & 21.36 & 17.35 & 15.09 & 12.56 & 29.45\\ 
 \bfseries 29.48 & \bfseries 16.07 & \bfseries 11.47 & \bfseries 9.10 & \bfseries 5.10 & \bfseries 13.40 \\ 
 \bottomrule
 \end{tabular}
 
 \\
 \\
 \begin{tabular}{c}
    \multirow{3}{*}{\rotatebox{90}{Quarter res.}} \\ 
 \end{tabular}
 \begin{tabular}{l}
 \\
 \toprule
 \\
 Model \\
 \midrule
 LEAStereo \\
 LEAStereo (ft) \\
 \midrule
 CFNet \\
 CFNet (ft) \\
 \midrule
 RAFT-Stereo \\
 RAFT-Stereo (ft) \\
  \midrule
 CREStereo \\ 
 CREStereo (ft)  \\ 
 \bottomrule
 \end{tabular}
 
 &
 
 \begin{tabular}{rrrr | rr }
 \multicolumn{6}{c}{All pixels} \\
 \toprule
 bad-2 & bad-4 & bad-6 & bad-8 & MAE & RMSE \\
 $\downarrow$ (\%) & $\downarrow$ (\%) & $\downarrow$ (\%) & $\downarrow$ (\%) & $\downarrow$ (px.) & $\downarrow$ (px.) \\
 \midrule
 42.21 & 30.23 & 24.37 & 20.43 & 6.89 & 12.92 \\ 
 \bfseries 26.21 & \bfseries 16.13 & \bfseries 12.47 & \bfseries 10.46 & \bfseries 5.15 & \bfseries 11.80 \\
 \midrule
 38.31 & 29.53 & 24.70 & 21.34 & 6.89 & 12.89 \\ 
 \bfseries 29.64 & \bfseries 19.93 & \bfseries 15.59 & \bfseries 12.73 & \bfseries 4.78 & \bfseries 10.42 \\ 
 \midrule
 17.43 & 13.49 & 11.59 & 10.10 & 4.07 & 8.64 \\ 
 \bfseries 10.72 & \bfseries 6.79 & \bfseries 5.28 & \bfseries 3.87 & \bfseries 1.29 & \bfseries 3.13 \\ 
 \midrule
 15.13 & 10.70 & 8.91 & 7.57 & 3.15 & 7.40 \\ 
 \bfseries 8.99 & \bfseries 5.30 & \bfseries 3.98 & \bfseries 2.81 & \bfseries 1.25 & \bfseries 3.18 \\ 
 \bottomrule
 \end{tabular}
 \end{tabular}
 }
 
 \caption{\textbf{Results on the Booster Balanced benchmark after fine tuning.} We fine-tune several stereo networks using the Booster training split, processing quarter-resolution images. We evaluate on full-resolution ground truths or downsampled to quarter resolution.}
 \label{tab:ft_bal}
\end{table}

\begin{table*}[t]
\centering
\setlength{\tabcolsep}{3pt}
\scalebox{0.7}{
\begin{tabular}{cccccccc}
 && \multicolumn{2}{c}{Full Res.} & & \multicolumn{2}{c}{Quarter Res.}\\
 \begin{tabular}{c}
    \multirow{3}{*}{\rotatebox{90}{Pre Ft.}} \\ 
 \end{tabular}
 \begin{tabular}{l}
 \\
 \toprule
 \\
 Category \\
 \midrule
 All \\
 \midrule
 Class 0 \\
 Class 1 \\
 Class 2 \\
 Class 3 \\
 \bottomrule
 \end{tabular}
 &&
 \begin{tabular}{rrrr | rr }
 \multicolumn{6}{c}{All pixels} \\
 \toprule
 bad-2 & bad-4 & bad-6 & bad-8 & MAE & RMSE \\
 (\%) & (\%) & (\%) & (\%) & (px.) &  (px.) \\
 \midrule
 35.64 & 23.62 & 19.61 & 17.43 & 16.28 & 34.64 \\ 
 \midrule
 28.14 & 11.86 & 6.44 & 3.86 & 2.50 & 6.70 \\ 
 39.16 & 24.54 & 19.37 & 16.50 & 9.25 & 18.60 \\ 
 71.33 & 57.32 & 47.21 & 40.75 & 39.24 & 47.55 \\ 
 79.94 & 70.06 & 63.80 & 58.35 & 48.68 & 60.34 \\ 
 \bottomrule
 \end{tabular}
 &
 \begin{tabular}{rrrr | rr }
 \multicolumn{6}{c}{All pixels} \\
 \toprule
 bad-2 & bad-4 & bad-6 & bad-8 & MAE & RMSE \\
 (\%) & (\%) & (\%) & (\%) & (px.) &  (px.) \\
  \midrule
    33.07 & 21.36 & 17.35 & 15.09 & 12.56 & 29.45 \\
    \midrule     
    25.55 & 10.62 & 5.81 & 3.45 & 2.12 & 5.06 \\            
    36.31 & 21.38 & 15.95 & 12.82 & 5.62 & 12.13 \\           
    62.77 & 45.31 & 34.68 & 28.81 & 20.89 & 26.87 \\           
    79.50 & 69.29 & 62.08 & 56.19 & 53.34 & 66.73 \\ 
 \bottomrule
 \end{tabular}
 &&
 \begin{tabular}{rrrr | rr }
 \multicolumn{6}{c}{All pixels} \\
 \toprule
 bad-2 & bad-4 & bad-6 & bad-8 & MAE & RMSE \\
 $\downarrow$ (\%) & $\downarrow$ (\%) & $\downarrow$ (\%) & $\downarrow$ (\%) & $\downarrow$ (px.) & $\downarrow$ (px.) \\
 \midrule
 17.43 & 13.49 & 11.59 & 10.10 & 4.07 & 8.64 \\ 
 \midrule
 3.91 & 1.82 & 1.22 & 0.88 & 0.62 & 1.61 \\ 
 16.54 & 10.18 & 6.95 & 5.85 & 2.31 & 4.62 \\ 
 40.74 & 29.28 & 26.45 & 25.38 & 9.81 & 11.90 \\ 
 58.37 & 43.55 & 35.53 & 29.36 & 12.16 & 15.06 \\ 
 \bottomrule
 \end{tabular} 
 &
 \begin{tabular}{rrrr | rr }
 \multicolumn{6}{c}{All pixels} \\
 \toprule
 bad-2 & bad-4 & bad-6 & bad-8 & MAE & RMSE \\
 $\downarrow$ (\%) & $\downarrow$ (\%) & $\downarrow$ (\%) & $\downarrow$ (\%) & $\downarrow$ (px.) & $\downarrow$ (px.) \\
 \midrule
    15.13 & 10.70 & 8.91 & 7.57 & 3.15 & 7.40 \\
    \midrule            
    3.49 & 1.55 & 0.82 & 0.57 & 0.54 & 1.30 \\           
    12.93 & 6.92 & 4.19 & 3.41 & 1.42 & 3.12 \\           
    28.90 & 18.76 & 16.66 & 15.66 & 5.25 & 6.82 \\           
    56.25 & 40.42 & 33.55 & 28.26 & 13.34 & 16.73 \\
 \bottomrule
 \end{tabular}
 \\\\
 \begin{tabular}{c}
    \multirow{3}{*}{\rotatebox{90}{Post Ft.}} \\ 
 \end{tabular}
 \begin{tabular}{l}
 \\
 \toprule
 \\
 Category \\
 \midrule
 All \\
 \midrule
 Class 0 \\
 Class 1 \\
 Class 2 \\
 Class 3 \\
 \bottomrule
 \end{tabular}
 &&
 \begin{tabular}{rrrr | rr }
 \multicolumn{6}{c}{All pixels} \\
 \toprule
 bad-2 & bad-4 & bad-6 & bad-8 & MAE & RMSE \\
 (\%) & (\%) & (\%) & (\%) & (px.) &  (px.) \\
 \midrule
 34.67 & 19.22 & 13.68 & 10.82 & 5.26 & 13.04 \\ 
 \midrule
 30.94 & 14.85 & 8.77 & 5.85 & 2.71 & 6.72 \\ 
 37.20 & 18.86 & 12.61 & 9.24 & 4.54 & 11.41 \\ 
 56.26 & 35.72 & 26.58 & 21.03 & 7.15 & 11.01 \\ 
 64.64 & 46.79 & 36.25 & 29.26 & 12.31 & 18.15 \\ 
 \bottomrule
 \end{tabular}
 &
 \begin{tabular}{rrrr | rr }
 \multicolumn{6}{c}{All pixels} \\
 \toprule
 bad-2 & bad-4 & bad-6 & bad-8 & MAE & RMSE \\
 $\downarrow$ (\%) & $\downarrow$ (\%) & $\downarrow$ (\%) & $\downarrow$ (\%) & $\downarrow$ (px.) & $\downarrow$ (px.) \\
 \midrule
    29.48 & 16.07 & 11.47 & 9.10 & 5.10 & 13.40 \\ 
    \midrule           
    24.38 & 11.09 & 7.00 & 5.02 & 3.09 & 6.51 \\ 
    30.19 & 13.82 & 8.46 & 6.21 & 3.30 & 9.14 \\ 
    53.00 & 31.48 & 21.09 & 15.86 & 4.79 & 7.76 \\ 
    67.40 & 50.33 & 40.02 & 32.94 & 18.93 & 27.03 \\ 
 \bottomrule
 \end{tabular}
 &&
 \begin{tabular}{rrrr | rr }
 \multicolumn{6}{c}{All pixels} \\
 \toprule
 bad-2 & bad-4 & bad-6 & bad-8 & MAE & RMSE \\
 $\downarrow$ (\%) & $\downarrow$ (\%) & $\downarrow$ (\%) & $\downarrow$ (\%) & $\downarrow$ (px.) & $\downarrow$ (px.) \\
 \midrule
 10.72 & 6.79 & 5.28 & 3.87 & 1.29 & 3.13 \\ 
 \midrule
 5.73 & 1.68 & 1.13 & 0.82 & 0.65 & 1.45 \\ 
 9.09 & 5.08 & 4.04 & 3.26 & 1.10 & 2.65 \\ 
 20.83 & 12.02 & 8.31 & 5.73 & 1.76 & 2.64 \\ 
 29.26 & 17.86 & 13.07 & 8.29 & 3.05 & 4.42 \\ 
 \bottomrule
 \end{tabular} 
 &
 \begin{tabular}{rrrr | rr }
 \multicolumn{6}{c}{All pixels} \\
 \toprule
 bad-2 & bad-4 & bad-6 & bad-8 & MAE & RMSE \\
 $\downarrow$ (\%) & $\downarrow$ (\%) & $\downarrow$ (\%) & $\downarrow$ (\%) & $\downarrow$ (px.) & $\downarrow$ (px.) \\
 \midrule
    8.99 & 5.30 & 3.98 & 2.81 & 1.25 & 3.18 \\ 
    \midrule            
    4.88 & 1.55 & 1.17 & 1.00 & 0.74 & 1.37 \\ 
    6.01 & 3.36 & 2.40 & 1.73 & 0.78 & 2.02 \\ 
    15.65 & 6.04 & 3.20 & 1.33 & 1.16 & 1.80 \\ 
    32.86 & 20.63 & 15.29 & 11.11 & 4.70 & 6.62 \\ 
 \bottomrule
 \end{tabular}
 \\\\
 && RAFT-Stereo \cite{lipson2021raft} & CREStereo \cite{li2022practical }& & RAFT-Stereo \cite{lipson2021raft} & CREStereo \cite{li2022practical} \\
 \end{tabular}
 }
 \caption{\textbf{Results on the Booster Balanced benchmark -- material segmentation.} We run RAFT-Stereo \cite{lipson2021raft} and CREStereo \cite{li2022practical} on the Booster Balanced test split. Top row: results obtained using official weights. Bottom row: results after fine-tuning on the Booster Balanced training split. We process quarter-resolution images. We evaluate on full-resolution ground truths, or by downsampling them to quarter resolution.}
 \label{tab:segmentation_balanced}
\end{table*}

\textbf{Evaluation metrics.} According to the testing split, stereo or mono, we select a set of metrics appropriate to the specific scenario. 
For Test Stereo, we take inspiration from Middlebury 2014 \cite{scharstein2014high} and compute the percentage of pixels having disparity errors larger than a threshold $\tau$ (bad-$\tau$). Since our ground-truth maps are inferred at half the input resolution, we assume 2 pixels as the lowest threshold. Additionally, we compute bad-4, bad-6, and bad-8 error rates, given the very high resolution featured by our dataset. Finally, we also measure the Mean Absolute Error (MAE) and Root Mean Squared Error (RMSE). 
For Test Mono, we instead follow Eigen \etal \cite{eigen2014depth} to compute metrics. In particular, we compute the absolute error relative to the ground value (ABS Rel.), and the percentage of pixels having the maximum between the prediction/ground-truth and ground-truth/prediction ratios lower than a threshold ($\delta_i$, with $i$ being 1.05, 1.15, and 1.25). Moreover, also in this case, we estimate the mean absolute error (MAE) and Root Mean Squared Error (RMSE).
All the metrics introduced so far are computed on any valid pixel (\textit{All}), or on pixels belonging to a specific material class $i$ (\textit{Class $i$}), to evaluate the impact of non-Lambertian objects. For the balanced setup, we also evaluate on non-occluded pixels only (\textit{Cons}) -- \ie those considered consistent by the left-right check step performed by our annotation pipeline -- producing the occlusion mask shown as the bottom image in the third column of Fig. \ref{fig:booster_sample}.
For any metrics considered for Test Stereo, the lower, the better -- annotated with $\downarrow$ in tables. The same applies to metrics used for Test Mono except for $\delta_i$, resulting in the higher, the better -- with $\uparrow$ being reported in tables.

\section{Experiments}

We now report the outcome of our experiments on Booster, respectively, on the Balanced Stereo, Unbalanced Stereo, and Monocular benchmarks.

\subsection{Balanced Stereo Benchmark} 

We first focus on the Balanced benchmark, \ie the one closer to classical stereo datasets in terms of setup. Over it, we run a set of different experiments enabled by the peculiar properties of Booster.

\textbf{Off-the-shelf deep networks.} 
We run state-of-the-art stereo architectures on Test Stereo to measure their effectiveness in facing the challenges posed by Booster.
We select the top-performing models from the Middlebury 2014 benchmark, \ie the most challenging among existing datasets, and evaluate them if their weights are publicly available.
This constraints us to selecting HSMNet \cite{yang2019hierarchical}, LEAStereo \cite{cheng2020hierarchical}, CFNet \cite{shen2021cfnet}, RAFT-Stereo \cite{lipson2021raft}, Neural Disparity Refinement \cite{aleotti2021neural}, and CREStereo \cite{li2022practical}.
RAFT-Stereo is employed with 20 or 32 GRU iterations, indicated in \cref{tab:stereo_tournament} and \cref{tab:unbalanced} as RAFT-Stereo$^*$ and RAFT-Stereo, respectively.
As a reference, we also include the classical Semi-Global Matching (SGM) algorithm \cite{hirschmuller2007stereo}, using the OpenCV implementation, and MC-CNN \cite{zbontar2016stereo} -- \ie the first attempt using deep learning for stereo, using the original Lua implementation\footnote{\url{https://github.com/jzbontar/mc-cnn}}. For the latter, we select its fast variant because of memory constraints.
We collect the outcome of this evaluation in \cref{tab:stereo_tournament}. 
On top, we compare the predictions by any network with the full resolution ground-truth maps on \textit{All} (left) and \textit{Cons} (right) pixels, respectively. 
Depending on the memory requirements of the method, input images are processed at the original resolution (F) or scaled to half (H) or quarter (Q) resolution, always with a single NVIDIA 3090 RTX GPU. 
As we can notice, most models can process only Q resolution images due to memory requirement constraints. Consequently, predicted disparity maps are upsampled with nearest-neighbor interpolation to compare them with the full resolution ground-truth maps, with disparity values multiplied by the same upsampling factor. 
At the bottom of Tab. \ref{tab:stereo_tournament}, the evaluation is carried out using ground-truth disparity maps downsampled to a quarter (Q) of its original resolution. 

It is evident how any stereo networks struggle with our benchmark.
On the one hand, error rates computed on quarter-resolution ground-truth maps are generally much lower than those measured at full resolution, confirming that high-resolution represents a challenge to existing models.
On the other hand, the former results are still far behind those observed on other popular datasets \cite{Geiger2012CVPR,Menze2015CVPR,scharstein2014high,schoeps2017cvpr}, e.g., 33.07 vs 8.13 bad-2 for CREStereo on all-pixels in the Middlebury benchmark. This evidence confirms that high resolution is not the only challenge in our benchmark, but networks also struggle due to the presence of transparent and specular surfaces in our dataset.
We have similar scores comparing errors calculated on \textit{All} and \textit{Cons} pixels. This fact highlights that occlusions are not the main difficulties in our benchmark.
Finally, similar to the Middlebury benchmark, we can notice that CREStereo achieves the best results. Moreover, the superior accuracy of CREStereo compared to RAFT-Stereo proves the absence of any bias in favor of the latter, despite ground-truth labels having been obtained by processing the textured images with our deep space-time framework based on RAFT-Stereo. 

\begin{figure*}[t]
    \centering
    \renewcommand{\tabcolsep}{1pt}
    \begin{tabular}{cccccccccc}
        \tiny \textit{RGB \& GT} & \tiny \textit{MC-CNN} \cite{zbontar2016stereo} & \tiny \textit{LEAStereo} \cite{cheng2020hierarchical} & \tiny \textit{CFNet} \cite{shen2021cfnet} & \tiny \textit{HSMNet} \cite{yang2019hierarchical} & \tiny \textit{Neural Ref.} \cite{aleotti2021neural} & \tiny \textit{RAFT-Stereo} \cite{lipson2021raft} & \tiny \textit{CREStereo} \cite{li2022practical} & \tiny \textit{RAFT-Stereo (ft)} \cite{lipson2021raft} & \tiny \textit{CREStereo (ft)} \cite{li2022practical} \\
        \tiny Input Res. & \tiny Q & \tiny Q & \tiny Q & \tiny H & \tiny H & \tiny Q & \tiny Q & \tiny Q & \tiny Q \\    
        
        \includegraphics[width=0.095\textwidth]{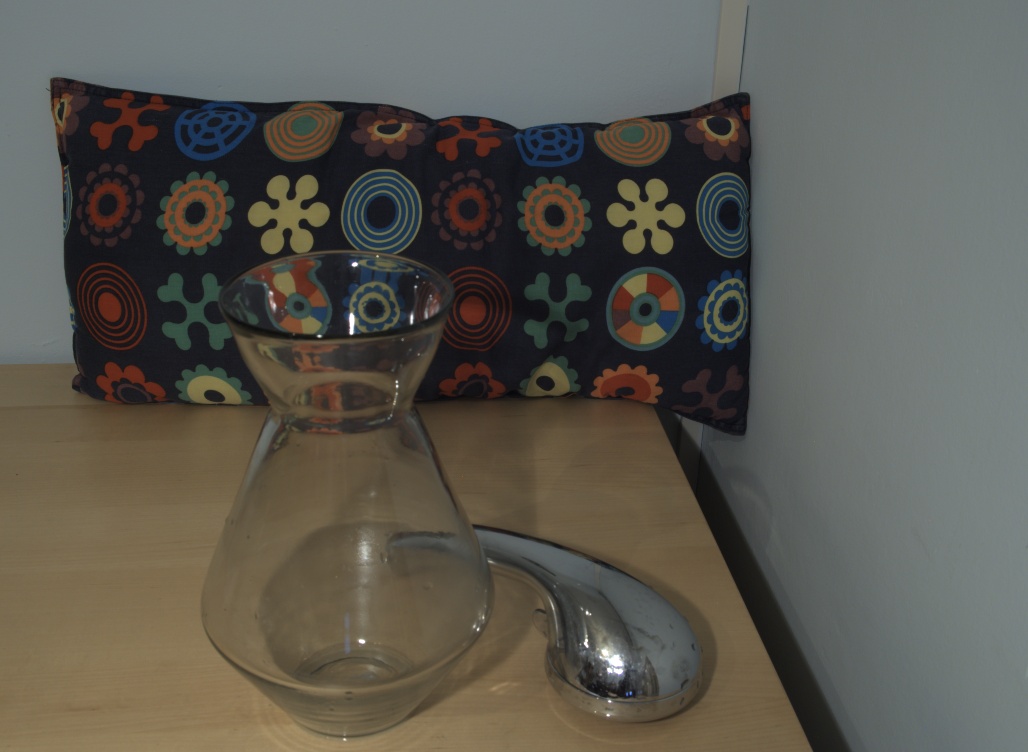} &
        \includegraphics[width=0.095\textwidth]{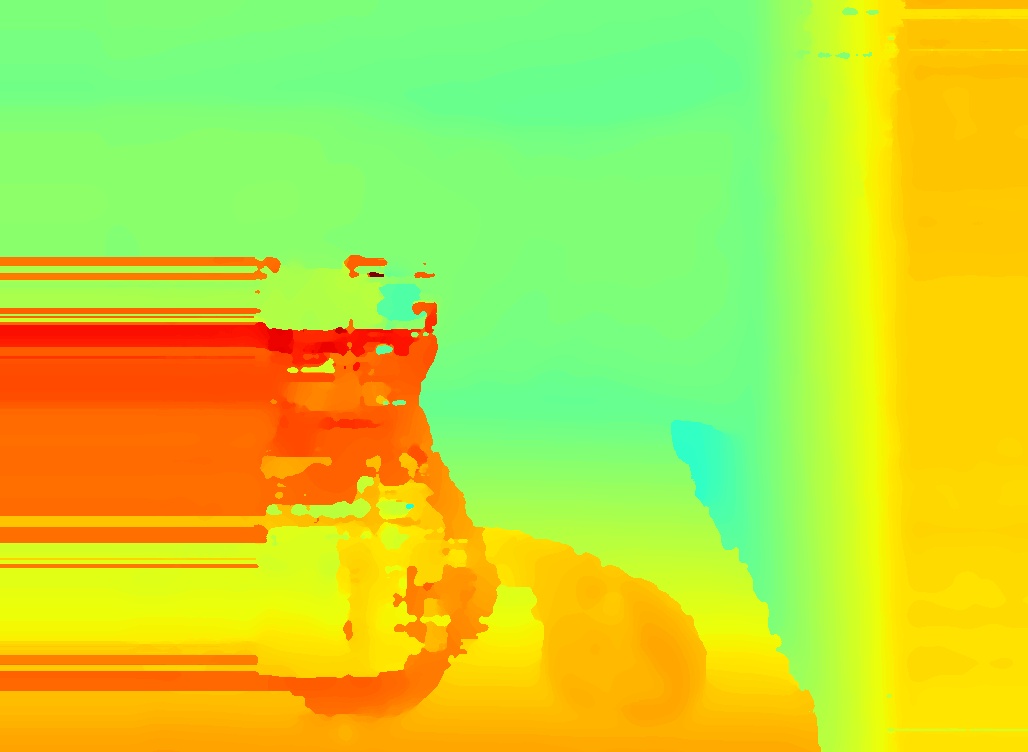} &
        \includegraphics[width=0.095\textwidth]{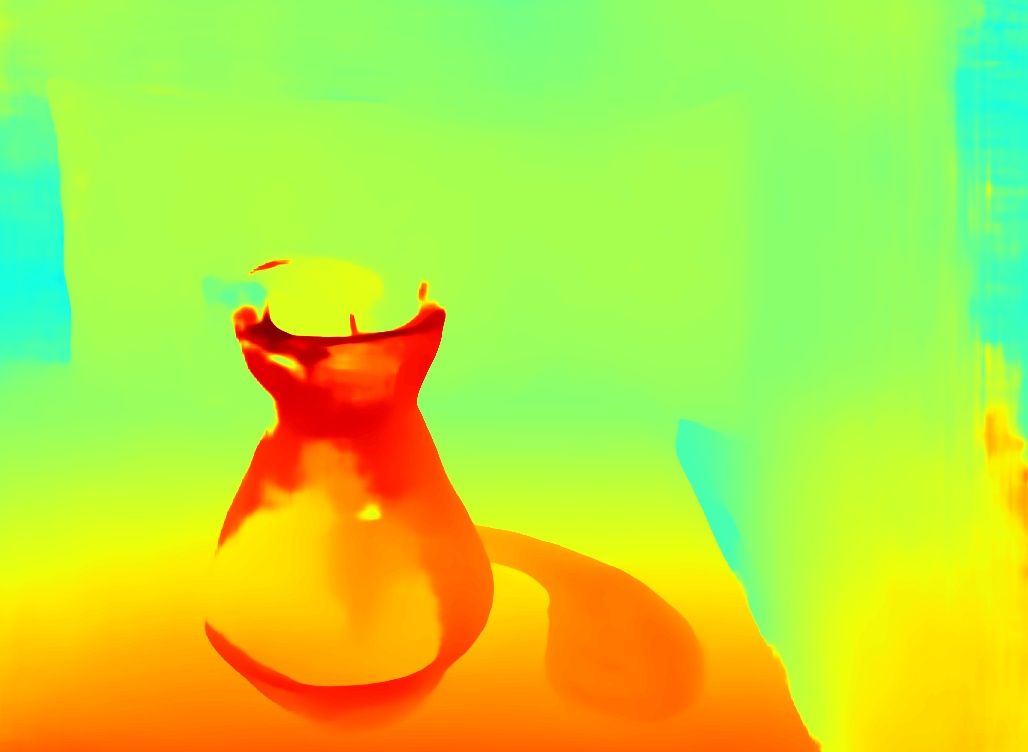} &
        \includegraphics[width=0.095\textwidth]{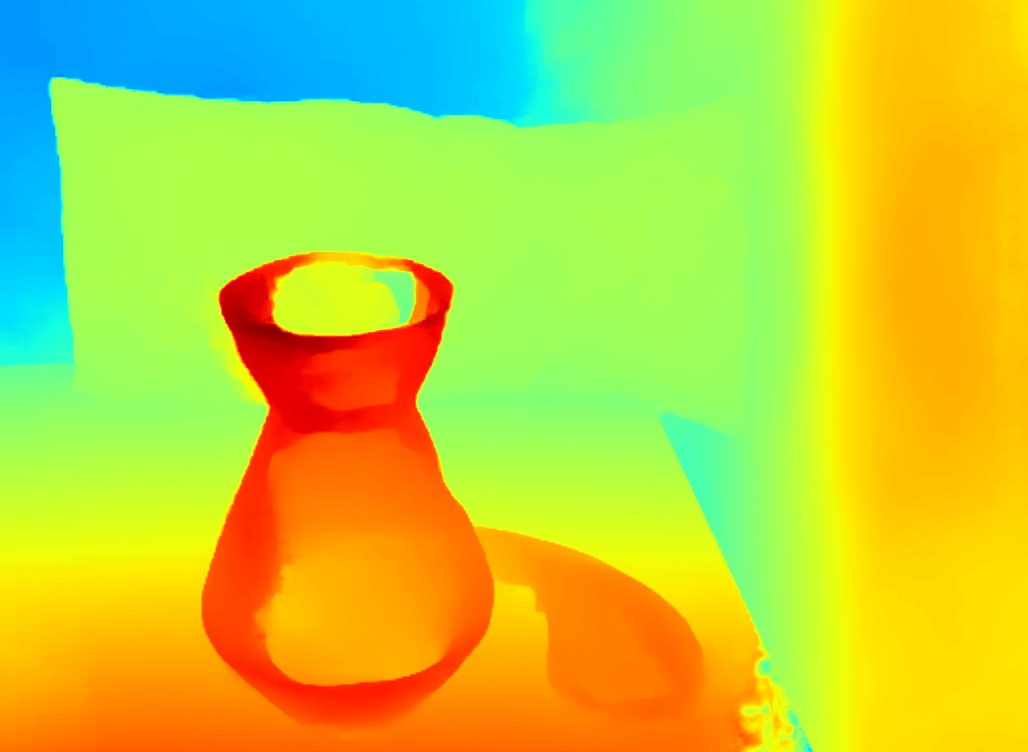} &
        \includegraphics[width=0.095\textwidth]{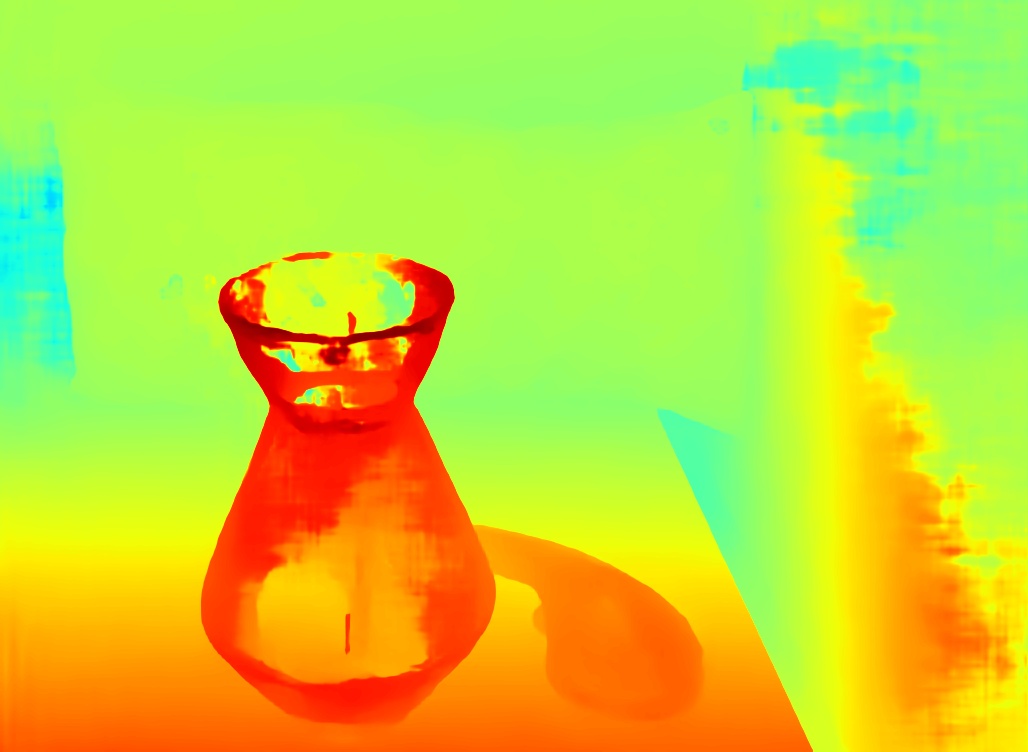} &
        \includegraphics[width=0.095\textwidth]{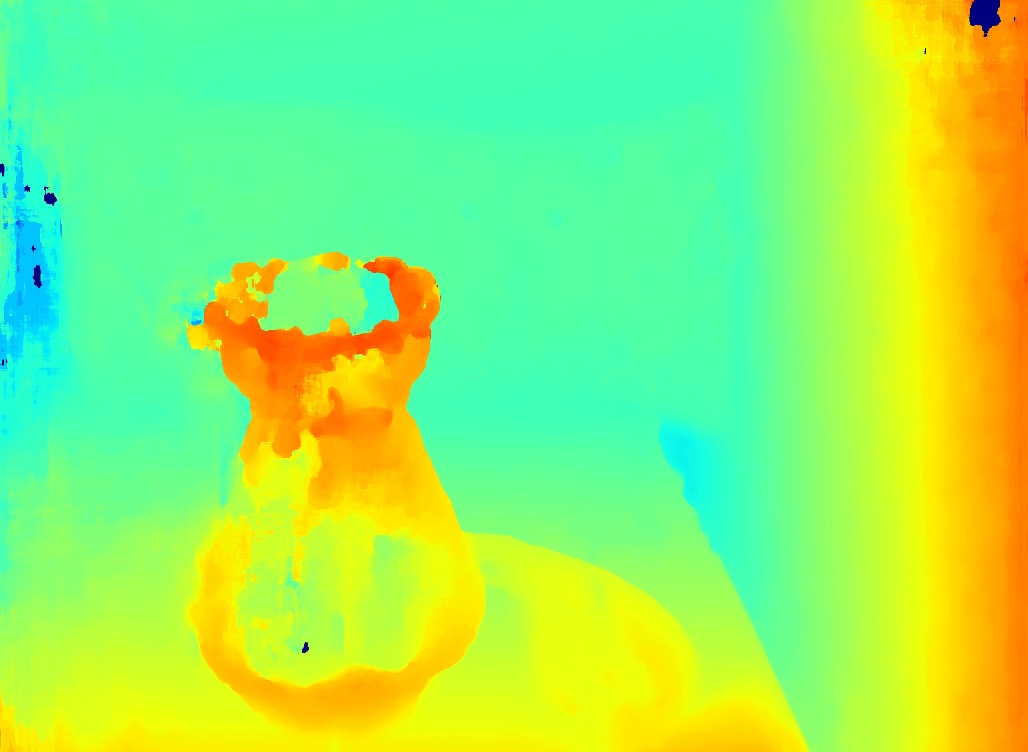} &
        \includegraphics[width=0.095\textwidth]{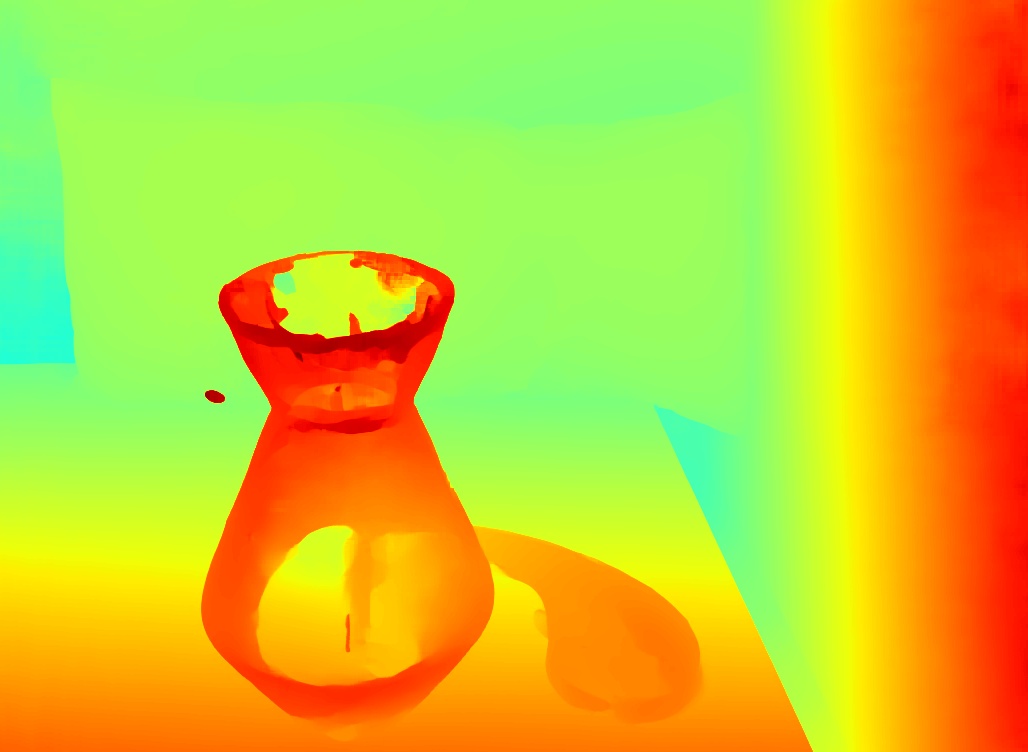} &
        \includegraphics[width=0.095\textwidth]{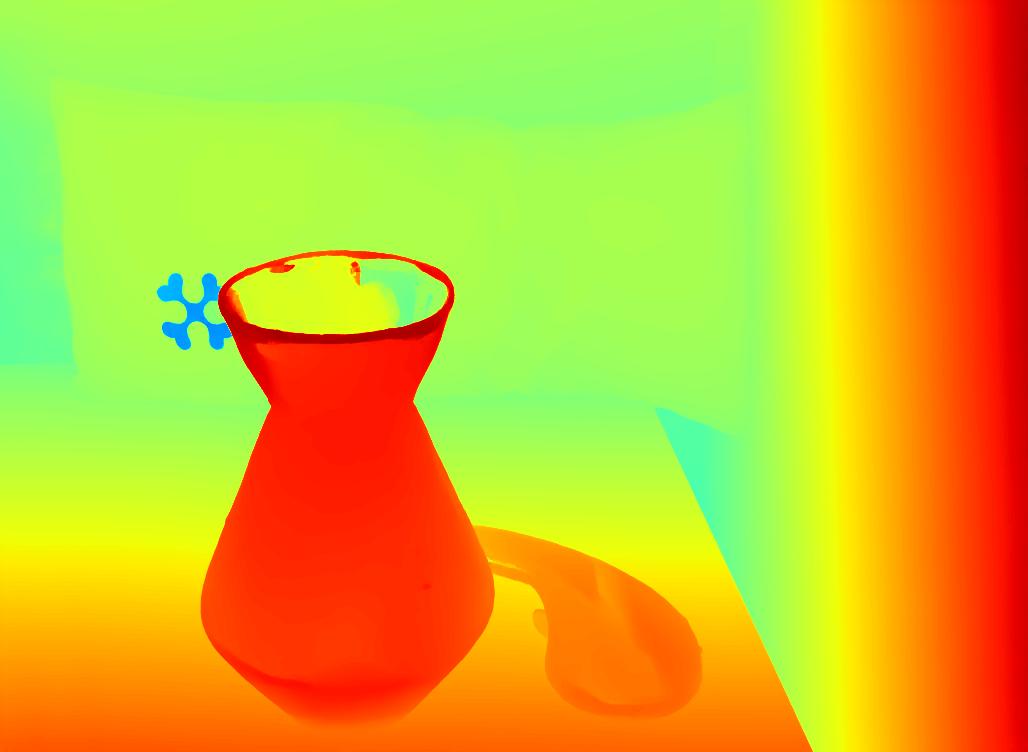} &
        \includegraphics[width=0.095\textwidth]{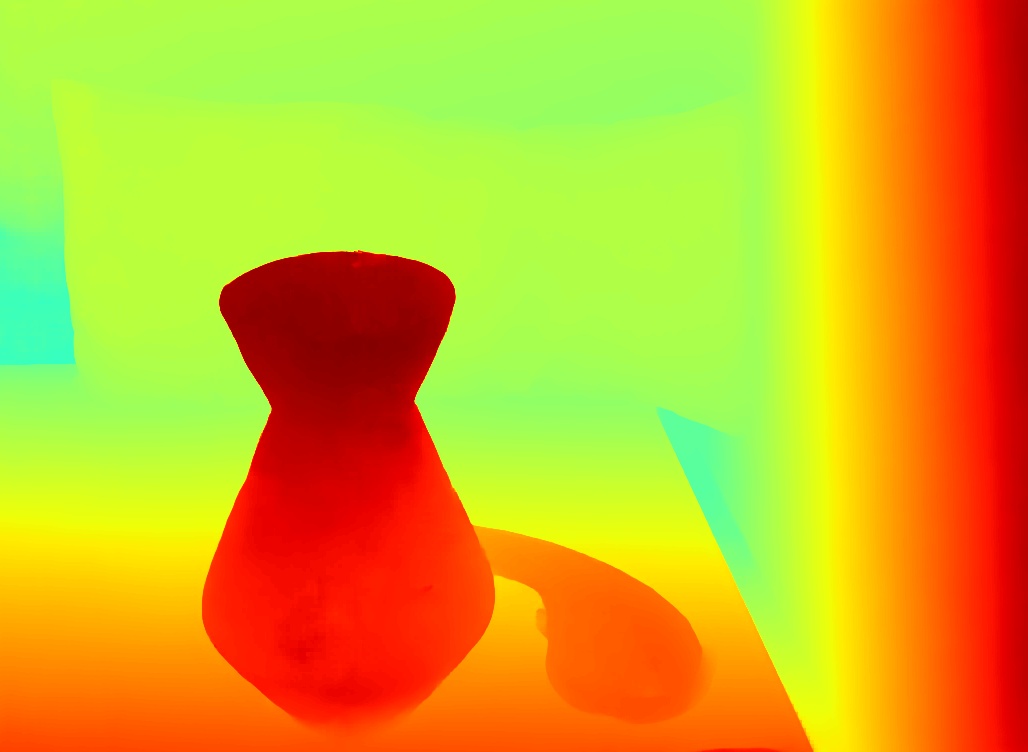} &
        \includegraphics[width=0.095\textwidth]{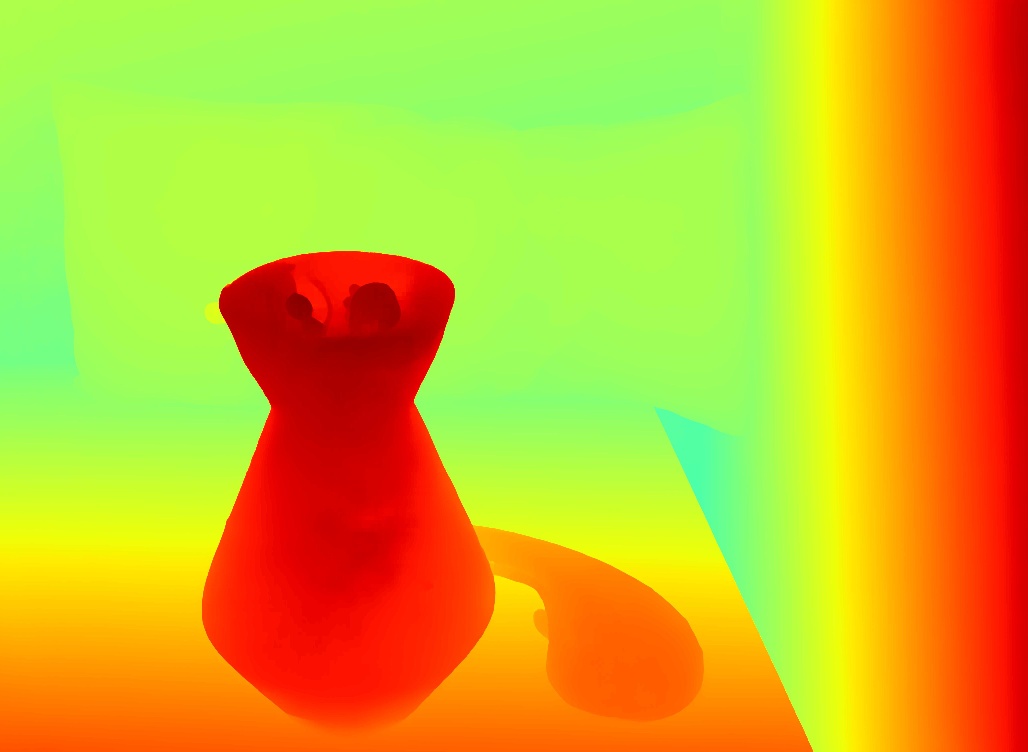} \\
        \includegraphics[width=0.095\textwidth]{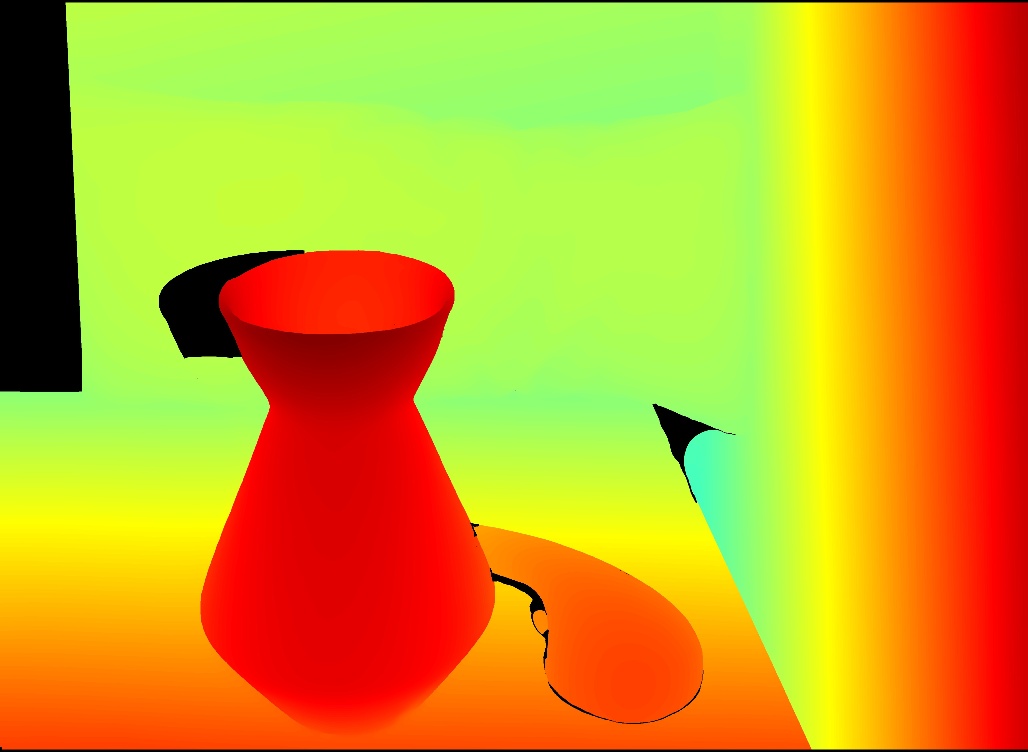} &
        \includegraphics[width=0.095\textwidth]{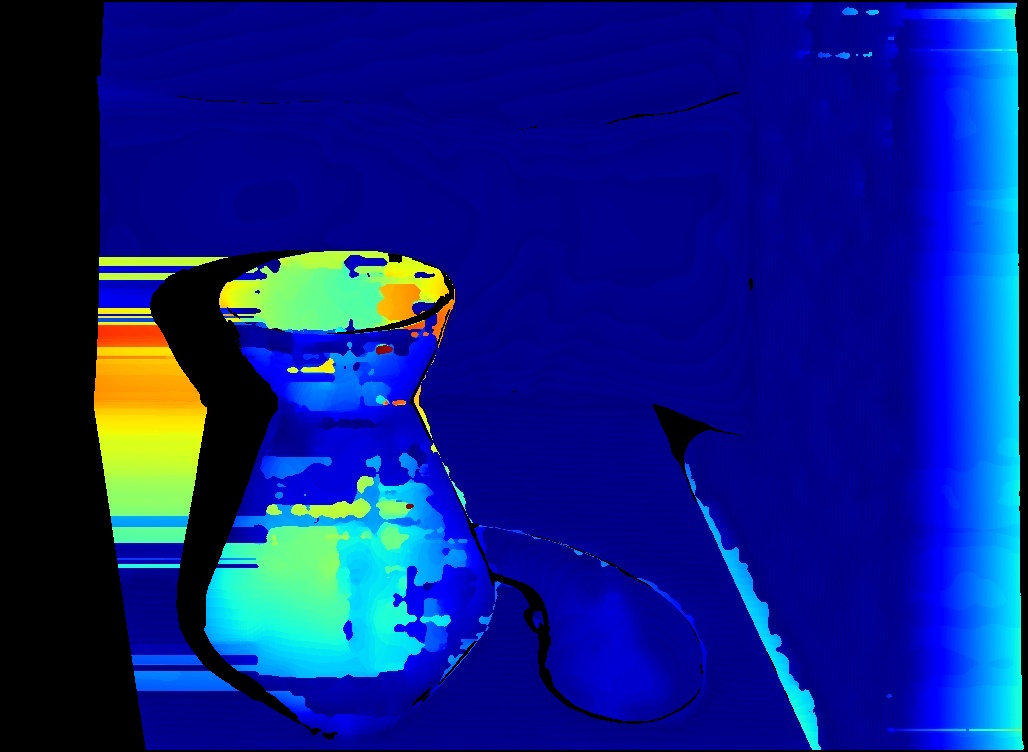} &
        \includegraphics[width=0.095\textwidth]{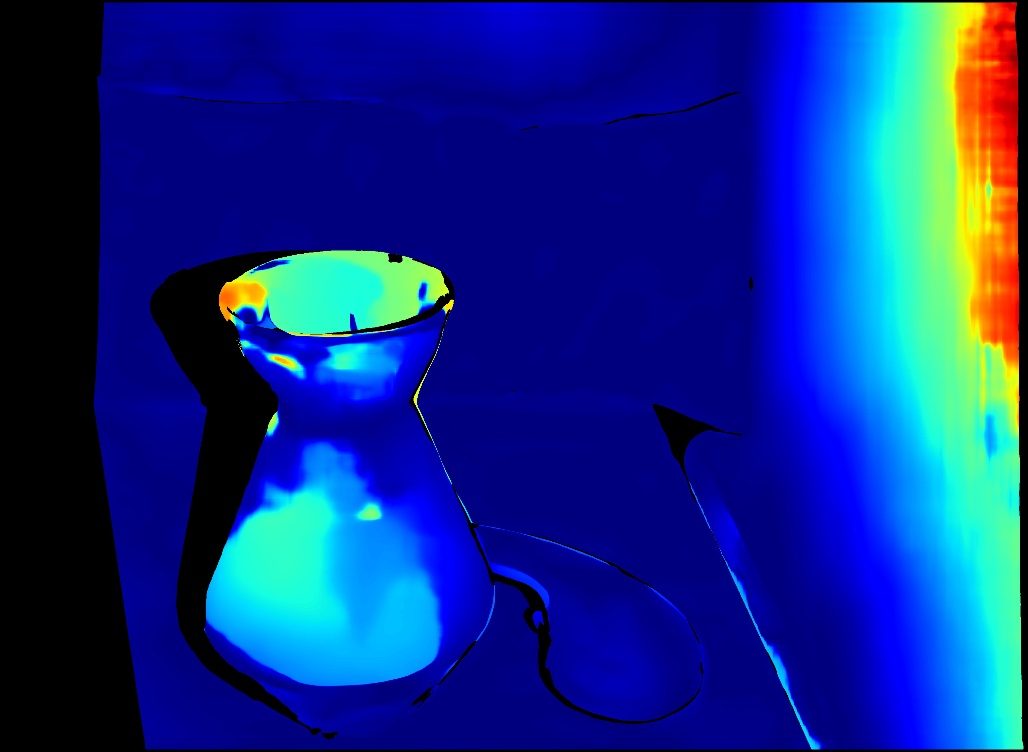} &
        \includegraphics[width=0.095\textwidth]{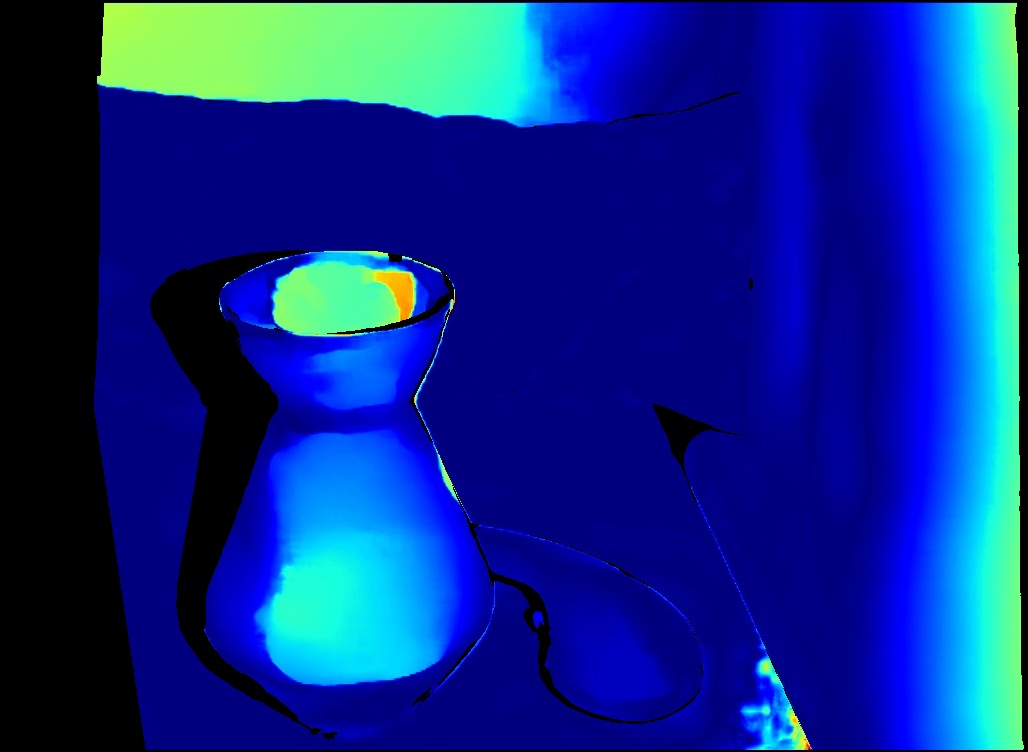} &
        \includegraphics[width=0.095\textwidth]{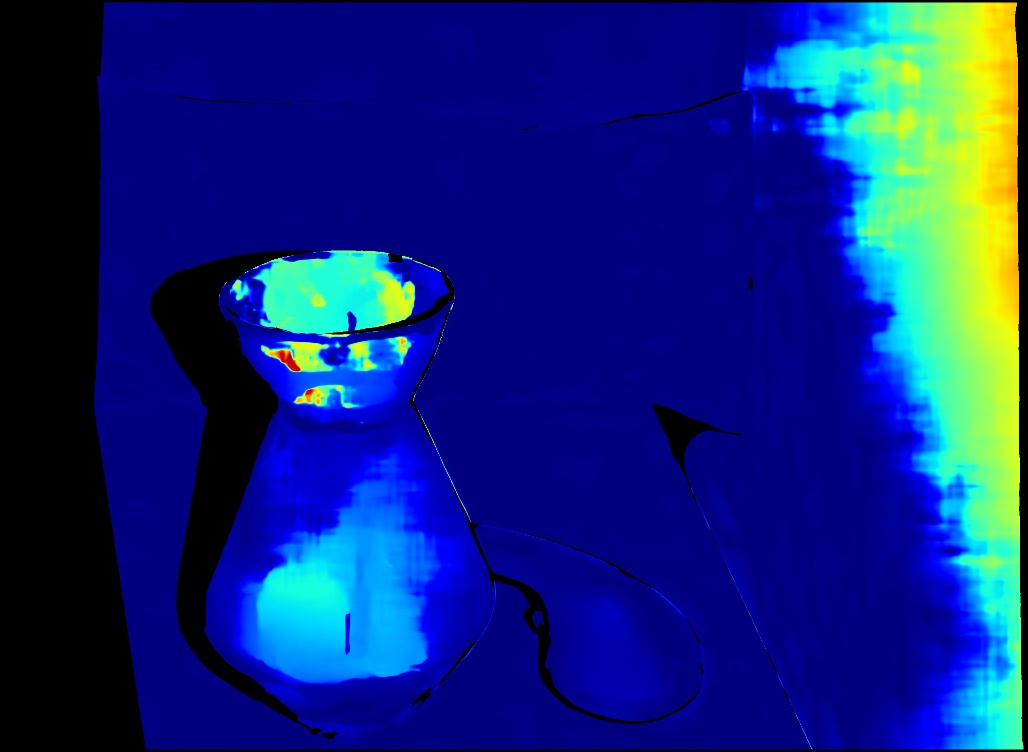} &
        \includegraphics[width=0.095\textwidth]{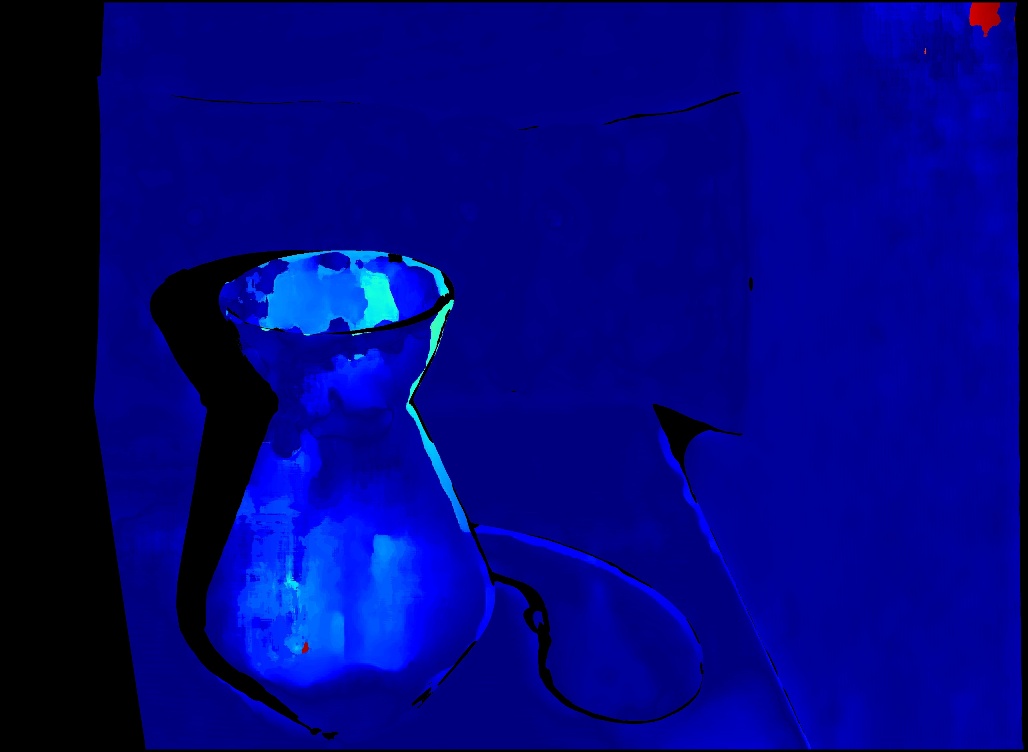} &
        \includegraphics[width=0.095\textwidth]{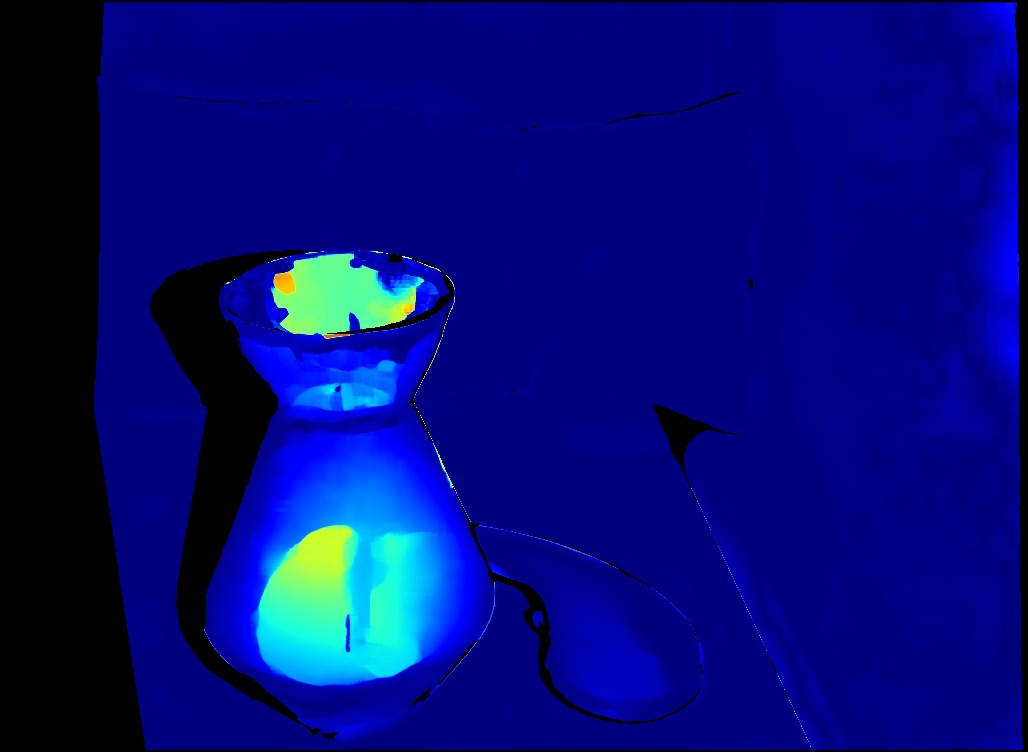} &
        \includegraphics[width=0.095\textwidth]{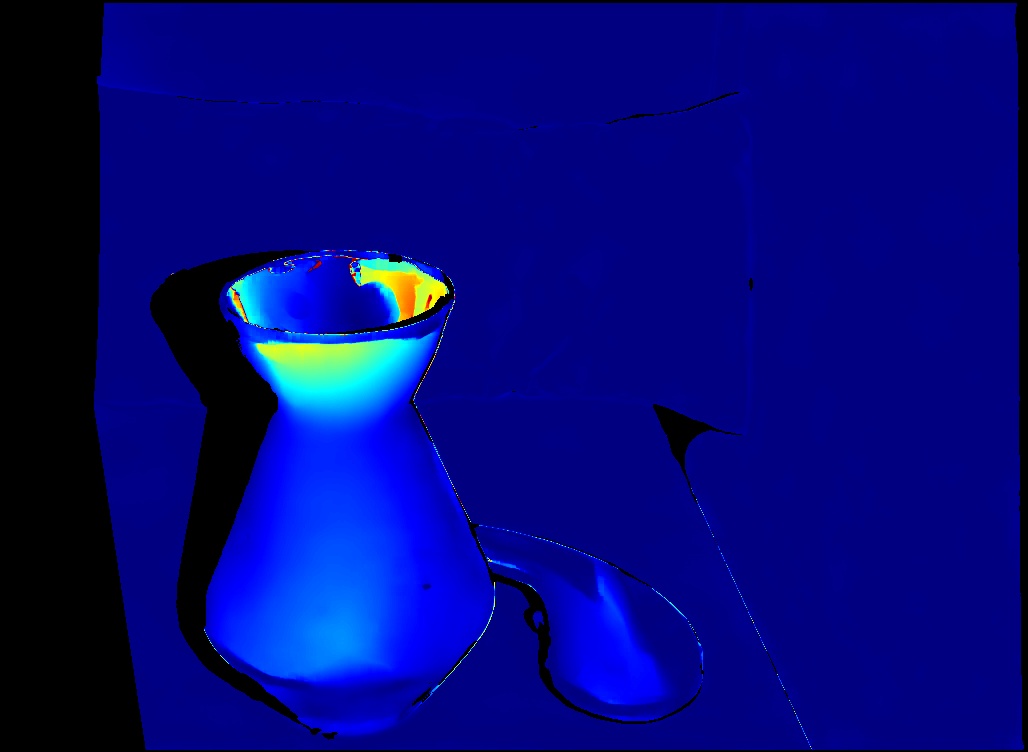} &
        \includegraphics[width=0.095\textwidth]{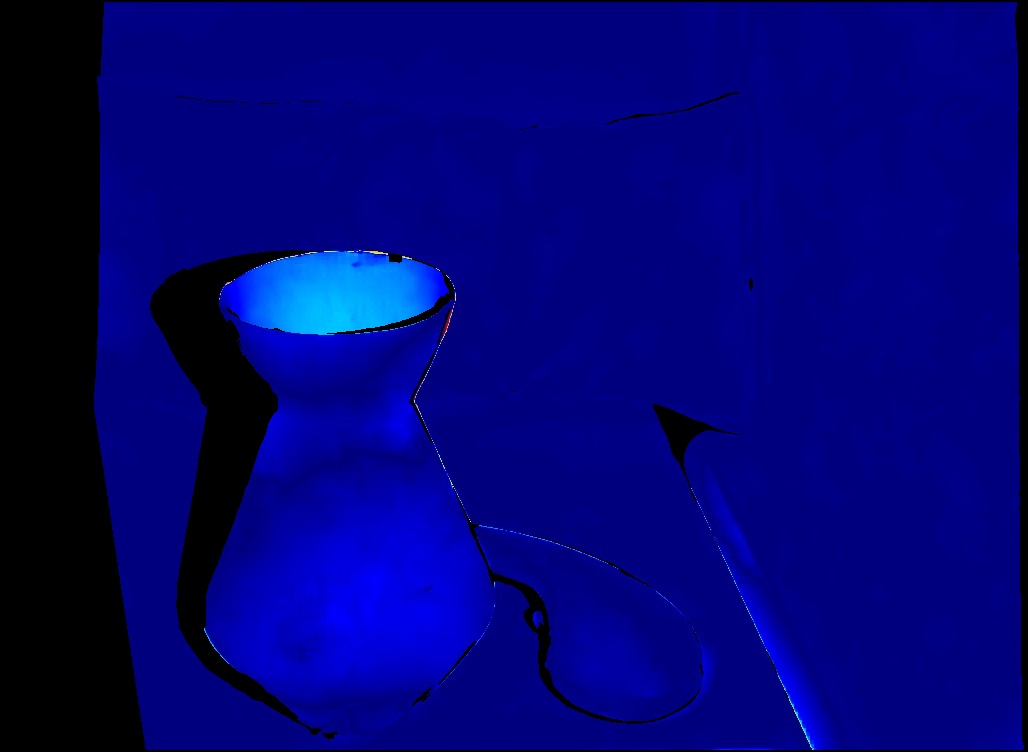} &
        \includegraphics[width=0.095\textwidth]{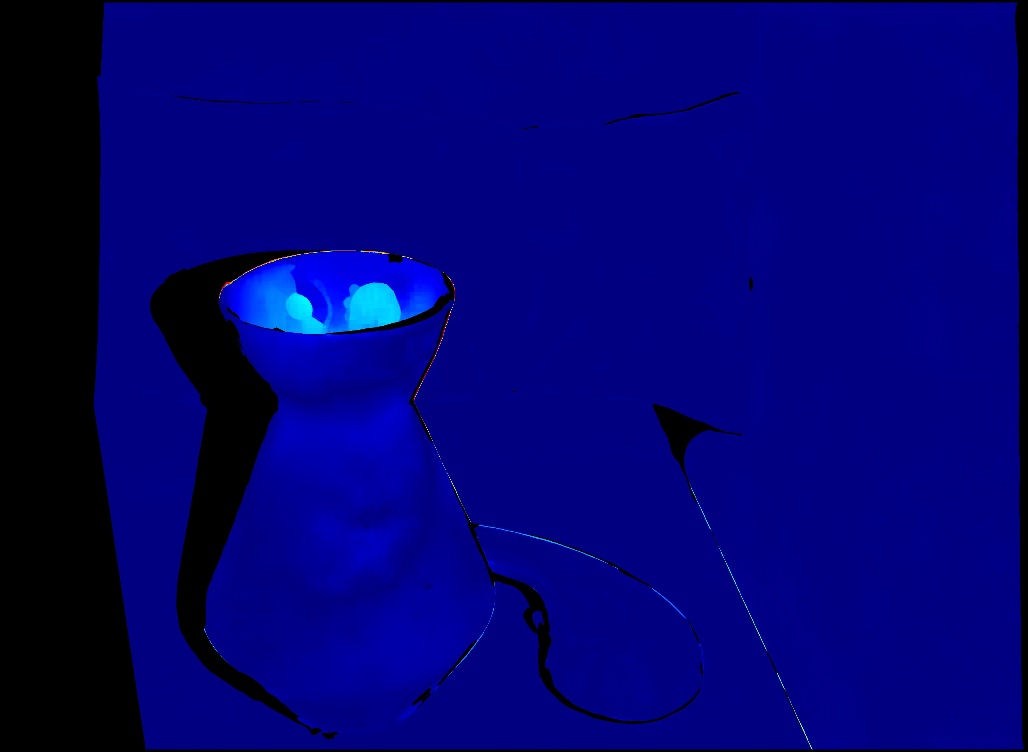} \\
    \end{tabular}
    \vspace{-0.25cm}\caption{\textbf{Qualitative results on the Balanced stereo test split.} We show the reference image (top) and the ground-truth map (bottom) on the leftmost column, followed by disparity (top) and error maps (bottom) for the deep models evaluated in our benchmark. (ft) denotes models fine-tuned on Booster training data.}
    \label{fig:balanced_qualitatives}
\end{figure*}

\textbf{Finetuning by the Booster training data.}
We fine-tune RAFT-Stereo \cite{lipson2021raft}, LEAStereo \cite{cheng2020hierarchical}, CFNet \cite{shen2021cfnet},
and CREStereo \cite{li2022practical} on the Booster training set to prove that the availability of annotated scenes can compensate for most errors due to the open challenges addressed in this paper.
Purposely, we run 100 epochs of training, with batches of two 884$\times$456 crops, extracted from images randomly resized to half or quarter of the original resolution, using the optimization procedure from RAFT-Stereo \cite{lipson2021raft} and initial learning rate set to 1e-5. We augment the Booster training set by adding images from the Middlebury 2014 dataset.
In \cref{tab:ft_bal} we report results at full and quarter resolution on \textit{All} pixels. Fine-tuning effectively improves the performance of all networks over the test split, confirming that the annotations provided with our dataset help to address the open challenges highlighted in the paper. However, network errors are still much higher than those of other benchmarks, highlighting the need for further investigations into this problem. 
In \cref{fig:balanced_qualitatives}, we provide some qualitative results on a sample from test stereo split, obtained by the networks evaluated in \cref{tab:stereo_tournament}, with the two rightmost columns showing predictions by fine-tuned RAFT-Stereo and CREStereo models. We highlight how the two models can handle much better the transparent object depicted in the scene thanks to the fine-tuning carried out on Booster. 

\textbf{Evaluation on challenging regions.} 
We now evaluate the accuracy of the networks in regions of increasing difficulty levels by using the material segmentation masks provided in our dataset.
We select the top-performing networks from the previous evaluation, \ie RAFT-Stereo, and CREStereo, and evaluate them on subsets of pixels defined by our manually annotated masks, and report results in \cref{tab:segmentation_balanced}. In the top tables, we show the results of the two networks before finetuning, while at the bottom we report results after finetuning. For each table, we also recall the results achieved on all valid pixels as a reference.
First, we point out that errors are lower when focusing on the least challenging material (class 0), and comparable with those of existing benchmarks  \cite{scharstein2014high} -- in particular, by evaluating quarter resolution ground-truth maps, as in the four rightmost tables. Then, errors increase when going towards more challenging classes. This fact confirms both our claims on the open challenges in deep stereo, and the significance of our segmentation masks.
In the four bottom tables, we can notice how, after fine-tuning, 
both networks achieve lower errors, independently of the resolution. 
More specifically, the metrics improve significantly for the most challenging materials, sometimes at the cost of a minimal decrease in accuracy within the simpler regions -- as on Class 0 for RAFT-Stereo. 
Even though our experiments suggest that the availability of more annotated data can help to handle transparent/specular surfaces better, the accuracy is still far from what is achieved on opaque materials. This evidence highlights that the open challenge represented by non-Lambertian objects for stereo is still unsolved. We hope that the availability of Booster will foster further progresses on this track.

\begin{table}[t]
\centering
\scalebox{0.75}{
\begin{tabular}{ccc}

 \begin{tabular}{l}
 \\
 \toprule
 \\
 Model \\
 \midrule
 SGM \cite{hirschmuller2007stereo} \\
 MC-CNN \cite{zbontar2016stereo} \\
 LEAStereo \cite{cheng2020hierarchical} \\
 CFNet \cite{shen2021cfnet} \\
 HSMNet \cite{yang2019hierarchical} $\dag$ \\
 SGM+Neural Ref. \cite{aleotti2021neural} $\dag$ \\
 RAFT-Stereo$^*$ \cite{lipson2021raft} \\
 RAFT-Stereo \cite{lipson2021raft} \\
 CREStereo \cite{li2022practical} \\
 \bottomrule
 \end{tabular}
 &
 \begin{tabular}{rrrr | rr }
 \multicolumn{6}{c}{All pixels} \\
 \toprule
 bad-2 & bad-4 & bad-6 & bad-8 & MAE & RMSE \\
 $\downarrow$ (\%) & $\downarrow$ (\%) & $\downarrow$ (\%) & $\downarrow$ (\%) & $\downarrow$ (px.) & $\downarrow$ (px.) \\
 \midrule
 78.47 & 62.74 & 52.62 & 45.97 & 42.63 & 97.62 \\ 
 86.30 & 68.67 & 54.20 & 44.78 & 23.64 & 45.46 \\ 
 74.31 & 57.70 & 47.11 & 39.88 & 17.68 & 31.29\\ 
 70.22 & 53.20 & 43.61 & 37.10 & 16.19 & 28.78 \\ 
 63.20 & 43.22 & 32.87 & 26.55 & 11.96 & 22.82 \\ 
 70.90 & 52.15 & 41.71 & 35.40 & 24.27 & 52.52 \\ 
 55.96 &  36.81 &  27.87 &  22.33 &  9.86 &  19.36 \\ 
 \bfseries 54.47 & 34.97 & 25.63 & 20.18 & 9.52 & 18.43 \\ 
 56.00 & \bfseries 34.81 & \bfseries 24.26 & \bfseries 18.65 & \bfseries 8.27 & \bfseries 15.89 \\
 \bottomrule
 \end{tabular}
 
 \end{tabular}
 }
 \caption{\textbf{Results on Booster Unbalanced benchmark.} We run stereo networks, using weights made available by their authors. We evaluate on full-resolution ground-truth maps. $\dag$ denotes images being resized to half the reference resolution (about 6 Mpx). Other networks process images at the lowest camera resolution.}
 \label{tab:unbalanced}
\end{table}

\subsection{Unbalanced Stereo Benchmark}
We now consider the Unbalanced benchmark and focus on this peculiar setting widely available on mobile phones.


\textbf{Off-the-shelf deep networks.} 
In Tab. \ref{tab:unbalanced}, we evaluate the same stereo methods considered in the previous experiments on the Unbalanced testing split.
For most methods, we follow the baseline approach defined in \cite{aleotti2021neural} and downsample the high-resolution reference image to the same resolution as the second image. The only exception is HSMNet, specifically designed to handle high-resolution images. In this case, we upsample images to full resolution.
We point out that the baseline length of the Unbalance setup is shorter than the Balanced one, making the results not directly comparable with those in \cref{tab:stereo_tournament}. Moreover, as smaller baselines correspond to smaller search ranges, finding corresponding pixels is easier. Nonetheless, the errors are larger than those in the Balanced split, highlighting the difficulty of this scenario. 
The networks rank similarly to what we observed on the balanced split, with CREStereo obtaining the best results with almost all metrics except for bad-2, which is comparable with RAFT-Stereo.

\begin{table}[t]
\centering
\scalebox{0.75}{
\begin{tabular}{ccc}
 \begin{tabular}{l}
 \\
 \toprule
 \\
 Model \\
 \midrule
 LEAStereo \\
 LEAStereo (ft) \\
 \midrule
 CFNet \\
 CFNet (ft) \\
 \midrule
 RAFT-Stereo \\
 RAFT-Stereo (ft) \\
 \midrule
 CRESTereo \\
 CRESTereo (ft) \\
 \bottomrule
 \end{tabular}
 &
 \begin{tabular}{rrrr | rr }
 \multicolumn{6}{c}{All pixels} \\
 \toprule
 bad-2 & bad-4 & bad-6 & bad-8 & MAE & RMSE \\
 $\downarrow$ (\%) & $\downarrow$ (\%) & $\downarrow$ (\%) & $\downarrow$ (\%) & $\downarrow$ (px.) & $\downarrow$ (px.) \\
 \midrule
 74.31 & 57.70 & 47.11 & 39.88 & 17.68 & 31.29 \\ 
 \bfseries 67.96 & \bfseries 44.90 & \bfseries 32.86 & \bfseries 26.38 & \bfseries 14.34 & \bfseries 29.27 \\
 \midrule
 70.22 & 53.20 & 43.61 & 37.10 & 16.19 & 28.78 \\ 
 \bfseries 67.31 & \bfseries 46.18 & \bfseries 35.18 & \bfseries 28.69 & \bfseries 12.99 & \bfseries 27.16 \\ 
 \midrule
 \bfseries 54.47 & 34.97 & 25.63 & 20.18 & 9.52 & 18.43 \\ 
 58.55 & \bfseries 32.06 & \bfseries 21.62 & \bfseries 16.17 & \bfseries 5.68 & \bfseries 9.57 \\ 
 \midrule
 56.00 & 34.81 & 24.26 & 18.65 & 8.27 & 15.89 \\ 
 \bfseries 53.52 & \bfseries 27.65 & \bfseries 18.56 & \bfseries 13.34 & \bfseries 4.80 & \bfseries 8.29 \\ 
 \bottomrule
 \end{tabular}
 \\
 \end{tabular}
 }
 
 \caption{\textbf{Results on the Booster Unbalanced benchmark after fine tuning.} We fine-tune several stereo networks using the Booster training split, downsampling  images at the lower camera resolution. We evaluate on full-resolution ground truths.
 }
 \label{tab:ft_unbal}
\end{table}


\begin{table*}[t]
\centering
\scalebox{0.8}{
\begin{tabular}{cccccc}
 \begin{tabular}{c}
    \multirow{3}{*}{\rotatebox{90}{Pre Ft.}} \\ 
 \end{tabular}
 \begin{tabular}{l}
 \\
 \toprule
 \\
 Category \\
 \midrule
 All \\
 \midrule
 Class 0 \\
 Class 1 \\
 Class 2 \\
 Class 3 \\
 \bottomrule
 \end{tabular}
 &
 \begin{tabular}{rrrr | rr }
 \multicolumn{6}{c}{All pixels} \\
 \toprule
 bad-2 & bad-4 & bad-6 & bad-8 & MAE & RMSE \\
 $\downarrow$ (\%) & $\downarrow$ (\%) & $\downarrow$ (\%) & $\downarrow$ (\%) & $\downarrow$ (px.) & $\downarrow$ (px.) \\
 \midrule
 54.47 & 34.97 & 25.63 & 20.18 & 9.52 & 18.43 \\ 
 \midrule
 52.25 & 29.58 & 18.59 & 11.42 & 6.27 & 8.12 \\ 
 62.80 & 36.91 & 24.70 & 17.93 & 6.28 & 10.16 \\ 
 77.00 & 53.85 & 38.34 & 30.57 & 17.21 & 22.34 \\ 
 81.26 & 67.48 & 57.93 & 50.56 & 28.06 & 34.64 \\ 
 \bottomrule
 \end{tabular}
 & &
 \begin{tabular}{rrrr | rr }
 \multicolumn{6}{c}{All pixels} \\
 \toprule
 bad-2 & bad-4 & bad-6 & bad-8 & MAE & RMSE \\
 $\downarrow$ (\%) & $\downarrow$ (\%) & $\downarrow$ (\%) & $\downarrow$ (\%) & $\downarrow$ (px.) & $\downarrow$  (px.) \\
 \midrule
 56.00 & 34.81 & 24.26 & 18.65 & 8.27 & 15.89 \\
 \midrule           
 52.33 & 30.42 & 19.40 & 12.93 & 5.21 & 6.85 \\      
 63.86 & 37.25 & 22.74 & 16.43 & 5.36 & 8.01 \\  
 66.79 & 39.00 & 20.49 & 14.80 & 7.36 & 9.36 \\      
 86.39 & 68.30 & 56.57 & 47.72 & 28.43 & 35.12 \\
 \bottomrule
 \end{tabular}
 \\
 \\
 \begin{tabular}{c}
    \multirow{3}{*}{\rotatebox{90}{Post Ft.}} \\ 
 \end{tabular}
 \begin{tabular}{l}
 \\
 \toprule
 \\
 Category \\
 \midrule
 All \\
 \midrule
 Class 0 \\
 Class 1 \\
 Class 2 \\
 Class 3 \\
 \bottomrule
 \end{tabular}
 &
 \begin{tabular}{rrrr | rr }
 \multicolumn{6}{c}{All pixels} \\
 \toprule
 bad-2 & bad-4 & bad-6 & bad-8 & MAE & RMSE \\
 $\downarrow$ (\%) & $\downarrow$ (\%) & $\downarrow$ (\%) & $\downarrow$ (\%) & $\downarrow$ (px.) & $\downarrow$ (px.) \\
 \midrule
 58.55 & 32.06 & 21.62 & 16.17 & 5.68 & 9.57 \\ 
 \midrule
 57.94 & 27.68 & 16.44 & 11.81 & 4.22 & 6.42 \\ 
 53.58 & 30.88 & 21.78 & 16.38 & 4.94 & 7.59 \\ 
 61.82 & 37.36 & 27.18 & 21.32 & 5.72 & 7.99 \\ 
 61.07 & 41.98 & 32.74 & 26.98 & 9.04 & 12.07 \\ 
 \bottomrule
 \end{tabular}
 & &
 \begin{tabular}{rrrr | rr }
 \multicolumn{6}{c}{All pixels} \\
 \toprule
 bad-2 & bad-4 & bad-6 & bad-8 & MAE & RMSE \\
 $\downarrow$ (\%) & $\downarrow$ (\%) & $\downarrow$ (\%) & $\downarrow$ (\%) & $\downarrow$ (px.) & $\downarrow$ (px.) \\
 \midrule
 53.52 & 27.65 & 18.56 & 13.34 & 4.80 & 8.29 \\ 
 \midrule
 52.87 & 28.36 & 18.95 & 12.04 & 3.75 & 5.24 \\ 
 52.31 & 23.89 & 14.47 & 9.85 & 3.95 & 6.14 \\ 
 60.21 & 25.75 & 14.11 & 8.99 & 3.39 & 4.55 \\ 
 63.37 & 41.01 & 29.73 & 23.18 & 8.84 & 12.31 \\ 
 \bottomrule
 \end{tabular}
 \\ \\
 & RAFT-Stereo \cite{lipson2021raft} & & CREStereo \cite{li2022practical} \\ 
 \end{tabular}
 } 
\caption{\textbf{Results on the Booster Unbalanced benchmark -- material segmentation.} We run RAFT-Stereo \cite{lipson2021raft} (left column), and CREStereo \cite{li2022practical} (right column), on the Unbalanced Stereo test set of Booster. We process images at the lowest camera resolution. We evaluate on full resolution ground truths. Top tables: generalization results using official weights. Bottom tables: results after fine-tuning using the Booster training set.}
 \label{tab:segmentation_unbalanced}
\end{table*}

\begin{figure*}
    \vspace{-0.2cm}
    \centering
    \renewcommand{\tabcolsep}{1pt}
    \begin{tabular}{cccccccccc}
        \tiny \textit{RGB \& GT} & \tiny \textit{MC-CNN} \cite{zbontar2016stereo} & \tiny \textit{LEAStereo} \cite{cheng2020hierarchical} & \tiny \textit{CFNet} \cite{shen2021cfnet} & \tiny \textit{HSMNet} \cite{yang2019hierarchical} & \tiny \textit{Neural Ref.} \cite{aleotti2021neural} & \tiny \textit{RAFT-Stereo} \cite{lipson2021raft} & \tiny \textit{CREStereo} \cite{li2022practical} & \tiny \textit{RAFT-Stereo (ft)} \cite{lipson2021raft} & \tiny \textit{CREStereo (ft)} \cite{li2022practical} \\
        \includegraphics[width=0.095\textwidth]{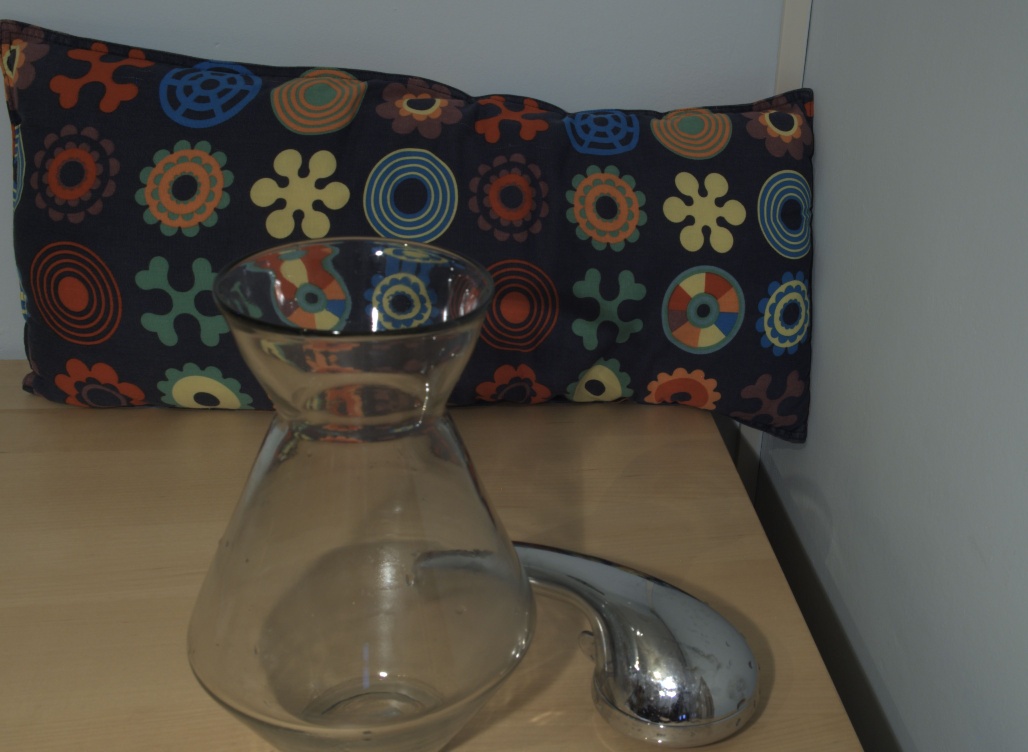} &
        \includegraphics[width=0.095\textwidth]{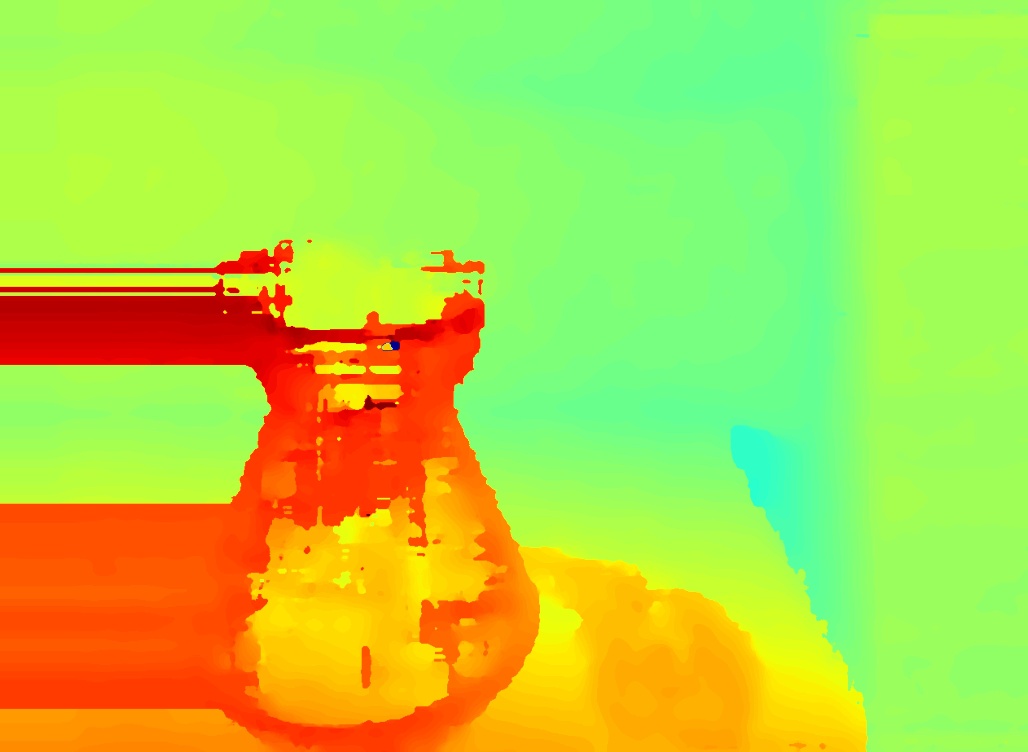} &
        \includegraphics[width=0.095\textwidth]{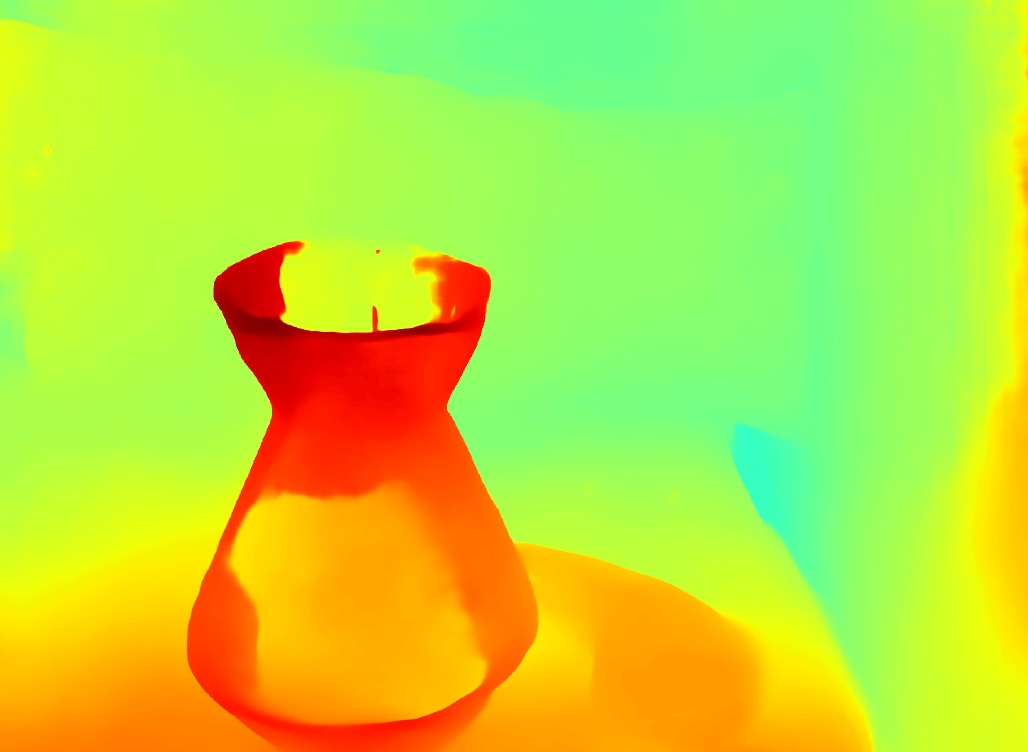} &
        \includegraphics[width=0.095\textwidth]{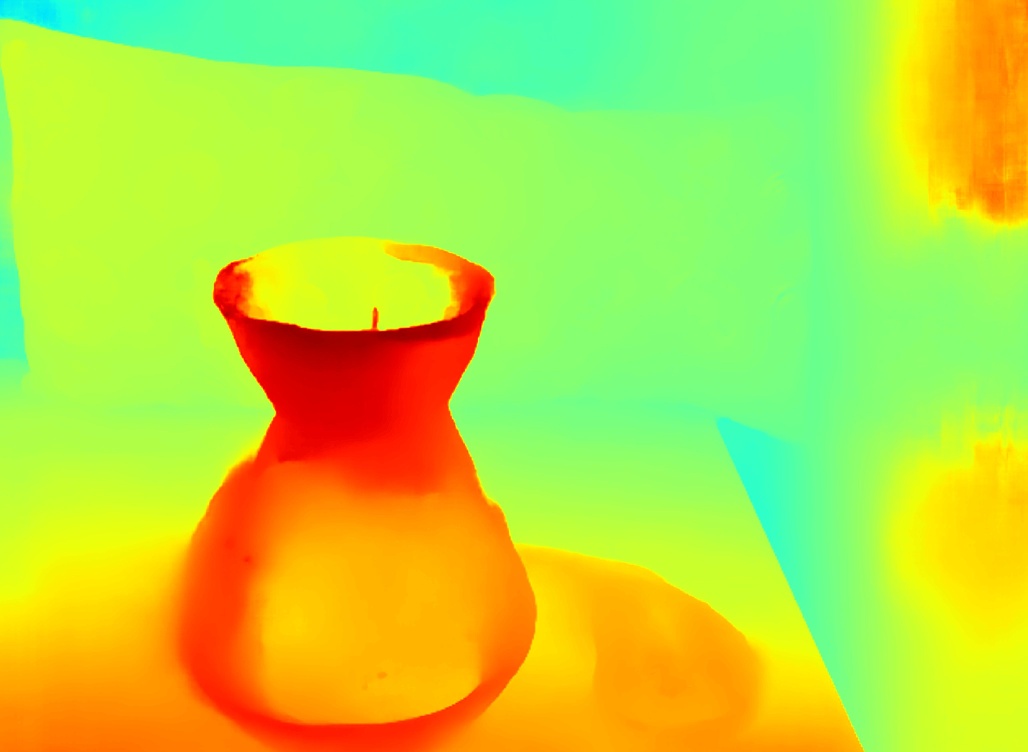} &
        \includegraphics[width=0.095\textwidth]{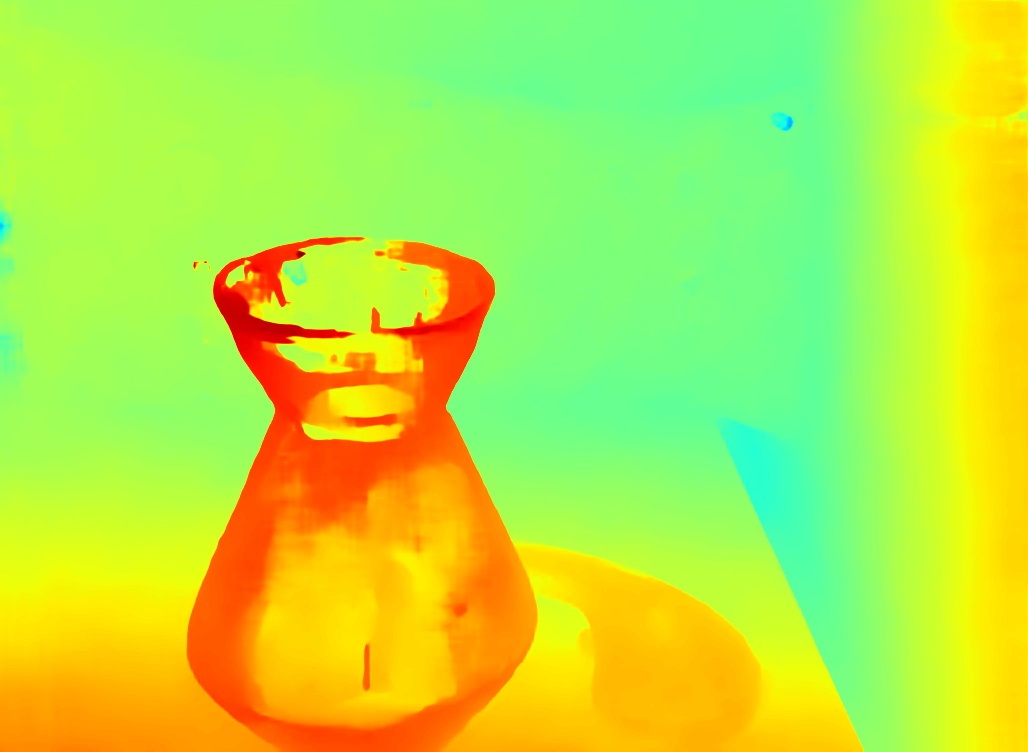} &
        \includegraphics[width=0.095\textwidth]{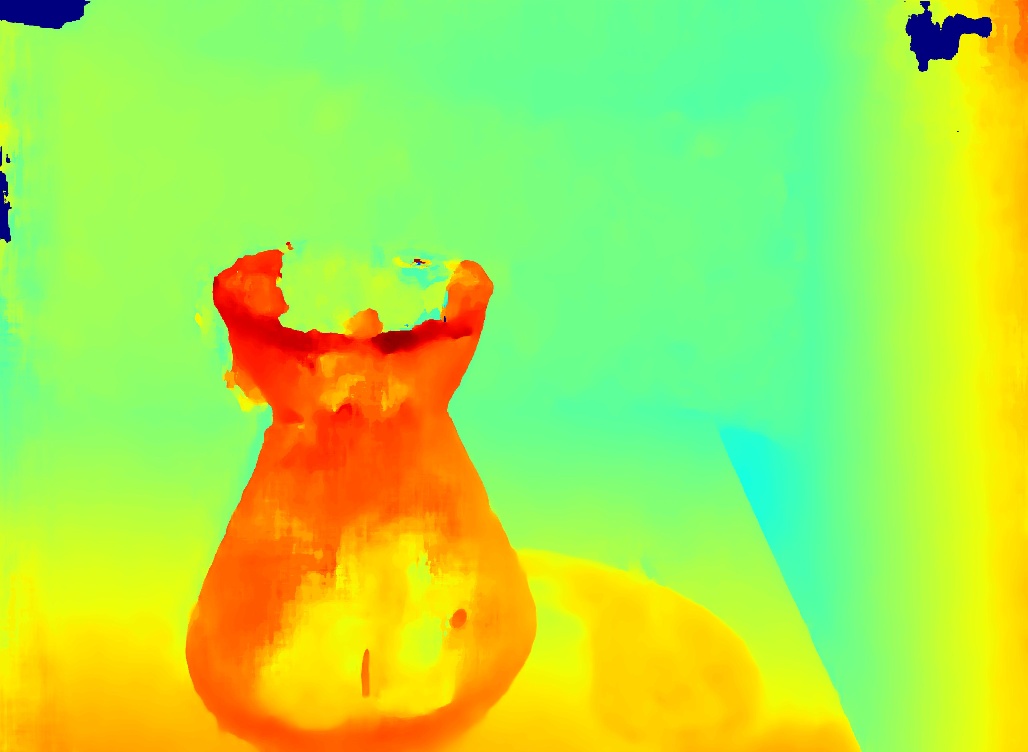} &
        \includegraphics[width=0.095\textwidth]{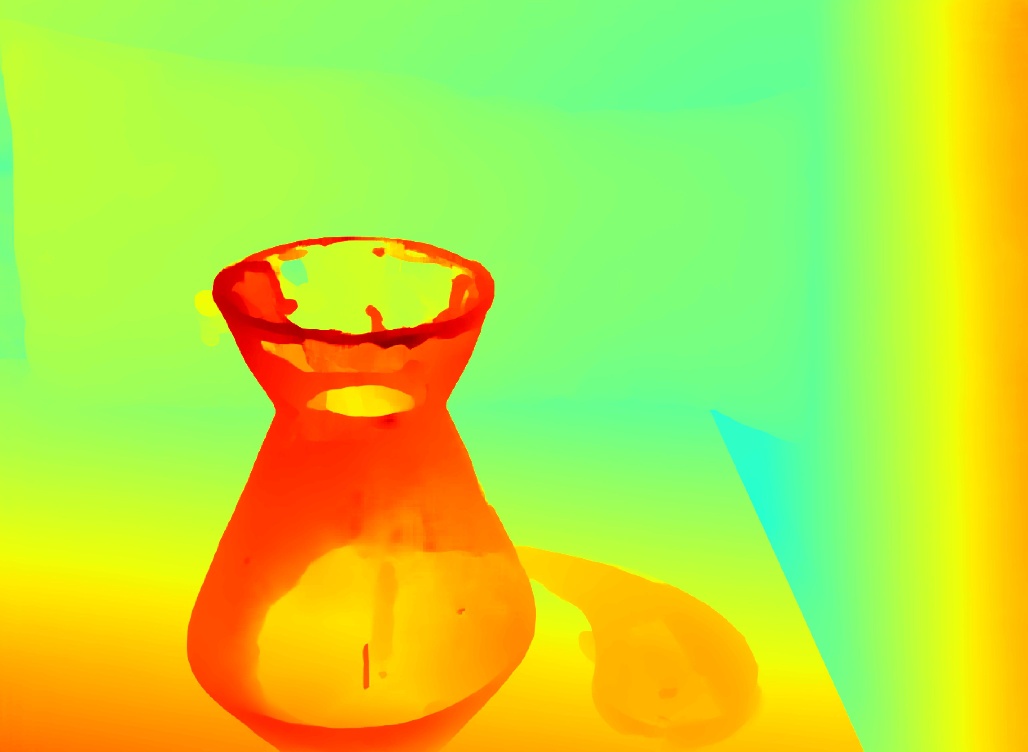} &
        \includegraphics[width=0.095\textwidth]{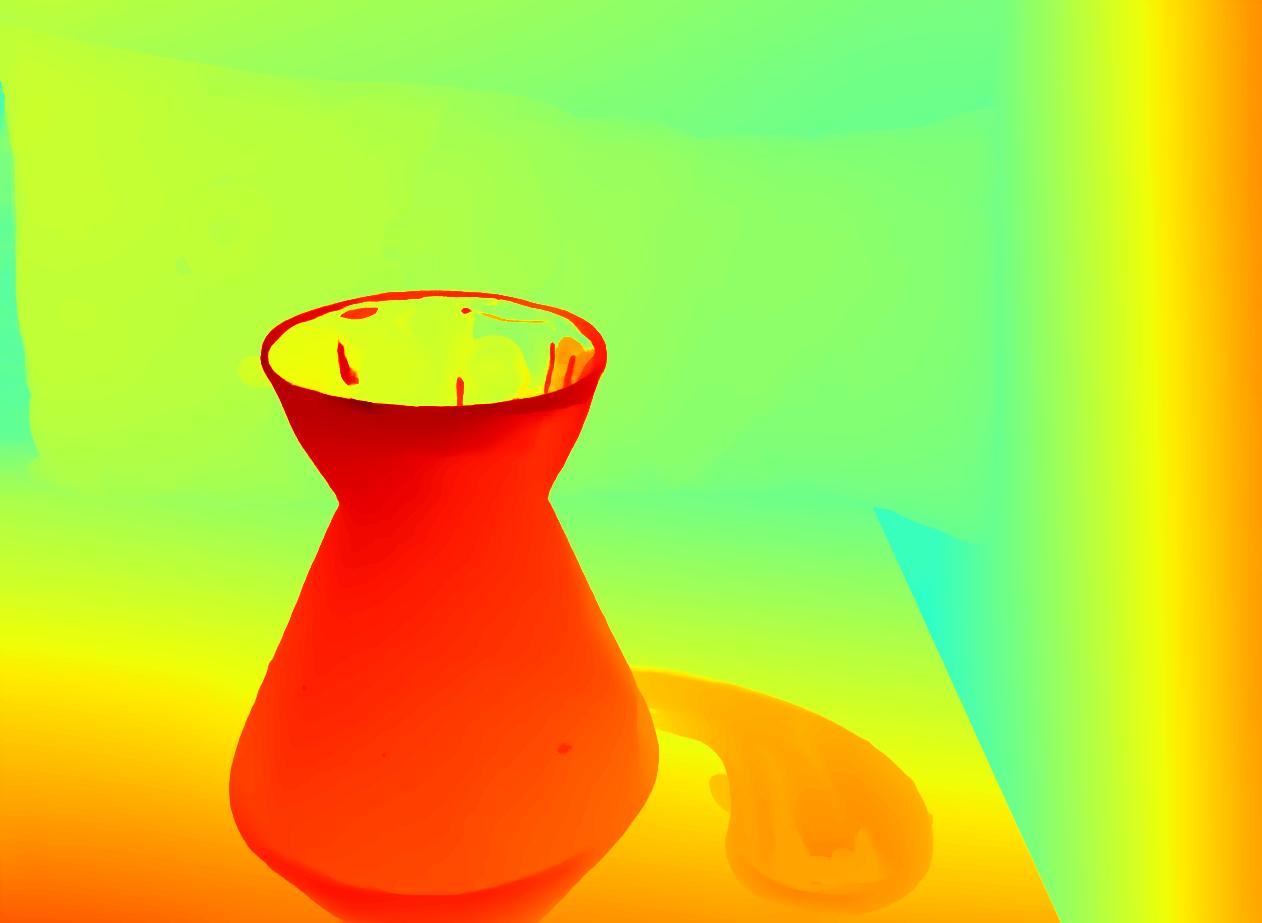} &
        \includegraphics[width=0.095\textwidth]{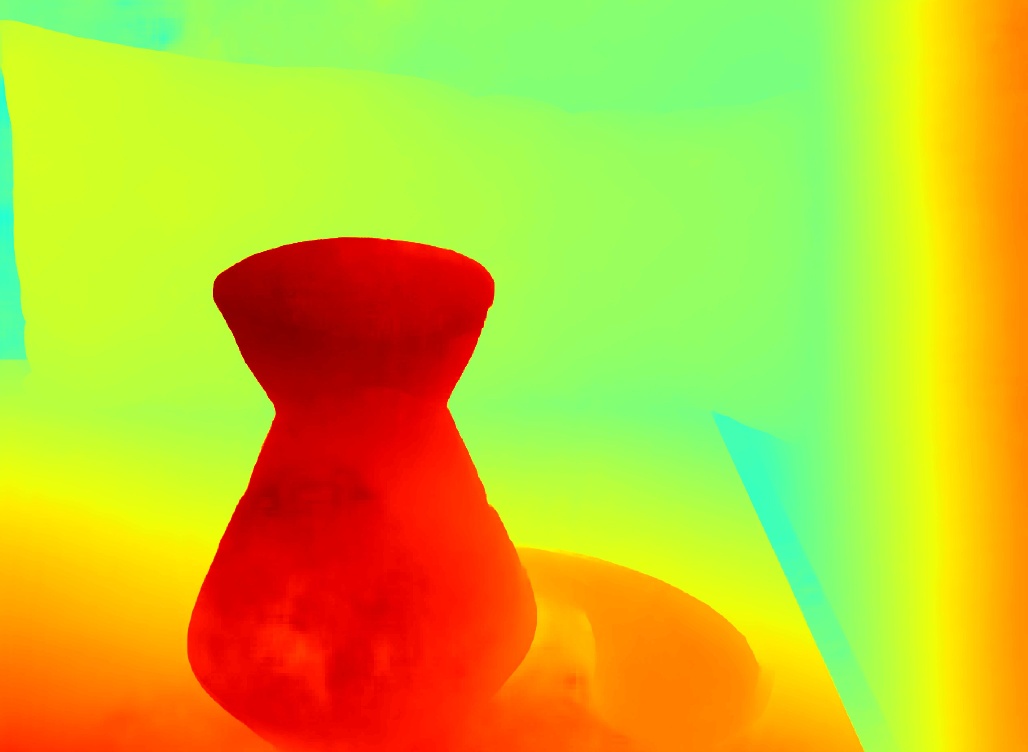} &
        \includegraphics[width=0.095\textwidth]{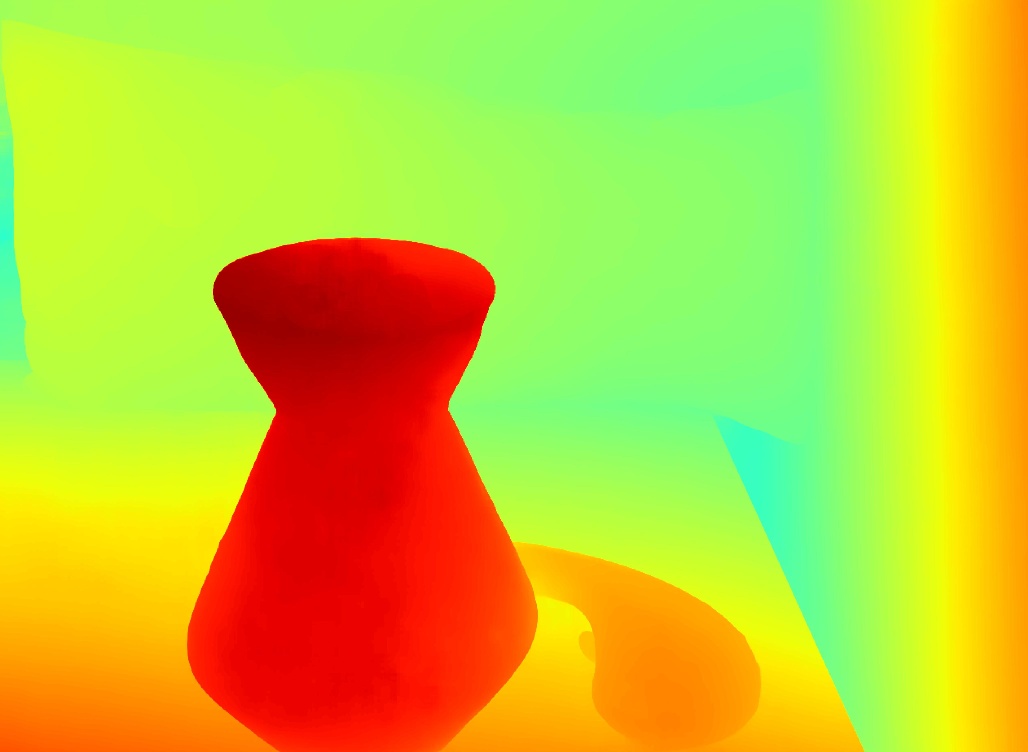} \\
        \includegraphics[width=0.095\textwidth]{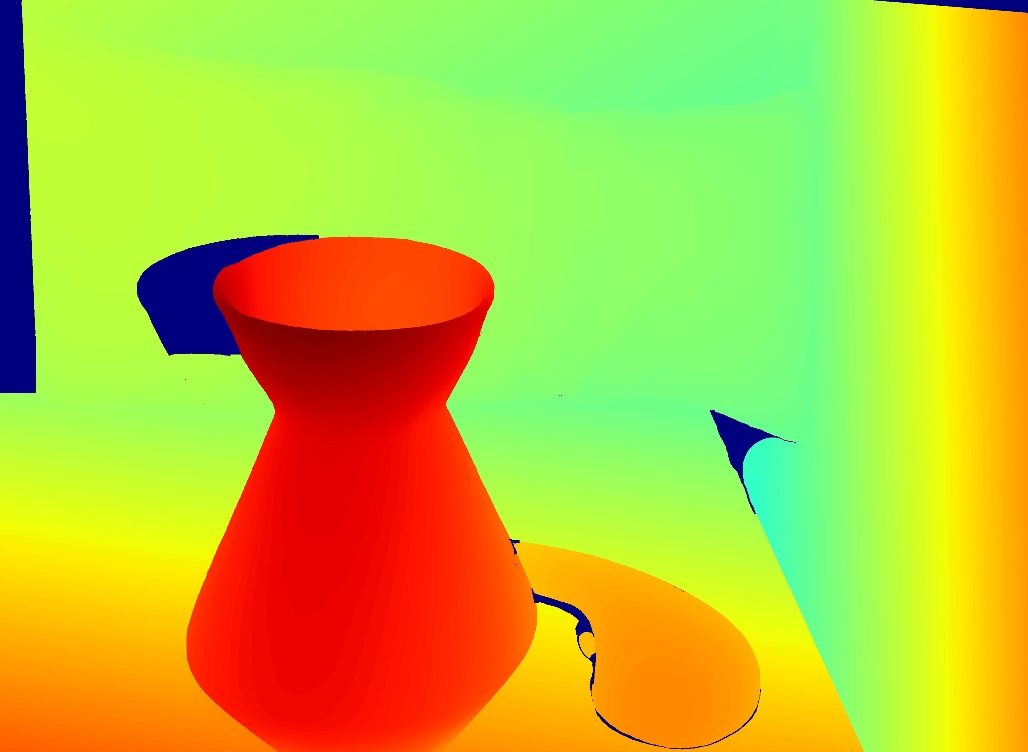} &
        \includegraphics[width=0.095\textwidth]{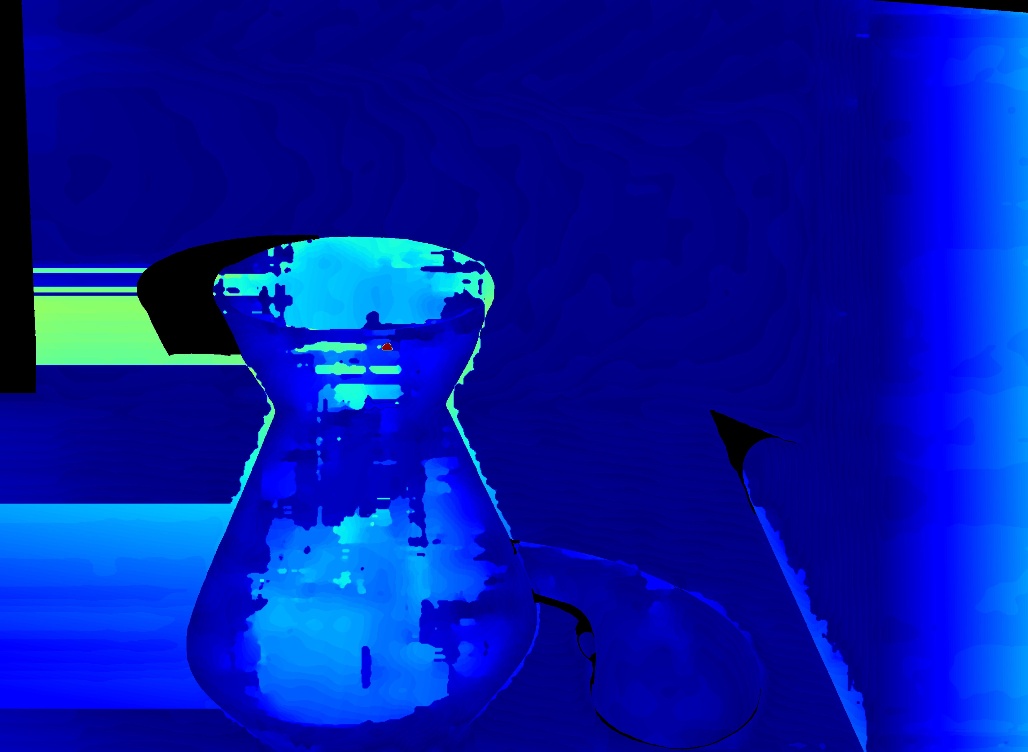} &
        \includegraphics[width=0.095\textwidth]{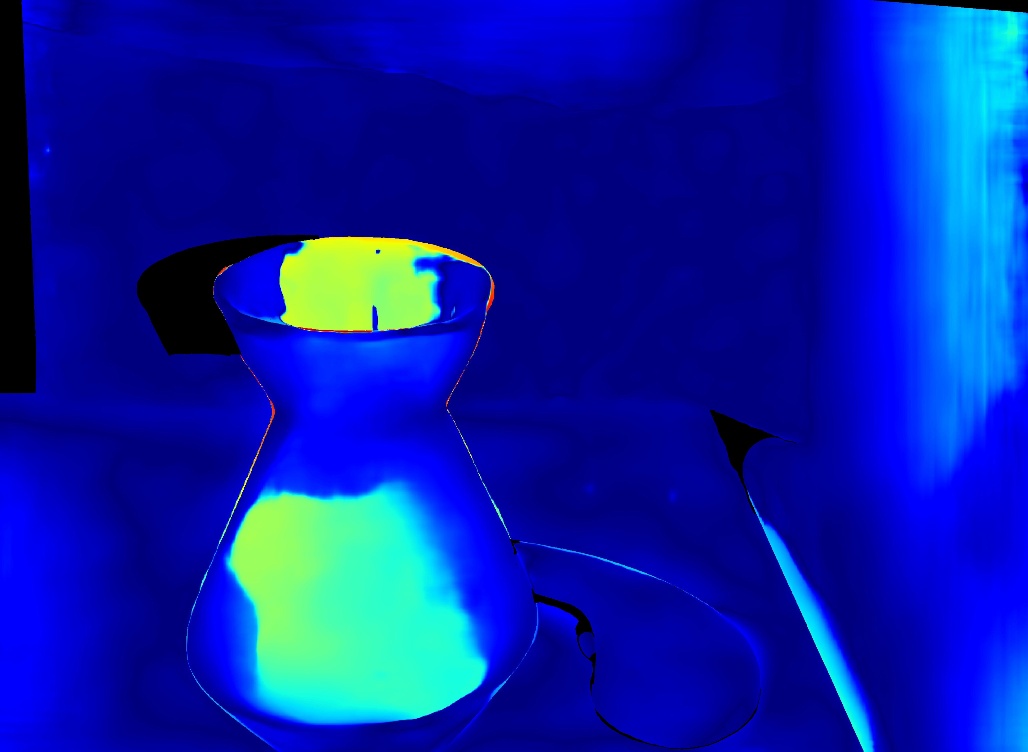} &
        \includegraphics[width=0.095\textwidth]{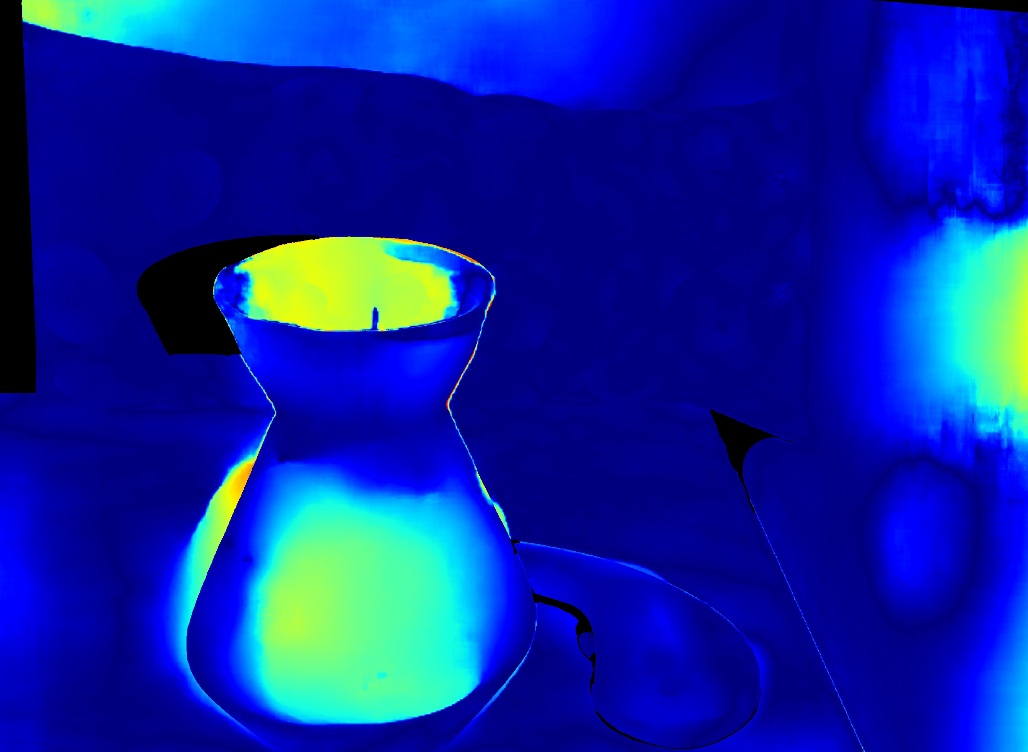} &
        \includegraphics[width=0.095\textwidth]{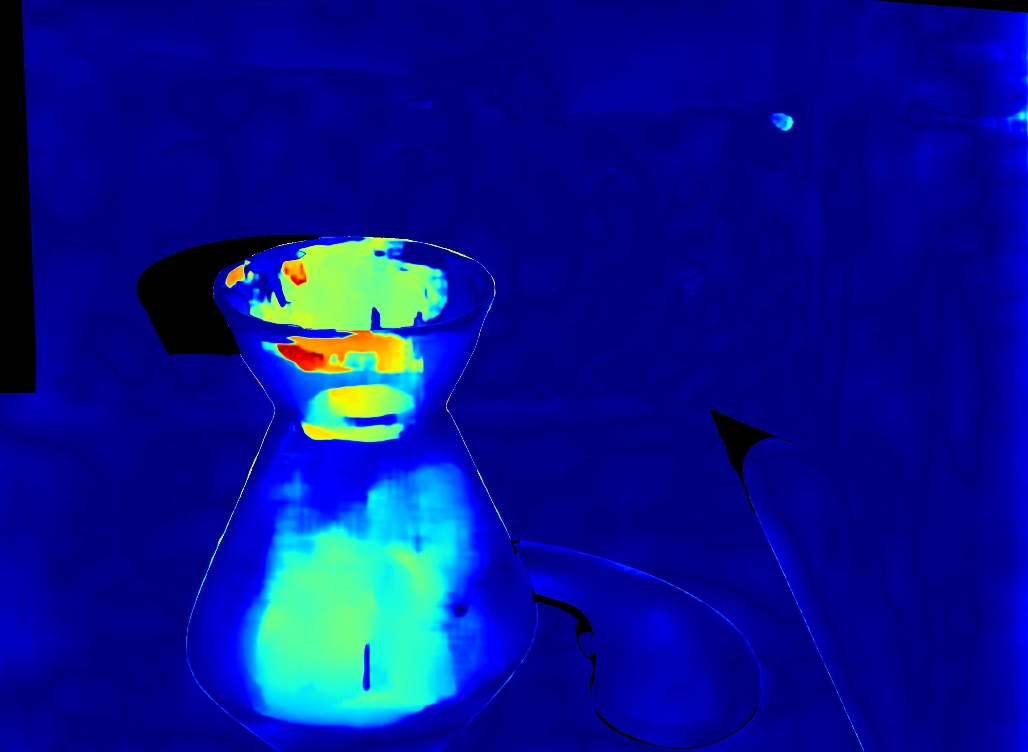} &
        \includegraphics[width=0.095\textwidth]{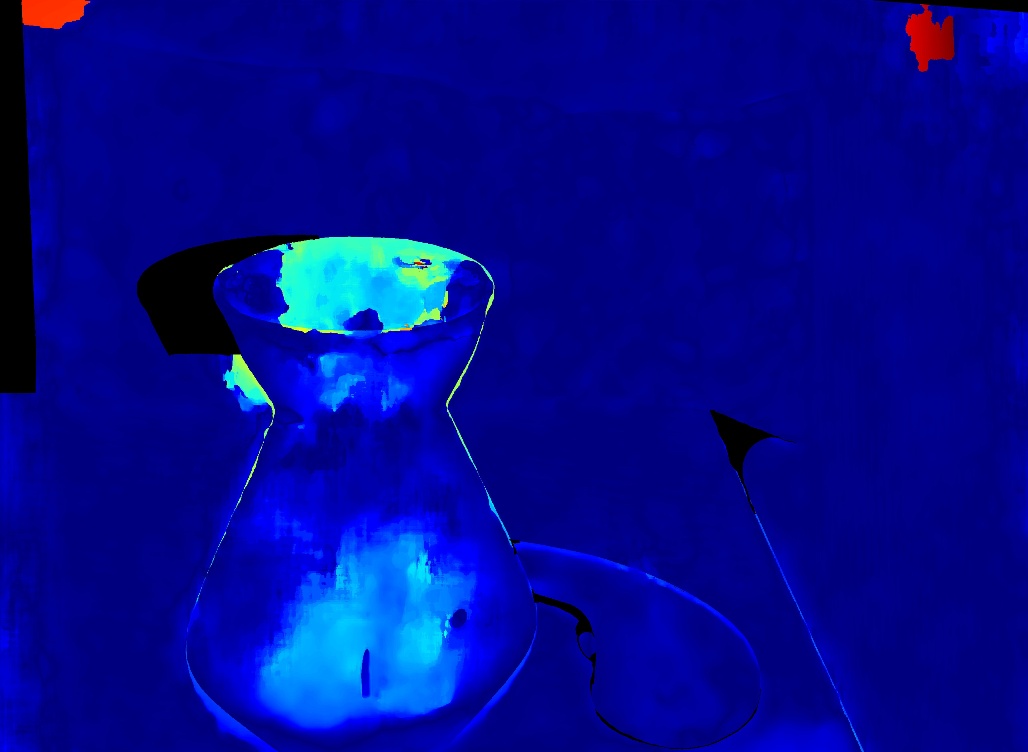} &
        \includegraphics[width=0.095\textwidth]{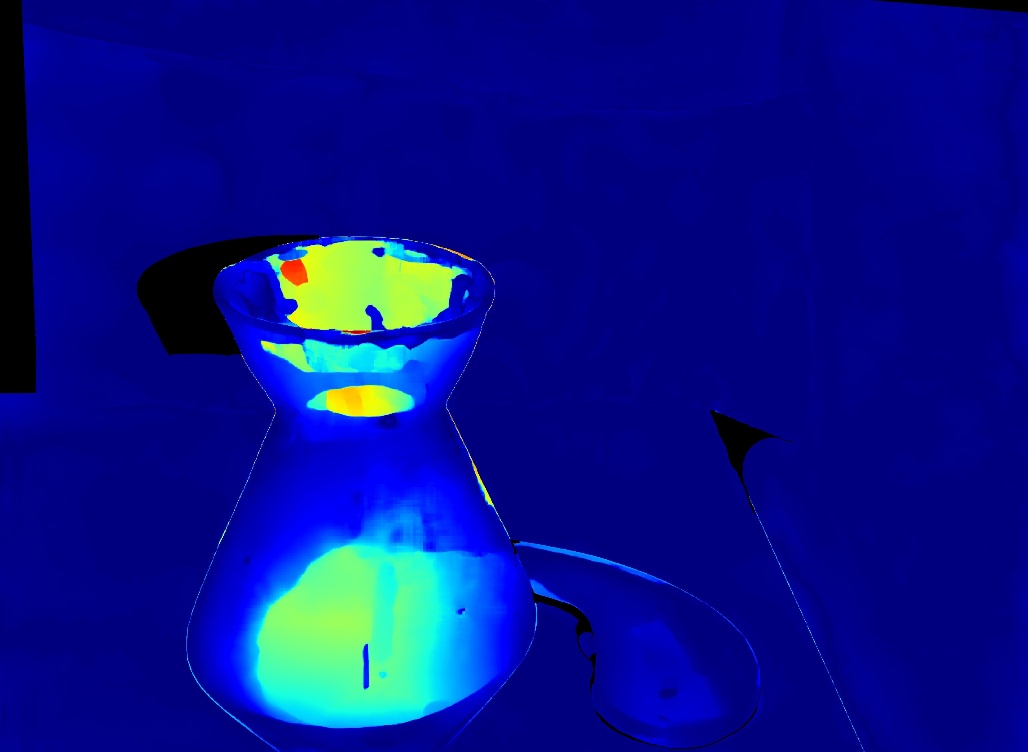} &
        \includegraphics[width=0.095\textwidth]{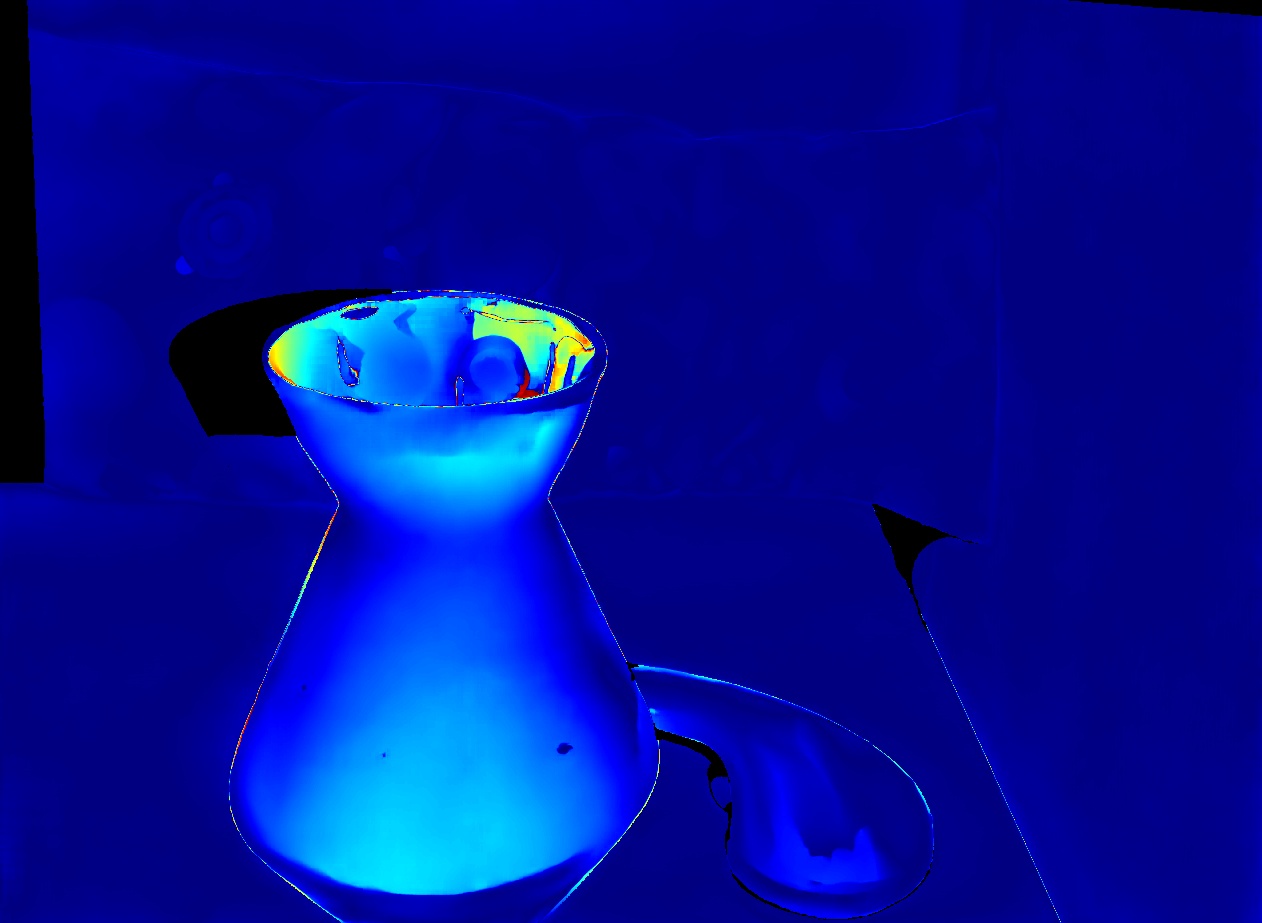} &
        \includegraphics[width=0.095\textwidth]{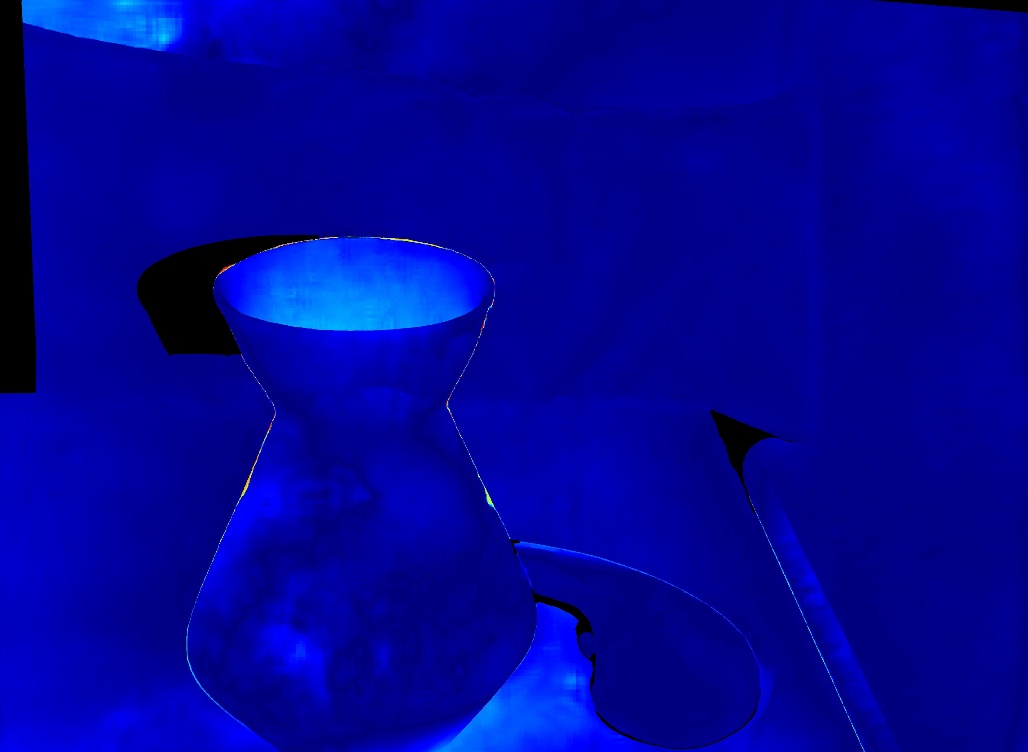} &
        \includegraphics[width=0.095\textwidth]{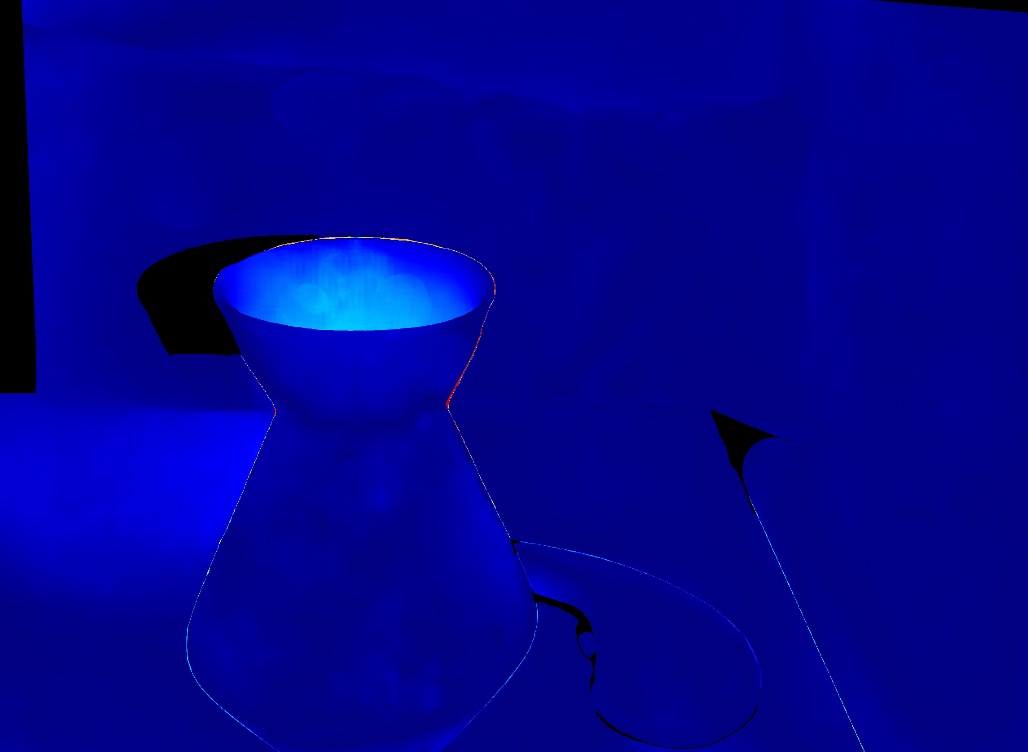} \\
    \end{tabular}
    \caption{\textbf{Qualitative results on Booster Unbalanced Stereo test split.} We show the reference image (top) and ground-truth map (bottom) on the leftmost column, followed by disparity (top) and error maps (bottom) for the deep models evaluated in our benchmark.}
    \label{fig:unbalanced_qualitatives}
\end{figure*}

\textbf{Finetuning by the Booster training data.}
In \cref{tab:ft_unbal}, we report results achieved by fine-tuning stereo networks on the Booster unbalanced training data. Given the peculiarity of this setup, we fine-tune four networks this time:
LEAStereo, CFNet, RAFT-Stereo, and CREStereo. We can notice that all networks improve their performance on almost all metrics after fine-tuning.
Thus, future research on stereo may leverage the finding that state-of-the-art deep models already hold the potential to learn better how to match specular/transparent surfaces, even in unbalanced settings, when properly fine-tuned with carefully annotated data. 
In \cref{fig:unbalanced_qualitatives}, we show qualitative results of the networks evaluated in \cref{tab:unbalanced} and \cref{tab:ft_unbal}, highlighting that transparent regions represent one of the main causes of failure for stereo networks on the unbalanced split too, showing promising results after fine-tuning (two rightmost columns).

\textbf{Evaluation on challenging regions.}
In \cref{tab:segmentation_unbalanced}, we report error rates over the different materials, sorted in increasing degree of difficulty employing the material segmentation masks, properly warped to be aligned with the unbalanced stereo pairs.
As done for the Balanced setup, we employ the two best models, RAFT-Stereo (left tables) and CREStereo (right tables), and process images at the lowest resolution ($C$ camera). Moreover, we report results before (top tables) and after (bottom tables) fine-tuning using the Booster training split to ease comparison. All metrics are computed over all valid pixels on ground-truth maps at full resolution. We note consistent results with the same experiment performed on the balanced split (\cref{tab:segmentation_balanced}), with error metrics increasing when going from less challenging materials (class 0) to the most difficult one (class 3).
Moreover, after fine-tuning, both stereo networks improve their performances by large margins in almost all metrics, especially for more challenging classes (1 to 3). For instance, the bad-2 metric on CREStereo vastly improves, from 86.32\% to 59.65\%.

\begin{table*}[t]
\centering
\renewcommand{\tabcolsep}{15pt}
\scalebox{0.8}{
\begin{tabular}{cc}
 \begin{tabular}{c} 
    \multirow{3}{*}{\rotatebox{90}{All pixels}} \\ 
 \end{tabular}
 \begin{tabular}{lc} 
 \\
 \toprule
 & Input \\
 Model & Res.\\
 \midrule
 MiDaS \cite{midas} & 384$\times$280 \\
 LeRes \cite{yin2021learning} & 224$\times$224 \\
 DPT \cite{Ranftl_2021_ICCV} & 524$\times$384 \\
 Boosting MiDaS \cite{miangoleh2021boosting} & F \\
 Boosting LeRes \cite{miangoleh2021boosting} & F \\
 \midrule
 MiDaS (ft) \cite{midas} & 384$\times$280 \\
 DPT (ft) \cite{Ranftl_2021_ICCV} & 524$\times$384 \\
 \bottomrule
 \end{tabular} 
 &
 \begin{tabular}{ rrr | rrr } 
 \multicolumn{6}{c}{Depth} \\
 \toprule
 $\delta$ $<$ 1.25 & $\delta$ $<$ 1.15 & $\delta$ $<$ 1.05 & MAE & Abs. Rel & RMSE \\
 $\uparrow$ (\%) & $\uparrow$ (\%) & $\uparrow$ (\%) & $\downarrow$ (m) & $\downarrow$ & $\downarrow$ (m)  \\
    \midrule
    93.70 & 84.36 & 51.13 & 0.090 & 0.073 & 0.122 \\ 
    93.48 & 83.77 & 49.77 & 0.090 & 0.077 & 0.117 \\ 
    92.89 & \bfseries 86.20 & \bfseries 57.04 & \bfseries 0.083 & \bfseries 0.069 & \bfseries 0.112 \\ 
    \bfseries 94.49 & 85.63 & 49.23 & 0.088 & 0.072 & 0.119 \\ 
    93.38 & 83.43 & 49.41 & 0.090 & 0.078 & 0.118 \\ 
    \midrule
    92.70 & 84.23 & 53.25 & 0.090 & 0.073 & 0.124  \\ 
    93.86 & 87.54 & 57.30 & 0.079 & 0.065 & 0.106 \\ 
    \bottomrule
 \end{tabular} 
 \end{tabular}
 }

 \caption{\textbf{Results on the Booster Mono benchmark.} We run off-the-shelves mono networks, using weights provided by their authors. We evaluate on full resolution ground-truth maps.
 Best scores in \textbf{bold}.}
 \label{tab:mono_comparison}
 
\end{table*}

\begin{table*}[t]
\centering
\scalebox{0.75}{
\begin{tabular}{ccccc}

 \begin{tabular}{c}
    \multirow{3}{*}{\rotatebox{90}{Pre ft.}} \\ 
 \end{tabular}
 \begin{tabular}{l}
 \\
 \toprule
 \\
 Category \\
 \midrule
 All \\
 \midrule
 Class 0 \\
 Class 1 \\
 Class 2 \\
 Class 3 \\
 \bottomrule
 \end{tabular}
 &
 \begin{tabular}{rrr | rrr}
 \multicolumn{6}{c}{All pixels} \\
 \toprule
 $\delta$ $<$ 1.25 & $\delta$ $<$ 1.15 & $\delta$ $<$ 1.05 & MAE & Abs. Rel & RMSE \\
 $\uparrow$ (\%) & $\uparrow$ (\%) & $\uparrow$ (\%) & $\downarrow$ (m) & $\downarrow$ & $\downarrow$ (m) \\
 \midrule
 93.70 & 84.36 & 51.13 & 0.090 & 0.073 & 0.122 \\
 \midrule
 92.08 & 81.85 & 45.82 & 0.100 & 0.082 & 0.132 \\ 
 86.91 & 75.86 & 47.46 & 0.104 & 0.092 & 0.118 \\ 
 99.02 & 90.55 & 52.36 & 0.059 & 0.063 & 0.072 \\ 
 90.32 & 78.91 & 46.61 & 0.087 & 0.094 & 0.100 \\  
 \bottomrule
 \end{tabular}
 & &
 \begin{tabular}{rrr | rrr}
 \multicolumn{6}{c}{All pixels} \\
 \toprule
 $\delta$ $<$ 1.25 & $\delta$ $<$ 1.15 & $\delta$ $<$ 1.05 & MAE & Abs. Rel & RMSE \\
 $\uparrow$ (\%) & $\uparrow$ (\%) & $\uparrow$ (\%) & $\downarrow$ (m) & $\downarrow$ & $\downarrow$ (m) \\
 \midrule
 92.89 & 86.20 & 57.04 & 0.083 & 0.069 & 0.112 \\ 
 \midrule
 94.19 & 87.68 & 52.73 & 0.085 & 0.068 & 0.114 \\ 
 90.60 & 82.00 & 52.18 & 0.090 & 0.078 & 0.104 \\ 
 89.87 & 77.29 & 39.84 & 0.092 & 0.097 & 0.112 \\ 
 88.17 & 78.95 & 45.47 & 0.093 & 0.100 & 0.108 \\  
 \bottomrule
 \end{tabular}
 \\
 \begin{tabular}{c}
    \multirow{3}{*}{\rotatebox{90}{Post ft.}} \\ 
 \end{tabular}
 \begin{tabular}{l}
 \\
 \toprule
 \\
 Category \\
 \midrule
 All \\
 \midrule
 Class 0 \\
 Class 1 \\
 Class 2 \\
 Class 3 \\
 \bottomrule
 \end{tabular}
 &
 \begin{tabular}{rrr | rrr }
 \multicolumn{6}{c}{All pixels} \\
 \toprule
 $\delta$ $<$ 1.25 & $\delta$ $<$ 1.15 & $\delta$ $<$ 1.05 & MAE & Abs. Rel & RMSE \\
 $\uparrow$ (\%) & $\uparrow$ (\%) & $\uparrow$ (\%) & $\downarrow$ (m) & $\downarrow$ & $\downarrow$ (m) \\
 \midrule
 92.70 & 84.23 & 53.25 & 0.090 & 0.073 & 0.124 \\
 \midrule
 92.12 & 82.45 & 47.78 & 0.097 & 0.079 & 0.130  \\
 86.32 & 74.86 & 46.42 & 0.107 & 0.095 & 0.122  \\
 99.40 & 89.59 & 54.49 & 0.059 & 0.062 & 0.074  \\
 88.24 & 77.81 & 47.71 & 0.093 & 0.099 & 0.106  \\ 
 \bottomrule
 \end{tabular}
 & &
 \begin{tabular}{rrr | rrr }
 \multicolumn{6}{c}{All pixels} \\
 \toprule
 $\delta$ $<$ 1.25 & $\delta$ $<$ 1.15 & $\delta$ $<$ 1.05 & MAE & Abs. Rel & RMSE \\
 $\uparrow$ (\%) & $\uparrow$ (\%) & $\uparrow$ (\%) & $\downarrow$ (m) & $\downarrow$ & $\downarrow$ (m) \\
 \midrule
 93.86 & 87.54 & 57.30 & 0.079 & 0.065 & 0.106 \\ 
 \midrule
 95.01 & 88.12 & 52.39 & 0.082 & 0.066 & 0.108  \\ 
 91.12 & 82.32 & 49.40 & 0.091 & 0.079 & 0.102  \\ 
 93.94 & 84.12 & 48.09 & 0.074 & 0.078 & 0.089  \\ 
 91.59 & 81.80 & 47.52 & 0.078 & 0.082 & 0.090  \\  
 \bottomrule
 \end{tabular}
 \\\\
 & MiDaS \cite{midas} && DPT \cite{Ranftl_2021_ICCV } 
 \end{tabular}
 }
 \caption{\textbf{Results on the Booster Mono benchmark -- material segmentation.} We run MiDaS \cite{midas} (left column) and DPT \cite{Ranftl_2021_ICCV} (right column) by processing images at the resolution suggested by the authors. Top: results using official weights. Bottom: results after fine-tuning using the Booster training split. We evaluate on full-resolution ground-truth maps.}
 \label{tab:segmentation_mono}
\end{table*}

\begin{figure*}[t]
    \centering
    \renewcommand{\tabcolsep}{5pt}
    \begin{tabular}{cccccccc}
        \tiny \textit{RGB \& GT} & \tiny \textit{MiDaS} \cite{midas} & 
        \tiny \textit{LeRes} \cite{yin2021learning} & 
        \tiny \textit{DPT} \cite{Ranftl_2021_ICCV} & \tiny \textit{Boosting MiDaS} \cite{miangoleh2021boosting} & \tiny \textit{Boosting LeRes} \cite{miangoleh2021boosting}   & \tiny \textit{MiDaS (ft)} \cite{midas} & \tiny \textit{DPT (ft)} \cite{Ranftl_2021_ICCV} \\
        \tiny Input Res. & \tiny 384$\times$280 & \tiny 224$\times$224 & \tiny 524$\times$384 & \tiny F & \tiny F & \tiny 384$\times$280 & \tiny 524$\times$384 \\    
        \includegraphics[width=0.095\textwidth]{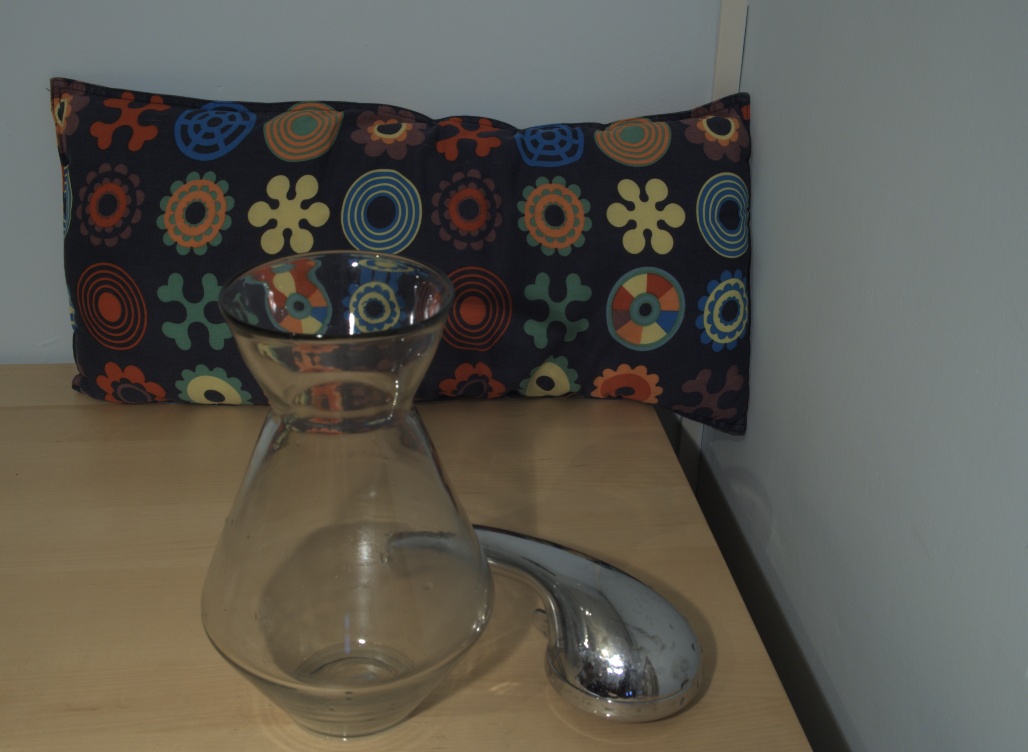} &
        \includegraphics[width=0.095\textwidth]{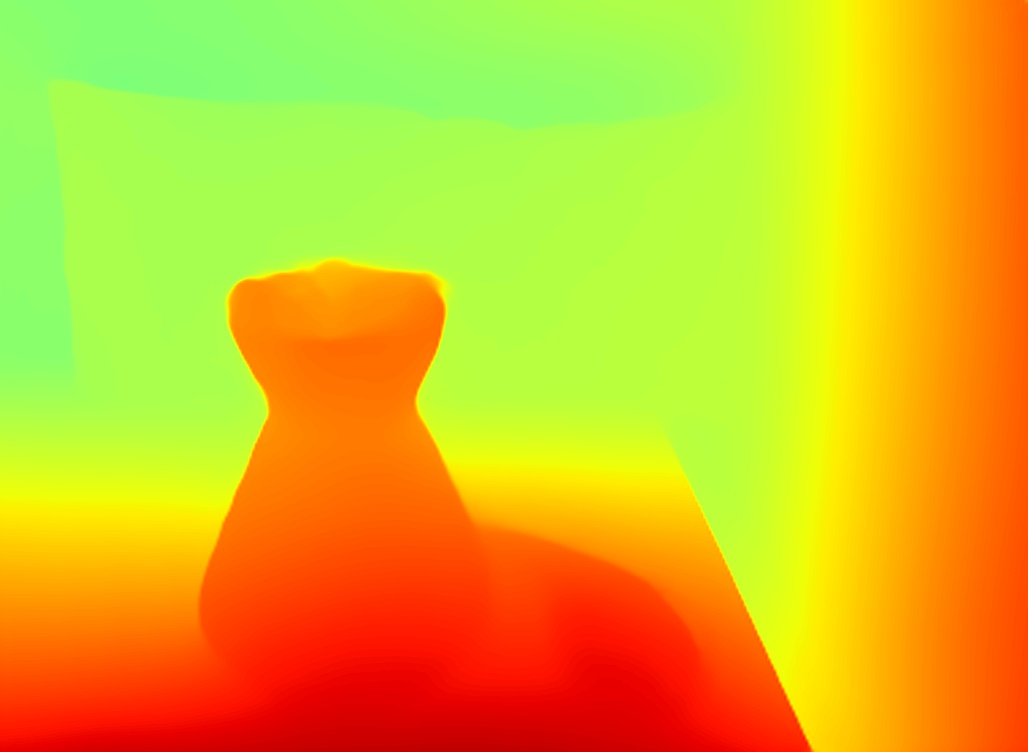} &
        \includegraphics[width=0.095\textwidth]{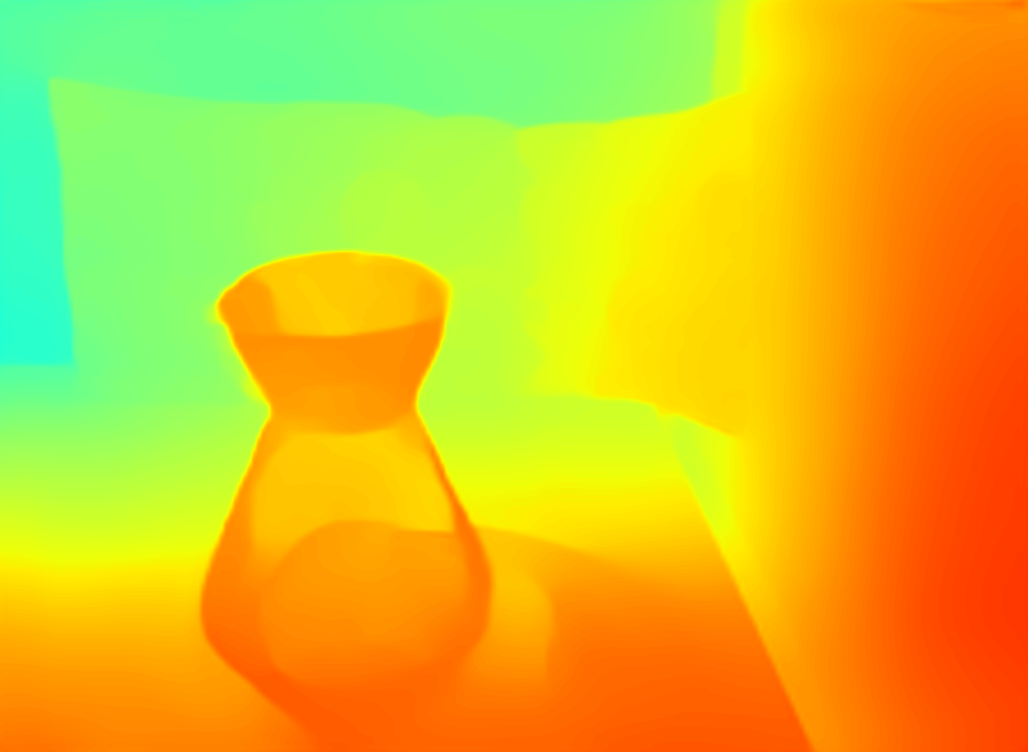} &
        \includegraphics[width=0.095\textwidth]{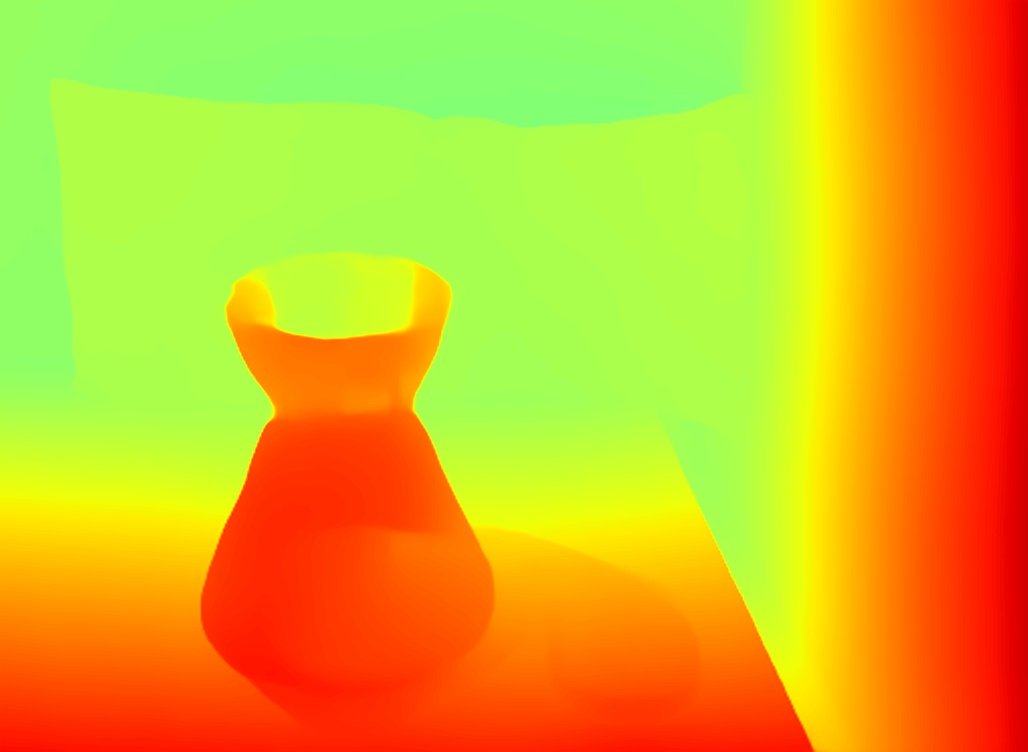} &
        \includegraphics[width=0.095\textwidth]{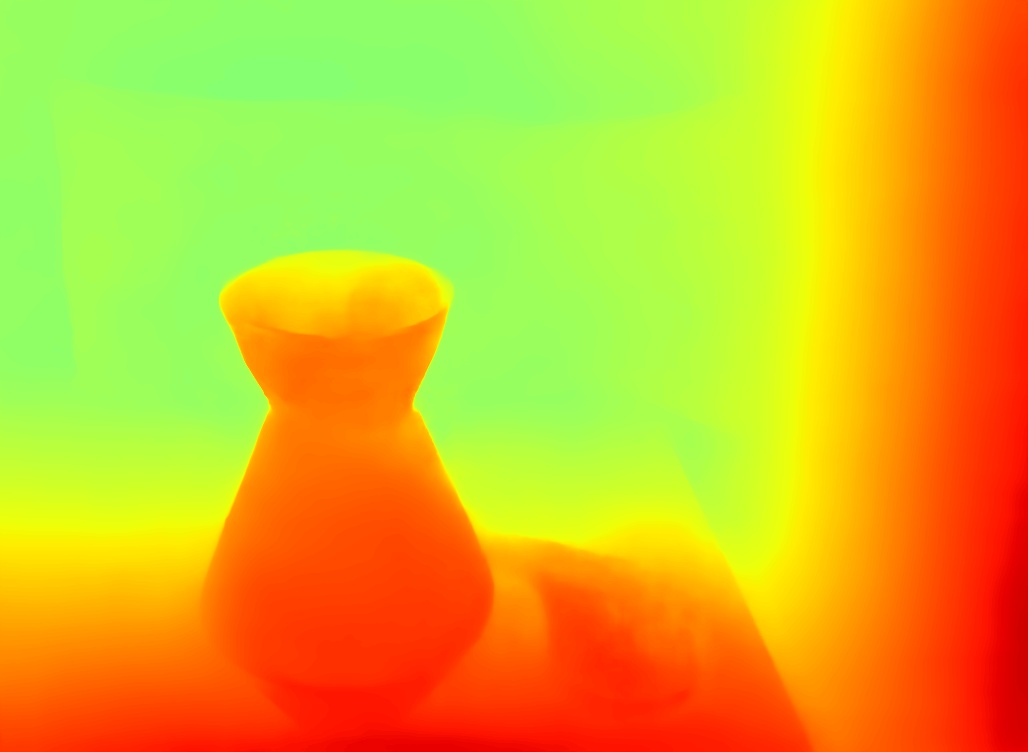} &
        \includegraphics[width=0.095\textwidth]{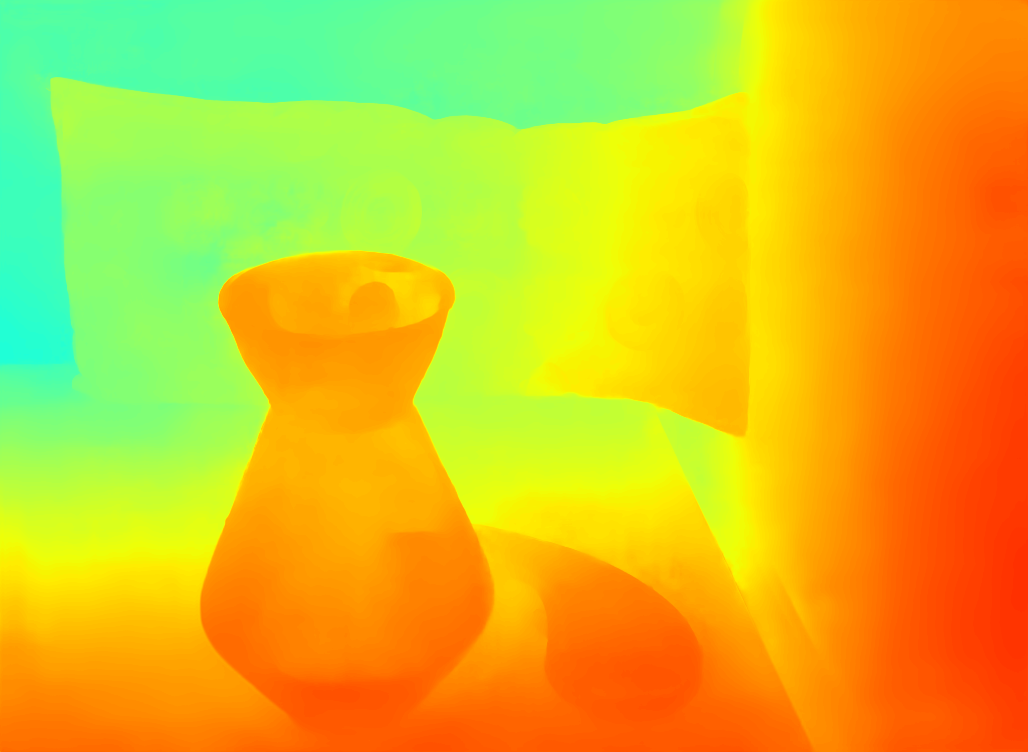} &
        \includegraphics[width=0.095\textwidth]{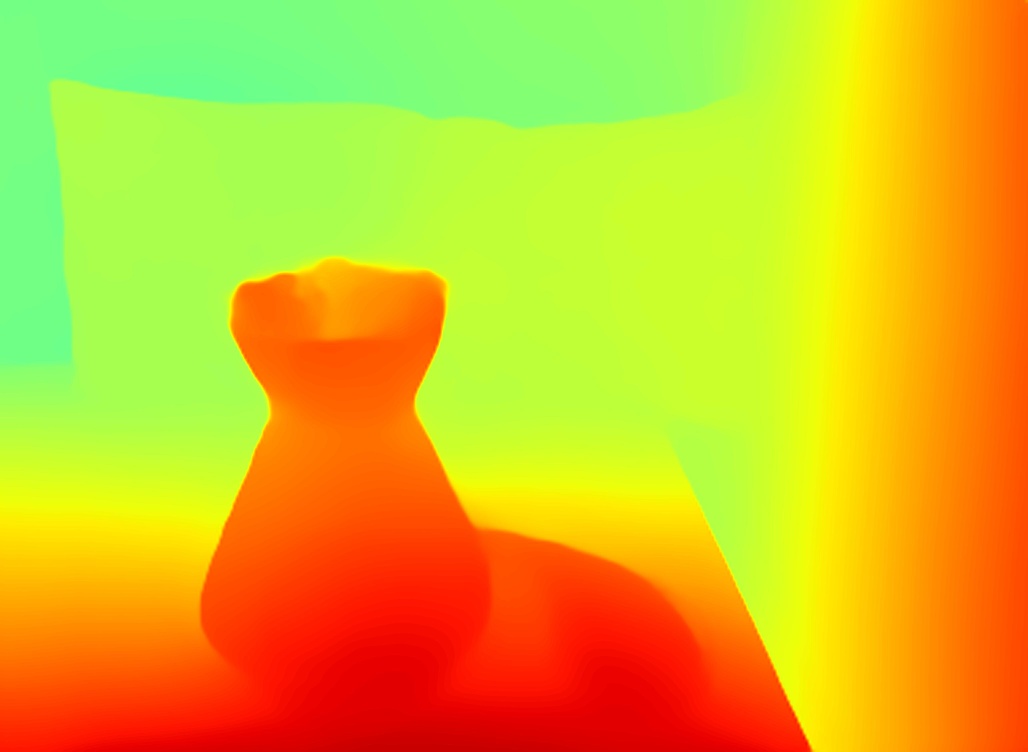} &
        \includegraphics[width=0.095\textwidth]{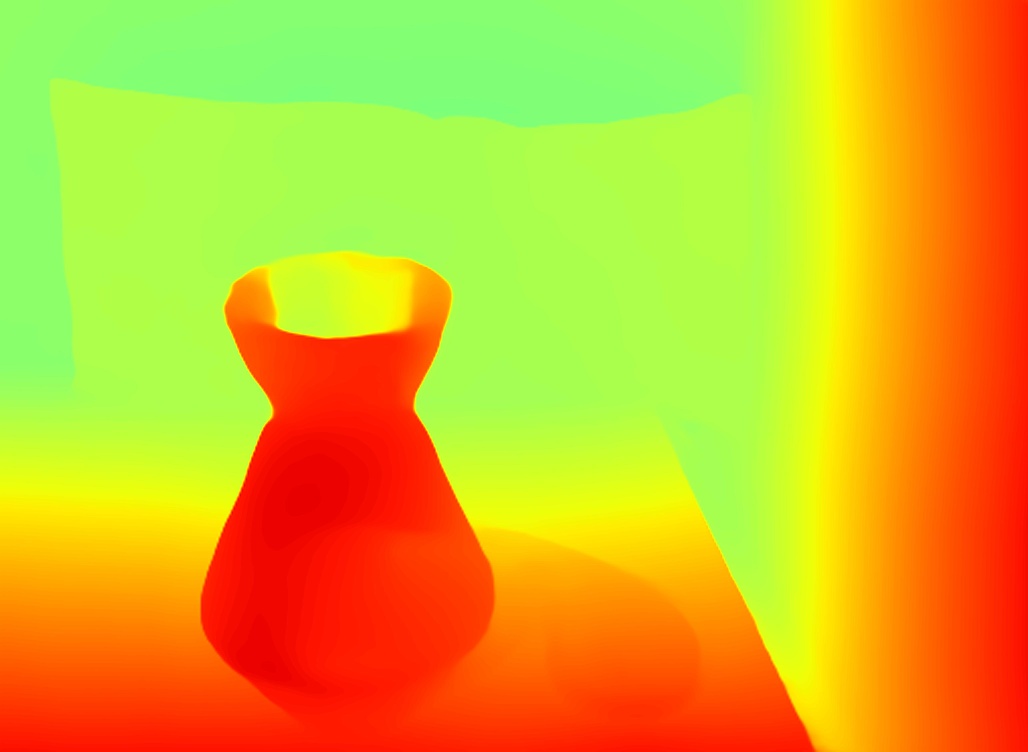} \\
        \includegraphics[width=0.095\textwidth]{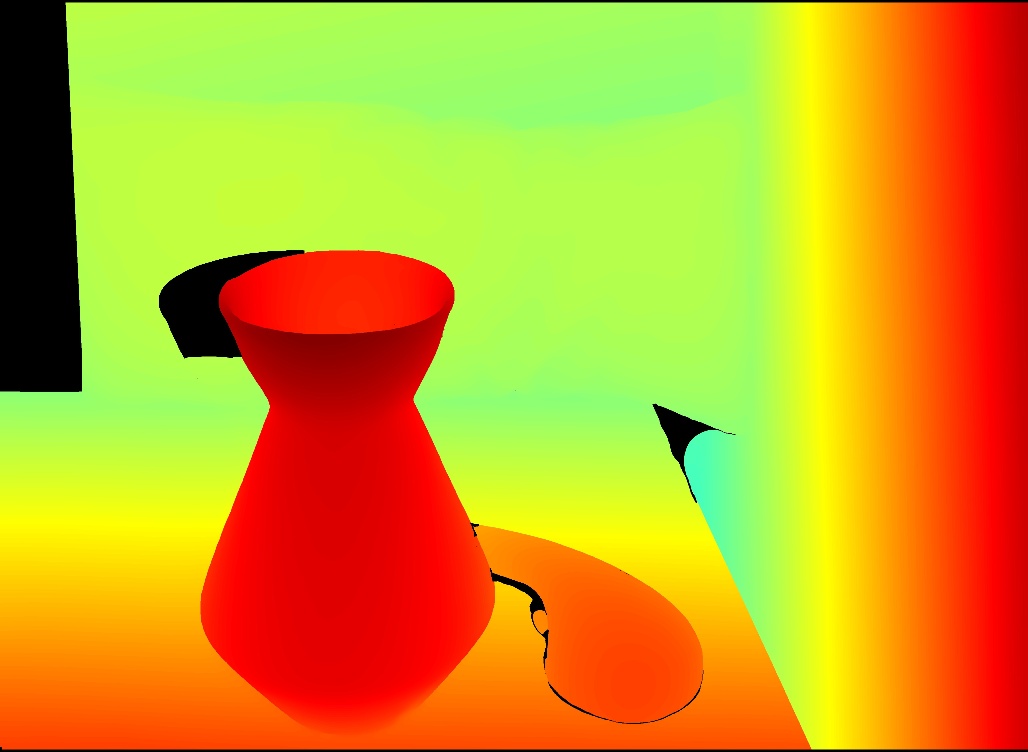} &
        \includegraphics[width=0.095\textwidth]{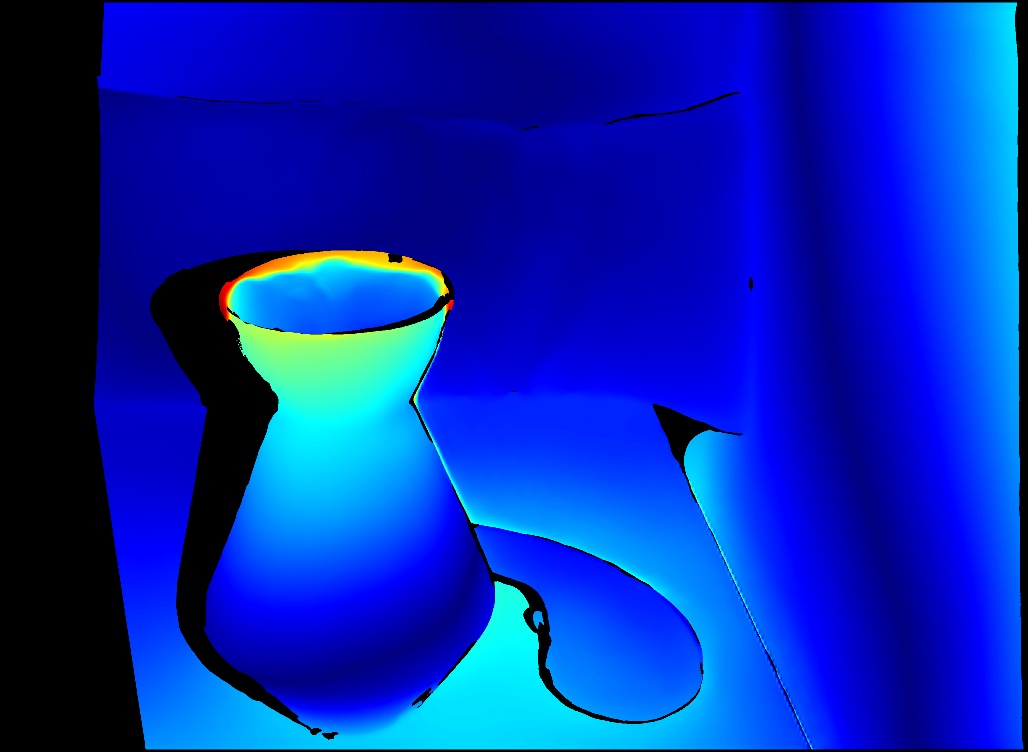} &
        \includegraphics[width=0.095\textwidth]{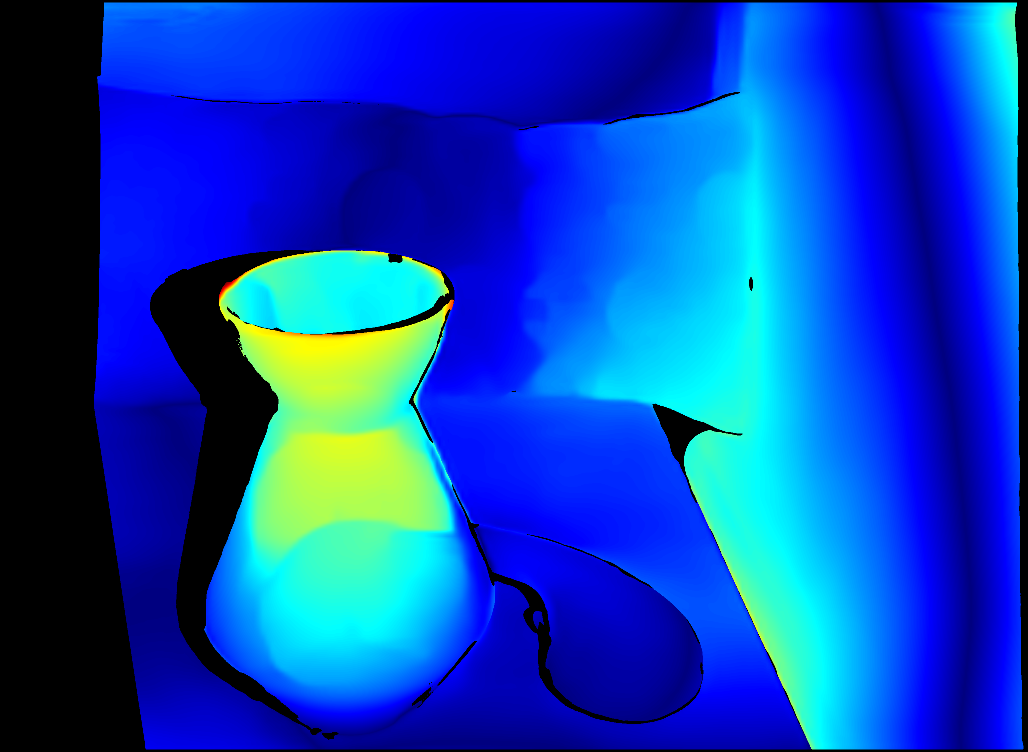} &
        \includegraphics[width=0.095\textwidth]{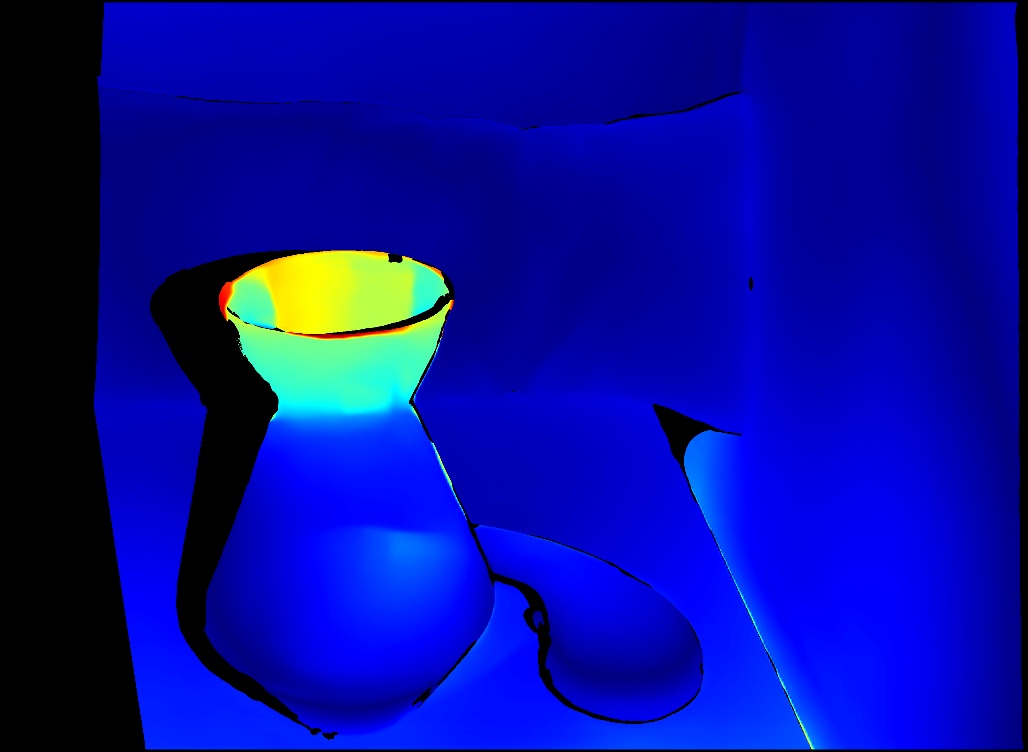} &
        \includegraphics[width=0.095\textwidth]{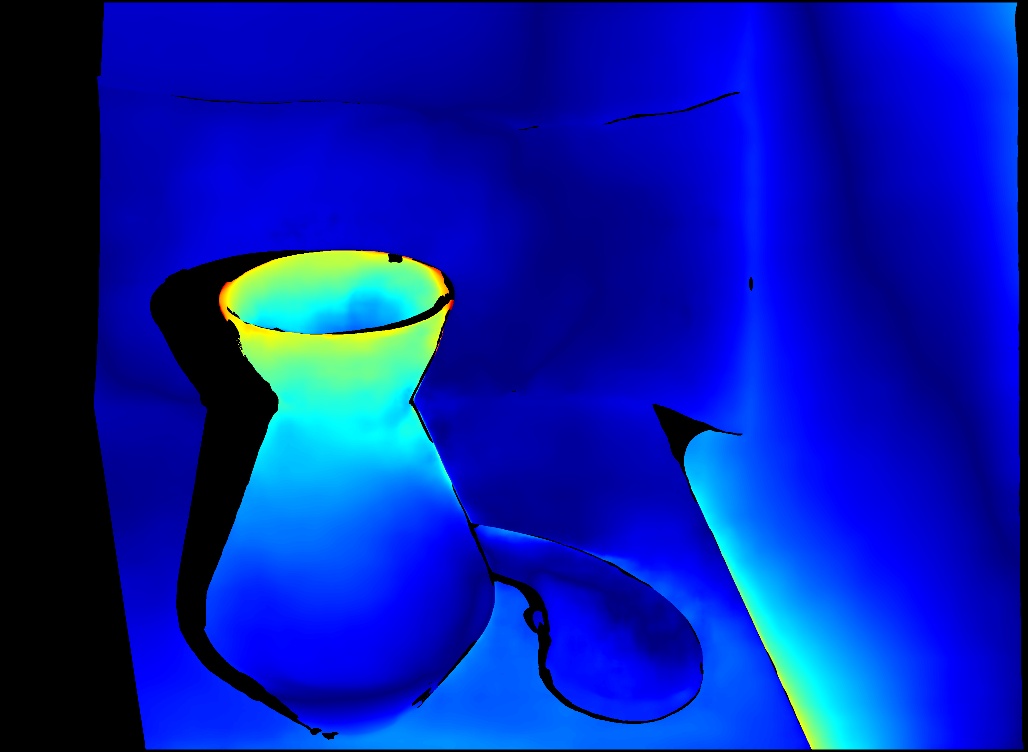} &
        \includegraphics[width=0.095\textwidth]{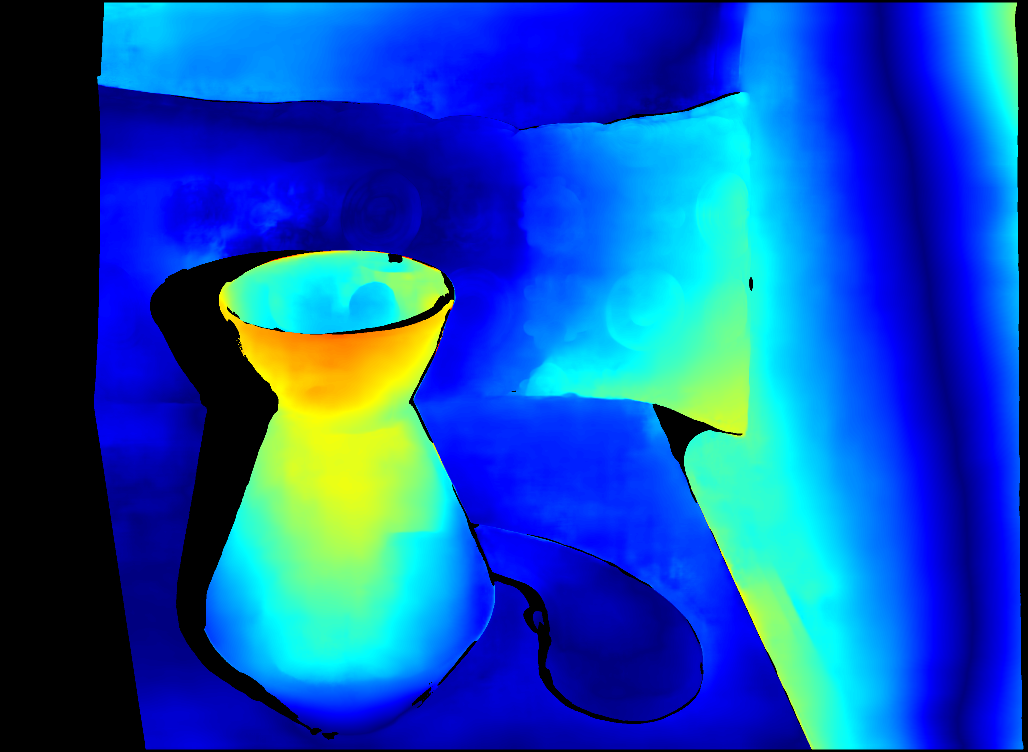} &
        \includegraphics[width=0.095\textwidth]{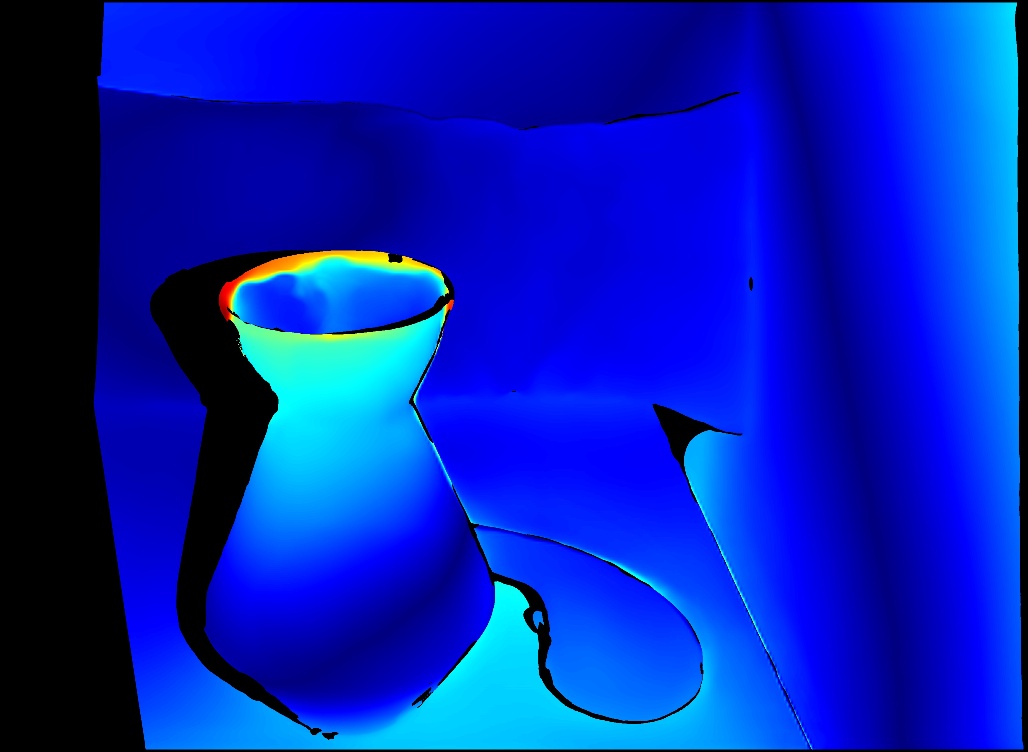} &
        \includegraphics[width=0.095\textwidth]{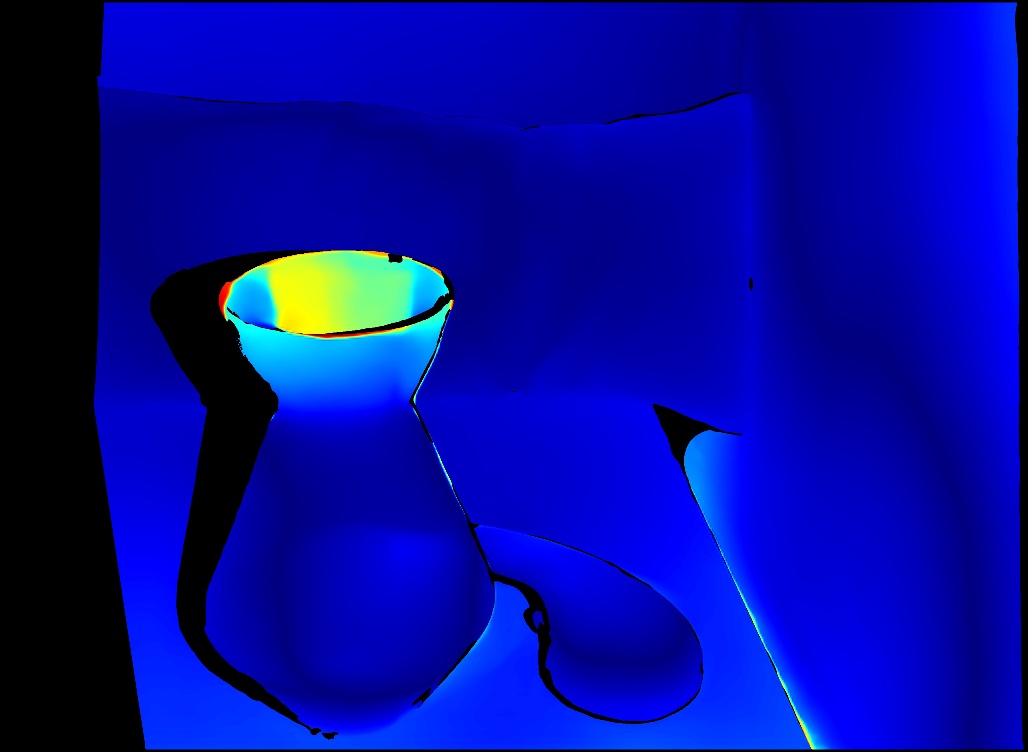} \\
    \end{tabular}
    \caption{\textbf{Qualitative results on Monocular networks.} We show the reference image (top) and the ground-truth map (bottom) on the leftmost column, followed by disparity (top) and error maps (bottom) for the monocular models evaluated in the Stereo test set.\vspace{-0.5cm}}
    \label{fig:mono}
\end{figure*}

\begin{figure}[t]
    \centering
    \setlength{\tabcolsep}{1pt}
    \begin{tabular}{cccc}
        \scriptsize RGB & \scriptsize MiDaS \cite{midas} & \scriptsize Error Map & \scriptsize GT \\
         \includegraphics[width=0.23\linewidth]{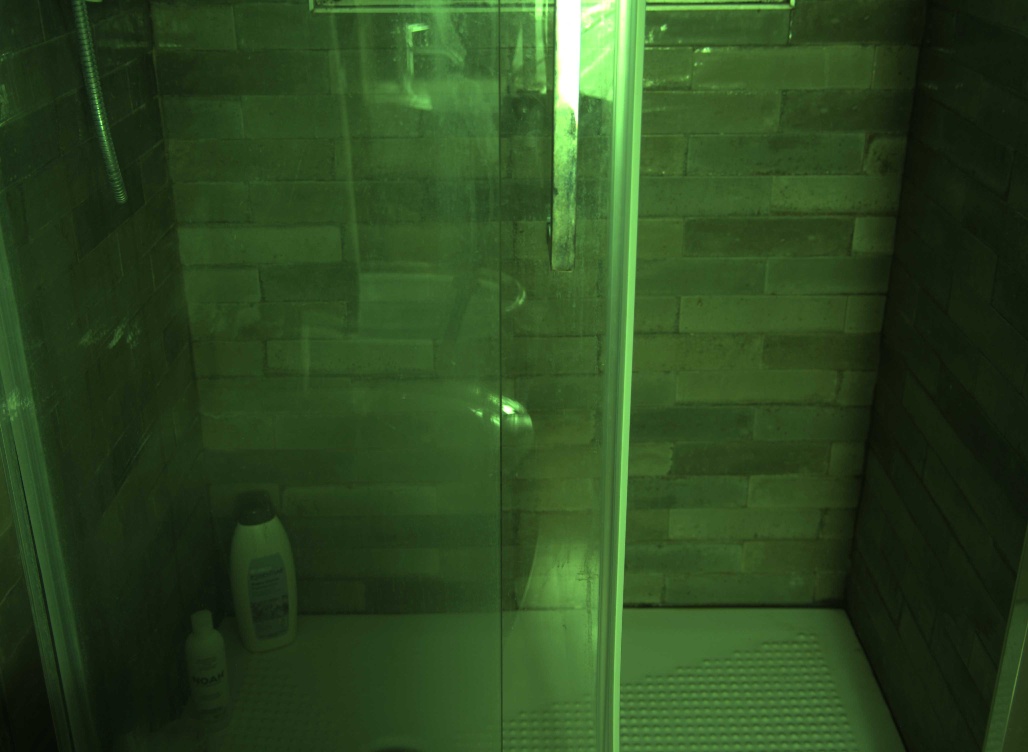} & \includegraphics[width=0.23\linewidth]{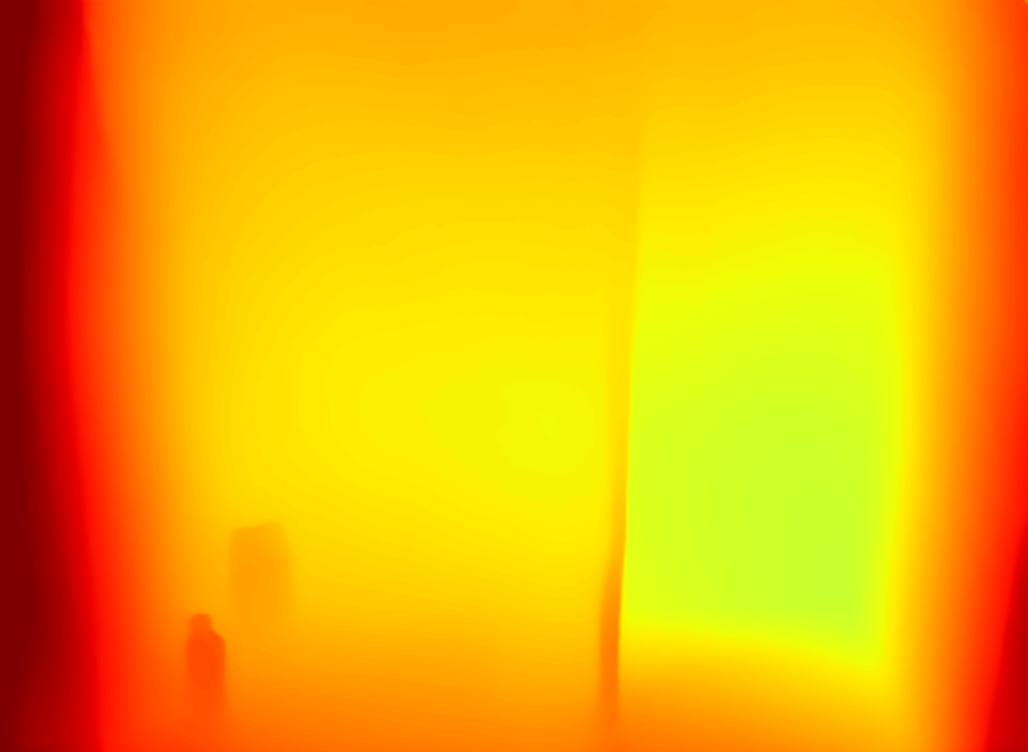} & \includegraphics[width=0.23\linewidth]{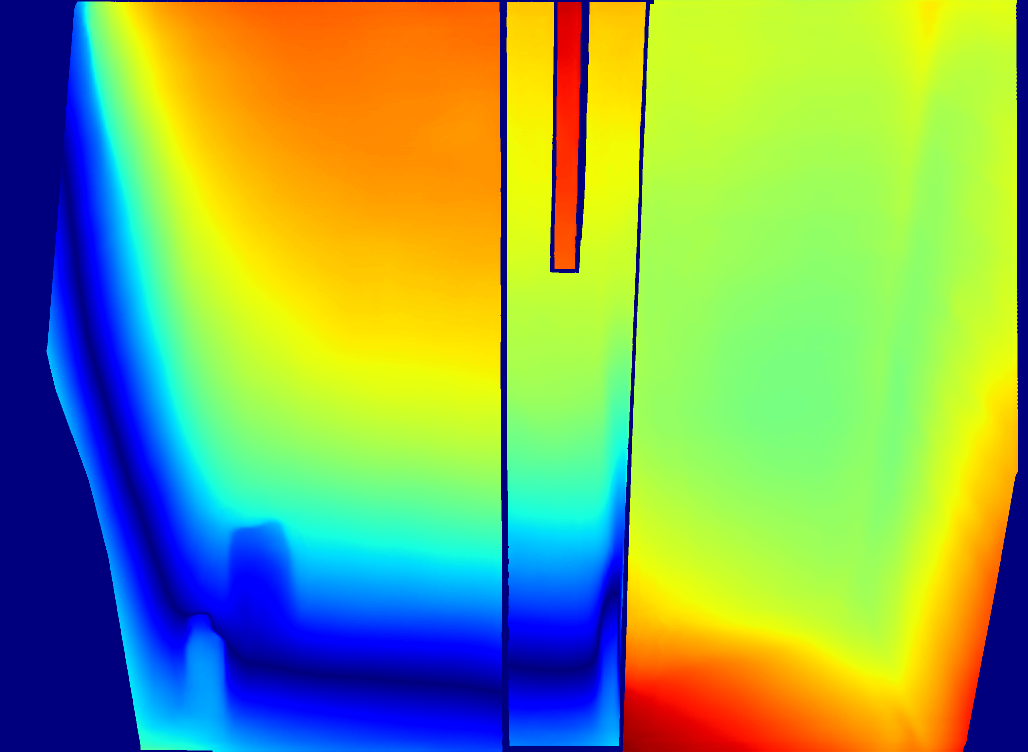} & \includegraphics[width=0.23\linewidth]{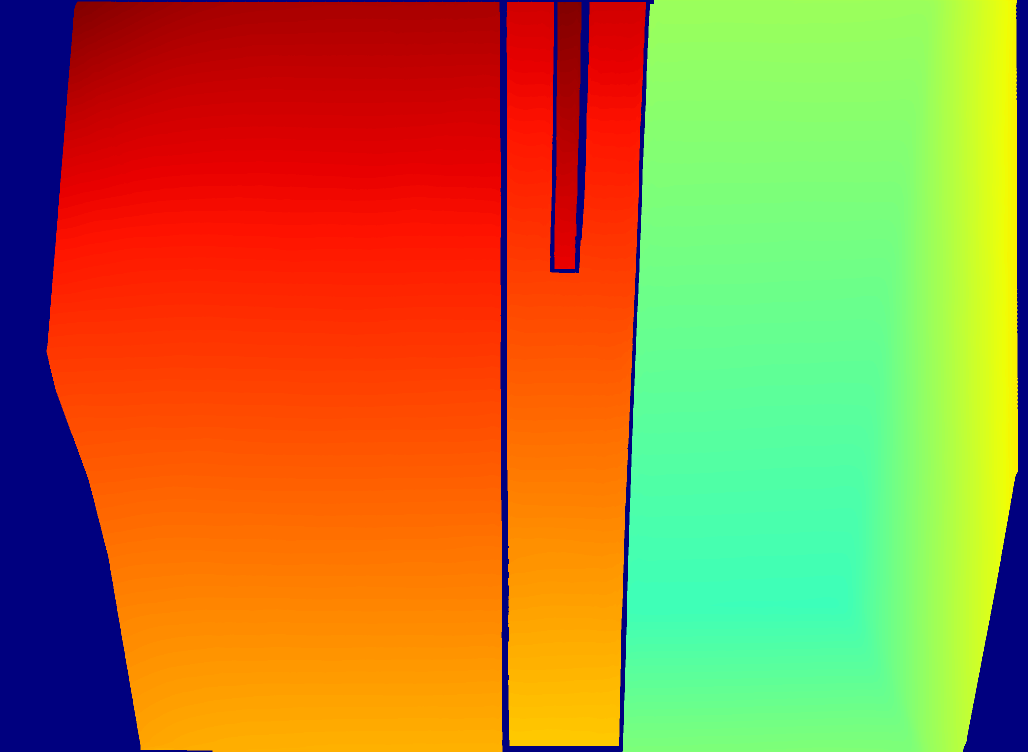} \\
    \end{tabular}
    \caption{\textbf{Monocular network failure case.} MiDaS fails in estimating the right depth in a challenging test image with a large transparent surface.}
    \label{fig:mono_failures}
\end{figure}

\begin{table*}[ht]
\centering
\scalebox{0.75}{
\begin{tabular}{ccccc}

 \begin{tabular}{c}
    \multirow{3}{*}{\rotatebox{90}{Pre ft.}} \\ 
 \end{tabular}
 \begin{tabular}{l}
 \\
 \toprule
 \\
 Category \\
 \midrule
 Class Transparent \\
 Class Mirror \\
 \bottomrule
 \end{tabular}
 &
 \begin{tabular}{rrrr | rr }
 \multicolumn{6}{c}{Stereo} \\
 \toprule
 bad-2 & bad-4 & bad-6 & bad-8 & MAE & RMSE \\
 (\%) & (\%) & (\%) & (\%) & (px.) &  (px.) \\
  \midrule
 83.70 & 74.25 & 66.15 & 59.34 & 37.93 & 49.94 \\
 81.26 & 73.26 & 71.62 & 70.66 & 133.29 & 152.81 \\
 \bottomrule
 \end{tabular}
 & &
 \begin{tabular}{rrr | rrr}
 \multicolumn{6}{c}{Mono} \\
 \toprule
 $\delta$ $<$ 1.25 & $\delta$ $<$ 1.15 & $\delta$ $<$ 1.05 & MAE & Abs. Rel & RMSE \\
 $\uparrow$ (\%) & $\uparrow$ (\%) & $\uparrow$ (\%) & $\downarrow$ (m) & $\downarrow$ & $\downarrow$ (m) \\
 \midrule
  87.47 & 78.53 & 45.32 & 98.25 & 0.11 & 114.73 \\
  99.69 & 93.03 & 51.26 & 65.13 & 0.06 & 77.22 \\
 \bottomrule
 \end{tabular}
 \\
 \begin{tabular}{c}
    \multirow{3}{*}{\rotatebox{90}{Post ft.}} \\ 
 \end{tabular}
 \begin{tabular}{l}
 \\
 \toprule
 \\
 Category \\
 \midrule
 Class Transparent \\
 Class Mirror \\
 \bottomrule
 \end{tabular}
 &
 \begin{tabular}{rrrr | rr }
 \\
 \toprule
 bad-2 & bad-4 & bad-6 & bad-8 & MAE & RMSE \\
 $\downarrow$ (\%) & $\downarrow$ (\%) & $\downarrow$ (\%) & $\downarrow$ (\%) & $\downarrow$ (px.) & $\downarrow$ (px.) \\
 \midrule
 69.27 & 53.42 & 42.08 & 34.25 & 14.90 & 22.36 \\
 68.19 & 48.11 & 42.53 & 38.75 & 33.69 & 42.43 \\
 \bottomrule
 \end{tabular}
 & &
 \begin{tabular}{rrr | rrr }
 \\
 \toprule
 $\delta$ $<$ 1.25 & $\delta$ $<$ 1.15 & $\delta$ $<$ 1.05 & MAE & Abs. Rel & RMSE \\
 $\uparrow$ (\%) & $\uparrow$ (\%) & $\uparrow$ (\%) & $\downarrow$ (m) & $\downarrow$ & $\downarrow$ (m) \\
 \midrule
  91.12 & 81.36 & 47.27 & 80.94 & 0.09 & 96.25 \\
  99.94 & 95.22 & 55.31 & 59.99 & 0.05 & 69.93 \\
 \bottomrule
 \end{tabular}
 \\\\
 & CREStereo \cite{li2022practical} && DPT \cite{Ranftl_2021_ICCV } 
 \end{tabular}
 }
 \caption{\textbf{Results on the Booster Stereo and Mono benchmark -- Transparent vs Mirror.} We run CREStereo \cite{li2022practical} (left column) and DPT \cite{Ranftl_2021_ICCV} (right column) by processing images at the resolution suggested by the authors. Top: results using official weights. Bottom: results after fine-tuning using the Booster training split. We evaluate on full-resolution ground-truth maps.}
 \label{tab:transp_vs_mirror}
\end{table*}

\subsection{Monocular Depth Benchmark}

We conclude by introducing the leaderboard of the Monocular benchmark, on top of which we study how monocular depth estimation networks perform when dealing with the challenges introduced by Booster.

\textbf{Off-the-shelf deep networks.} Following the protocol defined for the previous benchmarks, we run a set of off-the-shelf, state-of-the-art monocular networks on the Booster Test Mono split to assess their performance.
We pick networks with freely accessible implementations and pre-trained weights, selecting those with strong generalization performance on various datasets.
Hence, we restrict our selection to MiDaS \cite{midas}, DPT \cite{Ranftl_2021_ICCV}, LeRes \cite{yin2021learning}, Boosting monocular depth \cite{miangoleh2021boosting} -- this latter powered by either MiDaS (Boosting MiDaS) or LeRes (Boosting LeRes).
Tab. \ref{tab:mono_comparison} collects the outcome of our evaluation, obtained by comparing, on \textit{All} pixels at the full resolution, the predicted depths maps with the ground-truth.
Each method processes input images either at their original resolution (F) or resizes them to match the resolution used in the authors' code on a single 3090 RTX GPU. Since the selected models produce up-to-scale depth maps, we rescale each prediction by estimating the scale and shift factors \cite{midas} according to the corresponding ground-truth depth. For models predicting depth maps at a lower resolution, we upsampled them with nearest-neighbor interpolation to match the ground-truth size.
We point out that DPT \cite{Ranftl_2021_ICCV} achieves the best results in almost all metrics, with a large gap w.r.t. competitors for the most challenging $\delta$. 
Interestingly, Boosting \cite{miangoleh2021boosting} generally does not improve the performance of the base network, independently from the one employed, \ie MiDaS \cite{midas} or LeRes \cite{yin2021learning}. We believe this depends on the fact that Boosting itself merges predictions by the base network coupled with it at different resolutions, with the main goal of increasing high-frequency details in the prediction. However, if the base network presents large diffuse errors, such as in the third column of \cref{fig:mono} (transparent jar), these will also be present in the Boosting prediction (sixth column).

\textbf{Finetuning by the Booster training data.} 
As we did for stereo networks, we select two competitive methods (here MiDaS  \cite{midas} and DPT \cite{Ranftl_2021_ICCV}) for fine-tuning them on the Booster Stereo training set. As for previous cases, this allows us to check whether the availability of annotated scenes can effectively improve the result in the presence of the aforementioned open challenges, even when dealing with monocular depth estimation. We run 50 epochs on batches of random 2878$\times$2105 crops, $\sim$70\% of the full resolution to preserve enough context information, extracted from randomly horizontal flipped and random color jittered images. Crops are resized to the network resolutions suggested in the original papers, with an initial learning rate set to 1e-5, modulated by an exponential decay scheduler. Results are reported in the last row of \cref{tab:mono_comparison}. 
As we can see, MiDAs achieves comparable results before and after fine-tuning, with minor drops on most metrics, except for $\delta<1.05$ showing a $2\%$ increase in accuracy, while DPT shows some small yet consistent improvement on any metric.
This behavior is in stark contrast to what was observed for stereo networks, with these latter being improved by large margins. We believe that, as monocular networks reason on high-level cues such as relative object size and scene context information to estimate depth, they require a much larger amount of supervised data to consistently improve their performance. This fact again highlights how this problem remains open, outlining an interesting yet challenging future research direction.

\textbf{Evaluation on challenging regions.}
Finally, in \cref{tab:segmentation_mono} we evaluate the accuracy of the predicted disparities in regions of increasing difficulty. 
Purposely, we select networks from the previous evaluation, \ie MiDaS, and DPT, and evaluate them on subsets of pixels defined by our manually annotated masks before and after fine-tuning. \cref{tab:segmentation_mono} collects the outcome of this experiment, together with numbers measured on all valid pixels, reported on top as a reference.
The results are variable, with some metrics being better after fine-tuning and others being not.

Starting from the original models, MiDaS shows a much less intuitive behavior in challenging regions. Indeed, scores for class 2 pixels are much better than those for class 0 and 1. On the contrary, DPT behaves much more similar to what we observed with stereo networks, with metrics progressively dropping from class 0 to class 3 except in a few cases (RMSE). We ascribe this effect to the architectural difference between the two: MiDaS indeed features a local receptive field, whereas DPT exploits Transformers to globally process the image. Being context crucial for estimating depth out of a single image, the way such a context is accessed by the network affects its capacity to deal with different classes. As such, MiDaS probably results less affected by the ambiguities of the challenging surfaces since it lacks global context, yet at the expense of the performance on class 0 pixels.

We believe this also affects the effectiveness of fine-tuning: on the one hand, MiDaS seems less prone to benefit from fine-tuning, with minor improvements on $\delta<1.05$ metric alone for classes 0 to 2, and a slightly larger improvement on class 3.  
On the other, DPT shows consistent improvements in any of the classes for $\delta <$ 1.25 and 1.15, with a minor drop for $\delta < 1.05$ in classes 1 and 2. The same occurs for error metrics such as MAE, Abs. Rel. and RMSE. 
This suggests that, probably, the lack of global perception hinders the fine-tuning process, whereas DPT can properly benefit from it.
Overall, we observe worse performance on the most challenging class (Class 3) w.r.t. the easiest one (Class 0) for what concerns relative metrics $\delta$s and Abs. Rel.; 
This behaviour is not always evident from MAE and RMSE errors. However, we argue they are largely more affected by outliers in the monocular case.
Finally, Figure \cref{fig:mono} reports qualitative results for the monocular networks employed pre and post-finetuning (ft). Specifically, we use the same image shown in \cref{fig:balanced_qualitatives} to ease comparison with results achieved by stereo networks, although it is not part of the test mono split.
We can see that in general monocular networks can better handle transparent surfaces than stereo networks. Nonetheless, they still fail dramatically in challenging situations such as the one depicted in \cref{fig:mono_failures} with a large transparent surface covering half of the scene.

\section{Transparent vs Mirror Surfaces}
We further analyze here stereo and monocular networks when dealing with completely transparent surfaces, such as glass panels, or with those that are entirely reflective, such as mirrors.  
To do so, we select the test set images containing these kinds of surfaces and we manually segment mirrors and completely transparent surfaces. Then, we evaluate the performance of the two top-performing networks for stereo and monocular setups, namely CREStereo \cite{li2022practical} and DPT\cite{Ranftl_2021_ICCV}, respectively, specifically on these types of surfaces. 
The results of this investigation are presented in Tab. \ref{tab:transp_vs_mirror}, shedding light on the respective capabilities of the methods in handling reflective and transparent scenes.
Concerning the stereo setup, we can observe how the mirror class yields dramatically higher average error -- with MAE and RMSE being 3 times higher with respect to the transparent class. We ascribe this to the different consequences the two have on the visual appearance of the objects in the images: indeed, in the presence of transparent surfaces, a stereo network will very likely match pixels in the background and thus estimate a disparity lower than the one of the surface in the foreground; on the contrary, when framing a mirror the two images expose some content violating the search assumptions made by stereo algorithms -- i.e., constraining the search range on the right image to be on the left of the pixel coordinates we aim at matching with on the left one -- causing the stereo network to produce large artifacts, heavily impacting on the final error metrics. 
This is evident even after fine-tuning, with CREStereo still yielding much larger errors on the mirror class, proving that the network learned to reduce the presence of these artifacts, yet not entirely to deal with these surfaces. For what concerns the monocular case, we observe the exact opposite behavior, with the errors on transparent objects being higher compared to those belonging to the mirror class. We ascribe this to the very different working principle behind monocular networks, mostly based on contextual information: indeed, the presence of a transparent object is potentially harder to perceive, since it let the background appear in the image as if the object itself were not there. On the contrary, a mirror shows some content being largely different from the one in the background, easing its perception based on contextual information. This behavior remains unaltered even after fine-tuning, showing that DPT can further improve its accuracy on both classes, yet achieving lower errors on the mirror one.

\section{Conclusion, Limitations and Future work}
 In this paper, we have presented Booster, a novel dataset consisting of 606 high-resolution images containing many transparent and specular surfaces.
It contains dense and accurate ground-truth labels obtained through a novel deep space-time stereo pipeline with manually annotated material segmentation masks. 
We define three benchmarks for different applications: balanced and unbalanced stereo matching and monocular depth estimation. 
Our experiments unveil some intriguing challenges in depth estimation from monocular or stereo images and suggest new promising research directions.

\textbf{Limitations.} One of the main limitations of the dataset concerns the limited number of labeled images, which might be not sufficient to train robust monocular networks.
Moreover, the deep space-time pipeline and the small baseline used for annotations constrain the collected scenes to frame indoor environments.

\textbf{Future directions.} Follow-up works may be devoted to i) investigating deep architectures and training schemes to perform robust depth estimation also on challenging materials, ii) investigating a different ground-truth acquisition pipeline to collect annotated data also in outdoor settings, iii) building a large-scale dataset focusing on non-Lambertian materials using either graphic simulations or more recent Neural Radiance Fields \cite{mildenhall2021nerf} techniques, iv) scanning scenes through successive depth layers, to gather multiple depths at transparent/reflective objects which can be useful for applications such as augmented reality.

\section*{Acknowledgements} We gratefully acknowledge the funding support of Huawei Technologies Oy (Finland).

\bibliographystyle{IEEEtran}
\bibliography{egbib,depth}

\vspace{-1.25cm}
\begin{IEEEbiography}
[{\includegraphics[width=1in,height=1.25in,clip,keepaspectratio]{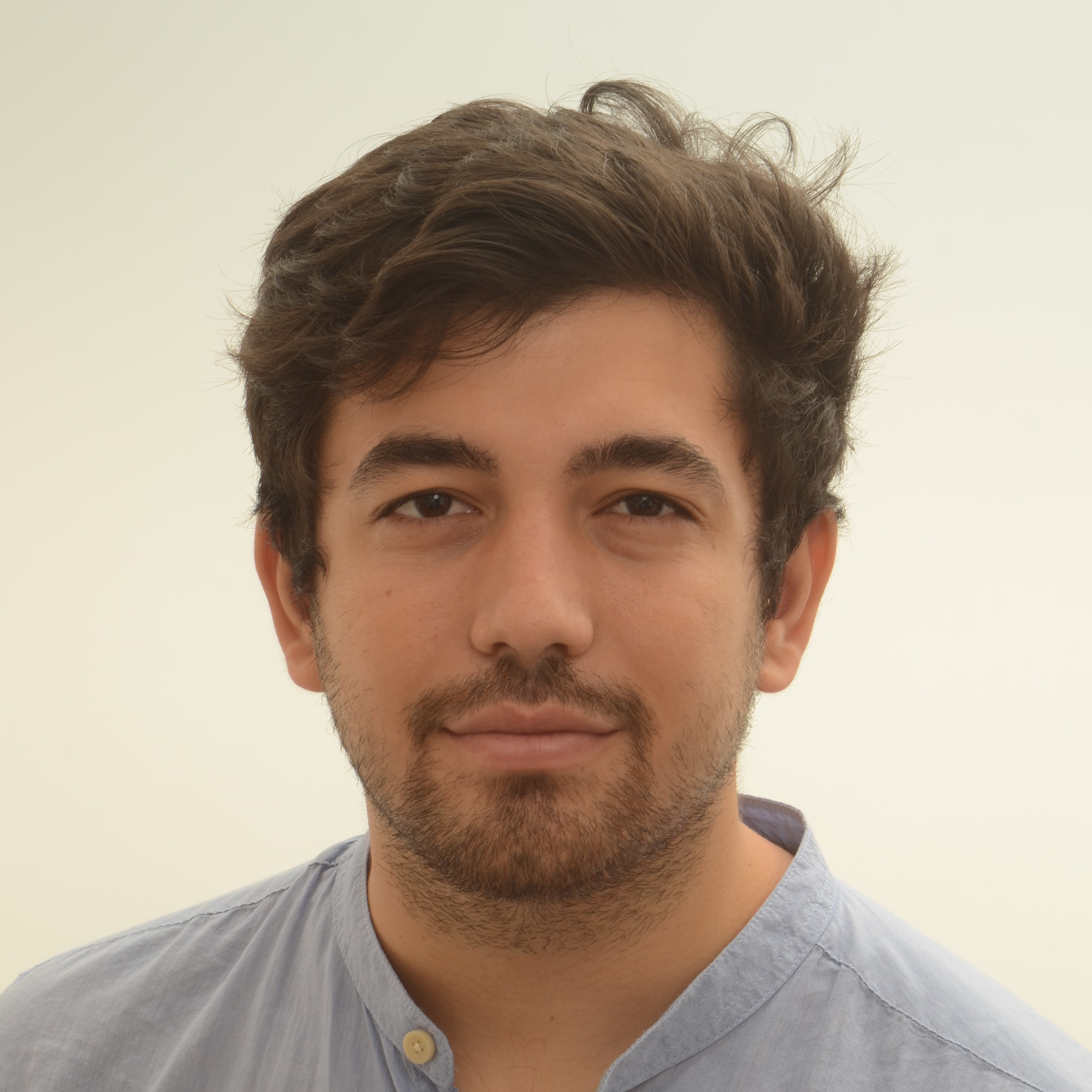}}]
{Pierluigi Zama Ramirez} received his PhD degree in Computer Science and Engineering in 2021. He has been a Research Intern at Google for 6 months and is currently a Post-Doc at the University of Bologna. He co-authored 15 publications on several computer vision research topics such as semantic segmentation, depth estimation, optical flow, domain adaptation, virtual reality, and 3D reconstruction.
\end{IEEEbiography}
\vspace{-1.75cm}
\begin{IEEEbiography}
[{\includegraphics[width=1in,height=1.25in,clip,keepaspectratio]{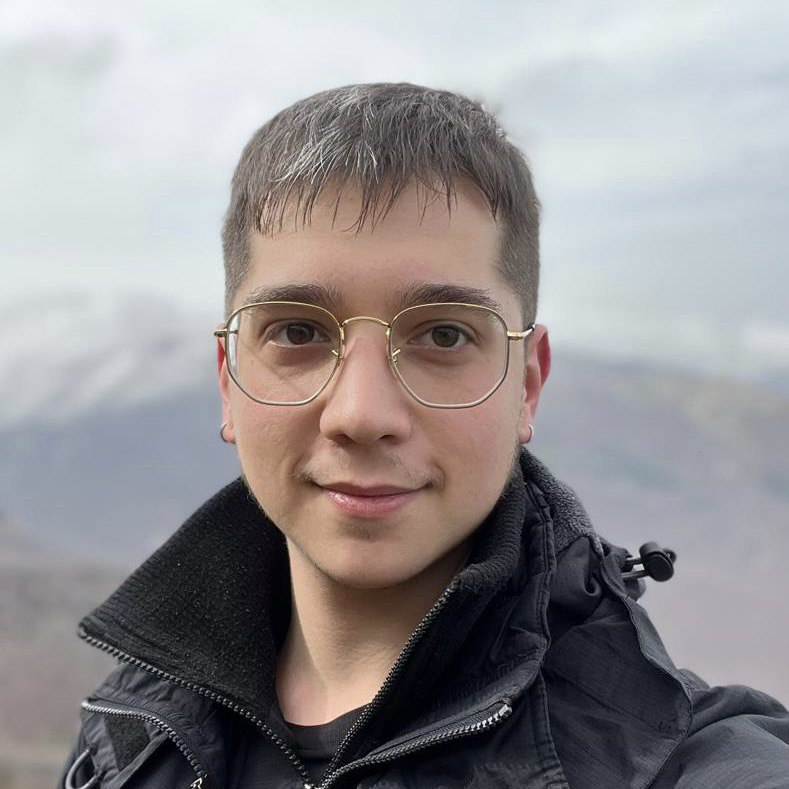}}]
{Alex Costanzino}
received his Master’s degree in Artificial Intelligence from University of Bologna in 2022. 
He is currently pursuing his PhD degree in Computer Science and Engineering at the Computer Vision Laboratory (CVLab), University of Bologna.
His research focuses on Artificial Intelligence and Deep Learning techniques for Computer Vision, in particular for Depth Estimation and Novel View Synthesis.
\end{IEEEbiography}
\vspace{-1.75cm}
\begin{IEEEbiography}
[{\includegraphics[width=1in,height=1.25in,clip,keepaspectratio]{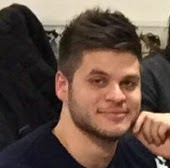}}]{Fabio Tosi} received his PhD degree in Computer Science and Engineering from University
of Bologna in 2021. Currently, he is a Post-doc
researcher at Department of Computer Science
and Engineering, University of Bologna. His re-
search interests include deep learning and depth sensing related topics.
\end{IEEEbiography}
\vspace{-1.75cm}
\begin{IEEEbiography}
[{\includegraphics[width=1in,height=1.25in,clip,keepaspectratio]{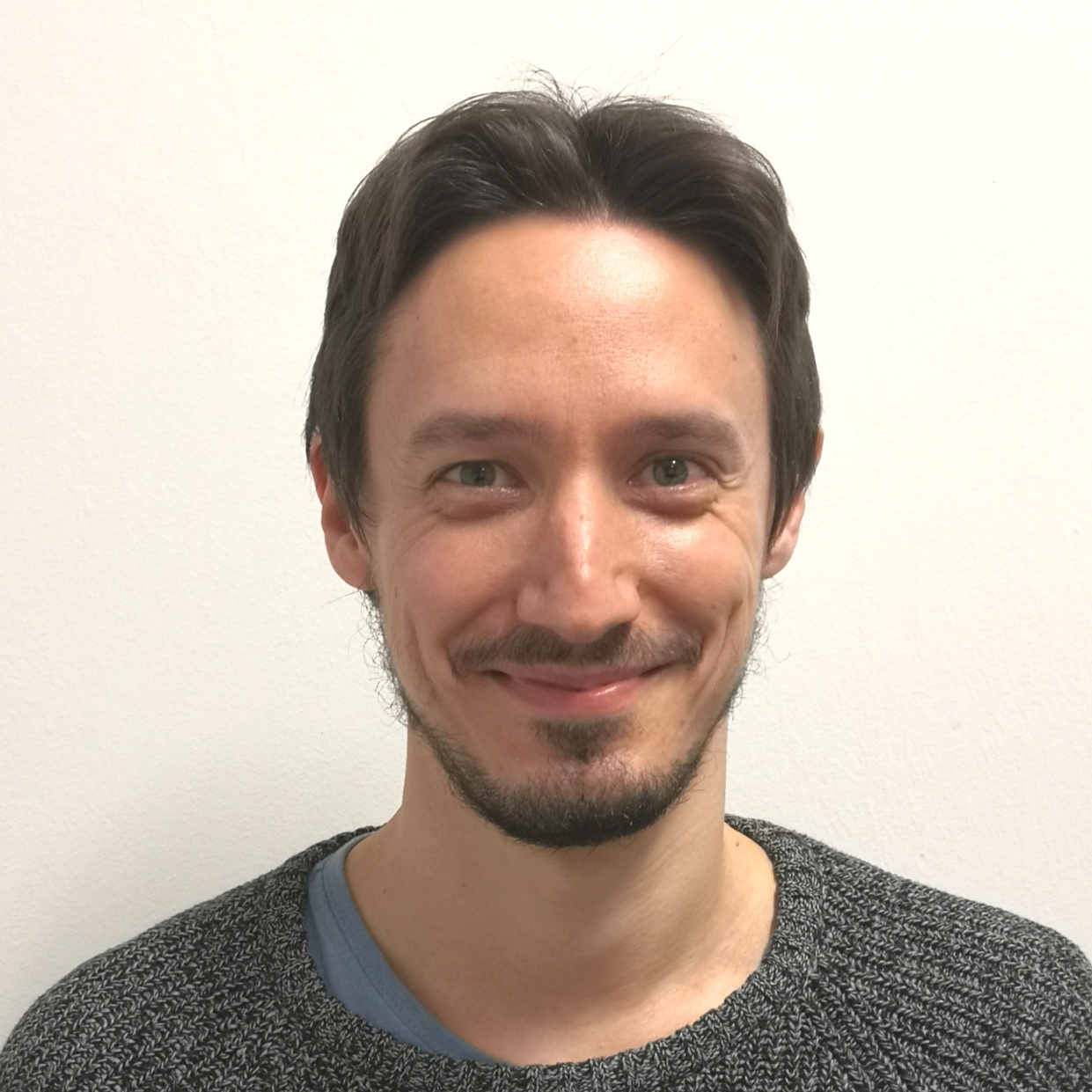}}]{Matteo Poggi}
received Master degree in Computer Science and PhD degree in Computer Science and Engineering from University of Bologna in 2014 and 2018 respectively. Currently, he is assistant professor at Department of Computer Science and Engineering, University of Bologna. He co-authored 65 papers, mostly about deep learning for depth estimation and related tasks.
\end{IEEEbiography}
\vspace{-1.75cm}
\begin{IEEEbiography}
[{\includegraphics[width=1in,height=1.25in,clip,keepaspectratio]{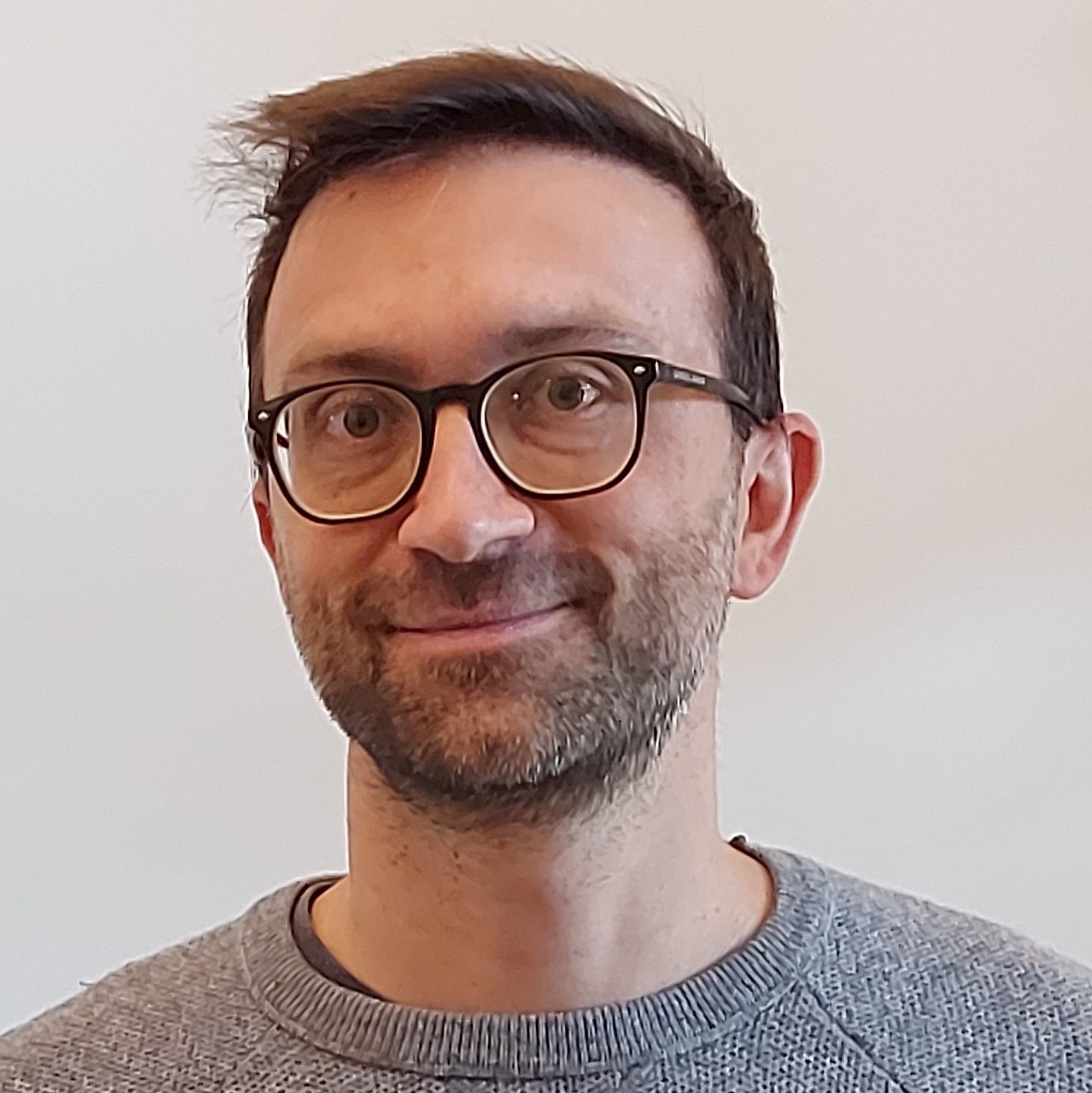}}]{Samuele Salti}
is currently associate professor at the Department of Computer Science and Engineering (DISI) of the University of Bologna, Italy. 
His main research interest is computer vision, in particular 3D computer vision, and machine/deep learning applied to computer vision problems.
Dr. Salti has co-authored more than 50 publications and 8 international patents. 
In 2020, he co-founded the start-up eyecan.ai. 
\end{IEEEbiography}
\vspace{-1.75cm}
\begin{IEEEbiography}
[{\includegraphics[width=1in,height=1.25in,clip,keepaspectratio]{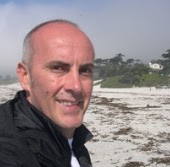}}]{Stefano Mattoccia}
is currently an associate professor at the Department of Computer Science and Engineering of the University of Bologna. His research activity concerns computer vision, mainly focusing on depth perception and related tasks. In these fields, he co-authored more than 130 scientific publications. He is Senior IEEE member.
\end{IEEEbiography}
\vspace{-1.75cm}
\begin{IEEEbiography}
[{\includegraphics[width=1in,height=1.25in,clip,keepaspectratio]{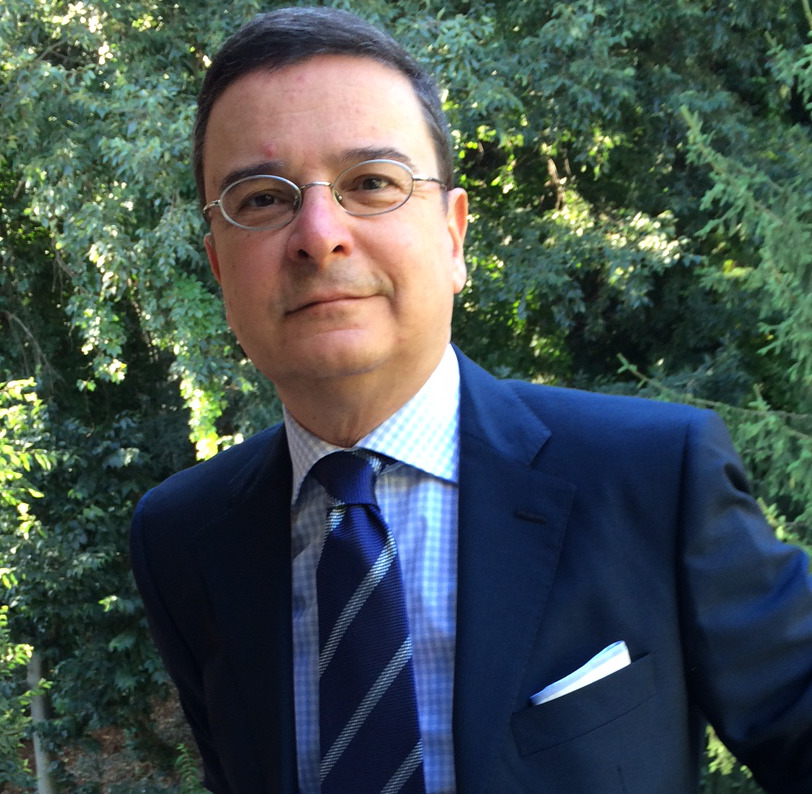}}]
{Luigi Di Stefano}
received the PhD degree in electronic engineering and computer science from the University of Bologna in 1994. He is currently a full professor at the Department of Computer Science and Engineering, University of Bologna, where he founded and leads the Computer Vision Laboratory (CVLab). His research interests include image processing, computer vision and machine/deep learning. He is the author of more than 150
papers and several patents. He has been scientific consultant for major companies in the fields of computer vision and machine learning. 
He is a member of the IEEE Computer Society and the IAPR-IC.
\end{IEEEbiography}

\end{document}